\documentclass[10pt]{article}
\usepackage[preprint]{tmlr}

\usepackage{amsmath,amsfonts,bm}

\def\eqref#1{equation~\ref{#1}}

\def\1{\bm{1}}

\DeclareMathAlphabet{\mathsfit}{\encodingdefault}{\sfdefault}{m}{sl}
\SetMathAlphabet{\mathsfit}{bold}{\encodingdefault}{\sfdefault}{bx}{n}

\usepackage{hyperref}

\usepackage{xstring}
\makeatletter
\DeclareRobustCommand{\autoref}[1]{%
  \IfBeginWith{#1}{app:}%
    {\hyperref[#1]{Appendix~\ref*{#1}}}%
    {\hyperref[#1]{Section~\ref*{#1}}}%
}
\makeatother
\usepackage{url}
\usepackage{booktabs}
\usepackage{multirow}
\usepackage{graphicx}
\graphicspath{{figures/}}
\usepackage{amsmath, amssymb}
\usepackage{xcolor}
\usepackage{subcaption}
\usepackage{tikz}
\usepackage{etoc}
\usepackage[edges]{forest}
\usetikzlibrary{arrows.meta,positioning,calc}
\usepackage{todonotes}
\usepackage{placeins}  %
\newcommand{\finding}[1]{$\mathcal{F}\!#1$}

\definecolor{darkgreen}{HTML}{005e19}
\definecolor{darkblue}{HTML}{240394}
\newcommand{\code}[1]{\texttt{#1}}

\begin{document}

\title{Many Circuits, One Mechanism: Input Variation and Evaluation Granularity in Circuit Discovery}

\author{\name Alireza Bayat Makou \email alireza.makou@tu-darmstadt.de \\
        \addr UKP Lab, Technical University of Darmstadt
        \AND
        \name Jingcheng Niu \email jingcheng.niu@tu-darmstadt.de \\
        \addr UKP Lab, Technical University of Darmstadt
        \AND
        \name Subhabrata Dutta \email subhabrata.dutta@tu-darmstadt.de \\
        \addr UKP Lab, Technical University of Darmstadt
        \AND
        \name Iryna Gurevych \email iryna.gurevych@tu-darmstadt.de \\
        \addr UKP Lab, Technical University of Darmstadt \\
        National Research Center for Applied Cybersecurity ATHENE}

\maketitle

\begin{abstract}
Circuit discovery methods identify subgraphs that explain specific model behaviors, and structural differences between discovered circuits are commonly interpreted as evidence of distinct mechanisms. We test this assumption by drawing input tokens from bands defined by their frequency in the pretraining data while holding the task fixed. The discovered circuits appear specialized by frequency when compared structurally, but functional and representational analyses show no reliable evidence of corresponding differences in their computations. We term this mismatch \emph{phantom specialization}. Using the Literal Sequence Copying task across four frequency bands plus a control sampled according to token frequency, we extract 75 circuits from five Pythia models (70M--1.4B). We find that structurally distinct circuits implement the same computation: band-specific edges transfer broadly across bands, a core shared across most bands recovers at least 99\% of circuit performance in models above 70M, and causal interchange interventions confirm that internal representations are interchangeable across frequency bands. A smaller replication on a subject-verb agreement task shows the same pattern: circuits differ structurally, transfer broadly across bands, and the core shared by most bands recovers nearly all of the circuit's accuracy in all five models. Repeated extractions within the same frequency band further suggest that discovery algorithms sample from an equivalence class of valid subgraphs rather than recovering a unique mechanism. Standard evaluation practice obscures this pattern: source-level evaluation inflates apparent faithfulness, while edge-level evaluation reveals the many-to-one mapping from structure to function. We find no reliable evidence that frequency bands are processed by different computations. In this work, we do not test specialization at the level of token positions or of features inside a component. Our results show that structural differences between circuits are not sufficient evidence for distinct mechanisms, and that exposing this requires edge-level evaluation and cross-condition transfer tests.\footnote{Our code and data are available at: \url{https://github.com/UKPLab/arxiv2026-phantom-specialization}.}
\end{abstract}

\section{Introduction}
\label{sec:introduction}

Mechanistic interpretability seeks to reverse-engineer the internal computations of neural networks, analogous to recovering the high-level algorithms of a compiled program~\mbox{\citep{olah2022mechanistic, elhage2021mathematical}}. An important object of study in this field is the \emph{circuit}: a subgraph of a model's computational graph that implements a specific behavior~\mbox{\citep{olah2020zoom, NEURIPS2023_34e1dbe9, rai2024practical}}. In this graph, nodes typically correspond to attention heads and MLP blocks, and edges represent the flow of information between components through the residual stream. Early circuits were identified manually through iterative causal interventions, ablating or patching individual components and tracing their effects back from the output, for behaviors such as in-context induction~\mbox{\citep{olsson2022context}}, indirect object identification (IOI)~\mbox{\citep{wang2022interpretability}}, and greater-than comparison~\mbox{\citep{hanna2023does}}. However, this process is slow and difficult to scale. To address this, automated methods have been proposed: ACDC iteratively prunes edges via activation patching~\mbox{\citep{NEURIPS2023_34e1dbe9}}; attribution-based methods such as EAP-IG approximate edge importance through gradients~\mbox{\citep{syed2024attribution, hanna2024have, kramar2024atp}}; and edge masking approaches formulate discovery as differentiable optimization~\mbox{\citep{bhaskar2024finding, yu2024functional}}. These methods are grounded in the causal abstraction framework, which provides a theoretical foundation for mechanistic interpretability and unifies activation and path patching within a common causal language~\mbox{\citep{geiger2023causal}}. Despite their differences, these methods share a common goal: to identify \emph{a} circuit responsible for a given behavior.

However, this goal assumes that each model--task pair has a single stable circuit to discover, an assumption increasingly challenged by recent work. Treating a discovered circuit as \emph{the} circuit for a task can therefore lead researchers to overlook other equally valid circuits produced under different extraction settings, prompts, or input distributions. Prior studies often refer to circuits as singular objects, such as ``the IOI circuit''~\mbox{\citep{wang2022interpretability}} and ``the induction circuit''~\mbox{\citep{olsson2022context}}. Several well-known challenges already complicate circuit discovery, including faithfulness metrics that are not robust~\mbox{\citep{miller2024transformer}}, sensitivity to extraction hyperparameters~\mbox{\citep{meloux2025mechanistic, NEURIPS2023_34e1dbe9}}, redundant backup behavior~\mbox{\citep{mcgrath2023hydra}}, and non-identifiability, among other open problems catalogued in recent surveys~\mbox{\citep{sharkey2025open}}. \mbox{\citet{meloux2025everything}} exhaustively enumerate candidate explanations in toy models and find that multiple structurally distinct circuits, interpretations, and causal alignments can satisfy current mechanistic interpretability criteria simultaneously. Moreover, \mbox{\citet{garriga2024adversarial}} show that the IOI circuit can fail systematically on benign inputs and certain input types, demonstrating that circuit behavior is not uniform across inputs even at evaluation time. These findings raise a further question: if discovered circuits are sensitive to both the extraction method and the evaluation inputs, might they also be sensitive to the input distribution used for extraction? Motivated by the \emph{universality hypothesis}~\mbox{\citep{olah2020zoom}}, existing work has tested circuit consistency across models and settings, finding recurring motifs in toy settings and across scales~\mbox{\citep{chughtai2023toy, tigges2024llm}}, across languages~\mbox{\citep{ferrando2024similarity}}, and across tasks~\mbox{\citep{merullo2023circuit, mondorf2025circuit}}, often relying primarily on structural overlap to assess similarity~\mbox{\citep{hanna2024have}}. Recent work has begun to probe circuit consistency within a single model: \mbox{\citet{franco2026finding}} show that per-prompt circuits for IOI cluster into families, and several studies document shared components across related tasks, as well as substantial overlap across reasoning subtasks within a task~\mbox{\citep{merullo2023circuit, lan2024towards, mondorf2025circuit, dutta2024think}}. However, these comparisons vary either the task or the prompt structure. It therefore remains unclear whether structural differences arise when only the token frequency distribution of the inputs changes while the task remains fixed, and whether any such differences are functionally meaningful. We address this gap by drawing input tokens from different frequency bands while holding the task fixed. Circuit discovery methods then recover structurally distinct circuits that implement the same computation.

\begin{figure*}[t]
\centering
\includegraphics[width=\textwidth]{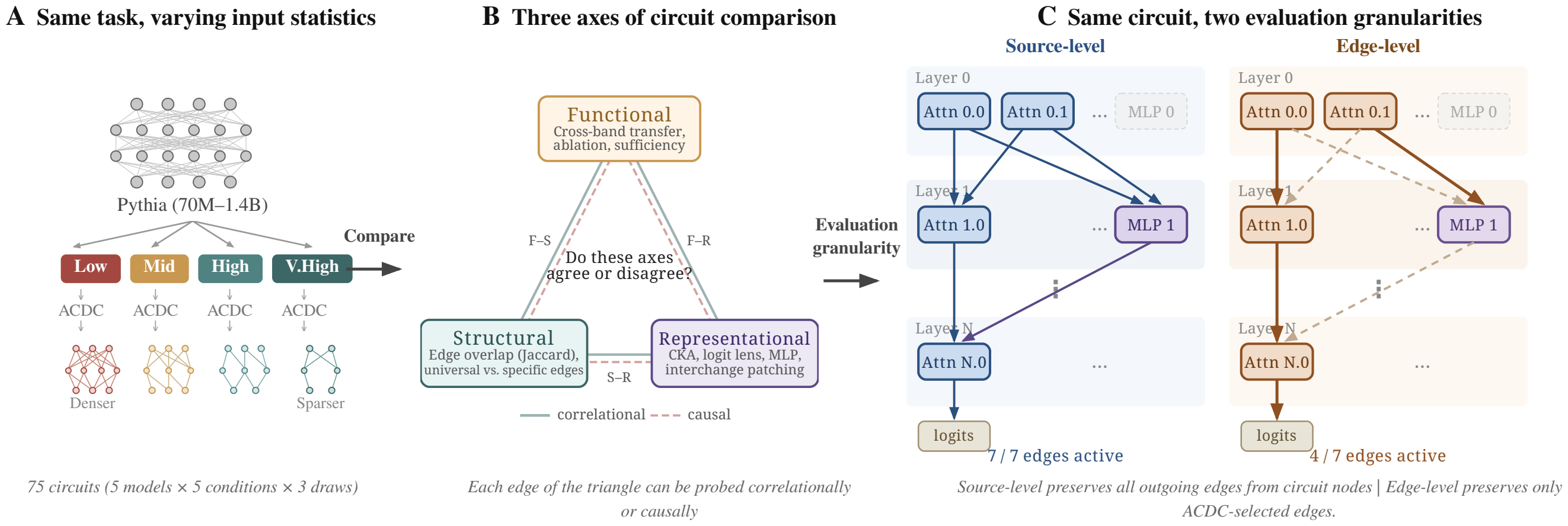}
\caption{Experimental design and analytical framework.
\textbf{(A)}~ACDC circuit discovery on Literal Sequence Copying across
four token frequency bands plus a control condition in five Pythia models (70M--1.4B) produces 75~circuits (5~models $\times$ 5~conditions $\times$ 3~draws).
\textbf{(B)}~Circuits are compared along three axes (structural, functional,
and representational), connected by both correlational and causal analyses.
\textbf{(C)}~The same circuit can be evaluated at two levels. Source-level evaluation preserves all outgoing edges from circuit nodes, whereas edge-level evaluation preserves only the edges selected by the discovery algorithm.
The figure depicts the primary LSC study. The SVA replication is described in \autoref{sec:sva_replication}.}
\label{fig:overview}
\end{figure*}

We approach the question of whether circuit structure depends on input statistics along two axes: token frequency and evaluation granularity. Token frequency, measured by how often a token occurs in the pretraining corpus, systematically affects language model behavior. Prior work finds effects on contextual representations~\mbox{\citep{zhou2021frequency}} and numerical reasoning behavior~\mbox{\citep{razeghi2022impact}}, even in tasks without semantic content~\mbox{\citep{niu2025illusion}}. This makes token frequency a natural probe for our first axis: it can be measured precisely from the training corpus, yet varies the input without changing the task definition. We use the Literal Sequence Copying (LSC) task from~\mbox{\citet{niu2025illusion}}, a non-semantic copying task in which the model must reproduce a token from an earlier occurrence in the sequence. For example, given the prompt \code{A B C D ... A B C}, the model must predict \code{D}. Because LSC minimizes semantic confounds and isolates a narrow copying behavior, it provides a relatively controlled setting for testing whether circuit structure changes even when the task remains fixed. To systematically test whether input statistics drive circuit differences, we partition the Pythia vocabulary~\mbox{\citep{biderman2023pythia}} into four frequency bands based on token frequency in The Pile~\mbox{\citep{gao2020pile}}, extract circuits with ACDC across five model scales (70M to 1.4B), and run three independent extractions per frequency band. This yields five conditions: four frequency bands plus a control sampled according to token frequency, and 75 circuits in total. We compare these circuits along structural, functional, and representational axes (Figure~\ref{fig:overview}). Our second axis is evaluation granularity. Discovered circuits can be evaluated at two levels: \emph{source-level} evaluation preserves all outgoing edges from any node that is the source of at least one selected circuit edge, while \emph{edge-level} evaluation preserves only the specific edges selected by the discovery algorithm. More broadly, the idea that causal analysis may depend on the chosen level of description~\mbox{\citep{noble2012theory, hoel2013quantifying}} motivates testing whether the causal picture changes across these two granularities. We report the following findings:
\begin{enumerate}
\setlength{\itemsep}{1pt}
\setlength{\parsep}{0pt}
\setlength{\topsep}{1pt}
\item[\finding{1}] Circuits extracted from different frequency bands are structurally distinct, with low-frequency circuits systematically larger.
\item[\finding{2}] These structural differences do not reflect functional specialization: band-specific edges transfer broadly across all frequency conditions, and a core shared across most bands suffices for nearly all circuit performance.
\item[\finding{3}] Extraction is noisy across repeated runs, yet this noise is functionally irrelevant: structurally different draws implement the same computation.
\item[\finding{4}] Source-level evaluation inflates apparent circuit accuracy by up to 85 percentage points relative to edge-level evaluation, masking the absence of specialization and obscuring the many-to-one mapping from structure to function.
\end{enumerate}

We call this mismatch \emph{phantom specialization}: the circuits look different, but evaluating each circuit on every frequency condition provides no reliable evidence of corresponding differences in what they compute. This mismatch reflects a many-to-one mapping from circuit structure to function. Multiple edges can be individually dispensable because computation can route through redundant or compensatory pathways~\mbox{\citep{mcgrath2023hydra}}, and many parameterizations can implement the same function~\mbox{\citep{bushnaq2024using}}. Greedy pruning must therefore choose which edges to keep; small differences in token frequency can tip these choices differently, producing structurally distinct circuits that implement the same computation. More broadly, discovery algorithms may sample from an equivalence class of structurally distinct yet functionally interchangeable subgraphs, a form of \emph{circuit degeneracy} analogous to degeneracy in biological systems~\mbox{\citep{edelman2001degeneracy}}. ACDC's band-specific edges are therefore better understood as byproducts of extraction than as adaptations. Because they improve performance across conditions rather than encode band-specific computations, they resemble \emph{spandrels} in the sense of \mbox{\citet{gould1979spandrels}}. Crucially, phantom specialization is detectable only at edge-level granularity. Source-level evaluation hides this by keeping many extra edges, so a circuit looks more faithful than it is at the edge level~\mbox{\citep{hoel2013quantifying}} (\finding{4}).

Taken together, our results show a many-to-one mapping from circuit structure to circuit function: structurally distinct circuits can be functionally interchangeable under cross-condition transfer, and discovery algorithms can sample from an equivalence class of functionally interchangeable subgraphs. Whether this degeneracy is visible depends on evaluation granularity: source-level evaluation systematically obscures it, whereas edge-level evaluation reveals it. More broadly, this dependence on granularity is consistent with the view that causal structure may appear different at different levels of description~\mbox{\citep{noble2012theory, hoel2013quantifying}}. For circuit discovery, the implication is that structural differences between circuits should not be interpreted as evidence of distinct mechanisms without first evaluating each circuit on every condition. A positive control supports this interpretation. We built a synthetic cue-controlled task and trained small transformers on it. Under one cue, the model must copy the token that followed the query's earlier occurrence in the prompt; under the other, it must ignore the prompt and answer from a memorized token-to-token mapping. Before circuit discovery, direct ablations confirmed that the two rules relied on different component groups in six trained models. For these models, our circuit discovery pipeline detects both structural and functional divergence (\autoref{sec:two_route_control}). This suggests that the frequency band result reflects the LSC setting, not an inability of the tests to detect real differences.
A smaller replication on a second task family, the syntactic task of subject--verb agreement, points in the same direction. Across the same frequency bands, circuits in all five models remain structurally distinct, yet their band-specific edges transfer broadly (\autoref{sec:sva_replication}). Because the SVA replication covers only the core structural and transfer analyses, we limit our claims accordingly.
The LSC negative result has two limitations. First, the transfer test can miss very small effects: depending on the model, it has 80\% power only for advantages on the extraction band of 0.013--0.026 accuracy points or larger (\autoref{app:transfer_power}). Second, our circuits are defined by edges between components rather than by token positions (\autoref{app:positional_attribution}) or features inside a component (Section~\ref{sec:conclusion}), so specialization at those finer levels is outside the scope of this study. Within those limits, we find no reliable evidence of functional specialization on LSC. In our setting, edge-level evaluation is the more informative faithfulness metric, and aggregating repeated extractions by retaining only edges selected in most runs provides a practical way to identify stable structure. While broader limitations of circuit discovery pipelines have been flagged as open problems~\mbox{\citep{sharkey2025open}}, and non-uniqueness has been demonstrated in toy settings~\mbox{\citep{meloux2025everything}}, our results show that the same phenomenon persists across five Pythia model scales, from 70M to 1.4B parameters, and across all frequency conditions we study. A single discovered circuit should therefore not be treated as uniquely identifying the underlying mechanism.

Section~\ref{sec:related_work} reviews background and related work. Section~\ref{sec:methodology} describes the experimental setup. Section~\ref{sec:results} presents evidence for apparent frequency-dependent specialization. Section~\ref{sec:phantom} tests whether the observed differences reflect genuine specialization, Section~\ref{sec:discussion} discusses implications and limitations, and Section~\ref{sec:conclusion} concludes.

\section{Background and Related Work}
\label{sec:related_work}
\label{sec:background_section}

\subsection{Circuit Discovery and Evaluation}
\label{sec:rw_circuit_discovery}

\begin{figure*}[t]
\centering
\scalebox{0.7}{%
\begin{tikzpicture}[
    box/.style={
      rounded corners=3pt,
      thick,
      align=center,
      font=\small,
      text width=2.2cm,
      minimum height=1.1cm,
      inner sep=3pt
    },
    arr/.style={-{Stealth[length=2.2mm]}, thick},
    lbl/.style={font=\scriptsize, align=center},
    panel/.style={font=\bfseries\small},
    node distance=4mm
]

\node[box, fill=teal!6, draw=teal!35] at (-7.8, 0) (a1)
  {\textbf{1. Select}\\[-1pt] {\scriptsize behavior \&\\ dataset}};
\node[box, fill=teal!10, draw=teal!35, right=of a1] (a2)
  {\textbf{2. Define}\\[-1pt] {\scriptsize graph \&\\ granularity}};
\node[box, fill=teal!16, draw=teal!40, right=of a2] (a3)
  {\textbf{3. Localize}\\[-1pt] {\scriptsize ablation /\\ patching}};
\node[box, fill=teal!22, draw=teal!40, right=of a3] (a4)
  {\textbf{4. Interpret}\\[-1pt] {\scriptsize hypothesize\\ component roles}};
\node[box, fill=teal!28, draw=teal!45, right=of a4] (a5)
  {\textbf{5. Evaluate}\\[-1pt] {\scriptsize faithfulness \&\\ sufficiency}};
\node[box, fill=teal!38, draw=teal!55, right=of a5] (a6)
  {\textbf{Circuit}\\[-1pt] {\scriptsize sparse faithful\\ subgraph}};

\foreach \i/\j in {a1/a2, a2/a3, a3/a4, a4/a5, a5/a6}
  \draw[arr, teal!50] (\i) -- (\j);

\draw[-{Stealth[length=2.2mm]}, thick, dashed, teal!35]
  (a4.south) .. controls +(0,-0.9) and +(0,-0.9) .. (a3.south);
\node[lbl, text=teal!60] at ($(a3)!0.5!(a4)+(0,-1.0)$) {iterate};

\node[panel] at (0, 2.0) (labA) {\textsf{(a) Manual Circuit Discovery}};

\node[box, fill=orange!6, draw=orange!35] at (-6.5, -4.8) (b1)
  {\textbf{1. Full graph}\\[-1pt] {\scriptsize all candidate\\ edges}};
\node[box, fill=orange!14, draw=orange!40, right=of b1] (b2)
  {\textbf{2. Score}\\[-1pt] {\scriptsize relevance /\\ masking}};
\node[box, fill=orange!20, draw=orange!40, right=of b2] (b3)
  {\textbf{3. Select}\\[-1pt] {\scriptsize sparse\\ subgraph}};
\node[box, fill=orange!26, draw=orange!40, right=of b3] (b4)
  {\textbf{4. Evaluate}\\[-1pt] {\scriptsize faithfulness \&\\ sufficiency}};
\node[box, fill=orange!38, draw=orange!55, right=of b4] (b5)
  {\textbf{Circuit}\\[-1pt] {\scriptsize sparse faithful\\ subgraph}};

\foreach \i/\j in {b1/b2, b2/b3, b3/b4, b4/b5}
  \draw[arr, orange!50] (\i) -- (\j);

\draw[-{Stealth[length=2.2mm]}, thick, dashed, orange!35]
  (b4.south) .. controls +(0,-0.9) and +(0,-0.9) .. (b2.south);
\node[lbl, text=orange!60] at ($(b2)!0.5!(b4)+(0,-1.0)$)
  {adjust threshold / objective};

\node[panel] at (0, -2.8) (labB) {\textsf{(b) Automated Circuit Discovery}};

\end{tikzpicture}%
}%

\caption{Two paradigms for circuit discovery. \textbf{(a)}~Manual discovery iterates between localizing components, interpreting their roles, and evaluating the resulting hypothesis~\mbox{\citep{rai2024practical}}. \textbf{(b)}~Automated methods score candidate edges, select a sparse subgraph, and evaluate its faithfulness across sparsity thresholds or objectives~\mbox{\citep{NEURIPS2023_34e1dbe9, syed2024attribution, bhaskar2024finding}}.}
\label{fig:discovery_pipelines}
\end{figure*}
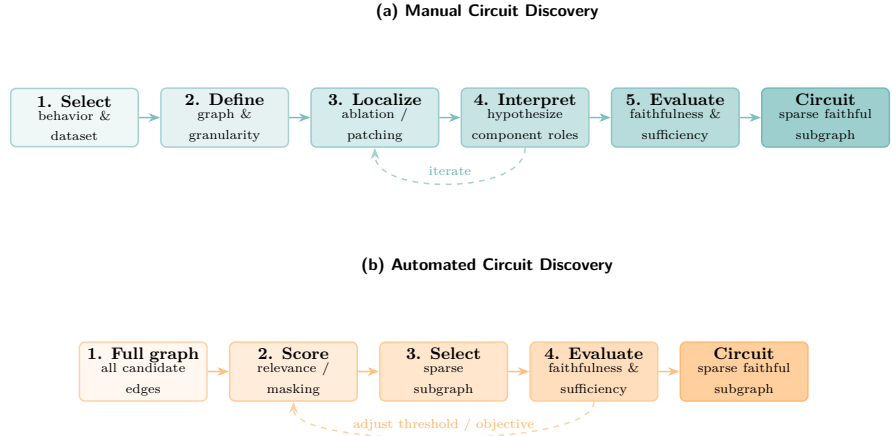

Circuit discovery aims to identify computational subgraphs, or \emph{circuits}, that explain how a neural network implements a specific behavior~\mbox{\citep{olah2020zoom, elhage2021mathematical}}. Throughout this paper, we use \emph{circuit} as an operational term: a sparse subgraph of components and connections that recovers a substantial fraction of the full model's performance on the target task. This definition intentionally depends on the method. Discovered circuits vary with how the computational graph is defined, how edges are selected, and how faithfulness is measured. In a transformer, attention heads and MLP blocks communicate through the residual stream: each component reads from the stream and adds its output back to it. This information flow can be represented as a directed acyclic graph (DAG)~\mbox{\citep{elhage2021mathematical}}. Nodes in this graph correspond to attention heads and MLP blocks, and edges indicate that one component reads an output written by another to the residual stream. A circuit is a subgraph of this DAG that is relevant to a target task~\mbox{\citep{NEURIPS2023_34e1dbe9, rai2024practical}}.

Most circuit discovery methods build on \emph{activation patching}~\mbox{\citep{vig2020investigating, meng2022locating}}. Activation patching replaces a component's activation with the corresponding activation from a counterfactual input and measures the resulting change in the model's output. The causal abstraction framework~\mbox{\citep{geiger2023causal}} subsequently formalized this practice as an interchange intervention within a causal model. Activation patching involves three forward passes: (1)~a \emph{clean run} on the original prompt, caching all intermediate activations; (2)~a \emph{corrupt run} on a modified input from which the signal relevant to the target behavior has been removed; and (3)~a \emph{patch run} in which the activation of a specific component is swapped between runs while the rest of the computation proceeds normally. Patching can be applied in two directions: noising (replacing a clean component with its corrupt value to test necessity) and denoising (restoring a corrupt component to its clean value to test sufficiency)~\mbox{\citep{rai2024practical, heimersheim2024use}}. The choice of corruption method (zero, mean, or resampling ablation) can materially affect patching results, and may therefore change the recovered circuit for the same model and task~\mbox{\citep{heimersheim2024use, meloux2025mechanistic}}. For an extended treatment of circuit discovery methods and evaluation, a brief review of activation patching for unfamiliar readers, as well as a detailed account of the corruption method, see \autoref{app:circuit_discovery}.

Automated methods search for sparse subgraphs relevant to the target task (Figure~\ref{fig:discovery_pipelines}b). \mbox{\citet{NEURIPS2023_34e1dbe9}} introduced ACDC, which automates circuit discovery by iteratively removing edges with the smallest measured effect on the model's output under activation patching. Subsequent automated methods fall into several families. Attribution-based methods approximate edge effects via gradient-based scores, including EAP~\mbox{\citep{syed2024attribution}}, EAP-IG~\mbox{\citep{hanna2024have}}, and AtP*~\mbox{\citep{kramar2024atp}}. On the MIB benchmark, EAP-IG-inputs achieves the best average performance, while other attribution-based methods remain competitive~\mbox{\citep{mueller2025mib}}. Edge pruning and masking approaches formulate discovery as differentiable optimization over computational graph masks, including Edge Pruning~\mbox{\citep{bhaskar2024finding}} and DiscoGP~\mbox{\citep{yu2024functional}}. Methods using one forward pass such as ACC++~\mbox{\citep{franco2026finding}} instead trace information flow without patching. Despite differences in search procedure, these methods are typically used to produce a single subgraph for a target behavior, an assumption increasingly questioned by recent work~\mbox{\citep{sharkey2025open, meloux2025everything}}.

This assumption has received limited scrutiny. \mbox{\citet{meloux2025mechanistic}} show that gradient-based attribution scores exhibit high variance across inputs. \mbox{\citet{NEURIPS2023_34e1dbe9}} note that ACDC's behavior is sensitive to threshold and metric choice. \mbox{\citet{sharkey2025open}} emphasize broader open problems in current circuit discovery pipelines. Using ACC++, a tracing method based on SVD of QK matrices that uses one forward pass rather than activation patching, \mbox{\citet{franco2026finding}} show that per-prompt circuits for IOI cluster into recurring ``prompt families.'' This result suggests that some task-level instability reflects systematic differences between prompt templates. Our work tests this concern systematically along the input distribution axis. By running multiple independent extractions on distinct frequency bands within the same task, we show that the instability and non-uniqueness suggested by prior work can appear as structurally distinct but functionally interchangeable circuits, precisely the scenario that makes structural comparison unreliable.

Evaluating circuit faithfulness presents a separate set of challenges. Faithfulness can be operationalized via logit difference~\mbox{\citep{wang2022interpretability}} or KL divergence~\mbox{\citep{NEURIPS2023_34e1dbe9, zhang2023towards}}, but aggregate faithfulness can mask per-input failures~\mbox{\citep{miller2024transformer}}, and many published circuits do not pass strict tests of both sufficiency and necessity~\mbox{\citep{shi2024hypothesis, garriga2024adversarial}}.

A key methodological distinction is evaluation granularity. Most circuit studies evaluate at the level of components or source nodes, preserving all outgoing connections from selected components rather than only the specific discovered edges. In this paper, we refer to this as \emph{source-level} evaluation, and reserve \emph{edge-level} evaluation for tests that preserve only the specific connections identified by the discovery algorithm. This distinction matters because circuit faithfulness scores can be sensitive to ablation methodology~\mbox{\citep{miller2024transformer}}. We show that source-level evaluation systematically inflates apparent circuit accuracy (Section~\ref{sec:inflation}). We also introduce cross-condition transfer as a complementary evaluation axis.

\subsection{Circuits Across Conditions}
\label{sec:rw_landmark}

Mechanistic interpretability studies have identified circuits for specific tasks, including Indirect Object Identification~\mbox{\citep{wang2022interpretability}}, in-context copying via induction heads~\mbox{\citep{olsson2022context}}, greater-than comparison~\mbox{\citep{hanna2023does}}, copy suppression~\mbox{\citep{mcdougall2024copy}}, successor operations~\mbox{\citep{ICLR2024_2722a0cc}}, and extractive question-answering~\mbox{\citep{basu2025mechanistic}}. A recurring question is whether the same motifs appear across models. \mbox{\citet{olah2020zoom}} proposed the \emph{universality hypothesis}, according to which analogous features and circuits recur across models. Subsequent empirical work found consistent IOI circuits across Pythia scales~\mbox{\citep{tigges2024llm}}, cross-lingual agreement circuits in Gemma~2B~\mbox{\citep{team2024gemma,ferrando2024similarity}}, and recurring successor heads across model families~\mbox{\citep{ICLR2024_2722a0cc}}.

Several studies have begun comparing circuits across conditions, but each addresses a different slice of the problem and leaves specific gaps that our design targets. \mbox{\citet{hanna2024have}} argued that structural overlap alone is insufficient for comparing circuits and advocated faithfulness-based comparison. However, they compare circuits found by different \emph{methods}, not circuits extracted by the same method from different \emph{inputs}, and do not quantify the source-level versus edge-level evaluation gap. We group the remaining studies by what they vary: the task, the model or training stage, or the inputs within a task.

\paragraph{Across tasks.}
\mbox{\citet{merullo2023circuit}} showed that IOI and Colored Objects circuits share ${\sim}78\%$ of attention heads in GPT-2~Medium~\mbox{\citep{radford2019language}}, but used source-level (head) granularity, a single extraction per task, no systematic transfer matrix, and no statistical controls.
\mbox{\citet{dutta2024think}} found that within chain-of-thought reasoning in Llama-2~7B~\mbox{\citep{touvron2023llama}}, the same induction-like attention heads serve decision-making, copying, and inductive reasoning subtasks, with heads identified for induction alone retaining $>$90\% accuracy across all subtask types. However, they did not test whether the \emph{structural differences} between subtask circuits are functionally meaningful.
\mbox{\citet{lan2024towards}} identified shared subcircuits for sequence continuation tasks (numerals, number words, months) in GPT-2~Small and Llama-2~7B with partial cross-task transfer, but at node-level granularity and without multiple extractions or causal confirmation.
\mbox{\citet{mondorf2025circuit}} found both shared and task-specific substructure for string edit operations, again at node level and across \emph{different tasks} rather than within a single task.

\paragraph{Across models, scales, and training time.}
\mbox{\citet{tigges2024llm}} tracked IOI and successor circuits at checkpoints spanning 300B tokens of Pythia training and across model scales up to 2.8B, finding algorithmic stability despite component-level fluctuations. However, they vary \emph{training stage and model scale}, not input distribution. They also use source-level (head) granularity without testing whether the structural variation they document is functionally irrelevant.
\mbox{\citet{ferrando2024similarity}} found consistent subject--verb agreement circuits across English and Spanish in Gemma~2B, including a language-independent number direction, but studied a single model with a single extraction per language and no edge-level evaluation.

\paragraph{Within a single task.}
Most directly, \mbox{\citet{franco2026finding}} showed that even within a single task (IOI), per-prompt circuits are not unique: different prompt templates induce systematically different circuits in GPT-2~Small, Pythia-160M, and Gemma-2~2B~\mbox{\citep{team2024gemma2}}. However, their prompt template variation changes the \emph{task structure} (e.g., name order), and they report systematically different prompt family mechanisms without testing cross-condition transfer or functional equivalence.
\mbox{\citet{sun-2025-circuit}} showed that circuit stability across input instances predicts generalization, but did not test whether structural \emph{instability} between conditions translates to functional difference.
\mbox{\citet{mahaut2025repetitions}} found that models can produce superficially similar repetition behavior using qualitatively different internal computations, providing an example of genuine specialization that contrasts with the phantom specialization we identify.

Across this body of work, three specific gaps persist. First, most comparisons vary the \emph{task} or \emph{prompt structure}, not the input \emph{statistics} within a fixed task. The prompt family analysis of \mbox{\citet{franco2026finding}} also varies prompt structure alongside prompt wording. Second, to our knowledge, all studies cited above evaluate at source or node level, and none systematically compare edge-level with source-level evaluation or quantify the resulting inflation. Third, none combine multiple independent extractions, cross-condition transfer matrices, and causal confirmation to distinguish genuine specialization from extraction artifacts. Our experimental design addresses all three gaps simultaneously.

\subsection{Token Frequency and Model Internals}
\label{sec:rw_frequency}

Token frequencies typically follow a Zipfian distribution~\mbox{\citep{zipf1949human}} and are known to affect language models at multiple levels of analysis. At the representation level, \mbox{\citet{zhou2021frequency}} showed that low-frequency words occupy more identifiable but less diverse regions of BERT's~\mbox{\citep{devlin2019bert}} embedding space, introducing systematic distortions in similarity metrics. \mbox{\citet{merullo2025linear}} found that linear representations of factual relations in OLMo~\mbox{\citep{groeneveld2024olmo}} and GPT-J~\mbox{\citep{gpt-j}} form only when subject-object co-occurrences exceed a frequency threshold, linking representational structure directly to pretraining statistics. At the behavioral level, \mbox{\citet{razeghi2022impact}} demonstrated that few-shot numerical reasoning accuracy correlates with operand frequency in pretraining data, and \mbox{\citet{niu2025illusion}} showed that copying by induction heads degrades for rare tokens even though the task logic is frequency-invariant. \mbox{\citet{pinto2024fair}} revealed that weight decay disproportionately depreciates low-frequency tokens, a bias invisible in aggregate training loss. At the neuron level, \mbox{\citet{liu2025distributed}} showed that the processing of rare tokens in Pythia and GPT-2 relies on distributed specialization. The relevant MLP neurons are coordinated but spatially scattered throughout the shared architecture, with no evidence of dedicated modules or attention routing. However, all of this work examines frequency effects at the level of representations, outputs, or individual neurons, not at the level of the computational graph. Our work extends to the circuit level, using token frequency as a controlled perturbation axis for circuit discovery and testing whether observed structural differences reflect genuine functional specialization.

\subsection{Redundancy, Non-Uniqueness, and Multiple Valid Mechanisms}
\label{sec:rw_degeneracy}

Several lines of evidence suggest that transformer computations are not implemented by a single, uniquely determined pathway. \mbox{\citet{meloux2025everything}} provide the most systematic demonstration of this problem. Using small MLPs trained on Boolean functions, they exhaustively enumerate valid explanations under two MI strategies: first finding a circuit and then interpreting its algorithm, or first proposing an algorithm and then locating it in the network. They find overwhelming non-identifiability. Valid explanations include multiple zero-error circuits, multiple interpretations per circuit, and multiple causally aligned algorithms per network, and their number grows dramatically with model size. \mbox{\citet{mcgrath2023hydra}} described the \emph{Hydra effect}. When important attention heads are ablated, backup heads compensate by increasing their contribution, indicating built-in redundancy. \mbox{\citet{wang2022interpretability}} observed similar backup behavior in the IOI circuit, and \mbox{\citet{ortu2024competition}} showed that factual recall and counterfactual reasoning engage competing mechanisms that dynamically trade off depending on context. \mbox{\citet{dutta2024think}} documented this redundancy at scale in Llama-2~7B. During chain-of-thought reasoning, multiple parallel pathways simultaneously write the answer token to the output residual stream, each collecting information from a different segment of the context (i.e., generated CoT, question, or few-shot examples). This provides a direct example of functionally interchangeable parallel circuits in a production-scale model. At the parameter level, \mbox{\citet{bushnaq2024using}} connected this redundancy to degeneracy in the loss landscape, showing that mechanistically distinct parameter configurations can achieve equivalent loss. This parallels the biological concept of \emph{degeneracy}: structurally different elements performing the same function~\mbox{\citep{edelman2001degeneracy}}.

An alternative to edge-level circuit discovery is to define circuits over learned interpretable features rather than architectural components, using sparse autoencoders~\mbox{\citep{marks2025sparse}} or transcoders~\mbox{\citep{dunefsky2024transcoders}}. Such \emph{feature-level} circuits may resolve ambiguities arising from polysemantic components, precisely the setting in which phantom specialization is most likely to occur. Whether feature-level circuits exhibit the same non-uniqueness is an important open question (Section~\ref{sec:conclusion}).

Our notion of \emph{phantom specialization} extends these observations from individual components or parameters to complete circuits: under controlled input variation, the discovery algorithm samples from an equivalence class of structurally distinct but functionally interchangeable circuits. This is distinct from the Hydra effect (compensation after intervention) and from loss-landscape degeneracy (parameter-level non-uniqueness). \mbox{\citet{franco2026finding}} provide independent evidence from a different method and perturbation axis. Their ACC++ traces reveal that the same canonical IOI component (e.g., the name-mover head) can carry entirely different internal signals across prompt templates while preserving its high-level role. This is a signal-level analog of our finding that structurally distinct circuits implement identical functions. Our work provides an empirical, large-scale answer to the open question posed by \mbox{\citet{meloux2025everything}}. They ask whether non-identifiability persists beyond toy models, and our findings in Pythia models up to 1.4B parameters suggest that it does.

\section{Experimental Setup}
\label{sec:background} %
\label{sec:preliminaries} %
\label{sec:brief_related_work} %
\label{sec:methodology} %

\subsection{Data and Task}
\label{sec:data}

\paragraph{Frequency bands.}
\label{sec:bands}

We measure each token's unigram frequency in The Pile, Pythia's pretraining
corpus, and report it as occurrences per million corpus tokens. This
normalizes each token's raw count by the corpus size and places frequencies
on an interpretable scale. After excluding the sparsely populated tails below
the 1st and above the 99th percentiles, we divide the remaining range into
five equal-width bands on a $\log_{10}$ scale. Each band therefore spans the
same multiplicative frequency ratio (approximately $3.5\times$), which makes
within-band heterogeneity comparable across the frequency spectrum while
keeping within-band variation below $4\times$ and retaining at least 500
tokens per band (boundary selection, ranges, and token counts are reported in
\autoref{app:band_design}).
The final design comprises five core bands (\textit{very\_low} through
\textit{very\_high}), two exploratory tail bands, and one \textit{control}
condition sampled according to token frequency (Table~\ref{tab:bands}).
Throughout, \emph{frequency band} refers to these token frequency strata of
the vocabulary; the term is unrelated to the rotation frequencies of rotary
positional embeddings (RoPE) or to any spectral decomposition of activations.

\paragraph{Confound control.}
\label{sec:confounds}

Token frequency covaries with other token properties. For example, rare tokens tend to be longer and more often capitalized (e.g., proper nouns), either of which could independently affect model behavior.
To isolate frequency as the primary axis of variation, we restrict the token pool to the 26{,}863 tokens that represent standalone words (identified by the BPE space prefix) and contain only unaccented Latin letters (a--z/A--Z), excluding subword fragments, digits, punctuation, and non-Latin scripts (see \autoref{app:vocab} for the full filtering pipeline).
Within this pool, character length and capitalization still vary across frequency bands, so we control both during task design by restricting tokens to lowercase and matching the character length distribution across bands (Section~\ref{sec:task}; \autoref{app:token_pools}).
Although character length is not directly visible to the model once a word is represented as a single token, it remains correlated with other token-level properties that may vary across frequency bands. We therefore match character length distributions across bands as a conservative proxy control to better isolate frequency as the variable of interest.
Our filtering and matching steps reduce, as much as possible, confounding from token properties that covary with rarity in natural text. The conclusions should therefore be interpreted as evidence about controlled frequency variation rather than about all naturally co-occurring effects of token rarity.

\paragraph{Task: Literal Sequence Copying.}
\label{sec:task}

\begin{figure}[t]
\centering
\resizebox{0.9\linewidth}{!}{%
\begin{tikzpicture}[
  tok/.style={minimum width=0.55cm, minimum height=0.5cm,
    font=\scriptsize, align=center},
]
  \node[font=\scriptsize\bfseries, text=blue!60] at (-1.1, 0) {Clean};
  \foreach \i/\lab/\col in {
    0/S1/blue!15, 1/S2/blue!15, 2/S3/blue!15,
    3/S4/blue!15, 4/S5/blue!15,
    5/T/orange!30,
    6/R1/gray!15, 7/R2/gray!15, 8/\dots/white,
    9/R10/gray!15,
    10/S1/blue!15, 11/S2/blue!15, 12/S3/blue!15,
    13/S4/blue!15, 14/S5/blue!15} {
    \node[tok, fill=\col, draw=gray!50] at (\i*0.7, 0) {\lab};
  }
  \node[tok, fill=orange!20, draw=orange!70!red, thick] at (15*0.7, 0) {\textbf{?}};
  \draw[thick, orange!70!red, -{Stealth[length=2mm]}] (15*0.7, -0.5) -- (5*0.7, -0.5)
    node[midway, below, font=\scriptsize] {copy \textbf{T}};

  \node[font=\tiny, text=gray] at (2*0.7, 0.55) {source};
  \node[font=\tiny, text=orange!70!red] at (5*0.7, 0.55) {target};
  \node[font=\tiny, text=gray] at (7.5*0.7, 0.55) {distractors};
  \node[font=\tiny, text=gray] at (12*0.7, 0.55) {repetition};

  \node[font=\scriptsize\bfseries, text=red!60] at (-1.1, -1.6) {Corrupt};
  \foreach \i/\lab/\col in {
    0/S1/blue!15, 1/S2/blue!15, 2/S3/blue!15,
    3/S4/blue!15, 4/S5/blue!15,
    5/T/orange!30,
    6/R1/gray!15, 7/R2/gray!15, 8/\dots/white,
    9/R10/gray!15,
    10/X1/red!15, 11/X2/red!15, 12/X3/red!15,
    13/X4/red!15, 14/X5/red!15} {
    \node[tok, fill=\col, draw=gray!50] at (\i*0.7, -1.6) {\lab};
  }

  \node[font=\tiny, text=red!60] at (12*0.7, -1.05) {random tokens};
  \node[font=\tiny, text=red!60] at (12*0.7, -2.35) {no repetition signal};
\end{tikzpicture}%
}%
\caption{LSC sequence structure and corruption. \textbf{Top (clean):} the model observes a source prefix
$S_{1\text{--}5}$ followed by target $T$, a distraction segment
$R_{1\text{--}10}$, and a repetition of the source prefix, then must predict $T$
at the final position. All 16 tokens are sampled from the same frequency band.
\textbf{Bottom (corrupt):} the repeated source prefix is replaced with random tokens $X_{1\text{--}5}$ from the same frequency band, destroying the repetition signal while leaving all other positions unchanged.}
\label{fig:lsc_sequence}
\end{figure}
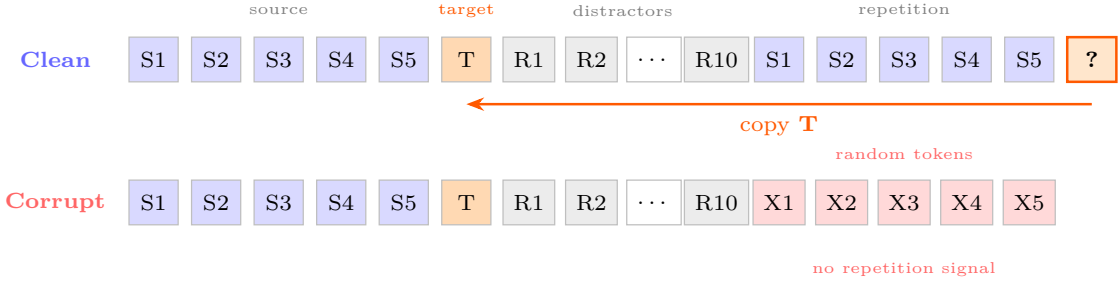

We use \textit{Literal Sequence Copying} (LSC), a non-semantic induction
task introduced by \mbox{\citet{niu2025illusion}}, who showed that LSC accuracy in the in-context learning setting decreases for rare tokens even though the task requires only pattern matching.
Each input sequence (Figure~\ref{fig:lsc_sequence}) contains a five-token
source prefix $S_{1\text{--}5}$, a target token $T$, a ten-token distraction
segment $R_{1\text{--}10}$, and a repetition of the source prefix.
The distraction segment requires the model to attend to the earlier source
prefix rather than rely on local bigram statistics.
The model's task is to predict $T$ by copying it from an earlier occurrence of the pattern (Figure~\ref{fig:lsc_sequence}). No instructions or demonstrations are provided, so the model must rely on its learned copying mechanism through autoregressive next-token prediction. All 16 unique tokens in each sequence (5 source tokens, 1 target, 10 distractors; the 5 source tokens are repeated, yielding 21 positions) are sampled randomly without replacement from the same frequency band, making accidental sequential patterns negligible.
Because LSC is non-semantic, frequency can vary across examples without
changing the task itself.

To control the confounds described above, we restrict the LSC sampling pools
to lowercase tokens and match their character length distributions across
bands, yielding 703 tokens per band for four core bands (low through
\texttt{very\_high}).
The \texttt{very\_low} band retains only 97 lowercase tokens after filtering.
Length matching uses the intersection of the character length counts
available in every participating band. This means that each band is
subsampled to the per-length counts shared by all bands. As a result,
including the sparse length profile of the \texttt{very\_low} band, which is
skewed toward longer tokens, would reduce each of the other four pools to far fewer
than 703 tokens. The band is therefore excluded from the matched pools and
the main analyses, but it is not dropped from the study. We extract and
evaluate it separately with the unmatched pool variant, without length
matching (\autoref{app:very_low}).
The control condition draws from all four matched bands (2{,}812 tokens total), weighting each token by its pretraining frequency so that the resulting sequences reflect the model's natural training distribution.
Circuit discovery thus operates on five conditions: four frequency bands plus
control (\autoref{app:control_detail} verifies that the control circuit
does not differ significantly from frequency-specific circuits).
We generate 1,500 sequences per band (70/15/15 train/validation/test) across three
independent \emph{draws}: each draw samples a fresh set of tokens from the pool and
generates new sequences, yielding both a new dataset and (after extraction) a new
circuit. Generation procedure and seeds in \autoref{app:lsc_generation}.

\paragraph{Second task family.}
To test whether the core findings are limited to LSC, we additionally
constructed a subject--verb agreement (SVA) task in which subject frequency
varies across bands. SVA has been used in prior causal and circuit analyses of
language models~\mbox{\citep{finlayson2021causal, ferrando2024similarity}}. Our
version uses six-token prompts in which a subject from a given frequency band
must agree with its verb across an intervening noun of the opposite number
(the \emph{attractor}). We score a prompt as correct when the grammatical verb
form (``~is'' or ``~are'') receives a higher logit than the ungrammatical
alternative. Exact character length matching is feasible only for the low,
medium, and high SVA noun pools: the very-low band contains no eligible noun
pairs, and the very-high pool cannot be matched by length under this construction.
For each condition and draw, we use 1,500 examples split 70/15/15 into 1,050
training, 225 validation, and 225 test examples, following the same protocol
using three draws as LSC (construction in \autoref{app:sva_generation}; results in
\autoref{sec:sva_replication}).
All analyses in the remainder of the paper concern the LSC circuits
unless SVA is named explicitly.

\subsection{Models}
\label{sec:models}

We use five models from the Pythia suite~\mbox{\citep{biderman2023pythia}}:
Pythia-70m, 160m, 410m, 1b, and~1.4b (Table~\ref{tab:model_and_threshold}, left).
Pythia is well suited for this study because both its tokenizer and its
pretraining corpus, The Pile, are public, allowing us to measure frequencies
for the model's exact vocabulary on the data used to pretrain it
(token frequency extraction details in \autoref{app:corpus}).
We use the standard (undeduped) variant throughout, as the deduped models
performed worse on LSC in preliminary evaluations.
The five scales span a $20\times$ parameter range, allowing us to test
whether circuit-level patterns hold across model capacity.
All larger models achieve accuracy close to the ceiling (${\geq}92.9\%$ top-1; Section~\ref{sec:base_performance}), confirming that the task is well within their capacity.
Pythia-70m is a boundary case because its base accuracy is low (30--67\%)
and several metrics diverge from those of the four larger models.
We include it to examine whether phantom specialization extends to low-capacity models, but qualify results throughout when Pythia-70m diverges from the larger models.

\subsection{Circuit Extraction}
\label{sec:pipeline}

We discover circuits using ACDC~\mbox{\citep{NEURIPS2023_34e1dbe9}},
as implemented in the AutoCircuit library~\mbox{\citep{miller2024transformer}}.
We use AutoCircuit rather than the original ACDC codebase because
the latter exhibited metric collapse on Pythia models during
preliminary testing (failure modes are documented in \autoref{app:algorithm_choice}).
We prefer ACDC to EAP and EAP-IG~\mbox{\citep{syed2024attribution, hanna2024have}}.
These methods rank edges independently using gradient-based scores and retain
the top $k$, which can produce disconnected edge sets that do not form
functional circuits. At circuit sizes matched to ACDC, EAP-IG recovers only
2--15\% of base model accuracy for models ${\geq}160$M, compared with 79--99\%
for ACDC (\autoref{app:method_comparison}).
ACDC requires paired clean and corrupted inputs. For LSC, the corrupted input
replaces the repeated source prefix with random tokens from the same frequency
band, destroying the repetition signal while leaving all other positions
unchanged (Figure~\ref{fig:lsc_sequence}; corruption procedure details in \autoref{app:corruption}).
In ACDC, an edge is pruned when applying resample ablation to it changes the
model's output by less than a threshold~$\tau$.
Formally, let $\mathcal{G} = (V, E)$ denote the full computational graph,
where $V$ comprises attention heads and MLP blocks and $E$ the connections
between them.
Given a clean input $x$ and its corrupted counterpart $\tilde{x}$,
\emph{resample ablation} of edge $(u, v)$ replaces the activation sent from
$u$ to $v$ with the value $u$ computed on~$\tilde{x}$.
For a set of retained (active) edges $C \subseteq E$, let $p_C(\cdot \mid x, \tilde{x})$
denote the model's output distribution when edges in~$C$ carry clean
activations and resample ablation is applied to all other edges.
ACDC iteratively prunes each edge $e$ from the current circuit~$C$ for which
\begin{equation}
\mathrm{KL}\!\bigl(p_{\text{base}}(\cdot \mid x) \;\big\|\;
  p_{C \setminus \{e\}}(\cdot \mid x, \tilde{x})\bigr) < \tau,
\label{eq:acdc_prune}
\end{equation}
where $p_{\text{base}} \equiv p_E$ is the output distribution of the full,
unablated model.
We use factorized patching, which allows ACDC to prune individual edge connections rather than all outputs of a component at once. We compute the KL divergence in Equation~\ref{eq:acdc_prune} at the final sequence position.
Following standard practice, we aggregate each edge across token positions,
assigning it a single keep-or-prune decision that applies jointly at all
positions rather than a separate decision at each position~\mbox{\citep{haklay2025position}}.
In our LSC corruption, the clean and corrupted prompts differ only at the five repeated prefix positions, so resample ablation directly changes activations only there. This does not mean the computation is confined to those positions: earlier tokens can still affect the prediction through causal attention (\autoref{app:positional_attribution}; \autoref{sec:limitations}).

The pruning threshold~$\tau$ balances minimality and faithfulness: increasing
$\tau$ permits more pruning and therefore produces smaller circuits, but
tolerates larger deviations from the full model's output.
Because no reference circuit for LSC on Pythia exists, we select~$\tau$ using
only circuit size and faithfulness.
Crucially, we perform threshold selection on the \emph{control} band rather
than on any frequency-specific band. Since the control condition samples
tokens with weights based on pretraining frequency, it is agnostic to the frequency
partition and prevents overfitting~$\tau$ to a particular band's
characteristics.
The selected~$\tau^*$ is then applied uniformly to all five conditions, so
any structural differences between the resulting circuits cannot be
attributed to per-band threshold tuning.
On the control band, we sweep 11 thresholds evenly spaced on a logarithmic
scale from $10^{-2}$ to~$10^{-6}$, using 256~training examples (effective
per-model counts in \autoref{app:circuit_extraction}) and evaluating on the
validation split (225~examples).
The Pareto frontier over edge fraction and KL~divergence
(Figure~\ref{fig:pareto_sweeps}) identifies thresholds for which no alternative
yields both a smaller circuit and lower KL divergence. From this set, we
select one threshold~$\tau^*$ per model
(Table~\ref{tab:model_and_threshold}; per-model selection criterion in \autoref{app:threshold_selection}).
All selected points achieve 0\% ablation accuracy, confirming that the
discovered edges are necessary for the task.
Using the thresholds immediately above and below~$\tau^*$ changes circuit size by factors of approximately $2$--$3$, but changes transfer efficiency by at most 5.5 percentage points (\autoref{app:threshold_robustness}), indicating that the conclusions are robust to threshold selection.
The full extraction pipeline requires approximately 736 GPU-hours, which limits the number of independent draws and threshold settings we can explore. Our robustness checks confirm stability across adjacent thresholds and three draws per frequency band, but additional extractions could further characterize the equivalence class of valid subgraphs.

\begin{table}[t]
\centering
\caption{Model architecture and selected ACDC thresholds for the primary LSC task. The SVA replication uses a separate threshold sweep, with selected values reported in \autoref{app:sva_generation}.
The left columns show fixed architectural properties; the right columns
show circuit extraction results.
Pythia-1b has fewer layers and heads than Pythia-410m despite having more
parameters, resulting in a much smaller computational graph (10K vs.\ 81K
edges) and a higher edge fraction.
Edge~\% and KL are reported for the circuit used to select the threshold on draw~1 of the control condition (Appendix~\ref{app:threshold_selection}). Extracting 75 circuits across three draws and five conditions with training splits of 1{,}050 examples yields slightly different averages, reported in Section~\ref{sec:circuit_size}.
See Table~\ref{tab:threshold_selection} in the appendix for additional
metrics including circuit accuracy, retention, and ablation accuracy.}
\label{tab:model_and_threshold}
\small
\begin{tabular}{l ccc c | ccc}
\toprule
 & \multicolumn{4}{c|}{Architecture} & \multicolumn{3}{c}{Circuit extraction} \\
\cmidrule(lr){2-5} \cmidrule(lr){6-8}
Model & Layers & Heads & $d_{\text{model}}$ & Graph edges & $\tau^*$ & Edge~\% & KL \\
\midrule
Pythia-70m  & 6  & 8  & 512  & 1{,}324   & $1.58 \times 10^{-3}$ & 32.9 & 0.24 \\
Pythia-160m & 12 & 12 & 768  & 11{,}467  & $6.31 \times 10^{-4}$ & 12.2 & 0.28 \\
Pythia-410m & 24 & 16 & 1024 & 80{,}581  & $2.51 \times 10^{-4}$ &  4.3 & 0.29 \\
Pythia-1b   & 16 & 8  & 2048 & 10{,}009  & $1.58 \times 10^{-3}$ &  9.4 & 0.48 \\
Pythia-1.4b & 24 & 16 & 2048 & 80{,}581  & $6.31 \times 10^{-4}$ &  2.6 & 0.50 \\
\bottomrule
\end{tabular}
\end{table}

With~$\tau^*$ fixed per model, we run ACDC across all five conditions
(four frequency bands plus control) and
three draws (75~circuits total),
evaluating each circuit on the held-out test
split (225~examples) using four standard metrics%
\footnote{Defined in \autoref{app:eval_metrics}.}
(Section~\ref{sec:evaluation}).
The validation split is used exclusively during threshold selection,
preventing information leakage between the two phases.
Each circuit is also evaluated on the test splits of the other four bands,
yielding a $5 \times 5$ cross-band transfer matrix.
As a robustness check, we compare each circuit against 100 random baselines, each containing the same number of edges sampled uniformly from the full graph (without connectivity constraints). Every discovered circuit outperforms all 100 random baselines (\autoref{app:random_baselines}).

\subsection{Evaluation Framework}
\label{sec:evaluation}
\label{sec:granularities}

We evaluate every circuit using four standard
metrics~\mbox{\citep{wang2022interpretability}}.
For a circuit $C \subseteq E$ and test set~$D$:
\begin{align}
\text{Faithfulness:}\quad
  & \mathbb{E}_{x \in D}\!\bigl[
    \mathrm{KL}\!\bigl(p_{\text{base}}(\cdot \mid x) \;\big\|\;
    p_C(\cdot \mid x, \tilde{x})\bigr)\bigr],
  \label{eq:faithfulness} \\
\text{Sufficiency:}\quad
  & \mathrm{Acc}(C, D) = \tfrac{1}{|D|}
    \textstyle\sum_{(x,y) \in D}
    \mathbf{1}\!\bigl[\arg\max p_C(\cdot \mid x, \tilde{x}) = y\bigr],
  \label{eq:sufficiency} \\
\text{Necessity:}\quad
  & \mathrm{Acc}_{\text{abl}}(C, D) = \mathrm{Acc}(E \setminus C,\, D),
  \label{eq:necessity} \\
\text{Minimality:}\quad
  & |C|\,/\,|E|.
  \label{eq:minimality}
\end{align}

Each metric is computed at two levels of granularity, which determine
how activations flow through the graph.
In \emph{edge-level} evaluation, for each $(u, v) \in E$ the activation
from $u$ to $v$ is clean (computed on~$x$) if $(u,v) \in C$; otherwise,
it is replaced with the corresponding activation computed on~$\tilde{x}$.
In \emph{source-level} evaluation, let
$V_C = \{u \in V : \exists\, v,\; (u,v) \in C\}$ be the set of nodes
contributing at least one circuit edge.
All outgoing edges from any $u \in V_C$ carry clean activations, even those
not in~$C$; only edges originating from nodes outside~$V_C$ are ablated.
Source-level evaluation is therefore more permissive because it allows
information to flow along unselected edges whenever their source component
contributes at least one selected circuit edge.
Throughout, we report edge-level as the primary metric and use
source-level comparisons to expose how evaluation granularity affects
apparent circuit accuracy (Section~\ref{sec:inflation}).
All statistical tests use nonparametric methods (Kruskal-Wallis~\mbox{\citep{kruskal1952use}}, Mann-Whitney~U~\mbox{\citep{mann1947test}}, Wilcoxon signed-rank~\mbox{\citep{wilcoxon1945individual}}, Jonckheere-Terpstra~\mbox{\citep{jonckheere1954distribution,terpstra1952asymptotic}}, Spearman~\mbox{\citep{spearman1904proof}}) with Benjamini-Hochberg false discovery rate correction~\mbox{\citep{benjamini1995controlling}} at $\alpha = 0.05$.\footnote{Test catalogue and FDR procedure in \autoref{app:stat_summary}.}

\section{Evidence for Frequency-Dependent Specialization}
\label{sec:results}

\label{sec:experiments_stage1}

We characterize the 75 extracted circuits through four analyses, presented in the subsections below together with a descriptive summary of circuit size and sparsity (a further 15 circuits without length matching for the excluded very\_low band are analyzed separately in \autoref{app:very_low}). First, we evaluate each of the five models on all five conditions (four frequency bands plus control) to measure \textbf{base model performance} and assess how accuracy varies across frequency bands. Second, we validate \textbf{circuit quality} by measuring faithfulness (KL divergence), sufficiency (circuit accuracy), necessity (ablation accuracy), and minimality (edge fraction), and compare each circuit against 100 random baselines. Third, we perform a \textbf{structural comparison} by measuring edge overlap between all pairs of circuits (Jaccard similarity: shared edges divided by total unique edges) and testing whether circuits from the same frequency band overlap more than circuits from different bands, with bootstrap confidence intervals to assess reliability. Fourth, we measure \textbf{cross-band transfer} by evaluating each circuit on all other bands' test data, yielding a $5 \times 5$ transfer matrix per model and draw, and test for directional asymmetry between low- and high-frequency circuits.

\subsection{Base Model Performance}
\label{sec:base_performance}

Before examining circuits, we measure the base model's sensitivity to
token frequency (full per-model breakdown in \autoref{app:base_performance_detail}).
Pythia-70m's top-1 accuracy rises with token frequency, from
30.1\% on the low band to 67.0\% on very\_high.
All larger models achieve at least 92.9\% across bands. Their
top-5 accuracy is close to the ceiling (${\geq}99\%$).
The frequency effect on base accuracy is statistically significant only for
Pythia-70m (Kruskal-Wallis $H = 13.5$, $\eta^2 = 0.95$,
$p_{\text{BH}} = 0.036$).
The task itself is therefore largely frequency-invariant for models with
sufficient capacity, so any circuit-level frequency effects in the larger
models cannot be attributed to differences in task difficulty.

\subsection{Circuit Size and Sparsity}
\label{sec:circuit_size}

Circuit sparsity increases with model scale: ACDC retains 30.9\% of edges
for Pythia-70m, 12.3\% for Pythia-160m, 4.4\% for Pythia-410m, 9.1\%
for Pythia-1b, and 2.8\% for Pythia-1.4b (Pythia-1b has fewer layers and heads than Pythia-410m, so its computational graph is much smaller, making the edge fraction higher despite retaining fewer edges in absolute terms; Table~\ref{tab:model_and_threshold}).
For models ${\geq}160$M, low-frequency circuits are consistently larger than
high-frequency circuits. For example, in Pythia-410m, low-frequency circuits retain 4.9\% of edges versus 4.1\% for very\_high, and in Pythia-1.4b, 3.2\% versus 2.5\%
(per-model size and minimality tables in \autoref{app:circuit_size_detail}).
Pythia-70m shows the opposite trend (low 29.4\% vs.\ very\_high 31.7\%), likely reflecting its low base accuracy rather than a frequency-specific effect.
Although low-frequency circuits are consistently larger across the four larger models, the absolute differences are small. As shown in Section~\ref{sec:phantom}, these structural differences do not correspond to functional differences.

\subsection{Circuit Quality Validation}
\label{sec:faithfulness}

Before comparing circuit structure and function, we verify that the 75 extracted circuits retain substantial task performance, outperform random subgraphs with the same number of edges, and are necessary for the task. When evaluated at edge level on the band used for extraction, each circuit for models ${\geq}160$M recovers at least 80\% of the base model's accuracy (two of fifteen Pythia-70m draws fall to 73--76\%; \autoref{app:faithfulness_detail}), and KL~divergence remains low
($0.21$--$0.71$ across all 75~circuits; Table~\ref{tab:model_and_threshold} reports per-model values for the control band).
Every circuit outperforms all 100 random baselines.
Ablation accuracy is 0\% for 69 of 75 circuits and at most 0.44\% (1/225) for the
remaining six. Therefore, removing the circuit eliminates task performance, confirming that
the discovered edges are necessary (per-circuit necessity values in \autoref{app:completeness_detail}).
These results indicate that the circuits are meaningful subgraphs rather
than artifacts of the extraction procedure, and that subsequent comparisons
across frequency bands start from a common baseline of circuit quality.

\subsection{Circuits Show Structural Differences}
\label{sec:structural_differences}

Having confirmed that all circuits are individually faithful, we now ask whether circuits extracted from different frequency bands differ structurally. We measure similarity using the Jaccard index~\mbox{\citep{jaccard1901etude}} over edge sets, $J(C_i, C_j) = |C_i \cap C_j|\,/\,|C_i \cup C_j|$, and compare two types of circuit pairs: \emph{within-band} (same frequency band, different draws) and \emph{between-band} (different frequency bands). Because Jaccard is symmetric and does not capture subset relationships, we complement it with directed containment analysis (\autoref{app:band_affinity_detail}), which confirms that low-frequency circuits contain more high-frequency edges.

Mean within-band Jaccard similarity exceeds mean between-band similarity in all five models (Table~\ref{tab:structural_jaccard}). Bootstrap 95\% confidence intervals~\mbox{\citep{efron1987better}}, based on 10{,}000 resamples, exclude zero for every model, so each gap is statistically distinguishable from zero. However, these differences are small in absolute terms. Among models ${\geq}160$M, the gaps range from 0.013 (95\% CI $0.003, 0.023$) to 0.032 ($0.024, 0.041$). Overall Jaccard similarities range from 0.37 to 0.79, and within-band pairs share only 8--55 more edges than between-band pairs, while individual circuits contain 400--3{,}600 edges. Pythia-70m also has a gap of 0.032 (95\% CI $0.023, 0.042$). Given its lower base accuracy, structural variation may have greater functional consequences in this model (Section~\ref{sec:generic_boost}).

The between-band gaps are only 0.51--1.56 times the standard deviation of Jaccard similarity across draws. This is comparable to the variation observed when ACDC is run twice on the same data. This indicates that structural differences between bands should be interpreted with caution. A formal power analysis (\autoref{app:power_analysis}) confirms that 2--5~draws per frequency band are needed to reliably detect gaps of this magnitude; our 3-draw design achieves ${\geq}0.89$ power for four of five models. Attention edges carry the largest share of the structural gap (75--93\% of band-specific edges are attentional; \autoref{app:component_jaccard_detail}).

We next group edges by the number of conditions in which they appear. Edges present in all five conditions form the \emph{universal core}; edges appearing in only one condition are \emph{band-specific}. The universal fraction (computed over the union of all draws) decreases from 65.5\% in Pythia-70m to 15.4\% in Pythia-1.4b, meaning that larger models appear to have more band-specific circuitry (Figure~\ref{fig:sharing_spectrum} visualizes the full edge sharing spectrum across all five tiers). Taken together, these results suggest that ACDC recovers structurally different circuits for different frequency bands, apparent evidence of frequency-dependent specialization.

\begin{table}[t]
\centering\caption{Edge overlap within a band versus between bands, per
model. $J_{\text{within}}$ compares two independent extractions from the same band, whereas $J_{\text{between}}$ compares extractions from different bands. A claim of structure specific to a frequency band depends on the difference between them. That gap is small, 0.013 to 0.032, but its confidence interval
excludes zero in every model, so band identity does change which edges
get selected.}\label{tab:structural_jaccard}\small
\begin{tabular}{lccc}\toprule
Model & $J_{\text{within}}$\;[95\% CI] & $J_{\text{between}}$\;[95\% CI] & Gap\;[95\% CI] \\\midrule
Pythia-70m  & 0.795\;[.787,\,.803] & 0.763\;[.756,\,.769] & 0.032\;[.023,\,.042] \\
Pythia-160m & 0.589\;[.582,\,.596] & 0.557\;[.552,\,.562] & 0.032\;[.024,\,.041] \\
Pythia-410m & 0.446\;[.442,\,.452] & 0.430\;[.427,\,.434] & 0.016\;[.010,\,.022] \\
Pythia-1b   & 0.478\;[.470,\,.487] & 0.465\;[.460,\,.470] & 0.013\;[.003,\,.023] \\
Pythia-1.4b & 0.385\;[.375,\,.395] & 0.366\;[.361,\,.371] & 0.019\;[.008,\,.030] \\\bottomrule
\end{tabular}\end{table}

\subsection{Asymmetric Transfer}
\label{sec:asymmetric_transfer}

Cross-band transfer is high overall, with small generalization gaps (same-band accuracy minus mean cross-band accuracy) that are often negative, meaning circuits frequently perform better on other bands than on their own, particularly when transferring from harder low-frequency to easier high-frequency bands (Detailed generalization gaps in \autoref{app:generalization_detail}). However, the transfer is asymmetric, meaning  low-frequency circuits generalize more effectively to high-frequency inputs than the opposite.
Transfer asymmetry ranges from 4.7 percentage points for Pythia-410m to 20.4 points for Pythia-1.4b. In Pythia-1.4b, low-frequency circuits achieve 92.1\% accuracy on high-frequency inputs, whereas high-frequency circuits achieve 71.7\% on low-frequency inputs. Results for all models are reported in Table~\ref{tab:asymmetry_summary}, with additional details in \autoref{app:asymmetry_detail}.
\begin{table}[t]
\centering
\caption{Transfer between the pooled low-frequency (low, medium) and high-frequency (high, very\_high) band groups, per model. LF$\to$HF is the accuracy of a low-frequency circuit tested on high-frequency inputs, whereas HF$\to$LF is the accuracy of a high-frequency circuit tested on low-frequency inputs. Asymmetry is the difference between these values, and Ratio is their quotient. Transfer is better from low to high in all five models, by 4.7
to 20.4 percentage points, consistent with low-frequency circuits
containing most of the edges the high-frequency circuits use.}
\label{tab:asymmetry_summary}
\small
\begin{tabular}{lcccc}
\toprule
Model & LF$\to$HF (\%) & HF$\to$LF (\%) & Asymmetry (pp) & Ratio \\
\midrule
Pythia-70m  & 43.1 & 29.3 & 13.8 & 1.47 \\
Pythia-160m & 91.8 & 86.1 & 5.7 & 1.07 \\
Pythia-410m & 96.5 & 91.7 & 4.7 & 1.05 \\
Pythia-1b   & 92.5 & 83.4 & 9.1 & 1.11 \\
Pythia-1.4b & 92.1 & 71.7 & 20.4 & 1.28 \\
\bottomrule
\end{tabular}
\end{table}
This pattern is consistent across all models above 70m and follows from the circuit sizes reported in Section~\ref{sec:circuit_size}: larger (low-frequency) circuits contain a superset of edges used by smaller (high-frequency) circuits, so they transfer well to easier inputs.
The structural asymmetry, where low-frequency circuits contain high-frequency edges more than the reverse, is statistically significant for Pythia-160m, 410m, and 1b, and the same pattern holds for Pythia-1.4b (\autoref{app:transfer_detail} analyzes predictors of this asymmetry).
The asymmetry appears consistent with specialization: low-frequency circuits include more of the computation needed for high-frequency inputs, but not vice versa.
In the very\_low extension without length matching, the same direction appears at the
three largest scales, where very\_low circuits retain more accuracy on
core-band tests than core-band circuits retain on very\_low, but not at 70m or
160m (\autoref{app:very_low}).
The LSC transfer asymmetry does not generalize to SVA. HF$\to$LF transfer is higher than LF$\to$HF transfer in four of five models, and low-frequency SVA circuits are not systematically larger (\autoref{sec:sva_replication}). Since the explanation above turns on
that size difference, the asymmetry looks like a consequence of how
large LSC's low-frequency circuits are rather than a general property
of circuits extracted from frequency bands.

Taken together, Sections~\ref{sec:base_performance}--\ref{sec:asymmetric_transfer}
present what appears to be strong evidence for frequency-dependent circuit
specialization (\finding{1}): circuits differ structurally, the differences are
systematic (low-frequency circuits are larger and more inclusive), and
these differences produce asymmetric transfer.
We next test whether this structural divergence reflects genuine functional specialization.

\section{Phantom Specialization}
\label{sec:phantom}

\label{sec:experiments_stage2}

We now test whether the apparent specialization described in Section~\ref{sec:results} reflects genuine frequency-dependent mechanisms or instead arises from the extraction process. The section builds one argument across its subsections: we first show \emph{functionally} that band-specific edges provide generic rather than band-specific computation (\autoref{sec:generic_edges}); we then identify the methodological factors that have made this pattern easy to miss (\autoref{sec:confounds_section}); and we complete the picture on the remaining axis with \emph{representational} similarity analysis across five metrics (\autoref{sec:confirmation}) and \emph{causal} interchange interventions that directly test whether band representations are interchangeable (\autoref{sec:causal_confirmation}). As a cross-method check, we repeat the core analysis using EAP and EAP-IG circuits, which select largely different edges (Jaccard 0.28--0.60 with ACDC), providing a method-independent control (developed in \autoref{sec:generic_edges}, ``Three additional controls''). The section closes with a positive control (\autoref{sec:two_route_control}), a replication on a second task (\autoref{sec:sva_replication}), and a synthesis of the evidence (\autoref{sec:convergence}).

\begin{figure}[t]
\centering
\begin{tikzpicture}[
  scale=0.80, transform shape,
  vertex/.style={
    rounded corners=4pt, draw=gray, thick,
    minimum width=4.2cm,
    text width=3.8cm, align=center, font=\small,
    inner sep=6pt
  },
]
  \node[vertex, fill=orange!12] (func) at (0, 5.2) {
    \textbf{Functional}\\[3pt]
    {\scriptsize Faithfulness \& completeness\\[1pt]
    Cross-condition transfer\\[1pt]
    Generalization \& asymmetry}
  };
  \node[vertex, fill=teal!12] (struct) at (-3.6, 0) {
    \textbf{Structural}\\[3pt]
    {\scriptsize Graph topology\\[1pt]
    Edge overlap (Jaccard)\\[1pt]
    Universal vs.\ specific edges\\[1pt]
    Circuit size by band}
  };
  \node[vertex, fill=blue!10] (repr) at (3.6, 0) {
    \textbf{Representational}\\[3pt]
    {\scriptsize Embedding, residual stream,\\[1pt]
    logit lens, attention, MLP\\[1pt]
    Information-theoretic\\[1pt]
    Base vs.\ circuit comparison}
  };

  \node[anchor=south west, font=\scriptsize] at (-6.8, 5.8) {
    \textcolor{teal!70}{\rule{0.8cm}{2pt}}
    {\;\small correlational}
  };
  \node[anchor=south west, font=\scriptsize] at (-6.8, 5.3) {
    \textcolor{red!50}{\rule[1pt]{0.4cm}{0pt}%
      \tikz[baseline=-0.5ex]{\draw[red!50, dashed, thick]
        (0,0) -- (0.5,0);}%
    }
    {\;\small causal}
  };

  \draw[thick, teal!70] ([yshift=2pt]struct.east) -- ([yshift=2pt]repr.west);
  \draw[thick, red!50, dashed]
    ([yshift=-2pt]struct.east) -- ([yshift=-2pt]repr.west);
  \node[font=\tiny\bfseries, text=gray!70, below=6pt]
    at ($(struct.east)!0.5!(repr.west)$) {S-R};

  \draw[thick, teal!70, transform canvas={xshift=-1.5pt}]
    (struct.55) -- (func.235);
  \draw[thick, red!50, dashed, transform canvas={xshift=1.5pt}]
    (struct.55) -- (func.235);
  \node[font=\tiny\bfseries, text=gray!70, left=4pt]
    at ($(struct.55)!0.5!(func.235)$) {S-F};

  \draw[thick, teal!70, transform canvas={xshift=1.5pt}]
    (repr.125) -- (func.305);
  \draw[thick, red!50, dashed, transform canvas={xshift=-1.5pt}]
    (repr.125) -- (func.305);
  \node[font=\tiny\bfseries, text=gray!70, right=4pt]
    at ($(repr.125)!0.5!(func.305)$) {F-R};
\end{tikzpicture}
\caption{The similarity triangle compares circuits along three axes, each probed through correlational analyses of metric pairs and causal interventions.
\textbf{S--F:} do structurally different circuits produce different outputs? Tested via cross-band transfer (causal) and Spearman correlations between structural and functional metrics (correlational; 47/80 pairs exhibit Simpson's paradox).
\textbf{S--R:} does structural divergence predict representational divergence? Tested via interchange patching and Boundless DAS (causal) and correlations between pairs of metrics (correlational; 66/80 reversals).
\textbf{R--F:} do functional and representational metrics agree? Tested via the same causal interventions and 64 correlations between pairs of metrics (40/64 reversals).
Phantom specialization corresponds to low structural similarity alongside high functional and representational similarity. Genuine specialization would instead produce low similarity on all three axes.}
\label{fig:similarity_triangle}
\end{figure}
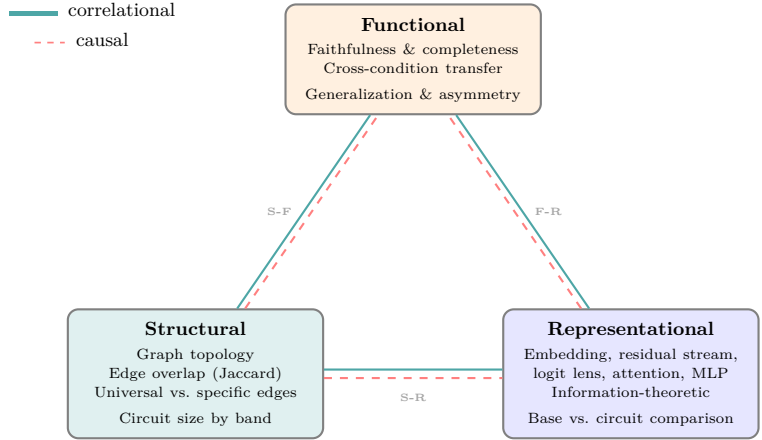

These analyses span three independent axes (Figure~\ref{fig:similarity_triangle}): \textbf{structural} (edge overlap via Jaccard indices; Section~\ref{sec:structural_differences}), \textbf{functional} (output equivalence via cross-band transfer; Sections~\ref{sec:faithfulness}--\ref{sec:asymmetric_transfer}), and \textbf{representational} (activations across layers). Each axis provides independent evidence for or against specialization. Genuine specialization would require low cross-band transfer, band-specific necessity under targeted ablation, \emph{and} robust representational divergence tied to distinct causal pathways. We diagnose phantom specialization when structurally different circuits transfer across bands, lack band-specific necessity, and use interchangeable representations. No single axis is conclusive, but consistent results across all three strengthen the diagnosis.
\label{sec:triangle}

\subsection{Band-Specific Edges Are Functionally Generic}
\label{sec:generic_edges}

Section~\ref{sec:results} showed that circuits extracted from different frequency bands differ structurally. We test whether this structural variation corresponds to functional differences. Band-specific edges improve performance across all frequency bands, and the core shared across most bands recovers nearly all circuit performance (\finding{2}). Independent extractions from the same band can also yield structurally different circuits that implement the same computation (\finding{3}). Together, these results show functional equivalence despite structural variation.

\subsubsection{Band-Specific Edges Boost All Bands Alike}
\label{sec:edge_sharing}

\begin{figure}[t]
\centering
\includegraphics[width=\linewidth]{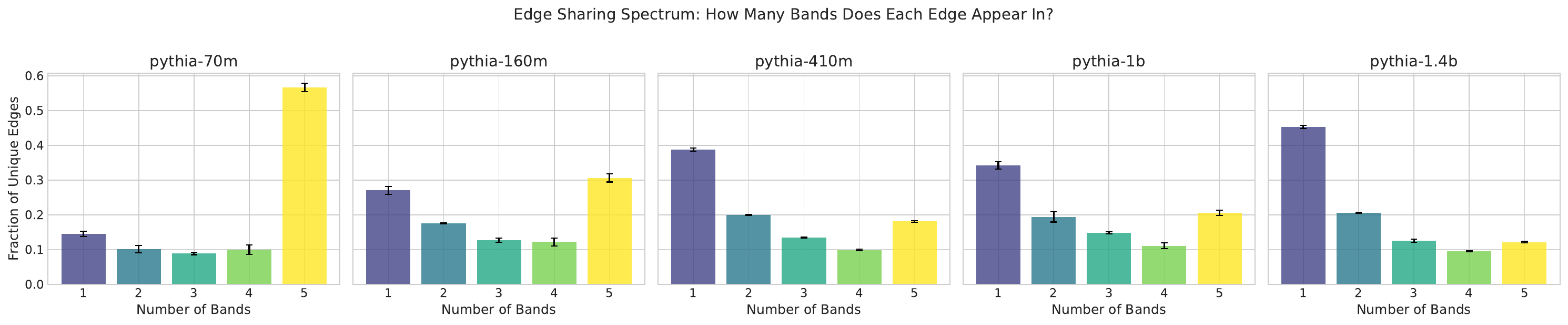}
\caption{Edge sharing spectrum by model.
Each bar shows the fraction of edges appearing in exactly $k$ of five
conditions.
Smaller models concentrate edges in the universal tier ($k{=}5$). Larger models distribute more edges across tiers shared by only some conditions.
Band-specific edges ($k{=}1$) are unstable across draws and do not recover task performance in isolation.}
\label{fig:sharing_spectrum}
\end{figure}

To test whether structural differences reflect genuine specialization, we
group edges by the number of conditions in which they are selected. For an
edge $e$ and the set of conditions~$\mathcal{B}$, let
$\kappa(e) = |\{b \in \mathcal{B} : e \in C_b\}|$
denote the number of conditions whose circuits contain~$e$.
Edges with $\kappa(e) = |\mathcal{B}|$ form the \emph{universal core}, whereas
edges with $\kappa(e) = 1$ are \emph{band-specific}.
Figure~\ref{fig:sharing_spectrum} shows the full sharing spectrum.
The universal core forms a single connected subgraph covering
48--91\% of circuit nodes and this connectivity matters because a disconnected
core would suggest structurally unrelated computations, whereas a
connected core implies a coherent shared mechanism. Edges in the universal core are highly stable across independent ACDC extractions. Across models, 57--82\% appear in all three extractions.
Band-specific edges, by contrast, are overwhelmingly attentional
(75--93\%), and only 1--2\% are stable across draws.\footnote{Per-component stability table in \autoref{app:draw_stability_detail}.}
This instability suggests that band-specific edges largely reflect
extraction noise rather than meaningful specialization.
The canonical copy mechanism, where previous token heads (which attend to the immediately preceding token) compose with induction heads (which match repeated prefixes and copy the following token)~\mbox{\citep{olsson2022context}}, is
85--100\% universal, while band-specific edges are disproportionately
BOS-sink heads and heads with diffuse attention whose contribution is substantial
but band-agnostic (\autoref{app:targeted_profiling}; BOS-sink ablation in \autoref{app:bos_sink_ablation}).

The universal core alone recovers 13.7--71.4\% of full circuit accuracy
(decreasing monotonically with model scale; see Table~\ref{tab:scaling_synthesis}), while band-specific edges in isolation
achieve 0\% accuracy, showing that the core is both partially
sufficient and strictly necessary (sufficiency evidence in \autoref{app:targeted_sufficiency}, ablation of the circuit complement in \autoref{app:targeted_necessity}).
The remaining gap is closed not by band-specific edges but by
edges shared by most bands (Section~\ref{sec:threshold_relaxation}).
Majority-vote aggregation across draws increases accuracy by ${\sim}5$ percentage points
without changing transfer efficiency (${\leq}0.012$ difference).This suggests that variation across draws primarily reflects extraction noise rather than stable differences between frequency bands.

\label{sec:generic_boost}
If band-specific edges encoded computations specialized by frequency, they
should preferentially benefit the band on which they were discovered.
We test this by adding each band's specific edges ($\kappa = 1$) to the
universal core and evaluating the resulting circuit on all five conditions.
Let $C_{\text{core}}$ denote the universal core and $C_b^{*}$ the
band-specific edges for condition~$b$.
The \emph{same-band boost} is
$\Delta_b = \mathrm{Acc}(C_{\text{core}} \cup C_b^{*},\, D_b) -
  \mathrm{Acc}(C_{\text{core}},\, D_b)$,
and the \emph{cross-band boost} to condition~$b'$ is
$\Delta_{b \to b'} = \mathrm{Acc}(C_{\text{core}} \cup C_b^{*},\, D_{b'}) -
  \mathrm{Acc}(C_{\text{core}},\, D_{b'})$.
We define \emph{transfer efficiency} as the ratio of mean cross-band boost
to same-band boost:
\begin{equation}
\mathrm{TE}(b) \;=\;
  \frac{\frac{1}{|\mathcal{B}|-1}
    \sum_{b' \neq b} \Delta_{b \to b'}}{\Delta_b}\,.
\label{eq:transfer_efficiency}
\end{equation}
$\mathrm{TE} = 1$ indicates that band-specific edges provide a purely
generic benefit and $\mathrm{TE} \ll 1$ would indicate genuine specialization.
The observed boost is nearly uniform across test bands
(Table~\ref{tab:transfer_efficiency}; Figure~\ref{fig:boost_heatmaps} in \autoref{app:targeted_transfer}):
transfer efficiency ranges from 81.4\% (Pythia-70m; 95\% CI [74, 91]) to 97.0\%
(Pythia-1.4b; [94, 99]).
In plain terms, edges labeled as ``specific to low-frequency tokens'' boost performance on high-frequency inputs nearly as much as on low-frequency inputs. If these edges truly encoded band-specific computation, one would expect cross-band benefit to be substantially smaller than same-band benefit, but this is not the case.
The same-band advantage is statistically significant (all $p \leq 0.028$) but
small in effect size~\mbox{\citep{cohen2013statistical}} (Cohen's $d = 0.16$--$0.51$), and random edges of
the same count contribute only 1--3\% boost.
The absolute same-band gap is approximately constant at 0.016--0.029
points across scales, even as the generic boost grows from
0.12 to 0.76, indicating a fixed structural offset rather than growing
specialization.
Under zero ablation this residual disappears: the gap between same-band and cross-band boosts is at most 0.012 points in every model, and Cohen's $d$ falls to $|d| \leq 0.18$, reversing sign in Pythia-410m and Pythia-1b (\autoref{app:zero_ablation}, Table~\ref{tab:zero_ablation}). The small same-band advantage therefore looks like a product of the resampling ablation rather than genuine band-specific computation.
Zero ablation is the stricter intervention and costs absolute accuracy,
with models ${\geq}$410M falling to near chance, so its comparison runs
on a compressed measurement scale (\autoref{app:zero_ablation}).
This zero ablation diagnostic is critical: it argues against the possibility that the small same-band advantage reflects a real but weak specialization signal, attributing it instead to a known property of the corruption procedure.
We also run mean ablation, replacing each ablated activation with its
mean at each token position over the corrupt input distribution. This
intervention is gentler as it keeps the models working (circuit accuracy on
the extraction band $0.14$--$0.68$), yet the same-band advantage is again not
significant in any model (all $p \geq 0.093$, against all $p \leq 0.028$
under resampling), and transfer efficiency stays at $66$--$96\%$
(Table~\ref{tab:zero_ablation}). One caveat is that the universal core itself
collapses under mean ablation for models ${\geq}$410M, so the boosts at
those scales are measured against a baseline close to zero
(\autoref{app:zero_ablation}).
The intervention sensitivity documented here is one of the two
evaluation artifacts behind the apparent specialization; the second,
source-level inflation, is quantified in \autoref{sec:inflation}.
The between-band structural differences are nonetheless statistically
reliable (bootstrap confidence intervals exclude zero;
Section~\ref{sec:structural_differences}), indicating that ACDC's
pruning is genuinely sensitive to input distribution differences, but
these structurally real differences are functionally inert (zero
standalone accuracy, ${\leq}2\%$ stable across draws).

\begin{table}[t]
\centering
\caption{Cross-band transfer under resample, zero, and mean ablation.
Mean ablation uses the mean at each token position over the corrupt input
distribution. Resample rows repeat the point estimates of Table~\ref{tab:transfer_efficiency} as the baseline for each model. Confidence intervals, the baseline from random edges, and $p$-values are reported there.
$^{\dagger}$Transfer efficiency undefined when same-band boost is
negative or below $0.01$ accuracy points.}
\label{tab:zero_ablation}
\small
\begin{tabular}{llcccc}
\toprule
Model & Ablation & Same Boost & Cross Boost & Transfer Eff.\ & Cohen's $d$ \\
\midrule
Pythia-70m  & resample & $+$0.122 & $+$0.099 & 81.4\% & 0.51 \\
            & zero     & $-$0.115 & $-$0.127 & ---$^{\dagger}$ & 0.18 \\
            & mean     & $+$0.115 & $+$0.099 & 86.6\% & 0.28 \\
\addlinespace
Pythia-160m & resample & $+$0.304 & $+$0.282 & 92.6\% & 0.29 \\
            & zero     & $+$0.127 & $+$0.127 & 99.8\% & 0.00 \\
            & mean     & $+$0.401 & $+$0.386 & 96.3\% & 0.10 \\
\addlinespace
Pythia-410m & resample & $+$0.450 & $+$0.434 & 96.4\% & 0.16 \\
            & zero     & $+$0.001 & $+$0.002 & ---$^{\dagger}$ & $-$0.16 \\
            & mean     & $+$0.361 & $+$0.293 & 81.1\% & 0.39 \\
\addlinespace
Pythia-1b   & resample & $+$0.513 & $+$0.485 & 94.4\% & 0.24 \\
            & zero     & $+$0.001 & $+$0.001 & ---$^{\dagger}$ & $-$0.14 \\
            & mean     & $+$0.427 & $+$0.358 & 83.8\% & 0.56 \\
\addlinespace
Pythia-1.4b & resample & $+$0.756 & $+$0.733 & 97.0\% & 0.26 \\
            & zero     & $+$0.004 & $+$0.003 & ---$^{\dagger}$ & 0.15 \\
            & mean     & $+$0.141 & $+$0.093 & 65.7\% & 0.56 \\
\bottomrule
\end{tabular}
\end{table}

Three additional controls support this conclusion.
First, for models ${\geq}160$M, circuit pairs agree on whether an example is answered correctly in 78--93\% of cases. Cross-band pairs agree 1--3 percentage points more often than same-band pairs (\autoref{app:failure_analysis}). Thus, aggregate transfer does not appear to mask systematic differences on individual examples. Second, an independent EAP-IG analysis~\mbox{\citep{hanna2024have}}\label{para:cross_method_check} finds no consistent same-band advantage across the same 75 extraction conditions. Across all five models and ten circuit sizes, the largest same-band advantage is 4.6 percentage points, and its direction varies across settings. This pattern holds when EAP-IG retains between 1\% and 100\% of the model's edges. Moreover, EAP-IG and ACDC select substantially different edge sets, with Jaccard overlap ranging from 0.28 to 0.60. To determine whether the result is caused by low EAP-IG faithfulness, we repeat the comparison using the smallest EAP-IG circuit that achieves at least 50\% faithfulness. These circuits contain $1.5\times$--$5\times$ as many edges as the corresponding ACDC circuits, yet their same-band advantage remains only 0.8--4.6 percentage points. Thus, the absence of a substantial same-band advantage cannot be attributed to evaluating unfaithful EAP-IG circuits (\autoref{app:method_comparison}). Third, adjacent ACDC thresholds ($2$--$3{\times}$ range in circuit size) yield at most 5.5 percentage points change in transfer efficiency, confirming that our conclusions are robust to the threshold selection. This is the defining signature of \emph{phantom specialization}: the structural differences documented in
Section~\ref{sec:structural_differences} do not translate into band-specific functional advantages. The same split between structure and function appears on the second task, subject-verb agreement, where the replication is reported in
\autoref{sec:sva_replication}.

\begin{table}[t]
\centering
\caption{Cross-band transfer of band-specific edges. Transfer efficiency is the ratio of the cross-band to the same-band accuracy boost. The column $d$ reports Cohen's $d$ for their difference, and $p$ reports the one-sided paired Wilcoxon signed-rank test.
95\% bootstrap percentile CIs ($N{=}10{,}000$) in brackets.}
\label{tab:transfer_efficiency}
\small
\begin{tabular}{lcccccc}
\toprule
Model & Same Boost & Cross Boost & Random & Transf.\ Eff.\ [95\% CI] & $d$ & $p$ \\
\midrule
Pythia-70m  & 0.122 $\pm$ 0.055 & 0.099 $\pm$ 0.035 & 0.014 & 0.814\;[.74,\,.91] & 0.49 & 0.004 \\
Pythia-160m & 0.304 $\pm$ 0.059 & 0.282 $\pm$ 0.048 & 0.028 & 0.926\;[.89,\,.96] & 0.42 & ${<}0.001$ \\
Pythia-410m & 0.450 $\pm$ 0.080 & 0.434 $\pm$ 0.064 & 0.024 & 0.964\;[.95,\,.98] & 0.22 & 0.004 \\
Pythia-1b   & 0.513 $\pm$ 0.091 & 0.485 $\pm$ 0.064 & 0.027 & 0.944\;[.92,\,.97] & 0.36 & 0.001 \\
Pythia-1.4b & 0.756 $\pm$ 0.027 & 0.733 $\pm$ 0.037 & 0.010 & 0.970\;[.94,\,.99] & 0.71 & 0.028 \\
\bottomrule
\end{tabular}
\end{table}

\begin{table}[t]
\centering
\caption{Cross-method comparison of same-band and cross-band accuracy at representative circuit sizes. Same-band accuracy is averaged over all five conditions and three draws in which the circuit is evaluated on its extraction band. Cross-band accuracy is averaged over evaluations on the other frequency bands. $\Delta$ is same-band minus cross-band accuracy, with positive values indicating specialization. Jaccard is the edge overlap with the corresponding ACDC circuit. EAP-IG circuits require $1.5$--$5\times$
more edges than ACDC to achieve comparable faithfulness (full
method comparison analysis in \autoref{app:method_comparison}).}
\label{tab:method_comparison}
\small
\begin{tabular}{llccccc}
\toprule
Method & Model & Size & Same Acc.\ & Cross Acc.\ & $\Delta$ & Jaccard \\
\midrule
ACDC    & 70m  & $1.0\times$ & 0.42 & 0.39 & $+$0.02 & --- \\
ACDC    & 160m & $1.0\times$ & 0.92 & 0.90 & $+$0.02 & --- \\
ACDC    & 410m & $1.0\times$ & 0.96 & 0.95 & $+$0.02 & --- \\
ACDC    & 1b   & $1.0\times$ & 0.93 & 0.90 & $+$0.03 & --- \\
ACDC    & 1.4b & $1.0\times$ & 0.88 & 0.86 & $+$0.02 & --- \\
\midrule
\multicolumn{7}{l}{\textit{EAP-IG at ACDC circuit size}} \\
EAP-IG  & 70m  & $1.0\times$ & 0.24 & 0.23 & $+$0.01 & 0.60 \\
EAP-IG  & 160m & $1.0\times$ & 0.15 & 0.14 & $+$0.01 & 0.42 \\
EAP-IG  & 410m & $1.0\times$ & 0.09 & 0.10 & $-$0.01 & 0.32 \\
EAP-IG  & 1b   & $1.0\times$ & 0.06 & 0.05 & $+$0.01 & 0.36 \\
EAP-IG  & 1.4b & $1.0\times$ & 0.02 & 0.05 & $-$0.03 & 0.28 \\
\midrule
\multicolumn{7}{l}{\textit{EAP-IG at faithful size (${\geq}50\%$ base accuracy)}} \\
EAP-IG  & 70m  & $1.5\times$ & 0.36 & 0.35 & $+$0.01 & --- \\
EAP-IG  & 160m & $2.0\times$ & 0.49 & 0.45 & $+$0.05 & --- \\
EAP-IG  & 410m & $3.0\times$ & 0.58 & 0.55 & $+$0.03 & --- \\
EAP-IG  & 1b   & $3.0\times$ & 0.58 & 0.54 & $+$0.04 & --- \\
EAP-IG  & 1.4b & $5.0\times$ & 0.60 & 0.58 & $+$0.02 & --- \\
\bottomrule
\end{tabular}
\end{table}

\subsubsection{Shared core Sufficiency}
\label{sec:threshold_relaxation}

\begin{figure}[t]
\centering
\includegraphics[width=\linewidth]{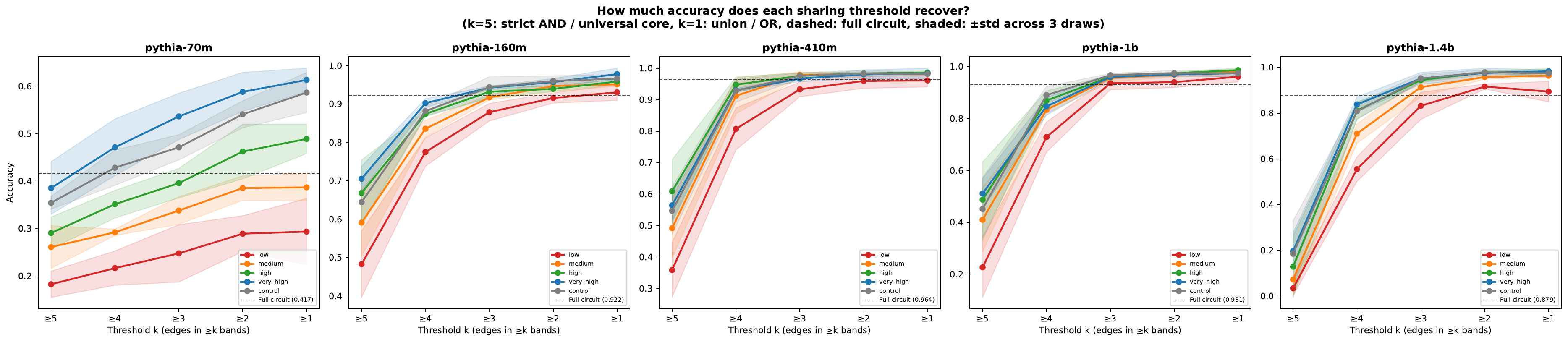}
\caption{Circuit accuracy as a function of sharing threshold~$k$.
Lower values of $k$ include more edges. The strict universal core corresponds to $k{=}5$.
Accuracy reaches at least 95\% of the full circuit at $k{=}3$ for most model$\times$band combinations. For models ${\geq}160$M, pooled gap closure at $k{\geq}3$ is at least 99\% (see text).}
\label{fig:threshold_curve}
\end{figure}

Relaxing the sharing requirement from strict universality ($k{=}5$, all conditions) to majority sharing ($k{\geq}3$, at least three of five conditions) recovers at least 99\% of the accuracy gap between the universal core and the full circuit for Pythia-160m and above (point estimate; at least 98\% at the 95\% CI lower bound; Figure~\ref{fig:threshold_curve}). We exclude Pythia-70m from this estimate because its low base accuracy (30--67\%) makes the estimated fraction of the gap recovered unstable. Its lower transfer efficiency (81\%) may reflect noisier extraction rather than a different mechanism.
The critical threshold is predominantly $k{=}3$;\footnote{Per-band critical-$k$ in \autoref{app:targeted_threshold}.} truly band-specific edges
($k{=}1$) contribute negligible additional accuracy.
Universal core retention ($k{=}5$) declines monotonically with model
size (71\% $\to$ 67\% $\to$ 53\% $\to$ 45\% $\to$ 14\% from 70M to
1.4B), consistent with increasing pathway redundancy: ACDC selects
different routes across extractions, reducing the strict intersection
while preserving the shared computation.
The core shared by most bands ($k{\geq}3$) closes this gap entirely,
and $k{\geq}3$ edge sets are themselves band-agnostic (pairwise Jaccard
$0.656$ for Pythia-1.4b, nearly double the full circuit between-band
Jaccard of $0.366$).\footnote{$k{\geq}3$ edge set composition in \autoref{app:k3_composition}.}
Pythia-1.4b shows this split between structure and function most
sharply: within-band draw sharing drops to 21--27\% (vs.\ 68--73\%
for Pythia-70m), yet cross-band
transfer matches or exceeds the other models, demonstrating that
maximal structural divergence is fully compatible with functional
interchangeability.

\subsubsection{The Shared Core Implements the Canonical Copy Mechanism}
\label{sec:mechanism}

The universal core contains a disproportionate number of attention heads associated with the canonical copying mechanism described by \mbox{\citet{olsson2022context}}.
Head roles are assigned by thresholded attention statistics computed on
LSC inputs (previous token attention, BOS attention, induction score, and
entropy; definitions and thresholds in \autoref{app:attention_detail};
see also the documented Pythia head roles
of~\mbox{\citealp{tigges2024llm}}).
Previous token attention heads are significantly enriched among universal
heads (odds ratio $3.7$--$14.4$, all $p \leq 0.044$), and heads labeled as induction heads are
enriched in the three largest models (all $p \leq 0.040$).
Notably, in Pythia-1b every head labeled as an induction head is classified as universal (OR~$= \infty$), and in Pythia-410m seven of eight are (OR~$= 10.3$). This pattern indicates that the core copying mechanism is largely shared across frequency bands.
Heads that attend to BOS (sink heads), which do not contribute to the copy
computation, are depleted (OR~$= 0.10$--$0.38$, all $p \leq 0.0017$).
Heads showing the canonical induction pattern, attention from the
prediction position to the answer token~\mbox{\citep{olsson2022context}},
are likewise concentrated in the universal core and classifying by that
criterion shows that heads showing the canonical induction attention pattern are enriched in every model above
70m (odds ratio ${\geq}\,6.3$, all $p \leq 0.0026$; in Pythia-1.4b all 25 such
heads are universal).
At the weight level, heads showing the canonical induction attention pattern show stronger K-composition~\mbox{\citep{elhage2021mathematical}} with earlier previous token heads than do other heads in the same layers (one-sided Mann--Whitney, all $p \leq 0.0083$). This result supports the canonical two-step sequence copying mechanism, in which previous token heads write a pattern that induction heads then read~\mbox{\citep{olsson2022context}}.

\begin{figure}[t]
\centering
\includegraphics[width=\linewidth]{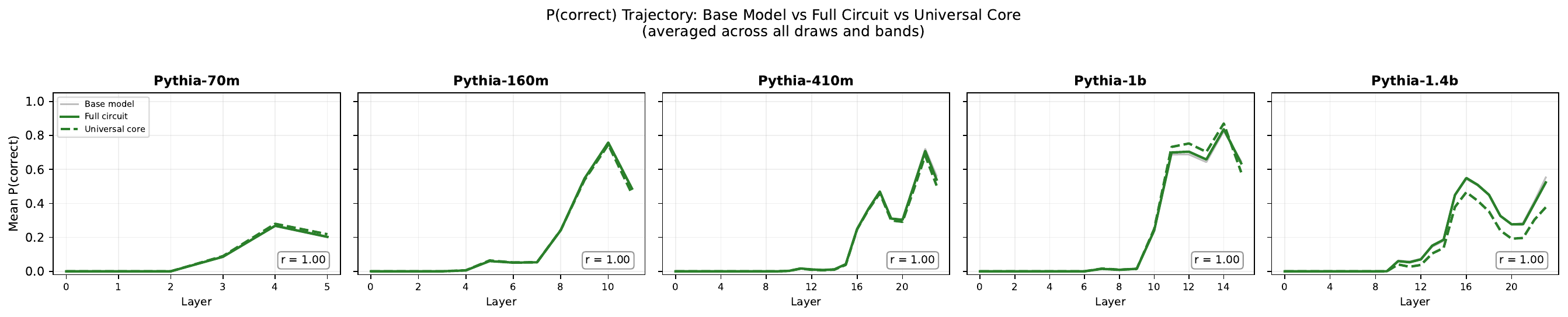}
\caption{Logit lens trajectories for the universal core (dashed), full circuit (solid), and base model (gray).
The core follows the shape and timing of the full circuit's trajectory. Its final probability of the correct token is 0.72--1.07 times that of the full circuit, with the largest reduction in Pythia-1.4b.
Correlation is high across all five models ($r = 0.99$--$1.00$).}
\label{fig:trajectory_overlay}
\end{figure}

The layerwise logit lens~\mbox{\citep{nostalgebraist2020interpreting}} trajectories of the universal core and the full circuit are highly correlated across all five models (Pearson~\mbox{\citep{pearson1895vii}} $r = 0.99$--$1.00$; Figure~\ref{fig:trajectory_overlay}). The universal core does not converge later than the full circuit, and its final probability assigned to the correct token is 72--107\% of the full circuit's value (\autoref{app:targeted_algorithm}). These similarities suggest that the universal core preserves the same computation rather than implementing a weaker or qualitatively different one.

\subsubsection{Same-Band Extractions Vary Structurally but Not Functionally}
\label{sec:cross_draw}

Phantom specialization also appears across independent circuit extractions from the same frequency band (\finding{3}). Circuits extracted from the same band differ substantially across draws. In Pythia-70m, 68--73\% of edges recur in all three extractions. In Pythia-1.4b, only 21--27\% do. Band-specific edges are especially unstable, with 87--93\% appearing in only one extraction. By comparison, 57--82\% of universal core edges appear in all three. These results are summarized in Table~\ref{tab:draw_stability_main}, with the full analysis reported in \autoref{app:targeted_draw}.
Despite this structural variation, circuit performance transfers across draws. When a circuit extracted from one draw is evaluated on test data from another draw of the same band, the ratio of cross-draw to same-draw accuracy ranges from 0.990 to 1.017. The aggregate 95\% CI is $0.990, 1.007$ and includes 1.0, consistent with no difference from same-draw evaluation.
This further demonstrates that structural variation does not imply
functional variation: circuits that share as few as 21\% of their edges
produce equivalent outputs.
The combination of high cross-draw instability and nearly perfect cross-draw transfer shows that many selected edges are not consistently required across functionally equivalent circuits. For models ${\geq}160$M, 52--79\% of edges are absent from at least one other extraction, even though the circuits achieve equivalent performance. This analysis does not show that each of these edges is causally irrelevant within its original circuit, but it does show that a single extraction cannot distinguish stable structure from edges specific to one extraction or substitutable edges. Multiple extractions and majority-vote aggregation are therefore necessary to distinguish stable circuit structure from extraction noise.

\begin{table}[t]
\centering
\caption{Draw stability by edge sharing level (\% of edges appearing in
1, 2, or 3 of 3 independent ACDC draws).
Band-specific edges are unique to one of the five conditions, whereas universal edges are shared across all five conditions.}
\label{tab:draw_stability_main}
\small
\begin{tabular}{l ccc ccc}
\toprule
 & \multicolumn{3}{c}{Band-specific} & \multicolumn{3}{c}{Universal} \\
\cmidrule(lr){2-4}\cmidrule(lr){5-7}
Model & 1 & 2 & 3 & 1 & 2 & 3 \\
\midrule
Pythia-70m  & 87 & 12 & 1 &  8 & 10 & 82 \\
Pythia-160m & 87 & 12 & 1 & 12 & 17 & 71 \\
Pythia-410m & 89 & 10 & 1 & 17 & 22 & 60 \\
Pythia-1b   & 93 &  6 & 1 & 16 & 25 & 59 \\
Pythia-1.4b & 90 & 10 & 1 & 19 & 24 & 57 \\
\bottomrule
\end{tabular}
\end{table}

\subsection{Why Standard Evaluation Misses the Phantom}
\label{sec:confounds_section}

Having seen that band-specific edges are functionally generic, we now identify two methodological factors that have potentially obscured this finding in prior work.

\subsubsection{Source-Level Evaluation Inflates Faithfulness}
\label{sec:evaluation_granularity}
\label{sec:inflation}

\begin{figure}[t]
\centering
\includegraphics[width=\linewidth]{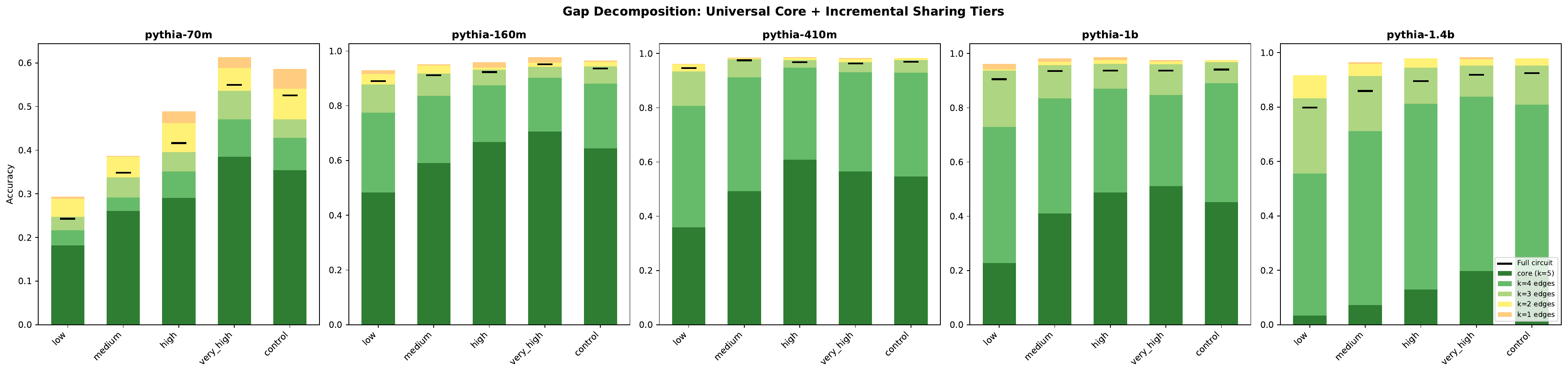}
\caption{Performance gap decomposition.
The gap between the universal core at edge level and the full circuit grows with model scale. It is 0.12 for Pythia-70m, 0.30 for Pythia-160m, 0.45 for Pythia-410m, 0.51 for Pythia-1b, and 0.76 for Pythia-1.4b.
Each bar decomposes this gap by sharing tier~$k$, the number of conditions containing an edge. Band-specific edges have $k{=}1$, and universal edges have $k{=}5$.
Including $k{\geq}3$ edges closes nearly all of the gap;
$k{=}1$ (band-specific) edges contribute negligibly.
Source-level evaluation, shown by the top bar for each model, yields substantially higher accuracy than edge-level evaluation because it preserves unselected outgoing edges.}
\label{fig:gap_decomposition}
\end{figure}

The standard evaluation granularity (source-level) used in most circuit discovery work systematically obscures the phantom by inflating apparent circuit accuracy (\finding{4}; Section~\ref{sec:granularities}). Together with the resampling artifact above (Section~\ref{sec:generic_boost}), this explains why discovered circuits can look both more specialized and more faithful than they are at edge level.
Source-level evaluation preserves all outgoing edges from components that contribute at least one edge to the universal core. Under this evaluation, the accuracy of the universal core is 0.51--0.99, compared with 0.12--0.62 when only the core edges selected by ACDC are preserved at edge level (Figure~\ref{fig:gap_decomposition}) and this inflation ranges from 0.22 (Pythia-70m) to 0.85 (Pythia-1.4b) accuracy
points.
At source level, even an incomplete circuit can appear nearly faithful because the evaluation preserves all outgoing edges from components that contribute at least one selected circuit edge, including edges that ACDC did not select.
This dependence on the granularity, connects to the broader principle that causal structure manifests differently at different scales of analysis~\mbox{\citep{noble2012theory, hoel2013quantifying}}. We develop this connection in Section~\ref{sec:why_phantom}. The implication is that any study comparing circuits across conditions must evaluate at edge level to avoid this confound.

\label{sec:scaling}
Table~\ref{tab:scaling_synthesis} reports the phantom specialization metrics for each model size.
As model size increases, the universal fraction generally decreases from
73.7\% (Pythia-70m) to 27.2\% (Pythia-1.4b), making circuits appear more
specialized. However, transfer efficiency increases from 81.4\% [74, 91]
to 97.0\% [94, 99], indicating that band-specific edges transfer more
effectively across frequency bands in larger models.
At the same time, source-level inflation grows with scale (the gap between
source-level and edge-level accuracy widens from 0.22 to 0.85), so the
methodological artifact that masks the phantom becomes more severe for
larger models.
Cross-draw transfer remains nearly perfect across all scales
(ratio $0.990$--$1.017$; \autoref{app:targeted_scaling}), confirming that circuits are functionally
reproducible even when structurally variable.

\begin{table}[t]
\centering
\caption{Scaling synthesis of phantom specialization metrics.
Univ.~\% is averaged over draws. For each ACDC draw, we compute the fraction of circuit edges present in all five conditions and then average this fraction over the three draws.
The stricter union variant across draws (an edge must appear in every condition
of every draw) is reported in Section~\ref{sec:structural_differences}
(65.5\%~$\to$~15.4\%); see Appendix~\ref{app:universal_edges_detail}.
Univ.~\% decreases with scale, so circuits appear more specialized, while transfer efficiency increases and cross-draw transfer remains close to 1.0.
Crit.~$k$: minimum sharing threshold at which accuracy reaches ${\geq}95\%$ of
the full circuit (mean over bands). 95\% bootstrap CIs in brackets.}
\label{tab:scaling_synthesis}
\small
\begin{tabular}{lccccc}
\toprule
Model & Univ.~\% & Retention & Transf.~Eff.\ [CI] & Draw Transf.\ [CI] & Crit.~$k$ \\
\midrule
Pythia-70m  & 73.7 & 0.714 & 0.814\;[.74,\,.91] & 0.990\;[.94,\,1.05] & 2.8 \\
Pythia-160m & 50.7 & 0.669 & 0.926\;[.89,\,.96] & 0.998\;[.99,\,1.01] & 3.0 \\
Pythia-410m & 36.4 & 0.533 & 0.964\;[.95,\,.98] & 0.999\;[.99,\,1.01] & 3.6 \\
Pythia-1b   & 38.9 & 0.447 & 0.944\;[.92,\,.97] & 1.004\;[.99,\,1.01] & 3.0 \\
Pythia-1.4b & 27.2 & 0.137 & 0.970\;[.94,\,.99] & 1.017\;[1.00,\,1.03] & 3.0 \\
\bottomrule
\end{tabular}
\end{table}

\subsubsection{Model Scale, Not Frequency Band, Drives Metric Variation}
\label{sec:analysis} %
\label{sec:variance_decomposition}

Across structural, functional, and representational metrics, model scale
explains more variance than frequency band. This indicates that the
apparent frequency dependence of circuit structure and function arises
from pooling results across model sizes rather than from variation within
individual models. The unified variance decomposition assigns the largest
share of variance to model scale for every metric examined (full results
in \autoref{app:integration_variance}). For functional metrics alone,
model scale accounts for 92.3\% of the variance, compared with 3.4\% for
frequency band. The band contribution is driven primarily by the increase in
Pythia-70m's accuracy with token frequency (Section~\ref{sec:base_performance}).

Model scale covaries with graph topology, baseline performance, and
pathway redundancy, and five models are insufficient to separate their
effects. However, baseline performance is likely the main driver.
Pythia-1b and Pythia-160m have similarly sized computational graphs
(10K and 11K edges) and show similar functional patterns despite their
different model sizes. By contrast, Pythia-70m differs primarily because
the task is difficult for it, with base accuracy ranging from 30\% to
67\%.

\begin{figure}[t]
\centering
\includegraphics[width=0.55\linewidth]{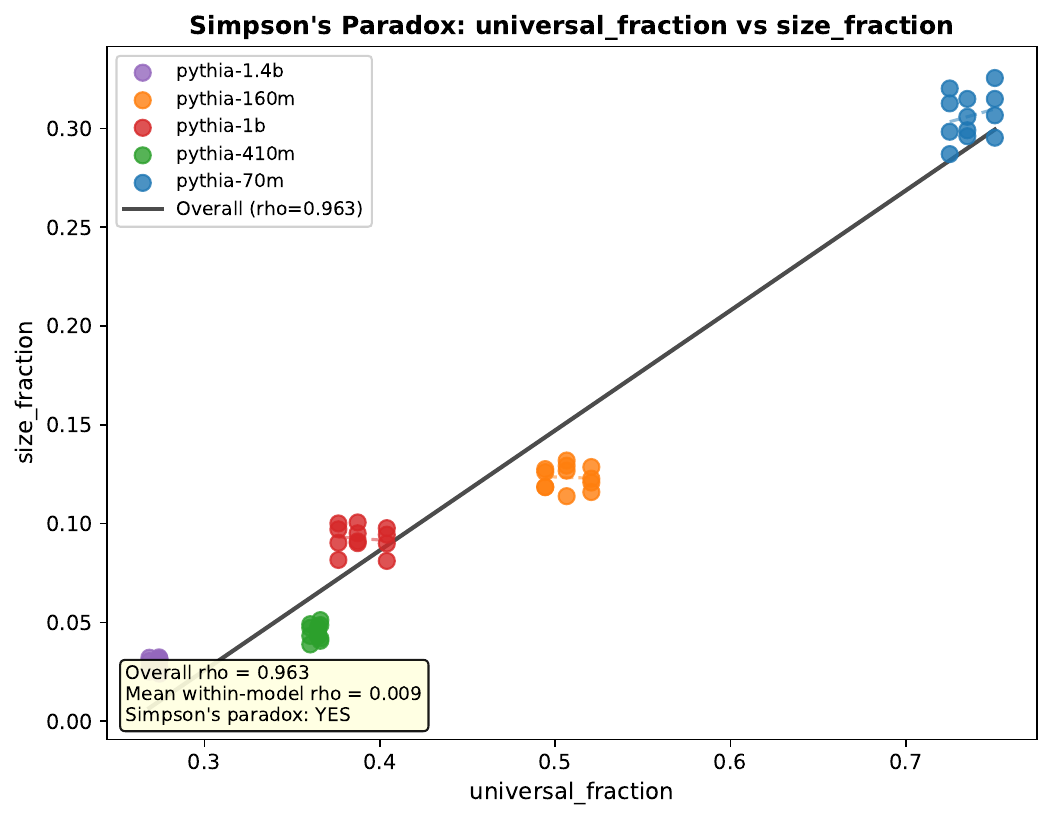}
\caption{Simpson's paradox in correlations across analysis types. Universal edge fraction and circuit size fraction correlate strongly when models are pooled ($\rho = 0.96$, black line), but the mean correlation within models is $\bar{\rho} = 0.01$ (colored lines). Model identity, not
frequency band, drives the apparent association. This pattern recurs in
47/80 structure--function and 66/80 structure--representation metric pairs.}
\label{fig:simpsons_paradox}
\end{figure}

We observe Simpson's paradox~\mbox{\citep{simpson1951interpretation}} in
47 of 80 structure--function metric pairs and 66 of 80
structure--representation metric pairs. In these cases, correlations
that appear strong after pooling across models reverse or disappear
within individual models. Complete results are reported in
\autoref{app:structure_function_detail} and
\autoref{app:integration_similarity_triangle}.
Figure~\ref{fig:simpsons_paradox} illustrates this pattern for universal
edge fraction and circuit size fraction. Their pooled correlation is
$\rho = 0.96$, whereas their mean within-model correlation is
$\rho = 0.01$. Studies comparing circuits across conditions should
therefore account for model identity.

\subsection{Representational and Causal Confirmation}
\label{sec:confirmation}

The preceding analyses traced the phantom along the structural--functional edge of the similarity triangle. We now examine the representational vertex for independent confirmation.

\subsubsection{Representational Similarity}
\label{sec:representational_summary}

We first compare the internal representations produced by each extracted circuit with those produced by the full model. Across five metrics, the extracted circuits preserve the representational geometry of their corresponding base models (Table~\ref{tab:repr_similarity_summary}). This indicates that differences among frequency bands are already present in the base model rather than introduced by circuit extraction. To measure residual stream similarity, we use centered kernel alignment (CKA)~\mbox{\citep{kornblith2019similarity}}. CKA compares two sets of representations computed for the same inputs and is invariant to rotations and isotropic scaling. Values near ~1 indicate that restricting the model to the extracted circuit leaves its residual stream geometry essentially unchanged.

\begin{table}[h]
\small\centering
\caption{Representational similarity between full circuit and base model
across five metrics. Each row independently supports the phantom specialization interpretation. Restricting the model to an extracted circuit preserves the base model's representational geometry.}
\label{tab:repr_similarity_summary}
\begin{tabular}{l l l}
\toprule
Metric & Result (circuit vs.\ base) & Detail \\
\midrule
Embedding geometry                                  & shared $k$NN/probe structure       & \autoref{app:embedding_detail} \\
Residual stream CKA~\mbox{\citep{kornblith2019similarity}} & median ${\geq}\,0.93$ (${\geq}\,0.98$ for ${\leq}1$B) & \autoref{app:residual_detail} \\
Attention entropy~\mbox{\citep{zhai2023stabilizing}}       & base/circuit aligned               & \autoref{app:attention_detail} \\
Band identity MI~\mbox{\citep{shannon1948mathematical}}    & ${<}\,10\%$ reduction        & \autoref{app:info_theoretic_detail} \\
MLP neuron selectivity                              & Spearman $\rho = 0.84$--$0.99$ (${\geq}160$M) & \autoref{app:mlp_detail} \\
\bottomrule
\end{tabular}
\end{table}

\subsubsection{Interchange Interventions}
\label{sec:causal_confirmation}

The representational metrics above are correlational and as a result we test
whether band representations are \emph{causally} interchangeable
using two complementary intervention methods.
Interchange patching~\mbox{\citep{geiger2024finding}} replaces one band's
residual stream activations with another's at the prediction position
and measures whether the model's output follows the source band's
target (Interchange Intervention Accuracy, IIA).
For models ${\geq}160$M, IIA is ${\geq}0.94$ across all 20~cross-band
pairs, with the same-band versus cross-band difference at most 0.004
(Table~\ref{tab:iia_matrix_summary}; Figure~\ref{fig:iia_matrix} in \autoref{app:interchange_patching}).
To verify that interchange patching responds to layer choice, we repeat the intervention at every layer using pairs of examples from the same band with different target tokens. At layers~0--2, the patched activations never cause the output to follow the source target (IIA${}=0$). At the peak layer, IIA rises to $0.92$--$0.98$ for models ${\geq}160$M. Across all five models, peak IIA approximately matches each model's unpatched accuracy on the same examples. Thus, the method distinguishes layers where patching has no effect from those where target information can be transferred (\autoref{app:positive_control}).
This shows that band representations are not merely similar but
causally interchangeable. Swapping one band's internal state for another's
preserves the model's computation with nearly perfect fidelity.
Boundless DAS~\mbox{\citep{wu2023boundless}} further localizes the causal
information distinguishing bands to just 12--32~dimensions
(0.6--6.3\% of $d_{\text{model}}$), and this fraction decreases with
model scale (Table~\ref{tab:bdas_dimensions}; full details are in \autoref{app:causal_interventions}).
Together, these results show that internal representations are largely interchangeable across frequency bands. The information distinguishing the bands is concentrated in a small subspace and does not indicate a separate task mechanism.

\begin{table}[t]
\centering
\caption{IIA at the peak layer for same-band and cross-band pairs.
The difference between same-band and cross-band pairs is at most 0.004 for every model, providing causal evidence that band identity does not determine representational format. Pythia-70m IIA is low overall (0.47), reflecting limited capacity
rather than band sensitivity.}
\label{tab:iia_matrix_summary}
\small
\begin{tabular}{lccccc}
\toprule
 & Pythia-70m & Pythia-160m & Pythia-410m & Pythia-1b & Pythia-1.4b \\
\midrule
Peak layer (of total) & L5/6  & L11/12  & L23/24  & L13/16  & L21/24 \\
Same-band IIA      & 0.466 & 0.962 & 0.984 & 0.986 & 0.968 \\
Cross-band IIA     & 0.466 & 0.962 & 0.984 & 0.985 & 0.972 \\
Difference         & $+$0.000 & $+$0.000 & $+$0.000 & $+$0.001 & $-$0.004 \\
\bottomrule
\end{tabular}
\end{table}

\begin{table}[t]
\centering
\caption{Boundless DAS effective dimensions at peak layer.
The subspace encoding frequency band information decreases from 5.7\% to 0.6\% of $d_{\text{model}}$ with scale. IIA is 1.0 for every model.
Percentages are for the low$\to$high direction; across both directions they
span 0.6--6.3\%.}
\label{tab:bdas_dimensions}
\small
\setlength{\tabcolsep}{5pt}
\begin{tabular}{lcccccc}
\toprule
Model & Peak layer & $d_{\text{model}}$ & Eff.\ dim (low$\to$high) & Eff.\ dim (high$\to$low) & \% of $d_{\text{model}}$ (low$\to$high) & IIA \\
\midrule
Pythia-70m  & L5  &  512 & 29 & 32 & 5.7 & 1.0 \\
Pythia-160m & L11 &  768 & 19 & 20 & 2.5 & 1.0 \\
Pythia-410m & L23 & 1024 & 16 & 20 & 1.6 & 1.0 \\
Pythia-1b   & L13 & 2048 & 13 & 12 & 0.6 & 1.0 \\
Pythia-1.4b & L21 & 2048 & 12 & 14 & 0.6 & 1.0 \\
\bottomrule
\end{tabular}
\end{table}

\subsubsection{A Genuine Frequency Effect: Processing Dynamics}
\label{sec:logit_lens}

Despite this broad representational invariance, one genuine
frequency effect emerges in processing
dynamics within the same circuit.
Logit lens analysis reveals that the base model's output distribution
converges to the final prediction later for low-frequency tokens. We define
normalized layer depth as the layer number divided by the model's total
number of transformer layers. Convergence occurs at normalized depths of
0.78--0.90 for low-frequency tokens and 0.61--0.72 for high-frequency
tokens. The same pattern occurs in all five models
(per-model convergence depths in \autoref{app:logit_lens_detail}).
The convergence depth differs by less than 0.05 of the model's total
depth between each extracted circuit and its corresponding base model.
Thus, the later convergence for low-frequency tokens is already present
in the base model and is not introduced by circuit extraction.

The difference in convergence depth between low- and high-frequency tokens increases from approximately 0.06 of the model's total depth for Pythia-70m to approximately 0.29 for Pythia-1.4b (Table~\ref{tab:repr_logit_lens}). After Tuned Lens calibration~\mbox{\citep{belrose2023eliciting}}, low-frequency tokens still converge later in all five models, and the difference remains approximately three times larger in Pythia-1.4b than in Pythia-70m (\autoref{app:tuned_lens_detail}). As a result, the delay does not appear to result from misalignment in the standard logit lens and differences in component use and token embedding norms do not explain the delay. Its mechanistic cause remains unclear, although low-frequency tokens receive slightly larger contributions from late-layer MLPs.

Token frequency affects convergence timing, with low-frequency tokens converging later. However, circuits extracted from different frequency bands remain functionally interchangeable. This combination of frequency-dependent processing dynamics, structural variation among extracted circuits, and functional equivalence is a concrete instance of phantom specialization.

\subsection{Positive Control: Detecting Genuine Specialization}
\label{sec:two_route_control}

The preceding analyses show structural differences without corresponding functional differences. This raises the question of whether the pipeline would detect genuine specialization if it was present. The sweep across layers in \autoref{app:positive_control} shows that interchange patching can detect causal differences. However, validating the full pipeline of ACDC extraction, structural comparison, and cross-condition transfer testing requires a setting in which different conditions are independently known to rely on different computational routes. We first show why LSC cannot provide such a setting and then construct one that can.

LSC cannot provide this positive control because it contains only one task mechanism. However, this same property makes LSC useful for the main experiment. The task holds the required computation fixed, and the transfer and interchange tests show that the frequency bands are functionally interchangeable. Structural differences among circuits extracted from different bands therefore cannot be attributed to different task mechanisms, even though structural comparison alone might interpret them as evidence of specialization.

Two LSC variants further support this interpretation. In the reverse-copy variant, the target appears before the prefix and requires a copy offset of $-5$ rather than $+1$. The model achieves 0\% accuracy, so there is no learned reverse-copy behavior from which to extract a circuit. In the zero-distractor variant, we remove the 10 distraction tokens while leaving the copying rule unchanged. Cross-condition transfer efficiency between the standard and zero-distractor circuits is 96.7\%, close to the cross-band accuracy ratio of 97.8\%. This high transfer occurs despite their lower structural overlap, with a Jaccard similarity of 0.539 compared with 0.557 between frequency bands. Results are reported in Table~\ref{tab:positive_control_main}, with further details in \autoref{app:lsc_single_mechanism}. The zero-distractor result therefore extends phantom specialization to a second type of input variation which is the sequence length.

\begin{table}[t]
\centering
\caption{Cross-condition transfer between the zero-distractor and standard LSC variants for Pythia-160m. Each entry shows edge-level circuit accuracy. Transfer efficiency is 96.7\%, and Jaccard similarity is 0.539.}
\label{tab:positive_control_main}
\small
\begin{tabular}{lcc}
\toprule
 & Zero-dist.\ test & Standard test \\
\midrule
Zero-dist.\ circuit & 82.7\% & 80.4\% \\
Standard circuit & 88.9\% & 92.4\% \\
Base model & 88.4\% & 98.7\% \\
\bottomrule
\end{tabular}
\end{table}

\begin{figure}[t]
\centering
\resizebox{0.94\linewidth}{!}{%
\begin{tikzpicture}[
  tok/.style={minimum width=0.62cm, minimum height=0.52cm,
    font=\scriptsize, align=center},
  ans/.style={minimum width=1.15cm, minimum height=0.52cm,
    font=\scriptsize, align=center, thick},
]
  \def\s{0.95}
  \def\ax{11.1}
  \def\dx{9.9}

  \node[font=\scriptsize\bfseries, text=blue!60, anchor=east] at (-0.5, 0) {Route A};
  \foreach \i/\lab/\col in {
    0/BOS/gray!12, 1/{$\mathrm{CUE_A}$}/blue!30,
    2/$x_1$/gray!15, 3/$x_2$/gray!15, 4/$x_3$/orange!30,
    5/$x_4$/green!25, 6/$x_5$/gray!15, 7/$x_6$/gray!15,
    8/{$q\,{=}\,x_3$}/orange!30} {
    \node[tok, fill=\col, draw=gray!50] at (\i*\s, 0) {\lab};
  }
  \node[ans, fill=green!25, draw=green!55!black] at (\ax*\s, 0) {$x_4$};
  \draw[thick, orange!70!red, -{Stealth[length=2mm]}] (8*\s, -0.45) -- (4*\s, -0.45);
  \node[font=\tiny, text=orange!70!red] at (6*\s, -0.78)
    {find the query's earlier occurrence};
  \draw[thick, green!45!black, -{Stealth[length=2mm]}] (5*\s, 0.45) -- (\ax*\s, 0.45)
    node[midway, above, font=\tiny] {answer with the next token, which is in the prompt};

  \node[font=\scriptsize\bfseries, text=purple!60, anchor=east] at (-0.5, -2.4) {Route B};
  \foreach \i/\lab/\col in {
    0/BOS/gray!12, 1/{$\mathrm{CUE_B}$}/purple!30,
    2/$x_1$/gray!15, 3/$x_2$/gray!15, 4/$x_3$/gray!15,
    5/$x_4$/gray!15, 6/$x_5$/gray!15, 7/$x_6$/gray!15,
    8/{$q\,{=}\,x_3$}/orange!30} {
    \node[tok, fill=\col, draw=gray!50] at (\i*\s, -2.4) {\lab};
  }
  \node[ans, fill=purple!22, draw=purple!60] at (\ax*\s, -2.4) {$\mathrm{PERM}[q]$};
  \draw[thick, purple!60, -{Stealth[length=2mm]}] (8*\s, -1.95) -- (\ax*\s, -1.95)
    node[midway, above, font=\tiny] {look $q$ up in a memorized table};
  \node[font=\tiny, text=purple!60] at (\ax*\s, -2.92) {not in the prompt};

  \draw[dashed, rounded corners=2pt, black!65]
    (1*\s-0.52, 0.72) rectangle (1*\s+0.52, -2.78);
  \node[font=\tiny, text=black!65, align=center] at (1*\s, 1.12)
    {the cue: one token,\\the only difference};

  \draw[dashed, gray!55] (\dx*\s, 0.95) -- (\dx*\s, -3.15);
  \node[font=\tiny, text=black!55] at (5.3*\s, 1.15) {prompt};
  \node[font=\tiny, text=black!55] at (\ax*\s, 1.15) {prediction};
\end{tikzpicture}%
}%
\caption{The learned two-route task. The cue is a single token sitting at
the start of the prompt, boxed above. Both rows are the same base example
under the two cues, so the two prompts differ in that one token and in
nothing else, the way a pair of LSC prompts differs only in the frequency
band its tokens are drawn from. Under $\mathrm{CUE_A}$, the model must find the query's earlier occurrence and answer with the following token, $x_{i+1}$, which is present in the prompt. Under $\mathrm{CUE_B}$, it must ignore the prompt and answer $\mathrm{PERM}[q]$ from a fixed mapping memorized during training; this answer is absent from the prompt. Each rule's answer is scored against the answer from the other rule, so the incorrect option identifies the competing condition. Corruption replaces the content tokens under
the same cue, destroying both the retrieval target and the table lookup's
input while leaving the cue intact.}
\label{fig:two_route_task}
\end{figure}
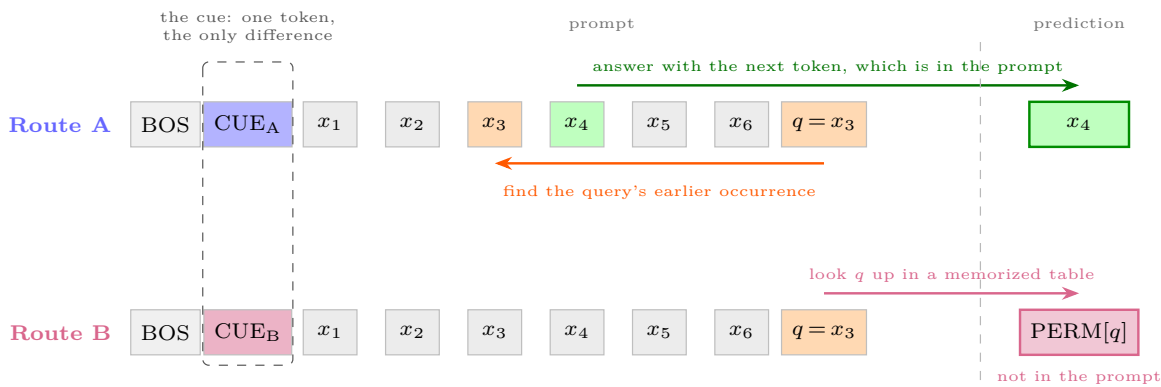

We therefore built a task with two ways of producing an answer, chosen
by a single token, the \emph{cue}, at the start of the prompt, 
(Figure~\ref{fig:two_route_task}).
Under one cue the model must look back in the prompt and answer with
the token that followed the query's earlier occurrence, the same
computation that LSC isolates.
Under the other cue, it must ignore the prompt and answer from a fixed
token-to-token table it memorized during training.
Two prompts can be identical apart from that one token and still have
different correct answers, so the cue plays the role that frequency
band plays in the main experiments.

We trained ten small transformers from scratch on an even mixture of the two cue conditions. Each model had four layers, $d_{\text{model}} = 128$, and four attention heads, and the models differed only in their random seeds. Seven models learned both rules, achieving at least $0.99$ accuracy on held-out prompts from each condition. Before circuit discovery, we used a direct ablation test independent of the pipeline to verify that the two rules depended on different component groups. A model passed if ablating all attention heads in layers~1--3 impaired the copying rule while preserving the memorized rule, whereas ablating the MLPs in those layers impaired the memorized rule while preserving the copying rule. Six of the seven models passed this check on a separate validation split, and only these six entered the circuit discovery pipeline. We refer to this test as the \emph{route separation check}. The full task, architecture, training procedure, and check are described in \autoref{app:two_route_control}.

We then ran the same pipeline on the six models that passed the route separation check. The setup follows the main experiments and uses ACDC with resample ablation, the same threshold selection rule, three conditions with three independent draws each, and the same structural and transfer analyses. The pipeline met all prespecified detection criteria in all six models (Table~\ref{tab:two_route_control}). Circuits extracted from repeated draws of the same route transfer almost perfectly (0.999) and overlap strongly (Jaccard 0.789), reproducing the within-condition phantom pattern of the main experiments. By contrast, circuits extracted for different routes overlap far less (Jaccard 0.211, compared with a baseline of 0.141 based only on circuit size), and the components specific to each route provide no benefit on the other route. Cross-route transfer efficiency is $-0.056 \pm 0.047$, compared with 0.81--0.99 across frequency conditions in LSC and SVA under the same resample ablation. Retrieval circuits are consistently larger than memorized-mapping circuits (73--91 vs.\ 29--47 edges), but the evidence for genuine specialization comes from the failure of cross-route transfer rather than from this structural difference alone.

\begin{table}[t]
\centering
\caption{Learned two-route positive control: aggregate results over the six
models that passed the route separation check (mean $\pm$ sample sd). The unchanged pipeline meets all detection criteria specified in advance in all six models. Transfer remains high within a route and falls across routes. For comparison,
transfer efficiency across frequency conditions in LSC and SVA is
0.81--0.99 under the same resample intervention.}
\label{tab:two_route_control}
\small
\begin{tabular}{lc}
\toprule
Quantity & Value \\
\midrule
Circuit accuracy on the extraction condition & $0.999 \pm 0.001$ \\
Within-condition cross-draw transfer & $0.999 \pm 0.001$ \\
Cross-route accuracy & $0.577 \pm 0.107$ \\
Cross-route transfer efficiency & $-0.056 \pm 0.047$ \\
Jaccard within condition (across draws) & $0.789 \pm 0.022$ \\
Jaccard across routes & $0.211 \pm 0.039$ \\
Jaccard baseline based only on circuit size & $0.141 \pm 0.015$ \\
\bottomrule
\end{tabular}
\end{table}

The same pipeline therefore distinguishes the two settings. On the two-route task, independent ablation confirms that the conditions depend on different component groups, and the pipeline detects both structural and functional divergence. Across the LSC frequency bands, it detects structural divergence without functional divergence. This contrast indicates that the absence of functional divergence on LSC is not simply caused by the pipeline's inability to distinguish circuits. The pipeline also detects the residual same-band advantage of 0.016--0.029 accuracy points in all five models, with $p \leq 0.028$, consistent with the nominal sensitivity estimate in \autoref{app:transfer_power}. However, this advantage disappears under zero ablation (\autoref{app:zero_ablation}). The limitations of this positive control are discussed in Section~\ref{sec:limitations}.

\subsection{Replication on a Second Task: Subject-Verb Agreement}
\label{sec:sva_replication}

The LSC results come from a non-semantic copying task. To test whether structural differences can coexist with broad cross-condition transfer on another task, we repeated the edge-level analyses on subject--verb agreement (SVA) using the frequency band design described in \autoref{sec:data}. We compare circuit structure and cross-condition transfer, test robustness under zero and mean ablation, and evaluate whether the core shared by most circuits is sufficient. We did not repeat the interchange interventions, EAP-IG comparison, source-level evaluation, or representational analyses. The claims in this subsection are therefore limited to the analyses reported here.

Briefly, each SVA prompt asks the model to choose between ``~is'' and ``~are'' according to the subject's number, while a nearer intervening noun has the opposite number. The subject and intervening noun are both drawn from pools defined by frequency band. The low, medium, and high pools are exactly matched for character length and define the primary matched analysis. The very-high band is included only in secondary unmatched analyses. The very-low band contains no eligible noun pairs and is therefore excluded from SVA. A separate very-low analysis for LSC is reported in \autoref{app:very_low}. We use the same circuit discovery, threshold selection, and edge-level evaluation procedures as for LSC. Across five models, five conditions, and three independent draws, this produces 75~circuits. Task construction and circuit discovery details are provided in \autoref{app:sva_generation}, while sensitivity analyses and controls are reported in \autoref{app:sva_replication}.

Before analyzing cross-condition transfer, we verify that the extracted SVA circuits are sufficient and necessary for task performance. Across models, the circuits retain an average of 89.5--93.8\% of base model's accuracy, and the lowest retention for any individual circuit is 80.9\%. When the selected circuit edges are removed while all other edges are retained, accuracy falls to at most 0.89\%. As a result, the extracted circuits preserve most of the model's performance, while their removal largely eliminates the behavior.

For the matched bands, circuits from different bands remain structurally distinct, but between-band overlap is comparable to within-band overlap across draws in all five models (Table~\ref{tab:sva_replication}), so band identity adds little edge-level structure beyond extraction noise. However, the Finding~1 pattern of systematically larger low-frequency circuits is absent in SVA because mean circuit sizes across the matched bands do not consistently decrease with frequency. More specifically, in Pythia-160m, mean circuit size increases from 92~edges at low frequency to 109 at high frequency. The replication claim is therefore limited to the split between structure and function, not circuit size. Functionally, band-specific edges transfer broadly, with transfer efficiency relative to the matched three-band universal core (25--230~edges) of $0.94$--$0.99$.

\begin{table}[t]
\centering
\caption{SVA replication. Matched scope restricts to the three
bands matched by length (low/medium/high). Gap is the mean accuracy on the extraction band minus
mean cross-band accuracy; $q$ comes from an exact directional permutation test that blocks by draw and enumerates 216 relabelings. We apply Benjamini--Hochberg correction across the five models. The transfer efficiency across five conditions (TE) is a
secondary check that includes the unmatched very-high and control
conditions. Jaccard compares between-band (within draw) to within-band
(across draws) edge overlap in the matched scope. Edges reports the mean circuit size over the nine circuits from matched bands and draws.}
\label{tab:sva_replication}
\small
\begin{tabular}{lcccccc}
\toprule
Model & Edges (matched) & TE (matched) & Gap & $q$ & TE (5-cond.) & Jaccard b/w \\
\midrule
Pythia-70m & 30 & 0.940 & $+$0.018 & 0.069 & 0.971 & 0.84 / 0.91 \\
Pythia-160m & 99 & 0.992 & $+$0.002 & 0.412 & 0.979 & 0.69 / 0.71 \\
Pythia-410m & 133 & 0.942 & $+$0.019 & 0.023 & 0.966 & 0.69 / 0.63 \\
Pythia-1b & 151 & 0.961 & $+$0.008 & 0.162 & 0.957 & 0.73 / 0.78 \\
Pythia-1.4b & 386 & 0.982 & $+$0.011 & 0.147 & 0.976 & 0.54 / 0.59 \\
\bottomrule
\end{tabular}
\end{table}

We restrict the main comparison to the three bands matched by length because the very-high pool cannot be matched by length, so including it would confound token frequency with word length.

For each model, we compute the advantage on the extraction band over the three
bands matched by length by averaging, across draws, each circuit's accuracy
on its own band minus its mean accuracy on the other two bands. Observations from the same draw share circuits and test data, so we do not use an independence test at that level. Instead we permute the three circuit labels within each
draw, enumerate all $6^3 = 216$ labelings, and apply Benjamini--Hochberg
correction across the five models.

The gaps are small ($+0.002$ to $+0.019$ accuracy points) and only
Pythia-410m shows an advantage on the extraction band that survives correction
($+0.019$, $q = 0.023$; its uncorrected $p$ sits at $1/216$, the
smallest value the enumeration can produce). It persists when the
mislabeled noun pairs are dropped from evaluation
(\autoref{app:sva_replication}). For comparison, the LSC residual same-band advantage is 0.016--0.029 accuracy points and is significant in all five models under its test for each condition and draw (Section~\ref{sec:phantom}).

The same comparison under the two LSC robustness protocols, zero
ablation and mean ablation
(\autoref{app:zero_ablation}), finds no advantage on the extraction band in any
model. Pythia-410m's gap falls from $+0.019$ under resample ablation
to $+0.002$ under zero ablation ($q = 0.687$), and to roughly zero
under mean ablation across all models $q \geq 0.57$ in the zero
family and $q \geq 0.79$ in the mean family.

This does not show that the resample result was an artifact. Comparing
the protocols directly, rather than testing each on its own, finds no
difference that survives correction (Pythia-410m: resample minus zero
$p = 0.22$, $q = 0.56$; resample minus mean $p = 0.037$, $q = 0.093$). A
result that is significant under one intervention and not under another is
not itself evidence that the protocols disagree. The alternative ablations
also leave little signal to measure: circuit accuracy falls to
$0.33$--$0.51$ under zero ablation ($0.35$ for Pythia-410m), and
mean ablation boosts are near zero (${\sim}0.001$). The SVA result is
therefore consistent with the resampling artifact found on LSC
(\autoref{app:zero_ablation}), but does not establish the same artifact on
SVA.

The shared core result also carries over to SVA. For each draw we keep the
edges that appear in at least three of the five condition circuits. These
cores shared by most bands contain 28--367 edges and recover 0.996--1.038 of
same-band circuit accuracy on average across models (per band--draw minimum 0.89), matching the LSC result
(Section~\ref{sec:threshold_relaxation}). The strict intersection of all
five condition circuits performs much worse on SVA, so the recoverable
shared computation is captured by the core shared by most bands, not by the
fully shared intersection.

Another LSC pattern that does not carry over as well is the transfer asymmetry from low to high frequency
(Section~\ref{sec:asymmetric_transfer}). We use the matched SVA analog of
the LSC statistic, comparing low and medium circuits evaluated on the high
test set against the high circuit evaluated on the low and medium test
sets. The asymmetry reverses in four of five models ($+0.037$ at
Pythia-70m; $-0.019$ to $-0.102$ in the others).

\subsection{Convergence of Evidence}
\label{sec:convergence}

The evidence developed in this section converges from every axis of the similarity triangle, and from the methodological checks, to a single conclusion: the structural specialization documented in Section~\ref{sec:results} is phantom.
\emph{Functionally}, band-specific edges transfer 81--97\% across all frequency bands, and edges shared by most bands ($k{\geq}3$) recover ${\geq}99\%$ of full circuit accuracy.
\emph{Representationally}, residual stream's CKA remains ${\geq}0.93$ (${\geq}0.98$ for models up to 1B) and MLP selectivity correlations between base model and circuit reach $\rho = 0.84$--$0.99$ for models ${\geq}160$M.
\emph{Causally}, interchange patching yields IIA~${\geq}0.94$ across all band pairs for models ${\geq}160$M, and Boundless DAS localizes information that distinguishes frequency bands to ${\leq}6.3\%$ of model dimensions.
\emph{Methodologically}, the zero ablation diagnostic eliminates the residual same-band advantage, source-level inflation is quantified at 0.22--0.85 points, an independent EAP-IG analysis finds a same-band advantage of at most 4.6 percentage points at any circuit size, and the two-route positive control confirms that the same pipeline reports functional divergence where an independent intervention verifies distinct routes (\autoref{sec:two_route_control}).
These patterns are not specific to LSC: under the core analyses, the SVA replication shows the same split between structural difference and broad functional transfer (\autoref{sec:sva_replication}).
No single test would be decisive. Instead, independent structural, functional, representational, and causal analyses converge on the same interpretation. Each line of evidence could have contradicted that interpretation if specialization had been genuine.

\section{Discussion}
\label{sec:discussion}

\subsection{Why the Phantom Exists}
\label{sec:why_phantom}

Phantom specialization arises because the mapping from circuit structure
to circuit function is many-to-one.
When multiple edges are individually dispensable because the model
implements the same function through redundant pathways~\mbox{\citep{mcgrath2023hydra, bushnaq2024using, rohweder2026hierarchical}}, greedy
pruning algorithms must choose which to keep, and small differences in
input statistics can tip these choices differently, producing
structurally distinct circuits that implement the same computation.
This resembles \emph{overdetermination}, a known limitation of counterfactual tests~\mbox{\citep{mueller2024missed}}. When several pathways are each sufficient on their own, ablating any one of them may have little effect, so single-component ablation can miss a cause that is really
there~\mbox{\citep{mueller2024missed}}. Greedy pruning faces the same ambiguity.
Each removal changes what the next measurement sees, so small
condition-dependent differences in KL can decide which of several
workable routes survives.
This marginality is directly measurable: ranked by an independent
importance measure (EAP-IG attribution), band-specific edges fall at or
beyond the selection boundary for the same number of edges (median rank ratio
1.1--3.3 across models; 59--84\% beyond it), while universal core edges
rank well inside (0.39--0.50; $p = 3.1 \times 10^{-5}$ in every model;
\autoref{app:edge_marginality}).
Band-specific edges can be framed as circuit \emph{
spandrels}~\mbox{\citep{gould1979spandrels}}: structural byproducts of the
[extraction] process rather than genuine functional adaptations.
Independent neuron-level evidence supports this conclusion as well. \mbox{\citet{liu2025distributed}} find that processing of rare tokens relies on coordinated subnetworks within shared MLP layers rather than modular separation, the neuron-level analog of our circuit-level finding that frequency does not route computation through different circuits.
The reverse can also happen. \mbox{\citet{franco2026finding}} show that the same canonical IOI head keeps its high-level role while carrying different internal signals across prompt templates (Section~\ref{sec:rw_degeneracy}). Shared components therefore do not guarantee identical computation, just as the divergent components here do not by themselves show different mechanisms.

To interpret these findings, we distinguish three sources of circuit variation: 
(i)~\emph{extraction instability} (same frequency band yields different edge
sets across runs), (ii)~\emph{non-uniqueness of minimal subgraphs}
(different conditions recover functionally equivalent circuits, the
empirical analog of the non-identifiability demonstrated exhaustively
in toy models by~\mbox{\citealp{meloux2025everything}}), and
(iii)~\emph{model-level degeneracy} (the model supports multiple
independent pathways;~\mbox{\citealp{edelman2001degeneracy}}).
Table~\ref{tab:evidence_mapping} maps our results to these sources.
Our evidence directly supports~(i) and~(ii); interpretation~(iii)
is consistent but not conclusively demonstrated.

\begin{table}[t]
\centering
\caption{Mapping of empirical results to sources of circuit variation:
(i)~extraction instability, (ii)~non-uniqueness of minimal subgraphs,
(iii)~model-level degeneracy.
A checkmark denotes direct support, while $\sim$ denotes consistency without decisive evidence. Results marked $\sim$ for source~(iii) are equally consistent with source~(ii), and only targeted ablation of multiple circuits would provide evidence specific to source~(iii). A dash indicates that the result does not bear on the source.}
\label{tab:evidence_mapping}
\small
\begin{tabular}{p{5.8cm}ccc}
\toprule
Empirical result & (i) & (ii) & (iii) \\
\midrule
Cross-draw Jaccard 0.21--0.73 (Sec.~\ref{sec:cross_draw}) & \checkmark & --- & --- \\
16--46\% edges selected in only one draw & \checkmark & --- & --- \\
Union circuits: $+$5 percentage points acc., ${\leq}$0.012 TE change & \checkmark & --- & --- \\
Cross-band transfer 81--97\% (Sec.~\ref{sec:generic_boost}) & --- & \checkmark & $\sim$ \\
Core shared by most bands ($k{\geq}3$) ${\geq}$99\% recovery (Sec.~\ref{sec:threshold_relaxation}) & --- & \checkmark & $\sim$ \\
IIA ${\geq}$0.94 across all band pairs, ${\geq}160$M (App.~\ref{app:interchange_patching}) & --- & \checkmark & $\sim$ \\
Boundless DAS: 0.6--6.3\% of $d_{\text{model}}$ (App.~\ref{app:boundless_das}) & --- & \checkmark & $\sim$ \\
EAP-IG: no same-band transfer advantage at any size (App.~\ref{app:method_comparison}) & --- & \checkmark & $\sim$ \\
Per-example agreement 78--93\%, no same-band advantage & \checkmark & \checkmark & --- \\
SVA task: transfer 0.94--0.99, recovery of the core shared by most circuits ${\approx}1$, between-band within-draw ${\approx}$ within-band Jaccard across draws (Sec.~\ref{sec:sva_replication}) & \checkmark & \checkmark & $\sim$ \\
\bottomrule
\end{tabular}
\end{table}
Cross-draw analysis confirms extraction instability is substantial
(16--46\% of edges in only one draw), and the EAP-IG analysis
in \autoref{sec:generic_edges}, which selects largely
different edges (Jaccard 0.28--0.60 with ACDC), likewise finds no band
specificity, providing partial evidence for non-uniqueness.
Per-example agreement (78--93\% for models ${\geq}160$M, with no
same-band advantage) confirms that agreement on coarse metrics is not
masking systematic divergence.
All three sources likely contribute, although their relative importance is unclear. Regardless of which source dominates, the evidence indicates a many-to-one mapping from extracted circuit structure to measured function.
As motivated in the Introduction, there is no privileged level of causation in complex systems~\mbox{\citep{noble2012theory}}. Noble argues, from cardiac cell modeling, that downward causation is not merely compatible with lower-level mechanisms but necessary for understanding multiscale systems; the causal structure one observes depends on the scale at which one analyzes it. Our results provide an empirical instance of this principle in neural networks. More specifically, at source-level granularity, circuits appear faithful and specialization appears genuine and at edge-level granularity, the phantom is revealed. The 0.22--0.85 gap between source-level and edge-level circuit accuracy shows how much source-level evaluation can overstate circuit faithfulness (Section~\ref{sec:inflation}).

This resembles causal emergence~\mbox{\citep{hoel2013quantifying}}, in which a coarse-grained description can provide a more effective account of a system's causal interactions than a fine-grained description. Source-level evaluation similarly groups many functionally equivalent edge configurations into a single apparently faithful circuit, hiding the differences among them. In Hoel et al.'s framework, this loss of detail is useful when the goal is to identify the scale with the most effective causal interactions. In circuit evaluation, however, those fine-grained differences are important because they show whether an extracted circuit is unique or one of many functionally equivalent subgraphs. This is, to our knowledge, the first quantitative demonstration of causal structure that depends on evaluation granularity in neural network circuits. For circuit discovery, this means that edge-level evaluation is not merely a stricter test but a fundamentally different window onto the model's causal structure, one that reveals redundancy and degeneracy that coarser analyses systematically obscure.
The practical consequence is that treating a single discovered circuit
as \emph{the} mechanistic explanation overstates what the evidence
supports. Methods that aggregate over circuit distributions may be
more robust than point estimates.

\subsection{Implications for Evaluating Circuits Across Conditions}
\label{sec:implications}

Our results yield four practical recommendations for the circuit
discovery and circuit analysis community.

\paragraph{Cross-condition transfer tests are necessary.}
Structural differences between circuits should not be taken as evidence of
functional specialization without cross-condition transfer tests.
In our study, circuits whose edge sets overlap by as little as $0.37$
(Jaccard; Table~\ref{tab:structural_jaccard}) produce equivalent outputs
across all frequency bands (Section~\ref{sec:generic_boost}).
Any future comparison of circuits across input conditions, model variants,
or training stages should verify whether observed structural differences
translate to functional differences before interpreting them as evidence
of distinct mechanisms.

\paragraph{Edge-level evaluation should be the primary metric.}
Source-level evaluation inflates apparent circuit accuracy by 0.22--0.85
points (Section~\ref{sec:inflation}) and can make even
incomplete circuits appear faithful.
We recommend reporting edge-level metrics as the primary faithfulness
measure and treating source-level results as an upper bound that
reflects the contribution of unselected pathways.
This confound is not specific to our task or perturbation axis; it
applies to any study comparing circuits extracted under different
conditions.

\paragraph{Multiple extractions are needed to identify stable structure.}
Even within the same frequency band, 27--79\% of edges appear in only one or
two of three extractions (Section~\ref{sec:cross_draw}).
A single ACDC run therefore cannot distinguish stable circuit structure from extraction noise.
Running at least three independent extractions and retaining only
majority-vote edges ($k{\geq}2$ of 3 draws, analogous to our
$k{\geq}3$ of 5 conditions) substantially reduces noise and yields a more
reliable circuit estimate.

\paragraph{Structural metrics require careful controls.}
Within models, Jaccard overlap and functional transfer are nearly uncorrelated (Section~\ref{sec:variance_decomposition}, with pair-level results in \autoref{app:structure_function_detail}). Even the largest within-band versus between-band Jaccard gap, 0.032 with $d = 1.22$--$1.45$, does not correspond to a functional difference. Evidence for genuine specialization should therefore come from an advantage on the extraction condition in cross-condition transfer tests, not from the magnitude of structural divergence alone.
More broadly, model scale dominates variance on every metric, producing pervasive
Simpson's paradox (Section~\ref{sec:variance_decomposition}).
Studies pooling observations across model sizes~\mbox{\citep[e.g.,][]{tigges2024llm, mondorf2025circuit, ferrando2024similarity}}
risk creating spurious associations; we recommend reporting both pooled
and per-model statistics.

\subsection{Limitations}
\label{sec:limitations}

Our findings come from a controlled setting using one model family, two tasks (non-semantic copying and subject--verb agreement), and one axis of variation which is token frequency. Circuit discovery relies primarily on ACDC. However, the result is not specific to ACDC. An EAP-IG edge ranking analysis~\mbox{\citep{hanna2024have}} finds a same-band advantage of at most 4.6 percentage points across circuit sizes (\autoref{app:method_comparison}). Replication with differentiable mask optimization methods~\mbox{\citep{bhaskar2024finding}} remains an important direction for future work. Separately, the causal interventions in Section~\ref{sec:causal_confirmation} test band interchangeability without relying on the discovered ACDC circuits.
On the subject-verb agreement replication (\autoref{sec:sva_replication}) we
ran only the core structural and transfer analyses. The interchange,
representational, source-level, and EAP-IG analyses are limited to LSC, and
extension to richer semantic tasks and to architectures with different
redundancy profiles remains open.

Our circuit-level analyses are also limited to prompts without examples. We evaluated accuracy in a few-shot variant by adding two examples of the copying task. This increased top-1 accuracy and reduced the accuracy gap between matched frequency bands in the three largest models (\autoref{app:fewshot}). We did not extract circuits because preliminary timing with Pythia-160m indicated that the longer prompts would approximately double the standard per-circuit extraction cost. Whether phantom specialization persists under few-shot prompting therefore remains open.

The length matching constraint excludes the \texttt{very\_low} band from the matched design, so we analyze it separately without controlling for character length. Transfer under resample, mean, and zero ablation and the EAP-IG cross-method check are reported in \autoref{app:very_low}. The representational and interchange analyses include all five bands.

Another limitation is that matching token length does not fully isolate frequency from other token properties. Frequency still correlates with co-occurrence patterns and token-level semantics (Section~\ref{sec:data}), which could contribute to the structural differences we observe. However, these correlations do not explain the functional equivalence, which we measure directly. The representational analyses also show that circuit extraction preserves the base model's embedding geometry (\autoref{app:embedding_detail}).

Position aggregation introduces another limitation~\mbox{\citep{haklay2025position}}. ACDC assigns each edge a single keep-or-prune decision across all token positions. If two bands use the same edge at different positions, transferring a circuit preserves that edge at every position. Cross-band transfer may therefore appear stronger than it would with position-specific masks. Our fixed prompt template and corruption scheme partly constrain this problem by limiting nonzero resample interventions to the five repeated prefix positions. However, position aggregation remains a limitation because information computed at earlier uncorrupted positions can reach those positions through causal attention, and the intervention cannot isolate these contributions by position. A position-resolved attribution check shows that bands differ slightly more than independent draws in how attribution to shared edges is distributed across these positions. Gaps in profile cosine similarity range from 0.002 to 0.006, with all $p \leq 0.0011$ (\autoref{app:positional_attribution}). Interchange interventions at the prediction position provide complementary position-specific evidence (\autoref{app:interchange_patching}), but neither analysis replaces position-aware circuit discovery.

The positive control uses a deliberately constructed setting rather than a naturally occurring one. It consists of small transformers trained on a two-route task, with route separation independently verified by ablating attention and MLP groups (Section~\ref{sec:two_route_control}). The control shows that the pipeline can detect specialization in this setting. However, it does not show that the pipeline would detect naturally occurring competition between mechanisms in pretrained models~\mbox{\citep{ortu2024competition}}, nor does it test whether the pipeline recovers planted ground-truth edges.

Using only three draws per frequency band limits the precision of our structural estimates. The functional results are nevertheless consistent across draws, with cross-draw transfer ratios of 0.990--1.017. Additional draws were impractical because extracting 75~circuits with ACDC required approximately 736 GPU-hours. Based on the power analysis, we recommend at least five draws for models ${\geq}1$B parameters (\autoref{app:power_analysis}). Pythia-70m requires additional caution because its base accuracy is low (30--67\%; \autoref{app:failure_analysis}). Despite this limitation, all primary conclusions hold within each model individually. We do not claim that all structural variation is phantom. Our claim is that structural divergence alone is insufficient evidence for specialization.

\section{Conclusion}
\label{sec:conclusion}

Our results indicate a many-to-one relationship between circuit structure and function. This degeneracy is visible only at edge-level evaluation
granularity. We term the resulting mismatch \emph{phantom specialization}, meaning
structural divergence that does not correspond to functional
specialization. 
In the Literal Sequence Copying task across four frequency bands
and five Pythia scales, ACDC finds structurally distinct circuits, but
band-specific edges act as generic boosters (81--97\% cross-band
transfer efficiency), edges shared by ${\geq}3$ bands recover
${\geq}99\%$ of full circuit accuracy in models above 70M, and the
universal core contains the canonical copy mechanism.
A smaller subject-verb agreement replication shows the same split. Circuits extracted from different frequency bands differ structurally but transfer broadly, and the edges shared by most circuits recover nearly all of a circuit's accuracy.

Our methodological recommendations include cross-condition transfer tests, edge-level evaluation, and multiple extractions with majority-vote aggregation. These practices address recurring challenges in circuit discovery and can be adopted by any study comparing circuits across conditions. A single discovered circuit should therefore not be treated as uniquely identifying the underlying mechanism.

Promising future directions include Bayesian circuit discovery, joint extraction across multiple conditions,
feature-level circuit discovery via sparse autoencoders~\mbox{\citep{marks2025sparse}} or transcoders~\mbox{\citep{dunefsky2024transcoders}}
(testing whether phantom specialization persists when circuits are
defined over monosemantic features), and developmental analysis using
Pythia's training checkpoints.
The structural variation itself is real and input-driven. What is phantom is the functional specialization inferred from that structural divergence.

\section*{Acknowledgments}

This work has been funded by the LOEWE Distinguished Chair ``Ubiquitous Knowledge Processing,'' LOEWE initiative, Hesse, Germany (Grant Number: LOEWE/4a//519/05/00.002(0002)/81), by the German Federal Ministry of Education and Research, and by the Hessian Ministry of Higher Education, Research, Science and the Arts within their joint support of the National Research Center for Applied Cybersecurity ATHENE. We also thank Falko Helm and Phu Hoang (UKP Lab, TU Darmstadt) for their feedback on an early draft of this work.

\bibliography{main}
\bibliographystyle{tmlr}

\clearpage
\appendix

{%
\makeatletter
\renewcommand{\@part}[1]{\noindent{\Large\bfseries #1}\par\medskip}
\makeatother
\part*{Appendix}
\addcontentsline{toc}{part}{Appendix}
}

\etocsettocstyle{\noindent\textbf{Contents}\smallskip\par}{}
\etocsettocdepth{subsection}
\localtableofcontents
\renewcommand{\textfraction}{0.05}
\renewcommand{\topfraction}{0.95}
\renewcommand{\bottomfraction}{0.5}
\renewcommand{\floatpagefraction}{0.85}
\FloatBarrier

\vspace{1em}

\setcounter{topnumber}{4}
\setcounter{bottomnumber}{2}
\setcounter{totalnumber}{6}

\section{Extended Background: Circuit Discovery and Evaluation}
\label{app:circuit_discovery}

This appendix provides an extended review of circuit discovery methods, evaluation practices, and open challenges in mechanistic interpretability, complementing the overview in Section~\ref{sec:rw_circuit_discovery}. For readers familiar with the field, the discussion in the main text is self-contained. This appendix offers additional depth on method families, metrics, and structural challenges.

\subsection{Circuit Discovery and Evaluation}

\begin{figure}[t]
\centering
\resizebox{0.5\columnwidth}{!}{%
\begin{forest}
for tree={
  draw,
  rounded corners=2pt,
  thin,
  align=center,
  grow=east,
  parent anchor=east,
  child anchor=west,
  anchor=west,
  minimum height=6.2mm,
  inner xsep=5pt,
  inner ysep=2pt,
  l sep=9mm,
  s sep=5mm,
  edge={->, semithick, >={Stealth[length=1.8mm]}},
  edge path={
    \noexpand\path[\forestoption{edge}]
      (!u.east) -- +(7pt,0) |- (.west)\forestoption{edge label};
  },
  font=\small
}
[Ablation/corruption, fill=gray!10, text width=3.2cm, font=\small\bfseries
  [Resampling, fill=red!5, text width=2.5cm, font=\small\bfseries
    [{needs counterfactuals}, fill=red!5, draw=red!50, text width=3.6cm, font=\scriptsize, minimum height=5.0mm, inner ysep=1.4pt]
    [{distribution-aware}, fill=green!5, draw=green!50!black, text width=3.6cm, font=\scriptsize, minimum height=5.0mm, inner ysep=1.4pt]
  ]
  [Random noise, fill=red!5, text width=2.5cm, font=\small\bfseries
    [{often unrealistic}, fill=red!5, draw=red!50, text width=3.6cm, font=\scriptsize, minimum height=5.0mm, inner ysep=1.4pt]
    [{stress test}, fill=green!5, draw=green!50!black, text width=3.6cm, font=\scriptsize, minimum height=5.0mm, inner ysep=1.4pt]
  ]
  [Mean, fill=red!5, text width=2.5cm, font=\small\bfseries
    [{distorts nonlinear structure}, fill=red!5, draw=red!50, text width=3.6cm, font=\scriptsize, minimum height=5.0mm, inner ysep=1.4pt]
    [{less extreme}, fill=green!5, draw=green!50!black, text width=3.6cm, font=\scriptsize, minimum height=5.0mm, inner ysep=1.4pt]
  ]
  [Zero, fill=red!5, text width=2.5cm, font=\small\bfseries
    [{off-manifold risk}, fill=red!5, draw=red!50, text width=3.6cm, font=\scriptsize, minimum height=5.0mm, inner ysep=1.4pt]
    [{simple}, fill=green!5, draw=green!50!black, text width=3.6cm, font=\scriptsize, minimum height=5.0mm, inner ysep=1.4pt]
  ]
]
\end{forest}%
}
\caption{Common ways to construct the corrupt run for activation patching. Simpler ablations are easy to apply but can move activations outside the data manifold or distort the representation. Corruption by resampling is usually more realistic but requires suitable counterfactual examples.}
\label{app:fig:corruption_tree}
\end{figure}
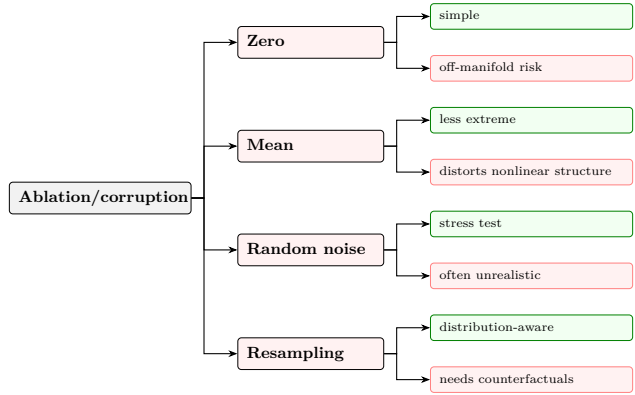

\subsubsection{Defining Circuits}

Mechanistic interpretability seeks to reverse-engineer neural networks, analogous to recovering interpretable source code from a compiled program~\mbox{\citep{olah2022mechanistic, elhage2021mathematical}}. Three concepts organize this work. \emph{Features} are directions in activation space that correspond to interpretable concepts~\mbox{\citep{elhage2022toy}}. \emph{Circuits} are sparse subgraphs of the model's computational graph that explain how it implements a specific behavior~\mbox{\citep{olah2020zoom, NEURIPS2023_34e1dbe9, rai2024practical}}. \emph{Universality} is the hypothesis that different models learn the same features and circuits for the same tasks~\mbox{\citep{olah2020zoom, chughtai2023toy}}.

Throughout this paper, we use \emph{circuit} as an operational term for a sparse subset of model components or connections whose preservation recovers a substantial fraction of the full model's performance on the target task under a specified intervention and evaluation procedure. This definition is intentionally method-dependent. Discovered circuits depend on how the computational graph is defined, which variables are intervened on, and how faithfulness is measured.

In a transformer language model, attention heads and MLP blocks communicate through the residual stream by reading from it and adding their outputs back to it. This information flow forms a directed acyclic graph (DAG). Nodes in this graph correspond to attention heads and MLP blocks, and edges indicate that one component reads an output written by another to the residual stream. Due to the additive nature of the residual stream, even components in nonadjacent layers are effectively connected by writing to and reading from the same stream~\mbox{\citep{elhage2021mathematical}}. A \emph{circuit} is a subgraph of this DAG selected for a target behavior~\mbox{\citep{NEURIPS2023_34e1dbe9, rai2024practical}}. Although circuits were originally defined as connections between features~\mbox{\citep{olah2020zoom}}, subsequent work has generalized them to connections between activation outputs of transformer components~\mbox{\citep{olsson2022context, wang2022interpretability}}. More recently, sparse autoencoders (SAEs)~\mbox{\citep{o2024sparse}} have enabled a finer decomposition by mapping activations into higher-dimensional sparse representations with more monosemantic units, and initial work has begun defining circuits over SAE features rather than architectural components~\mbox{\citep{o2024sparse}}. However, to our knowledge, whether feature-level circuits exhibit the same non-uniqueness as component-level circuits remains an open question.

A key challenge is that the naive decomposition into architectural components (individual neurons, attention heads, or layers) does not cleanly carve the network at its functional joints~\mbox{\citep{sharkey2025open}}. Neurons and attention heads are often \emph{polysemantic}, responding to multiple unrelated features~\mbox{\citep{elhage2022toy}}, which confounds circuit boundaries and motivates caution when interpreting circuits defined at this level of granularity.

\begin{figure}[!htbp]
\centering
\scalebox{0.55}{%
\begin{tikzpicture}[
    box/.style={
      rounded corners=3pt,
      thick,
      draw,
      align=center,
      font=\small,
      minimum height=0.95cm,
      inner sep=4pt
    },
    arr/.style={-{Stealth[length=2mm]}, thick},
    panel/.style={font=\bfseries\small}
]

\tikzset{
  flowbox/.style={
    rounded corners=3pt,
    thick,
    draw,
    align=center,
    font=\small,
    text width=2.55cm,
    minimum height=0.95cm,
    inner sep=3pt
  }
}

\node[panel] at (0,0) {(a) Activation Patching};

\node[flowbox, fill=blue!10, draw=blue!45] at (-6.55,-0.7) (cleanrun)
  {\textbf{Run clean}\\ input};
\node[flowbox, fill=blue!18, draw=blue!55] at (-3.35,-0.7) (cleancache)
  {\textbf{Cache activation}\\ at $C_i$};
\node[flowbox, fill=red!10, draw=red!45] at (-6.55,-2.4) (corrun)
  {\textbf{Run corrupt}\\ input};
\node[flowbox, fill=red!18, draw=red!55] at (-3.35,-2.4) (corcache)
  {\textbf{Cache activation}\\ at $C_i$};
\node[flowbox, fill=violet!10, draw=violet!45] at (0.15,-1.55) (patch)
  {\textbf{Insert activation}\\ at $C_i$};
\node[flowbox, fill=violet!18, draw=violet!55] at (3.35,-1.55) (patched)
  {\textbf{Rerun patched}\\ \textbf{model}};
\node[flowbox, fill=gray!8, draw=gray!45] at (6.55,-1.55) (measure)
  {\textbf{Compare}\\ \textbf{output}};

\draw[arr, blue!55] (cleanrun) -- (cleancache);
\draw[arr, red!55]  (corrun) -- (corcache);
\draw[arr, blue!55] (cleancache.east) -- ++(0.45,0) |- (patch.west);
\draw[arr, red!55]  (corcache.east) -- ++(0.45,0) |- (patch.west);
\draw[arr, violet!60] (patch) -- (patched);
\draw[arr, gray!65] (patched) -- (measure);

\node[panel] at (0,-3.6) {(b) Two Directions};

\node[box, fill=gray!8, draw=gray!45, text width=1.9cm] at (-5.8,-4.5) (n1)
  {\textbf{Noising}};
\node[box, fill=blue!10, draw=blue!45, text width=2.2cm] at (-3.2,-4.5) (n2)
  {\textbf{Base}\\ clean run};
\node[box, fill=red!10, draw=red!45, text width=2.5cm] at (-0.2,-4.5) (n3)
  {\textbf{Insert}\\ corrupt $C_i$};
\node[box, fill=gray!8, draw=gray!45, text width=2.5cm] at (2.8,-4.5) (n4)
  {\textbf{Performance}\\ drops};
\node[box, fill=gray!8, draw=gray!45, text width=2.1cm] at (5.7,-4.5) (n5)
  {\textbf{Necessary}};

\draw[arr, black!55] (n1) -- (n2);
\draw[arr, blue!55] (n2) -- (n3);
\draw[arr, red!55] (n3) -- (n4);
\draw[arr, black!55] (n4) -- (n5);

\node[box, fill=gray!8, draw=gray!45, text width=1.9cm] at (-5.8,-6.0) (d1)
  {\textbf{Denoising}};
\node[box, fill=red!10, draw=red!45, text width=2.2cm] at (-3.2,-6.0) (d2)
  {\textbf{Base}\\ corrupt run};
\node[box, fill=blue!10, draw=blue!45, text width=2.5cm] at (-0.2,-6.0) (d3)
  {\textbf{Insert}\\ clean $C_i$};
\node[box, fill=gray!8, draw=gray!45, text width=2.5cm] at (2.8,-6.0) (d4)
  {\textbf{Performance}\\ recovers};
\node[box, fill=gray!8, draw=gray!45, text width=2.1cm] at (5.7,-6.0) (d5)
  {\textbf{Sufficient}};

\draw[arr, black!55] (d1) -- (d2);
\draw[arr, red!55] (d2) -- (d3);
\draw[arr, blue!55] (d3) -- (d4);
\draw[arr, black!55] (d4) -- (d5);

\end{tikzpicture}%
}%
\caption{Activation patching and its two main directions. \textbf{(a)}~Run the model on a clean input and on a corrupted input, cache the activation of a chosen component $C_i$ from each run, insert one into the other, rerun the model with that patched activation, and compare the output. \textbf{(b)}~In \emph{noising}, a corrupt activation is inserted into a clean base run. A performance drop indicates that the component is necessary. In \emph{denoising}, a clean activation is inserted into a corrupt base run. A performance recovery indicates that the component is sufficient~\mbox{\citep{rai2024practical, heimersheim2024use}}.}
\label{fig:intervention}
\end{figure}
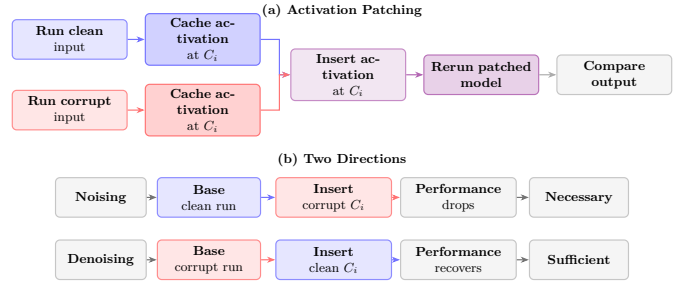

The causal abstraction framework of \mbox{\citet{geiger2023causal}} formalizes this approach using interchange interventions. These interventions replace a component's activation with the corresponding activation from a counterfactual input and measure the resulting change in the model's output. This idea underpins \emph{activation patching}~\mbox{\citep{vig2020investigating, meng2022locating}}, the causal tool used by most circuit discovery methods. An activation patching experiment involves three forward passes. First, a \emph{clean run} processes the original prompt and caches all intermediate activations. Second, a \emph{corrupt run} processes a modified prompt that disrupts the target behavior. Third, a \emph{patch run} swaps a specific component's activation between the two runs while the rest of the computation proceeds normally. Two complementary directions of patching are used~\mbox{\citep{rai2024practical, heimersheim2024use}}. In \emph{noising} patching from the clean input to the corrupted input, the model runs on the clean input but one component's activation is replaced with its value from the corrupt run. A drop in performance indicates that the component is necessary. In \emph{denoising} patching from the corrupted input to the clean input, the model runs on the corrupt input but one component is restored to its clean value. A recovery in performance indicates that the component is sufficient (Figure~\ref{fig:intervention}). \emph{Path patching}~\mbox{\citep{goldowsky2023localizing}} extends activation patching to isolate the effect of specific computational pathways rather than individual components.

The choice of corruption technique for the corrupt run is consequential. \emph{Zero ablation}~\mbox{\citep{olsson2022context}} replaces the component's output with a zero vector, but this can push the model far out of distribution, causing effects unrelated to the component's actual role. \emph{Mean ablation}~\mbox{\citep{wang2022interpretability}} replaces the output with its mean over a reference distribution, partially mitigating the out-of-distribution problem but failing when activation distributions are nonlinear (e.g., the mean of points on a circle lies at the center, not on the circle). \emph{Resampling ablation}~\mbox{\citep{chan2022causal}} replaces the output with the activation from a different, counterfactual input drawn from the same task distribution, avoiding the out-of-distribution issue at the cost of requiring paired counterfactual examples. Each choice can yield different circuits from the same model and task~\mbox{\citep{heimersheim2024use, meloux2025mechanistic}}; Figure~\ref{app:fig:corruption_tree} summarizes the options and their trade-offs.

\subsubsection{Manual Circuit Discovery}

The earliest circuits were identified through a hypothesis-driven, iterative process (Figure~\ref{fig:discovery_pipelines}a) that follows a common workflow~\mbox{\citep{rai2024practical}}: (1)~select a behavior for which the model performs well, ensuring the mechanism is reliably present; (2)~define the computational graph, choosing the granularity of nodes (e.g., attention heads and MLP blocks) and the edge structure; (3)~\emph{localize} important nodes and edges via intervention; (4)~\emph{interpret} each component by generating and validating hypotheses about its functional role; and (5)~\emph{evaluate} the resulting circuit. In practice, steps~3--4 are iterative: researchers inspect activations, hypothesize which components are relevant to the task, intervene via ablation or patching, measure the behavioral effect, and refine the hypothesized circuit until a stable, minimal subgraph emerges.

This manual approach has produced detailed mechanistic accounts of several core transformer behaviors: in-context copying via induction heads~\mbox{\citep{olsson2022context}}, indirect object identification (IOI) with 26 attention heads grouped into 7 classes in GPT-2 Small~\mbox{\citep{wang2022interpretability}}, greater-than comparison~\mbox{\citep{hanna2023does}}, copy suppression across the full training distribution~\mbox{\citep{mcdougall2024copy}}, successor operations~\mbox{\citep{ICLR2024_2722a0cc}}, docstring completion~\mbox{\citep{heimersheim2023circuit}}, three-letter acronym prediction~\mbox{\citep{garcia2024does}}, subject--verb agreement across languages~\mbox{\citep{finlayson2021causal, ferrando2024similarity}}, extractive question-answering~\mbox{\citep{basu2025mechanistic}}, and arithmetic via a ``bag of heuristics''~\mbox{\citep{nikankin2024arithmetic}}. These studies have revealed specialized component roles: previous token heads, duplicate token heads, induction heads, negative heads that suppress tokens that appeared earlier, successor heads, and multifunction heads that implement different algorithms depending on context~\mbox{\citep{heimersheim2023circuit}}. The compositional requirements for induction head formation (a core building block of many circuits) have also been studied~\mbox{\citep{singh2024needs, crosbie2025induction}}. The iterative description--validation loop is rarely made explicit in published work; only the final interpretation is presented~\mbox{\citep{sharkey2025open}}. However, the manual approach is labor-intensive, relies on subjective choices about which components to test, and does not scale to the thousands of edges present even in moderately sized models.

\subsubsection{Automated Circuit Discovery}

To overcome the scalability limitations of manual analysis, automated methods search for sparse subgraphs for a target task (Figure~\ref{fig:discovery_pipelines}b). These methods generally start from the full computational graph, score or mask each edge according to a relevance criterion, select a sparse subgraph, and validate it through patching or ablation. Existing methods fall into several families:

\paragraph{Iterative pruning.} \mbox{\citet{NEURIPS2023_34e1dbe9}} introduced ACDC, which systematizes the manual workflow into three steps (select behavior, define computational graph, patch activations) and automates the third step. ACDC greedily prunes edges in reverse topological order. For each candidate edge, it removes the edge, patches all pruned edges with corrupted activations, and measures the change in a divergence metric such as KL divergence. If the change falls below a threshold~$\tau$, the edge is permanently removed. ACDC validated its approach by rediscovering known circuits (e.g., all 5 component types in the Greater-Than circuit, selecting 68 of 32,000 edges in GPT-2 Small). However, ACDC automates only localization, not the subsequent interpretation of component roles~\mbox{\citep{sharkey2025open}}. It is greedy and sensitive to the threshold (the order of parent iteration can affect results) and can fail to recover all negative (suppressive) components when optimizing a single metric~\mbox{\citep{NEURIPS2023_34e1dbe9}}. It also requires an independent forward pass for every edge tested, making it computationally expensive for larger models~\mbox{\citep{rai2024practical}}.

\paragraph{Attribution-based methods.} Attribution patching approximates edge importance through gradients rather than testing each edge individually. EAP~\mbox{\citep{syed2024attribution}} computes first-order approximations of the effect of patching each edge. It scores all edges using only two forward passes and one backward pass, making it orders of magnitude faster than ACDC. Although the correlation between attribution and activation patching scores is modest ($R^2 \approx 0.27$ on the Docstring task~\mbox{\citep{syed2024attribution}}), EAP outperforms ACDC on circuit recovery because edge \emph{ranking} matters more than magnitude accuracy. EAP-IG~\mbox{\citep{hanna2024have}} refines these estimates using integrated gradients to mitigate zero gradients at the linearization point, and AtP*~\mbox{\citep{kramar2024atp}} further addresses failure modes from attention saturation and cancellation between direct and indirect effects. On the MIB benchmark, EAP-IG generally achieves the strongest performance among automated methods~\mbox{\citep{mueller2025mib}}. These methods trade some faithfulness for much greater scalability but rely on a first-order approximation of the true patching effect, and it remains unclear whether this is adequate for all settings~\mbox{\citep{sharkey2025open}}.

\paragraph{Differentiable masking.} Edge pruning methods formulate circuit discovery as differentiable optimization over continuous edge masks. \mbox{\citet{bhaskar2024finding}} learn a mask for each edge that is jointly optimized to minimize a faithfulness loss subject to a sparsity penalty, using a disentangled residual stream that retains all previous activations for per-edge masking. Their method finds circuits in GPT-2 with less than half the edges of ACDC/EAP circuits at equal faithfulness, and scales to CodeLlama-13B~\mbox{\citep{roziere2023code}}, 100$\times$ larger than models typically tackled by automated methods. \mbox{\citet{yu2024functional}} propose DiscoGP, which jointly prunes both edges and weight parameters at neuron-level granularity, and demonstrate that ``canonical circuits'' identified by prior work have very low functional faithfulness when evaluated in isolation. These approaches avoid the greedy ordering dependence of ACDC but introduce their own hyperparameters (sparsity weight, learning rate, mask initialization).

\paragraph{Methods using one forward pass.} \mbox{\citet{franco2026finding}} propose ACC++, which traces information flow through the QK attention matrices via SVD decomposition in a single forward pass, without requiring activation patching or multiple model evaluations. This yields per-prompt circuits at low computational cost but is limited to attention-mediated pathways.

\paragraph{Subspace-level methods.} Standard activation patching replaces entire hidden representations, implicitly assuming a localist mapping between causal variables and disjoint sets of neurons. Distributed Interchange Interventions (DII) instead intervene in rotated subspaces of the representation, enabling more fine-grained analysis when features are distributed across neurons~\mbox{\citep{geiger2024finding}}. Distributed Alignment Search (DAS) extends this by learning the rotation matrix and the $k$-dimensional subspaces that best align with high-level causal variables in a supervised fashion.

Despite their differences in search procedure, these methods share a common assumption: the discovered circuit is often treated as if it were the unique explanation for the target behavior~\mbox{\citep{sharkey2025open, meloux2025everything}}. The output of each method depends on the threshold, metric, patching direction, sparsity penalty, and data distribution. Nevertheless, the resulting circuit is typically reported as a single fixed object.

\subsubsection{Metrics and Evaluation}

Evaluating a discovered circuit requires assessing whether it faithfully captures the model's computation for the target behavior. Four properties are commonly considered~\mbox{\citep{wang2022interpretability, shi2024hypothesis}}:

\begin{itemize}
    \item \textbf{Faithfulness}: does the circuit reproduce the full model's behavior on the target task?
    \item \textbf{Completeness}: does the circuit capture all components the model uses for the task?
    \item \textbf{Sufficiency}: does the circuit alone produce correct outputs when all other components are ablated?
    \item \textbf{Minimality}: is the circuit as sparse as possible while remaining faithful?
\end{itemize}

\paragraph{Metric choice.} After performing an intervention, the resulting change in model output must be quantified. Three metrics are common~\mbox{\citep{rai2024practical, heimersheim2024use}}. \emph{Raw probability} or \emph{logit} measures the change in the model's confidence on the correct token before and after patching. However, this can fail to detect \emph{negative} components (those that suppress incorrect answers rather than promoting correct ones) because a component that boosts both the correct and incorrect token equally will show no change in raw probability~\mbox{\citep{zhang2023towards, heimersheim2024use}}. \emph{Logit difference}~\mbox{\citep{wang2022interpretability}} measures the change in the gap between the correct and incorrect logits, and is generally recommended because it controls for components that promote both targets and is linear in the residual stream. \emph{KL divergence}~\mbox{\citep{NEURIPS2023_34e1dbe9, zhang2023towards}} compares the full output distribution before and after intervention, capturing changes beyond a single token pair.

\mbox{\citet{meloux2025mechanistic}} show that the corruption method and evaluation metric materially affect the result. On IOI, different method--metric combinations detect different incomplete subsets of attention heads, and no single configuration recovers the full known circuit. \mbox{\citet{miller2024transformer}} further show that faithfulness scores are highly sensitive to seemingly insignificant changes in ablation methodology, concluding that such scores reflect the methodological choices of researchers as much as the actual components of the circuit.

\paragraph{Evaluation granularity.} Most studies evaluate at the \emph{source level}, preserving all outgoing connections from selected components, a weaker test than \emph{edge-level} evaluation, which preserves only the specific connections identified by the discovery algorithm. Source-level evaluation can inflate apparent circuit quality by allowing components to exploit connections that the discovery method did not identify~\mbox{\citep{miller2024transformer}}.

\paragraph{Overlap vs.\ faithfulness.} \mbox{\citet{hanna2024have}} demonstrate that high component \emph{overlap} between two circuits does not imply high \emph{faithfulness}: when overlap is moderate, it does not predict faithfulness, leading them to recommend that ``when comparing circuits, measuring overlap is no substitute for measuring faithfulness.'' Relatedly, \mbox{\citet{goldowsky2023localizing}} note that path patching measures \emph{sufficiency} but not \emph{completeness}: with redundant components, a compact subset can achieve zero unexplained effect while missing contributors.

\paragraph{Aggregate vs.\ per-input performance.} Aggregate faithfulness can mask failures on individual inputs. \mbox{\citet{miller2024transformer}} show substantial within-task variation in circuit performance, and \mbox{\citet{garriga2024adversarial}} find that circuits generalize poorly to adversarial evaluation examples. This observation led \mbox{\citet{sharkey2025open}} to question whether first selecting a human-defined task and then discovering its circuit is an effective approach, since the task definition may not align with the model's internal computational decomposition. Many published circuits do not pass strict sufficiency and necessity tests~\mbox{\citep{shi2024hypothesis, garriga2024adversarial}}. More broadly, seemingly convincing interpretations can prove false (\emph{interpretability illusions}~\mbox{\citep{bolukbasi2021interpretability, makelov2023subspace}}), underscoring the need for rigorous validation.

Standardized benchmarks are beginning to address these concerns: Tracr~\mbox{\citep{lindner2023tracr}} provides synthetic transformers with known ground-truth circuits, and MIB~\mbox{\citep{mueller2025mib}} evaluates on both synthetic and pretrained models. However, concerns remain that results on synthetic benchmarks may not transfer to naturally trained transformers~\mbox{\citep{rai2024practical, sharkey2025open}}.
For the operational definitions used in this study, see Appendix~\ref{app:eval_metrics}.

\subsubsection{Shortcomings and Open Challenges}

Beyond evaluation, circuit discovery faces several structural challenges that suggest recovered circuits may not uniquely identify the underlying mechanism.

\paragraph{Non-uniqueness and instability.} \mbox{\citet{sharkey2025open}} identify non-uniqueness as a key open problem. Multiple lines of evidence support this concern. At the method level, ACDC's output depends on threshold, metric, and the order of edge iteration~\mbox{\citep{NEURIPS2023_34e1dbe9}}, and gradient-based attribution scores exhibit high variance across inputs~\mbox{\citep{meloux2025mechanistic}}. At a more fundamental level, \mbox{\citet{meloux2025everything}} exhaustively enumerate valid explanations in small MLPs trained on Boolean functions and find substantial non-identifiability. Valid explanations include multiple zero-error circuits, multiple interpretations per circuit, and multiple causally aligned algorithms per network. Their number also grows substantially with model size. Empirically, \mbox{\citet{franco2026finding}} show that even within a single task (IOI), different prompt templates induce systematically different circuits that cluster into ``prompt families,'' with the dominant source of variation being model-dependent. \mbox{\citet{haklay2025position}} further show that circuits are position-specific: assuming position invariance, as most methods do, leads to low precision and recall when edge importance is aggregated across positions. Even on a single task (modular addition), small changes to hyperparameters and initialization can induce qualitatively different algorithms (e.g., the ``Clock'' vs.\ ``Pizza'' algorithms~\mbox{\citep{zhong2023clock}}), and models sometimes implement multiple imperfect copies in parallel.

However, the picture is not entirely negative. \mbox{\citet{tigges2024llm}} track circuits across 300 billion tokens of training in the Pythia suite (70M--2.8B parameters) and find that although individual components may change over time, the overarching \emph{algorithm} remains stable and tends to replicate across model scale. \mbox{\citet{chughtai2023toy}} reach a similar conclusion on group composition tasks: networks consistently implement the same representation-theoretic algorithm (weak universality), but the specific representations learned vary across random seeds (against strong universality).

\paragraph{Redundancy and backup behavior.} \mbox{\citet{mcgrath2023hydra}} described the \emph{Hydra effect}. When important attention heads are ablated, backup heads compensate by increasing their contribution, indicating built-in redundancy that complicates the notion of a unique circuit. \mbox{\citet{wang2022interpretability}} observed similar backup behavior in the IOI circuit, where backup name-mover heads use a qualitatively different mechanism from the negative heads they compensate for~\mbox{\citep{mcdougall2024copy}}. \mbox{\citet{ortu2024competition}} showed that factual recall and counterfactual reasoning engage competing mechanisms that dynamically trade off depending on context, and \mbox{\citet{dutta2024think}} documented functionally interchangeable parallel pathways at scale in Llama-2~7B, where multiple circuits simultaneously write the answer token during chain-of-thought reasoning. More broadly, models can implement multiple algorithms in tandem for the same task~\mbox{\citep{zhong2023clock, nanda2023progress}}, and superficially identical behavior can arise from qualitatively different internal processes~\mbox{\citep{mahaut2025repetitions}}. Noising and denoising interventions interact differently with redundant structure. In AND-like serial circuits, noising finds all components while denoising finds only the output. In OR-like parallel circuits, denoising finds all components while noising finds only the output~\mbox{\citep{heimersheim2024use}}.

These empirical observations have theoretical grounding. \mbox{\citet{rohweder2026hierarchical}} show that under hierarchical data generation, gradient descent's implicit bias toward symmetric solutions distributes predictive power across multiple components, guaranteeing that ablation of one can be compensated by others. At the parameter level, \mbox{\citet{bushnaq2024using}} connect this redundancy to degeneracy in the loss landscape, showing that mechanistically distinct parameter configurations can achieve equivalent loss, paralleling the biological concept of \emph{degeneracy}, in which structurally different elements perform the same function~\mbox{\citep{edelman2001degeneracy}}.

\paragraph{Component reuse across tasks.} Despite the non-uniqueness of circuits \emph{within} a task, there is growing evidence that models reuse circuit components \emph{across} tasks. \mbox{\citet{merullo2023circuit}} demonstrate that the IOI circuit and a Colored Objects circuit in GPT-2 Medium share approximately 78\% of their most important attention heads, despite having no linguistic overlap, suggesting algorithmic building blocks that are general across tasks (detect duplication $\to$ inhibit/gather $\to$ copy). \mbox{\citet{lan2024towards}} find that semantically related sequence continuation tasks rely on shared circuit subgraphs in both GPT-2 Small and Llama-2-7B. \mbox{\citet{mondorf2025circuit}} confirm both shared and task-specific substructure across string edit operations, and show that circuits can be composed via set operations to explain more complex behaviors. \mbox{\citet{ferrando2024similarity}} show that the SVA circuit in Gemma~2B is highly consistent across English and Spanish, driven by a language-independent ``subject number'' direction. \mbox{\citet{sun-2025-circuit}} show that a model's ability to apply consistent circuitry across input variants (circuit stability) predicts length, structural, and compositional generalization. \mbox{\citet{nainani2024adaptive}} further show that the IOI circuit generalizes to prompt variants where its hypothesized algorithm should fail, reusing 100\% of nodes with only additional input edges, though they also uncover an artifact of knockout using mean ablation (``S2 Hacking'') that can inflate circuit performance relative to the full model.

\paragraph{Scalability and scope.} Most circuit studies have been conducted on small models (GPT-2 Small, Pythia-70M, toy models), with only limited demonstrations on larger architectures~\mbox{\citep{lieberum2023does}}. The heavy reliance on human interpretation for hypothesis generation and validation poses a scalability bottleneck~\mbox{\citep{rai2024practical}}. Furthermore, the tasks studied have been deliberately simple and amenable to mechanistic analysis; \mbox{\citet{sharkey2025open}} caution that this ``streetlight interpretability'' risks developing methods and insights that do not transfer to more complex, safety-relevant settings.

\paragraph{The missing test.} Prior work has documented that circuits vary with extraction \emph{method}~\mbox{\citep{NEURIPS2023_34e1dbe9, syed2024attribution, bhaskar2024finding}}, with \emph{hyperparameters} such as metric and corruption method~\mbox{\citep{meloux2025mechanistic}}, with \emph{prompt template}~\mbox{\citep{franco2026finding}}, and with \emph{token position}~\mbox{\citep{haklay2025position}}. However, to our knowledge, whether circuits for the same task within a single model vary when the \emph{input distribution} changes (holding the task and method fixed) remains largely unexamined. Token frequency is a natural axis of variation for this question. Token frequencies follow a Zipfian distribution~\mbox{\citep{zipf1949human}}, influence embedding geometry~\mbox{\citep{zhou2021frequency, merullo2025linear}} and few-shot reasoning accuracy~\mbox{\citep{razeghi2022impact}}, and affect task performance even when the task logic is frequency-invariant~\mbox{\citep{niu2025illusion}}. At the neuron level, \mbox{\citet{liu2025distributed}} show that processing rare tokens in Pythia and GPT-2 relies on distributed specialization. The relevant MLP neurons are coordinated but spatially scattered throughout the shared architecture rather than organized into discrete modules. Whether this distributed specialization finding extends to the circuit level, i.e., whether structurally distinct circuits emerge for different frequency bands of the same task, remains an open question.

\FloatBarrier

\FloatBarrier
\section{Data Construction}
\label{app:data_construction}

\begin{figure}[t]
\centering
\resizebox{0.8\linewidth}{!}{%
\begin{tikzpicture}[
  node distance=0.35cm,
  stage/.style={
    rectangle, rounded corners=3pt, draw=black!50, fill=blue!6,
    minimum width=2.6cm, minimum height=0.75cm,
    font=\small, align=center, text=black
  },
  output/.style={
    rectangle, rounded corners=3pt, draw=black!50, fill=orange!10,
    minimum width=2.6cm, minimum height=0.75cm,
    font=\small, align=center, text=black
  },
  arr/.style={-{Stealth[length=1.8mm]}, thick, black!40},
  desc/.style={font=\scriptsize, text=black!55, align=center,
    text width=2.6cm},
]

\node[stage] (pile) {The Pile};
\node[stage, right=0.7cm of pile] (freq) {Corpus\\[-2pt]Frequencies};
\node[stage, right=0.7cm of freq] (cat) {Token\\[-2pt]Categorization};
\node[stage, right=0.7cm of cat] (bands) {Band\\[-2pt]Design};

\draw[arr] (pile) -- (freq);
\draw[arr] (freq) -- (cat);
\draw[arr] (cat) -- (bands);

\node[desc, below=0.1cm of pile]
  {300B tokens\\Pythia tokenization};
\node[desc, below=0.1cm of freq]
  {corpus-wide counting\\$\log_{10}$ per-million};
\node[desc, below=0.1cm of cat]
  {Unicode + BPE prefix\\$\to$ 26,863 \texttt{word\_en}};
\node[desc, below=0.1cm of bands]
  {$k{=}5$: ratio ${<}4{\times}$\\pool ${\geq}500$ tokens};

\node[stage, below=1.6cm of pile] (pools) {Base Token\\[-2pt]Pools};
\node[stage, right=0.7cm of pools] (valid) {External\\[-2pt]Validation};
\node[stage, right=0.7cm of valid] (confound) {Confound\\[-2pt]Profiling};
\node[output, right=0.7cm of confound] (out) {Validated\\[-2pt]Token Pools};

\draw[arr] (pools) -- (valid);
\draw[arr] (valid) -- (confound);
\draw[arr] (confound) -- (out);

\draw[arr, rounded corners=4pt]
  (bands.east) -- ++(0.7, 0) -- ++(0, -1.65) -| (pools.north);

\node[desc, below=0.1cm of pools]
  {\texttt{word\_en} filtered\\into 8 band pools};
\node[desc, below=0.1cm of valid]
  {FineWeb cross-check\\$85$--$100$\% retained};
\node[desc, below=0.1cm of confound]
  {length, case, composition\\identified per band};
\node[desc, below=0.1cm of out]
  {validated \texttt{word\_en} pools\\ready for task-specific control};

\end{tikzpicture}%
}%
\caption{Data construction pipeline. Each stage (blue) transforms the output of
the previous step into increasingly refined token pools. The final output (orange) is a set of validated token pools controlled for frequency, with remaining confounds identified. The pools are ready for controls specific to each task, such as restricting tokens to lowercase and matching character length distributions for LSC.}
\label{fig:data_pipeline}
\end{figure}
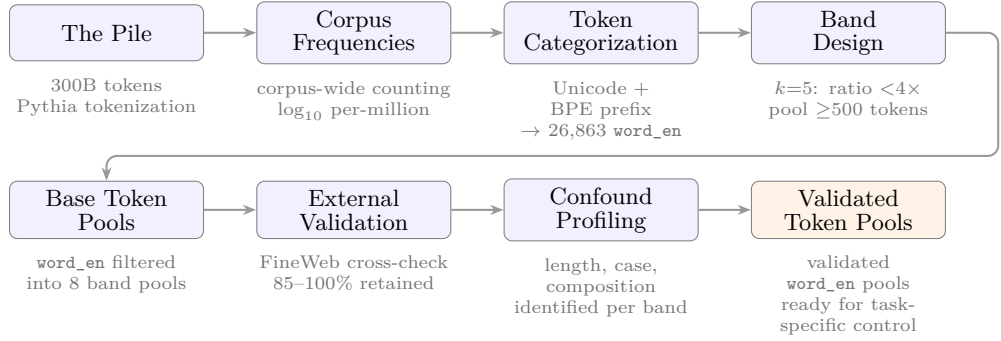

Figure~\ref{fig:data_pipeline} provides an overview of the full data construction
pipeline.  The remainder of this section describes each stage in detail.

\subsection{Corpus and Token Frequency Extraction}
\label{app:corpus}

Our frequency analysis requires exact token-level counts over the Pythia training
corpus.
We use the standard (undeduped) Pythia models~\mbox{\citep{biderman2023pythia}}, which
performed better in preliminary evaluations than the deduped variants.
Accordingly, we use the undeduped Pile in its preshuffled form
(\texttt{EleutherAI/pile-standard-pythia-preshuffled} on HuggingFace),
the exact tokenized corpus used for pretraining.%
\footnote{\url{https://huggingface.co/datasets/EleutherAI/pile-standard-pythia-preshuffled}}
The corpus contains approximately 300\,B tokens and uses the GPT-NeoX-20B
tokenizer~\mbox{\citep{gpt-neox-20b}}, whose vocabulary contains 50,277
entries. Of these, 50,063 token IDs have nonzero counts.

\paragraph{Token identity preservation.}
BPE encodes word boundaries through a leading-space prefix (\texttt{\.{G}}),
making \texttt{\.{G}the} and \texttt{the} distinct entries.
We preserve this distinction by operating on token IDs throughout and using
the \texttt{\.{G}} prefix for word boundary classification
(Appendices~\ref{app:vocab} and~\ref{app:token_pools}).

\subsection{Vocabulary Filtering and Categorization}
\label{app:vocab}

We transform raw counts into log-frequency space and categorize the vocabulary
to isolate a usable word-level token pool.

\paragraph{Frequency transformation.}
Raw counts are converted to
$f_{\log}(t) = \log_{10}(\text{count}(t) / N \times 10^6)$
($N = 300{,}039{,}168{,}000$), where $0$ corresponds to one occurrence per
million tokens. The resulting distribution is smooth and unimodal with no
natural clusters, motivating equal-width bands in log-frequency space
(\autoref{app:band_design}).

\paragraph{Token categorization.}
The raw vocabulary includes punctuation, digits, code fragments, and subword
pieces that cluster at particular frequencies.
To ensure frequency is the primary axis of variation, we classify each token
into eight categories using Unicode character analysis and the BPE
leading-space prefix \texttt{\.{G}}~\mbox{\citep{sennrich2016neural}}
(Table~\ref{tab:categorization_rules}).

\begin{table}[t]
\centering
\small
\begin{tabular}{lp{8.5cm}}
\toprule
Category & Rule \\
\midrule
\textbf{word\_en} & Space prefix + all letters + all Latin script + all ASCII \\
\textbf{subword\_en} & No space prefix + all letters + all Latin + all ASCII \\
word\_other & Space prefix + all letters + non-Latin script or non-ASCII Latin (e.g., accented characters) \\
subword\_other & No space prefix + all letters + non-Latin or non-ASCII Latin \\
numeric & All content characters are digits \\
punctuation & All content characters are punctuation or symbols \\
whitespace & Pure whitespace, control characters, or BPE whitespace markers \\
mixed & Multiple character groups present (e.g., letters + digits) \\
\bottomrule
\end{tabular}
\caption{Token categorization rules. Each token is classified based on its
BPE space prefix and the Unicode properties of its content characters
(after stripping the prefix). \textbf{word\_en} tokens form the primary
experimental pool. The distinction between word\_en and subword\_en, and
between word\_other and subword\_other, rests entirely on the space prefix
that marks word-initial position in BPE. This filter is heuristic and admits
loanwords, proper names, and tokens shared across Latin-script languages
rather than only English words.}
\label{tab:categorization_rules}
\end{table}

\begin{figure}[t]
\centering
\includegraphics[width=\linewidth]{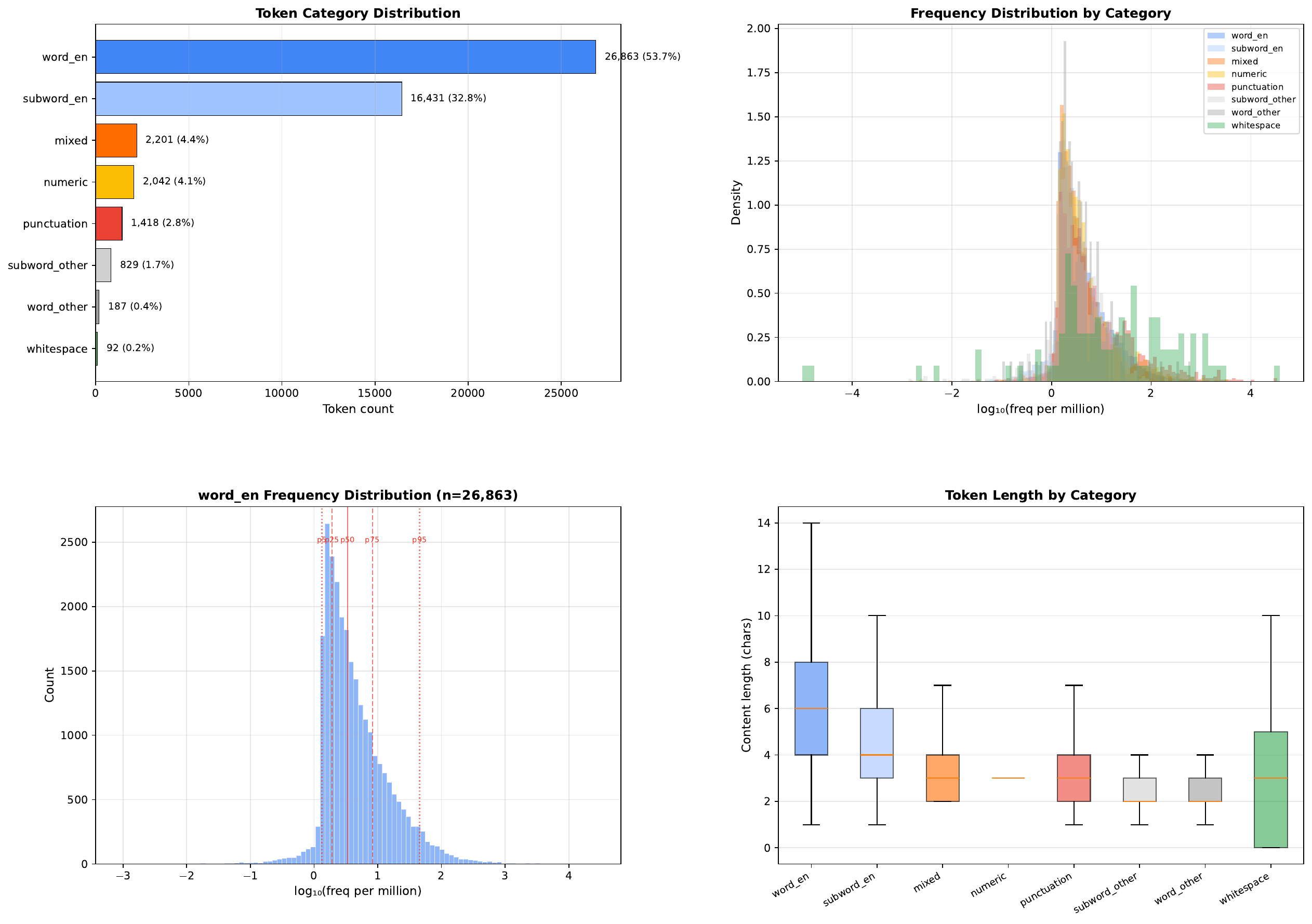}
\caption{Token vocabulary overview.
The top-left panel shows the category distribution, in which word\_en tokens dominate the vocabulary. The top-right panel shows log frequency by category. The bottom-left panel shows the word\_en frequency distribution and percentile markers defining the core range for band construction. The bottom-right panel shows character length distributions by category.}
\label{fig:categories}
\end{figure}

Table~\ref{tab:token_categories} summarizes the resulting category distribution,
and Figure~\ref{fig:categories} shows the frequency and length profiles across categories.

\begin{table}[t]
\centering
\small
\begin{tabular}{lrrcc}
\toprule
Category & Count & \% & Med.\ log-freq & Med.\ length \\
\midrule
\textbf{word\_en}  & 26,863 & 53.7 & 0.53 & 6 \\
subword\_en        & 16,431 & 32.8 & 0.46 & 4 \\
mixed              &  2,201 &  4.4 & 0.45 & 3 \\
numeric            &  2,042 &  4.1 & 0.52 & 3 \\
punctuation        &  1,418 &  2.8 & 0.65 & 3 \\
subword\_other     &    829 &  1.7 & 0.47 & 2 \\
word\_other        &    187 &  0.4 & 0.43 & 2 \\
whitespace         &     92 &  0.2 & 1.22 & 3 \\
\midrule
Total              & 50,063 & 100  &      &   \\
\bottomrule
\end{tabular}
\caption{Token vocabulary categories.
Each of the 50,063 tokens with nonzero training frequency is classified into one of eight
categories based on Unicode character content.
\textbf{word\_en} tokens begin with a space and contain only ASCII Latin characters. They form the majority of the vocabulary and the experimental pool used to construct frequency bands.}
\label{tab:token_categories}
\end{table}

\subsection{Frequency Band Design}
\label{app:band_design}

Although token frequency is a continuous variable, comparing circuits across the
spectrum requires discrete bands.
Since the log-frequency distribution is unimodal and lacks clear natural clusters
(\autoref{app:vocab}), we use equal-width bands in log-frequency space, so each
band spans the same multiplicative frequency ratio.

\paragraph{Core range.}
We define the core experimental range as the 1st to 99th percentile of the word\_en
log-frequency distribution, $[-0.43, 2.27]$. This range spans $2.69$
log units, corresponding to a $493\times$ frequency ratio.
The tails beyond this range contain qualitatively different populations. The bottom~1\%
contains 269 tokens spread across $2.6$ log units, making it too sparse for
controlled experiments, while the top~1\% contains 269 extremely common function words.
Both tails are retained as separate exploratory conditions.

\paragraph{Selecting $k$.}
We sweep $k = 3, \ldots, 8$ and select the largest value that keeps the within-band
frequency ratio below $4\times$ while still leaving a comfortably large token pool
(well above 500 word\_en tokens) in every band.
We select $k=5$. Each band spans $0.54$ log units, corresponding to a
$3.46\times$ within-band ratio, and the smallest core band
(\texttt{very\_high}) contains 821 word\_en tokens.
At $k=4$, the bands are too wide, with a $4.7\times$ ratio. At $k=6$, the
smallest pool shrinks to 513 tokens, reducing sampling diversity without
providing an analytical benefit.

\paragraph{Final scheme: eight conditions.}
The five core bands range from very\_low to very\_high.
Two exploratory tail bands and one control sampled according to token frequency complete the design
(Table~\ref{tab:bands}).

\begin{table}[t]
\centering
\small
\begin{tabular}{llccc}
\toprule
Band & Type & Log-freq range & Center & word\_en tokens \\
\midrule
bottom\_tail & exploratory & $[-3.05, -0.43)$ &        & 269 \\
very\_low    & core        & $[-0.43, 0.11)$  & $-0.16$ & 935 \\
low          & core        & $[0.11, 0.65)$   & $0.38$ & 14,768 \\
medium       & core        & $[0.65, 1.19)$   & $0.92$ & 6,917 \\
high         & core        & $[1.19, 1.73)$   & $1.46$ & 2,884 \\
very\_high   & core        & $[1.73, 2.27]$   & $2.00$ & 821 \\
top\_tail    & exploratory & $(2.27, 4.44]$   &        & 269 \\
control      & baseline    & $[-3.05, 4.44]$  &        & 26,863 \\
\bottomrule
\end{tabular}
\caption{Frequency band definitions.
Core bands use equal-width intervals of $0.54$ log units ($3.46\times$ ratio).
The control band samples from the full word\_en pool weighted by pretraining frequency,
representing the uncontrolled baseline.}
\label{tab:bands}
\end{table}

The center frequencies of the five core bands range from about $0.7$ to 100
occurrences per million, a ${\sim}145\times$ difference. The $493\times$ ratio above
instead compares the lower and upper boundaries of the full core range.

The control condition samples from the entire word\_en pool weighted by pretraining
frequency.
It reflects the natural frequency mix encountered during training and serves as an
uncontrolled baseline against which frequency-specific circuits can be compared.

\subsection{Token Pool Construction and Validation}
\label{app:token_pools}

Each band's word\_en tokens are exported to pool files.
Before use, we validate these pools against an external corpus and profile potential
confounds.

\paragraph{External validation.}
We validate word\_en tokens against FineWeb~\mbox{\citep{penedo2024the}} by comparing
standalone word frequency with substring occurrence. Tokens used mainly as
substrings are treated as likely subword fragments.
Core bands retain 98.4\%--99.7\% of tokens (85.1\%--100\% across all bands),
confirming that the vast majority are genuine English words.

\begin{figure}[t]
\centering
\includegraphics[width=\linewidth]{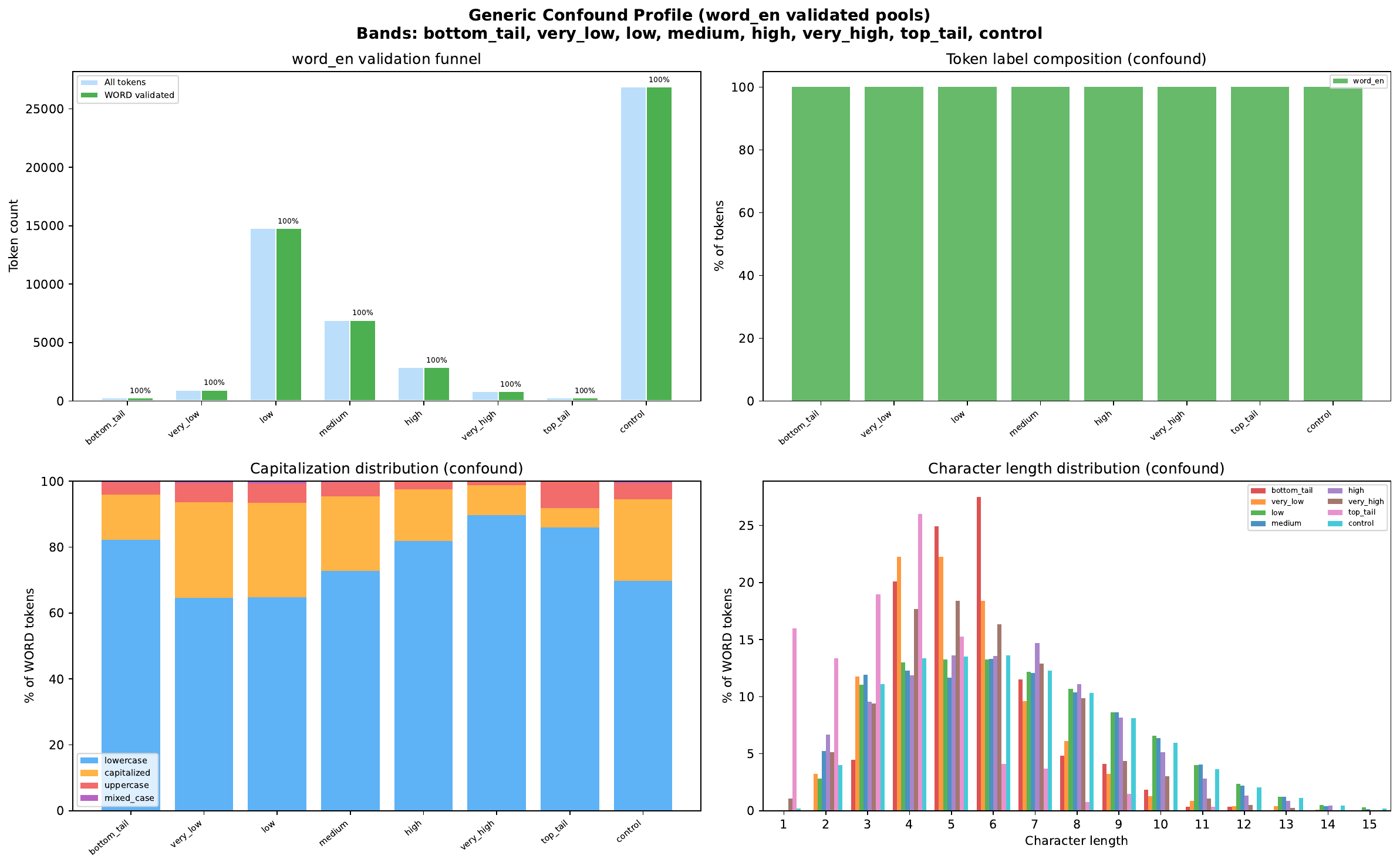}
\caption{Confound profile across frequency bands for word pools that passed validation.
The top-left panel shows FineWeb validation retention by band, and the top-right panel shows word status after validation. The bottom-left and bottom-right panels show capitalization and character length distributions, which motivate restricting LSC tokens to lowercase and matching their length distributions.}
\label{fig:confound_profile}
\end{figure}

\paragraph{Confound profiling.}
We profile three potential confounds in the token pools
(Figure~\ref{fig:confound_profile}). First, lower-frequency tokens tend to be
longer, so we match character length distributions during dataset generation
(\autoref{app:lsc_generation}). Second, we control capitalization by
restricting LSC to lowercase tokens. Third, the word\_en filter controls the
proportion of words and subword fragments.
The validated pools serve as input for all downstream task generators.

\subsection{LSC Task and Dataset Generation}
\label{app:lsc_generation}

\paragraph{Task definition.}
Literal Sequence Copying (LSC), introduced by \mbox{\citet{niu2025illusion}},
tests whether a language model can recognize a repeated token pattern and copy
the token that followed its earlier occurrence.
\mbox{\citet{niu2025illusion}} showed that the performance of induction heads
degrades for rare tokens when tokenizer index is used as a frequency proxy.
We adopt their task and extend it with the controlled frequency bands and
confound controls described in Appendices~\ref{app:band_design}--\ref{app:token_pools}.
Each sequence has the following structure.
\[
    \underbrace{S_1, S_2, S_3, S_4, S_5}_{\text{source}},\;
    T,\;
    \underbrace{R_1, R_2, \ldots, R_{10}}_{\text{distraction}},\;
    \underbrace{S_1, S_2, S_3, S_4, S_5}_{\text{repetition}}
\]
yielding 21 tokens per sequence.
At position~20 (the second occurrence of $S_5$), the model should predict $T$, the
token that followed $S_5$ in the source segment.
The 10-token distraction segment forces the model to use long-range attention
(induction heads) rather than local bigram statistics.
All 16 unique tokens in each sequence ($S_{1\text{--}5}$, $T$, $R_{1\text{--}10}$)
are drawn from the same frequency band, sampled without replacement.

\paragraph{Token pool preparation.}
From the validated pools (\autoref{app:token_pools}), we prepare
two variants of token pools for LSC.
The \textit{matched} variant applies two additional filters. It restricts the
pools to lowercase tokens and matches their character length distributions by
subsampling each band to the length counts available in every participating band.
Four of the five core bands (low through very\_high) satisfy this requirement, yielding
703 tokens per band.
The \texttt{very\_low} band contains only 97 lowercase word\_en tokens after
filtering, too few for reliable character length matching, and is therefore
excluded from the matched pools. It remains available in the unmatched variant
used for the separate analysis without character length control
(\autoref{app:very_low}).
The matched \textit{control} pool is the union of the four matched core bands
(2,812 tokens), sampled with weights based on pretraining frequency to reflect the model's
natural frequency mix.
The \textit{unmatched} variant applies only the lowercase filter to the full validated
pools, retaining all eight bands (pool sizes range from 97 for very\_low to 15,269 for
control) and serves as a robustness check.

\paragraph{Dataset generation.}
For each band, we generate 1,500 sequences split 70/15/15 into train (1,050),
validation (225), and test (225) sets.
We produce three independent draws, the minimum needed to measure circuit extraction
stability, using deterministic seeds to ensure exact reproducibility.
Comparing results across draws quantifies within-band sampling noise and tests whether
recovered circuits are robust to the specific tokens selected from each pool.
The matched variant generates datasets for five conditions, comprising the four
matched core bands and the control. The unmatched variant generates seven
conditions, comprising all five core bands, \texttt{top\_tail}, and the control.
The \texttt{bottom\_tail} condition is excluded because its pool contains only
11 tokens.

\paragraph{BOS handling.}
The raw dataset stores 21-token sequences without a beginning-of-sequence token.
BOS is prepended at inference time by the modeling framework (TransformerLens
\mbox{\citealp{nanda2022transformerlens}} uses \texttt{prepend\_bos=True}), shifting all
position indices by $+1$.
With BOS, the prediction target moves from position~20 to position~21.

\subsection{SVA Task and Dataset Generation}
\label{app:sva_generation}

\paragraph{Task definition.}
The second task (\autoref{sec:sva_replication}) is subject-verb agreement
across an attractor. Each prompt follows the template ``The \textit{subject} near
the \textit{attractor}'' and contains six tokens, including the BOS token prepended
by the model. The subject and attractor have opposite grammatical number, and both
are sampled from noun pools defined by frequency band. We score the model by comparing the logits
for \code{~is} and \code{~are}. Because these two answer tokens are fixed
high-frequency function words, frequency varies only through the input
nouns. The corruption swaps the subject's number, which flips the correct
answer while leaving the template, the attractor, and every token position
unchanged. This subject number corruption parallels the LSC corruption because
it removes the signal needed for the target prediction while preserving the
rest of the prompt.
For example, one clean prompt from the low band is \code{The curtains near the
scaffold}, where the correct answer is \code{~are}. Its corrupted version
is \code{The curtain near the scaffold}, where the correct answer is
\code{~is}. We also record top-1 and top-5 accuracy as secondary metrics.

\paragraph{Noun pair pools.}
We start from 3,060 lowercase \texttt{word\_en} tokens
(\autoref{app:vocab}) labeled \texttt{NOUN\_COMMON} by the project's POS
classifier. The classifier uses this label for common nouns, distinguishing
them from stems and other parts of speech. Of these, 848 have single-token
plurals, and 699 carry that label for both the singular and the plural
form.
A later token ID audit found that three pairs (\textit{brow}, \textit{disc},
and \textit{cap}) should have been excluded because the classifier labels
their singular forms as stems, leaving 696 valid pairs. \textit{Cap} was never
sampled into a prompt, so the verifier's documented exception concerns only
\textit{brow} and \textit{disc}. Their occurrence in the data and the
corresponding sensitivity analyses are documented in
\autoref{app:sva_replication}. Four further validated
pairs fall in the upper tail outside the study's five bands and are
unused (hence Table~\ref{tab:sva_pools} lists 695 of the 699).
Bands are assigned from the singular form's log-frequency using the
canonical boundaries (\autoref{app:band_design}). Plural forms
typically fall about one band lower, which we report as a limitation.

\begin{table}[t]
\centering
\caption{SVA noun pools. Matched pools have identical character length distributions, verified end to end. The very-high pool cannot be matched by length and is marked unmatched. The very-low pool has no eligible pairs.}
\label{tab:sva_pools}
\small
\begin{tabular}{lccl}
\toprule
Band & Validated pairs & Pool used & Matched by length \\
\midrule
very-low & 0 & --- & infeasible \\
low & 133 & 117 & yes \\
medium & 307 & 117 & yes \\
high & 209 & 117 & yes \\
very-high & 46 & 46 & no (flagged) \\
control & union of all four pools & 397 & subject weighted by token frequency \\
\bottomrule
\end{tabular}
\end{table}

\paragraph{Conditions and generation.}
We subsample the low, medium, and high pools so that they have identical
character length distributions, leaving 117~pairs per band
(Table~\ref{tab:sva_pools}). The control condition uses the union of these
three pools and the unmatched very-high pool. This union contains 397~pairs.
Subjects are sampled with weights proportional to token frequency, whereas
attractors are sampled uniformly. Consequently, very-high pairs supply
49--53\% of control subjects across draws.
This differs from the LSC control, which weights all sampled content tokens
by frequency. The primary analysis excludes the control and very-high
conditions because it is restricted to bands matched by length. The control
sampling scheme therefore affects only the sensitivity analysis and the
analysis of shared cores across all five conditions.
For each condition and draw, we generate 1,875 candidate examples and split
them 70/15/15 into 1{,}312 training, 281 validation, and 282 test examples.
We score each prompt by its negative log-likelihood under Pythia-1b and
independently retain the 1{,}050/225/225 most probable prompts in each split,
corresponding to a uniform 20\% trim. Because Pythia-1b is one of the evaluated
models, using it for this screen is a limitation.
We generate three independent draws and record the seed for each draw and
condition in every split file. The frozen datasets and their recorded seeds
define the data used in the analyses.
Four of the five models meet the base model criterion specified in advance at
least 0.90 two-way accuracy in every condition and a spread of at most 0.05.
Pythia-160m has a spread of 0.0548, although every condition reaches at least
0.915 and the spread across matched bands is 0.031. We therefore retain the
model and report this deviation from the criterion.

\FloatBarrier
\section{Circuit Discovery Details}
\label{app:circuit_details}

This appendix explains the details of the three phases of the primary LSC
circuit discovery pipeline summarized in Section~\ref{sec:pipeline}. The
pipeline produces 75~circuits across five Pythia models, five conditions
(four frequency bands plus control), and three draws per condition. The SVA
replication reuses the pipeline with its own data and the configuration changes
described in \autoref{app:sva_discovery_config}, while the \texttt{very\_low}
extension uses the unmatched pool without changing the pipeline
(\autoref{app:very_low}).
\begin{enumerate}
  \item \textbf{Threshold selection} (\autoref{app:threshold_selection}):
    a Pareto sweep on the control band identifies candidate ACDC thresholds,
    from which one threshold~$\tau^*$ is selected per model.
  \item \textbf{Circuit extraction} (\autoref{app:circuit_extraction}):
    ACDC runs at~$\tau^*$ across all five conditions and three draws per
    model, with evaluation on the held-out test split.
  \item \textbf{Post-hoc validation} (\autoref{app:random_baselines}):
    random baselines test whether the discovered circuits outperform
    matched arbitrary edge subsets.
\end{enumerate}

\subsection{Corruption Procedure}
\label{app:corruption}

Each clean sequence has the form
$[\textsc{bos}]~S_1 \ldots S_5~T~R_1 \ldots R_{10}~S_1 \ldots S_5$
(22~tokens including \textsc{bos}).
The corrupted sequence is identical through position~16 and replaces
positions~17--21 (the repeated prefix that triggers induction) with five tokens
sampled without replacement from the pool for the same band, excluding tokens in the
clean sequence.
Position~17 is therefore the divergence index, the first token position at
which the clean and corrupted sequences differ.
Each example is paired with an incorrect answer token from the band pool, as
required by the AutoCircuit \texttt{PromptDataset} API.

\subsection{ACDC Configuration}
\label{app:acdc_config}

\paragraph{Pythia's parallel architecture.}
Pythia models use the GPTNeoX architecture~\mbox{\citep{biderman2023pythia}}, which
employs \emph{parallel} transformer blocks. Attention and the MLP receive the
same layer-normalized input, and their outputs are added to the residual
stream as follows.
\[
\mathbf{x}_{l+1} = \mathbf{x}_l + \mathrm{Attn}(\mathrm{LN}(\mathbf{x}_l))
+ \mathrm{MLP}(\mathrm{LN}(\mathbf{x}_l)).
\]
Attention and the MLP within the same layer therefore cannot interact directly.
The computational graph used for circuit discovery must respect this constraint.

\paragraph{Algorithm and library choice.}
\label{app:algorithm_choice}
In preliminary experiments, EAP~\mbox{\citep{syed2024attribution}} and
EAP-IG~\mbox{\citep{hanna2024have}} produced unstable circuits on our LSC task,
consistent with the high intrinsic variance reported by
\mbox{\citet{meloux2025mechanistic}}.
ACDC~\mbox{\citep{NEURIPS2023_34e1dbe9}} has substantially lower variance and
allows us to control the random seed, making it appropriate for comparisons
across conditions. However, rerunning discovery with identical seeds in our
environment can still return slightly different masks. This variation,
documented for the SVA circuits in \autoref{app:sva_replication}, is part of
the extraction noise measured by the design with three draws.
We nevertheless include a cross-method analysis as a robustness check
(Appendix~\ref{app:method_comparison}).
We run ACDC through AutoCircuit~\mbox{\citep{miller2024transformer}} rather than the
original codebase, which produced metric collapse on Pythia models.
AutoCircuit correctly handles parallel blocks via the
\texttt{parallel\_attn\_mlp} flag and also makes threshold sweeps practical
through faster edge patching (\autoref{app:threshold_selection}).

\paragraph{Model loading and ACDC settings.}
Models are loaded through TransformerLens with \texttt{fold\_ln=True},
\texttt{center\_writing\_weights=True}, and \texttt{center\_unembed=True}
(folding layer norm into adjacent weights), with hook flags
\texttt{use\_attn\_result}, \texttt{use\_attn\_in}, and
\texttt{use\_hook\_mlp\_in} enabled.
ACDC uses factorized tree patching with resample ablation, joint QKV
treatment, and output slicing at the final sequence position. Edges outside
the circuit carry activations resampled from the corrupted distribution
(\autoref{app:corruption}). All computations use FP32; BF16 produced numerical discrepancies on Pythia.

\paragraph{SVA configuration.}
\label{app:sva_discovery_config}
The SVA replication uses the same threshold grid,
\texttt{minimal\_acceptable} selection rule, and edge-level tree patching
with resample ablation. The selected thresholds are $10^{-2}$ for 70m,
$1.58\times10^{-3}$ for 160m, 410m, and 1b, and $6.31\times10^{-4}$ for
1.4b. ACDC uses an effective first batch of 128 training examples for every
SVA circuit. The divergence index records the first token position at which
the clean and corrupted prompts differ. It is 2 for SVA because the corruption
replaces the subject, rather than 17 as in LSC, where the corruption replaces
the repeated prefix.

\subsection{Threshold Selection via Pareto Analysis}
\label{app:threshold_selection}

\paragraph{Motivation.}
ACDC's pruning threshold~$\tau$ controls the sparsity/faithfulness trade-off.
Because no reference circuit for LSC on Pythia exists, we select~$\tau$ using
circuit size, KL divergence from the base model, task performance, and
performance after removing the circuit. These criteria measure minimality,
faithfulness, sufficiency, and necessity, respectively.

\paragraph{Sweep design.}
On draw~1 of the \emph{control} band, we sweep 11 thresholds evenly spaced on
a logarithmic scale from $10^{-2}$ to $10^{-6}$.
For each threshold, ACDC samples 256~examples from the training
split (seed~42; the implementation consumes a single batch, giving the
per-model effective counts listed in \autoref{app:circuit_extraction}) and
the resulting circuit is evaluated on the full validation
split (225~examples, seed~123).
We use the control band because it samples tokens according to frequency from all
four core bands (\autoref{app:lsc_generation}).

\paragraph{Pareto frontier.}
For each model, we compute the Pareto frontier over edge fraction and
$\mathrm{KL}(p_{\text{base}} \| p_{\text{circuit}})$. A point lies on the
frontier when no alternative has both a smaller edge fraction and lower KL
divergence.

\begin{figure}[t]
\centering
\includegraphics[width=\linewidth, height=0.22\textheight, keepaspectratio]{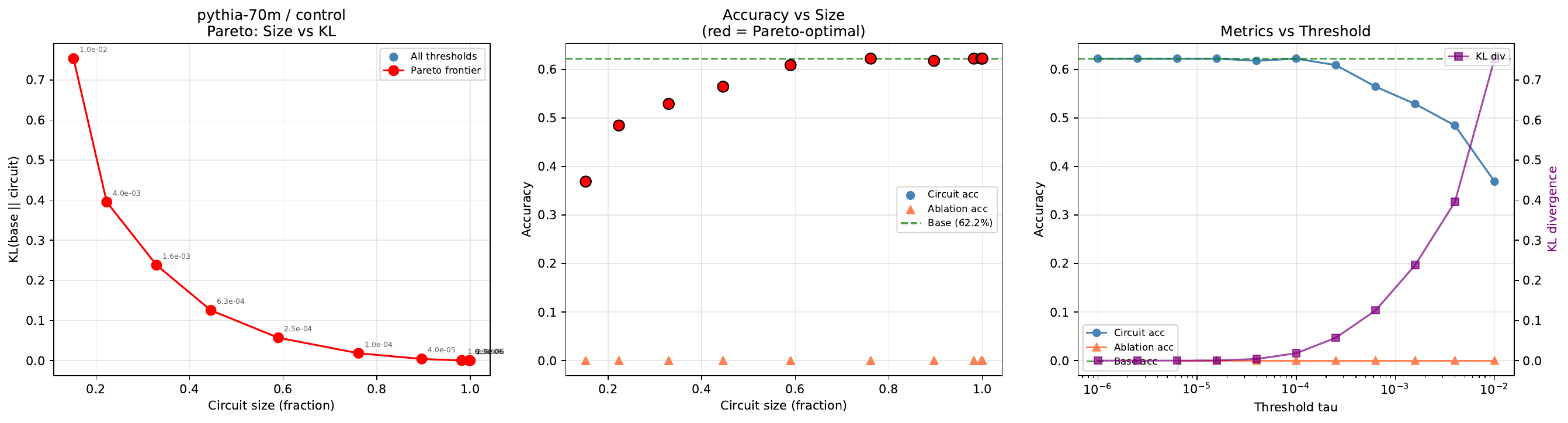}\\[4pt]
\includegraphics[width=\linewidth, height=0.22\textheight, keepaspectratio]{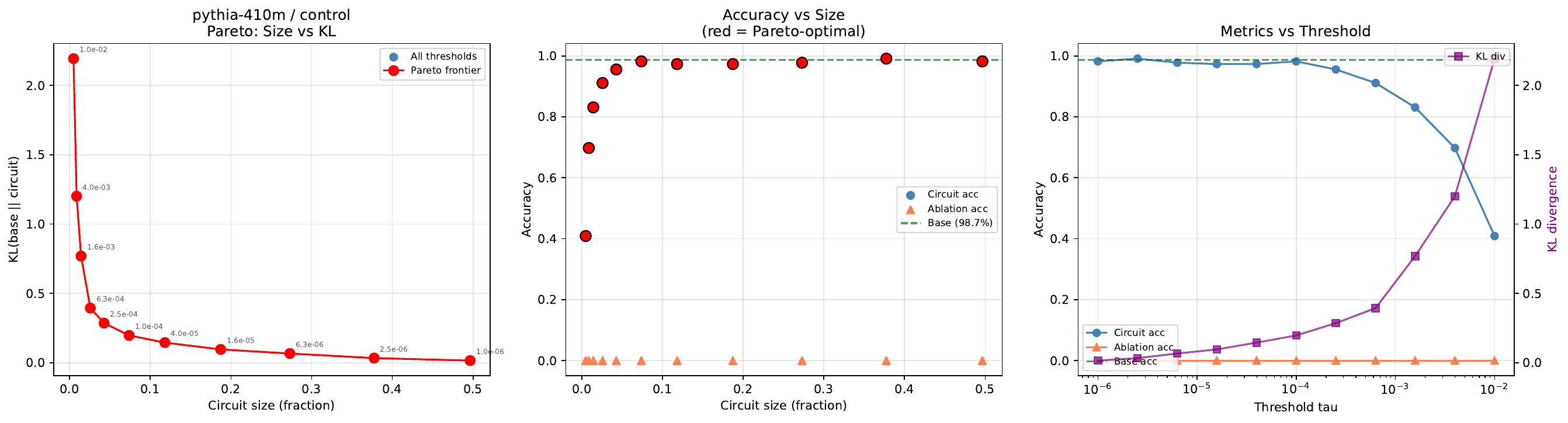}
\caption{Pareto sweep results for two representative Pythia models on the control band. The top panels show Pythia-70m, and the bottom panels show Pythia-410m. Results for 160m, 1b, and 1.4b are qualitatively similar.
The left panels show the Pareto frontier over circuit size and $\mathrm{KL}(p_{\text{base}} \| p_{\text{circuit}})$, with threshold values annotated. The center panels show circuit and ablation accuracy against circuit size, with Pareto-optimal points in red. The right panels show accuracy and KL against threshold~$\tau$.
Pythia-70m has substantially lower base accuracy; see the \textit{Pythia-70m} note below
Table~\ref{tab:threshold_selection}.}
\label{fig:pareto_sweeps}
\end{figure}

\paragraph{Semi-automated threshold selection.}
Automated heuristics filter Pareto-optimal points by minimum quality criteria
(${\geq}80\%$ accuracy retention, $\mathrm{KL} < 0.5$) and an expected
size range for each model.
A human reviewer then selects one threshold~$\tau^*$ per model from the
filtered candidates (Figure~\ref{fig:pareto_sweeps};
Table~\ref{tab:threshold_selection}).

\begin{table}[ht]
\centering
\caption{Selected ACDC thresholds per model. All points lie on the Pareto
frontier and achieve 0\% ablation accuracy.
Retention is circuit accuracy divided by base model accuracy on the control
band validation split.}
\label{tab:threshold_selection}
\small
\begin{tabular}{lccccccc}
\toprule
Model & $\tau^*$ & Edge~\% & $n_{\text{edges}}$ & KL & Circ.\ acc.\ (\%) & Retention (\%) & Abl.\ acc.\ (\%) \\
\midrule
Pythia-70m  & $1.58 \times 10^{-3}$ & 32.9 & 436   & 0.238 & 52.9 & 85.0 & 0.0 \\
Pythia-160m & $6.31 \times 10^{-4}$ & 12.2 & 1,396 & 0.284 & 93.3 & 96.3 & 0.0 \\
Pythia-410m & $2.51 \times 10^{-4}$ & 4.3  & 3,444 & 0.285 & 95.6 & 96.8 & 0.0 \\
Pythia-1b   & $1.58 \times 10^{-3}$ & 9.4  & 939   & 0.485 & 92.0 & 94.1 & 0.0 \\
Pythia-1.4b & $6.31 \times 10^{-4}$ & 2.6  & 2{,}097 & 0.499 & 92.4 & 94.1 & 0.0 \\
\bottomrule
\end{tabular}
\end{table}

\paragraph{Pythia-70m.}
Pythia-70m is a boundary case because its base accuracy is only 62.2\%,
compared with at least 95\% for the larger models. Its selected circuit is
also denser, retaining 32.9\% of edges and 85.0\% of base model accuracy.
Differences from larger models may reflect capacity limitations rather than
qualitative differences in circuit organization.

\subsection{Circuit Extraction}
\label{app:circuit_extraction}

Using the model-specific thresholds from
Table~\ref{tab:threshold_selection}, we run ACDC across all five conditions
(four frequency bands plus control) and three independent draws for each of
the five models, yielding 75~circuits in total.
Each circuit samples 256~examples from the condition-specific training split
(seed~42) and is evaluated on the test split (225~examples, seed~123). The
validation split is reserved for threshold selection, preventing information
leakage (\autoref{app:threshold_selection}).
The AutoCircuit ACDC implementation consumes a single training batch, so
the effective number of examples seen during discovery equals the batch
size. These counts are 256, 256, 128, 96, and 64 for Pythia-70m, 160m, 410m,
1b, and 1.4b, respectively.
Within each model, every condition and draw uses the same effective count,
so comparisons across conditions are balanced. Effective training set size
nevertheless differs across model scales.
The SVA replication, whose loader reduces its nominal batch differently,
has a uniform effective count of 128 for every model and circuit
(\autoref{app:sva_generation}).
Each circuit is also evaluated on the test splits of the other four bands,
yielding a $5 \times 5$ cross-band transfer matrix for each model and draw
(Section~\ref{sec:circuit_size}).

\subsection{Evaluation Metrics}
\label{app:eval_metrics}

We assess circuit quality along four dimensions
\mbox{\citep{wang2022interpretability}}:

\paragraph{Faithfulness.}
We measure faithfulness as
$\mathrm{KL}(p_{\text{base}} \| p_{\text{circuit}})$ at the final sequence
position~\mbox{\citep{NEURIPS2023_34e1dbe9}}.

\paragraph{Sufficiency.}
We measure sufficiency using circuit accuracy (top-1, top-5, and top-10) and
mean correct token probability while applying resample ablation to edges
outside the circuit.
Resample ablation provides an upper bound on standalone performance~\mbox{\citep{yu2024functional}}.

\paragraph{Necessity (completeness).}
Ablation accuracy measures task performance when only the circuit's complement
is retained. A necessary circuit yields ablation accuracy close to zero.

\paragraph{Minimality.}
Edge fraction $n_{\text{edges}} / n_{\text{total}}$. Base model metrics are
precomputed with HuggingFace Transformers and cached across phases.

\subsection{Random Baselines}
\label{app:random_baselines}

Each circuit is compared against $K = 100$ random edge sets of identical size,
reporting the $z$-score and percentile rank.
All 75~circuits achieve a percentile rank of 100\%.

\subsection{Implementation Details}
\label{app:implementation}

All experiments use fixed random seeds (seed~42 for training, seed~123 for
evaluation) and deterministic CUDA settings on NVIDIA A100 80GB GPUs.
Batch sizes are adapted to GPU memory (256 for 70m/160m, 128 for 410m, 96 for
1b, 64 for 1.4b). The full pipeline requires approximately 736~GPU-hours,
including about 289 for threshold sweeps and 447 for circuit discovery.
Individual ACDC runs range from about 80~s for Pythia-70m to 3.5~h for
Pythia-1b, and jobs are distributed across four GPUs.

\subsection{Threshold Robustness}
\label{app:threshold_robustness}

To test sensitivity to the selected $\tau^*$, we evaluate the circuit from the control condition
at the immediately lower threshold, $\tau^*$, and the immediately
higher threshold. These settings span a factor of 2--3 in circuit size.
For each threshold, the circuit is evaluated on the test splits of all
five conditions, and cross-band transfer efficiency is computed as
the ratio of mean accuracy on non-control conditions to accuracy on the control condition
(Table~\ref{tab:threshold_robustness}).

\begin{table}[ht]
\centering
\caption{Cross-band transfer efficiency of circuits extracted from the control condition at three
threshold levels. $\tau_{\text{low}}$ retains more edges than $\tau^*$, whereas $\tau_{\text{high}}$ retains fewer.
Transfer efficiency here is the ratio of mean accuracy on non-control conditions to accuracy on the control condition
circuit \emph{accuracy} (distinct from the \emph{boost} ratio for
band-specific edges of Eq.~\ref{eq:transfer_efficiency}).}
\label{tab:threshold_robustness}
\small
\begin{tabular}{lcccccc}
\toprule
 & \multicolumn{2}{c}{$\tau_{\text{low}}$} & \multicolumn{2}{c}{$\tau^*$} & \multicolumn{2}{c}{$\tau_{\text{high}}$} \\
\cmidrule(lr){2-3}\cmidrule(lr){4-5}\cmidrule(lr){6-7}
Model & Edges & Transf.~Eff. & Edges & Transf.~Eff. & Edges & Transf.~Eff. \\
\midrule
Pythia-70m  &  590 & 0.770 &  436 & 0.755 &  295 & 0.731 \\
Pythia-160m & 2138 & 0.996 & 1396 & 0.993 &  904 & 0.965 \\
Pythia-410m & 5949 & 0.993 & 3444 & 0.984 & 2073 & 0.968 \\
Pythia-1b   & 1425 & 0.967 &  939 & 0.954 &  551 & 0.919 \\
Pythia-1.4b & 3770 & 0.913 & 2097 & 0.935 & 1197 & 0.880 \\
\bottomrule
\end{tabular}
\end{table}

Transfer efficiency varies by at most 2.2~percentage points (looser) and 5.5~percentage points (stricter)
relative to $\tau^*$ across all models.
Pythia-70m shows lower absolute efficiency (73--77\%), consistent with its
capacity limitations (Section~\ref{sec:threshold_relaxation}). All other
models remain above 88\% even at $\tau_{\text{high}}$, indicating that the
phantom specialization conclusions are robust across the tested thresholds.

\subsection{Cross-Method Comparison: ACDC, EAP, and EAP-IG}
\label{app:method_comparison}

As an independent check on whether ACDC's iterative search is responsible for
the phantom specialization pattern, we compute EAP-IG~\mbox{\citep{hanna2024have}}
edge importance scores for each of the 75 extraction conditions
(5~models $\times$ 5~conditions $\times$ 3~draws), using integrated gradients
over ten interpolation steps.
For each condition, we rank edges by their continuous scores and retain the
edges with the highest scores to produce ten edge sets ranging in size from $0.1\times$
to $5.0\times$ the corresponding ACDC edge count (7{,}500 evaluations total).

\paragraph{Methodological caveat.}
Top-$k$ thresholding selects individually important edges but does not
guarantee a coherent subgraph. At the ACDC circuit size ($1.0\times$),
EAP-IG circuits achieve only $2$--$15\%$ of base model accuracy for the four
larger models, compared with $79$--$99\%$ for ACDC
(Table~\ref{tab:method_comparison} in the main text). We therefore compare methods via \emph{same-band advantage}
(same-band minus cross-band accuracy) rather than transfer efficiency ratios.
Even at faithful sizes EAP-IG needs $1.5$--$5\times$ more edges than ACDC, so
this is a cross-method check rather than a method for producing small circuits.

\begin{figure}[t]
\centering
\includegraphics[width=\linewidth]{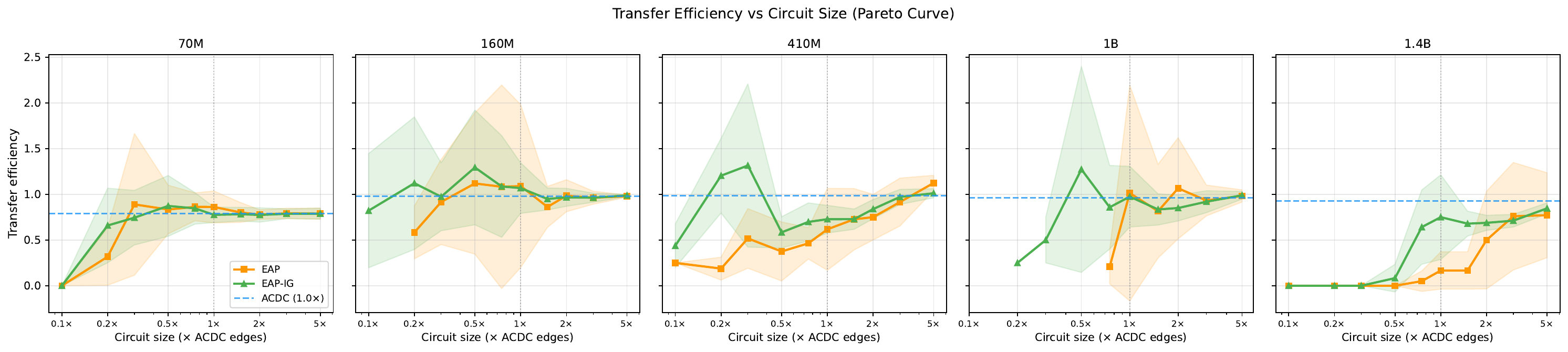}
\caption{Same-band and cross-band accuracy of EAP-IG circuits as a function
of circuit size (as a multiple of the ACDC edge count), averaged over draws
and circuit bands. Each panel shows one model. The vertical dotted line marks the point with the same number of edges as ACDC ($1.0\times$). The two curves overlap closely
at all sizes, showing that the absence of a same-band transfer advantage is stable across circuit sizes. Note that EAP-IG circuits require
substantially more edges than ACDC to achieve comparable faithfulness
(see text).}
\label{fig:pareto_eap}
\end{figure}

\paragraph{No same-band advantage at any circuit size.}
Across all five models and ten size multipliers, the same-band advantage
never exceeds 4.6~percentage points, with inconsistent sign across models
(Table~\ref{tab:method_comparison} in the main text).
Figure~\ref{fig:pareto_eap} traces same-band and cross-band accuracy
across the full size sweep for every model. The
marked points show the smallest sizes that meet the faithfulness criterion.
The small same-band advantages for ACDC at $1.0\times$ ($\Delta = +0.02$ to
$+0.03$ across models) are within measurement noise and consistent with the
absence of band-specific specialization. This pattern holds from $1\%$ to
$100\%$ of edges, ruling out threshold choice artifacts.

\paragraph{Comparison of points that meet the faithfulness criterion.}
At the smallest size where EAP-IG achieves ${\geq}50\%$ of base accuracy,
same-band advantage is $+0.8$ to $+4.6$~percentage points across all five models, consistent
with ACDC. Larger models require $3.0$--$5.0\times$ edges to reach this
threshold, at which point $14$--$27\%$ of all edges are included.

\paragraph{Structural overlap with ACDC.}
Jaccard overlap ranges from 0.28 to 0.60 at the $1.0\times$ point,
decreasing with model size (Spearman $r = -0.90$, $p = 0.037$ over the five per-model means).
The fact that two methods select largely different edges but find no consistent
band specificity provides further evidence of a many-to-one relationship
between circuit structure and function.

\begin{figure}[t]
\centering
\includegraphics[width=0.5\linewidth]{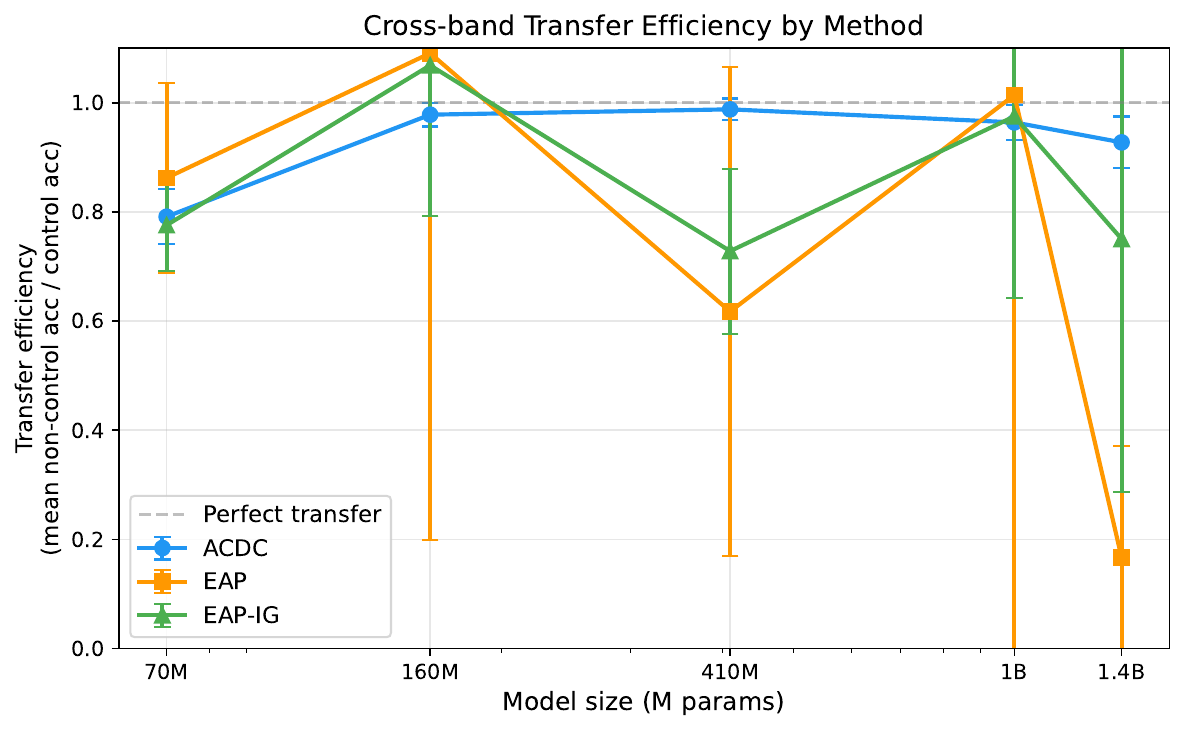}
\caption{Transfer efficiency across all three circuit discovery methods (ACDC, EAP, EAP-IG) at the ACDC circuit size ($1.0\times$). Despite selecting largely different edges (Jaccard 0.25--0.60), none shows a reliable same-band functional transfer advantage, though plain EAP is unstable on larger models.}
\label{fig:te_by_method}
\end{figure}

\paragraph{Plain EAP comparison.}
We also evaluate plain EAP~\mbox{\citep{syed2024attribution}} (single gradient at zero mask value, without integrated gradients) under the same protocol.
Table~\ref{tab:cross_method_summary} and Figure~\ref{fig:te_by_method} compare transfer efficiency and Jaccard overlap across all three methods.
Plain EAP shows erratic transfer efficiency on larger models (61.7\% for Pythia-410m, 16.7\% for Pythia-1.4b), reflecting the instability of estimates based on one gradient in deeper networks.
EAP-IG (with integrated gradients over ten steps) is substantially more stable, with transfer efficiency ranging from 72.8\% to 106.8\% for models ${\geq}160$M.
Despite these differences in circuit quality, neither gradient-based method shows a consistent same-band advantage at any circuit size, consistent with the ACDC result.

\begin{table}[t]
\centering
\caption{Cross-method comparison of circuit discovery algorithms on LSC.
Transfer efficiency here is cross-band circuit accuracy divided by same-band circuit accuracy, averaged over all circuit bands and draws at the ACDC circuit size ($1.0\times$). This accuracy ratio differs from the boost ratio for band-specific edges in Eq.~\ref{eq:transfer_efficiency}.
Jaccard is the edge overlap between EAP/EAP-IG and ACDC circuits.}
\label{tab:cross_method_summary}
\small
\begin{tabular}{llcc}
\toprule
Method & Model & Transfer Eff.\,(\%) & Jaccard vs.\,ACDC \\
\midrule
ACDC & 70m & 79.1 & --- \\
ACDC & 160m & 97.8 & --- \\
ACDC & 410m & 98.8 & --- \\
ACDC & 1b & 96.4 & --- \\
ACDC & 1.4b & 92.7 & --- \\
\midrule
EAP & 70m & 86.2 & 0.594 \\
EAP & 160m & 109.1 & 0.379 \\
EAP & 410m & 61.7 & 0.307 \\
EAP & 1b & 101.4 & 0.315 \\
EAP & 1.4b & 16.7 & 0.252 \\
\midrule
EAP-IG & 70m & 77.6 & 0.598 \\
EAP-IG & 160m & 106.8 & 0.416 \\
EAP-IG & 410m & 72.8 & 0.323 \\
EAP-IG & 1b & 97.5 & 0.361 \\
EAP-IG & 1.4b & 75.0 & 0.281 \\
\bottomrule
\end{tabular}
\end{table}

The causal interventions in Section~\ref{sec:causal_confirmation} test band
interchangeability without relying on circuit discovery.

\FloatBarrier\section{Additional Functional Results}
\label{app:functional}

This appendix reports the full functional evaluation of all 75~LSC
circuits (5~models $\times$ 5~conditions $\times$ 3~draws). Functional
results for the SVA replication are reported in
\autoref{sec:sva_replication}.

\subsection{Base Model Performance Across Frequency Bands}
\label{app:base_performance_detail}

\begin{figure}[t]
\centering
\includegraphics[width=\linewidth]{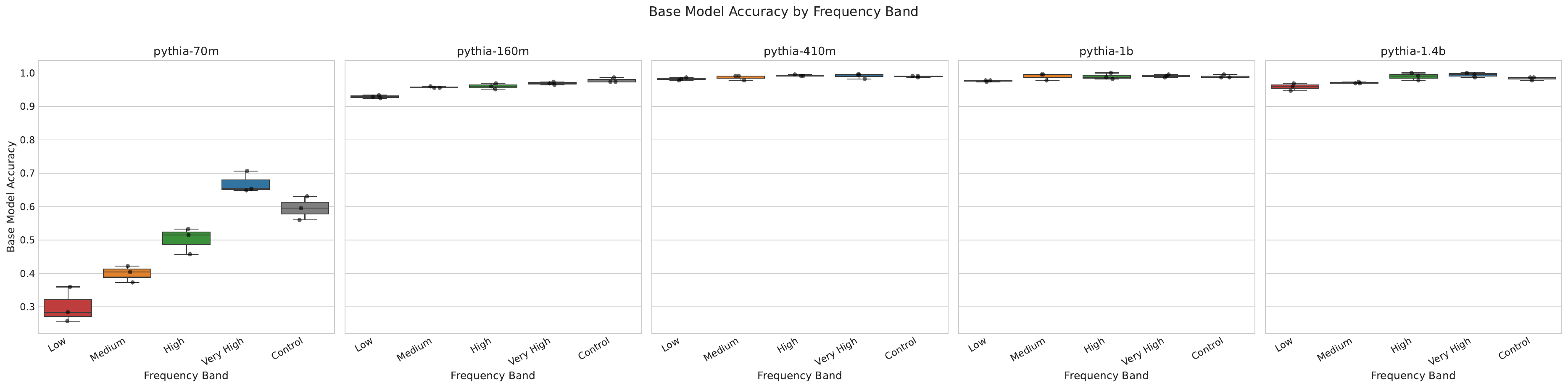}
\caption{Base model top-1 accuracy across frequency bands.
Pythia-70m's accuracy increases with token frequency. The larger models achieve accuracy close to the ceiling.}
\label{fig:base_model_accuracy}
\end{figure}

Pythia-70m's top-1 accuracy increases with token frequency (30--67\%),
whereas all larger models achieve ${\geq}$92.9\%, with top-5 accuracy close to the ceiling
(Figure~\ref{fig:base_model_accuracy}, Table~\ref{tab:base_model_stats}).

\begin{table}[t]
\centering
\caption{Base model performance on LSC by model and frequency band.
Values are top-1 accuracy (\%) $\pm$ std over three draws.
Top-5 accuracy is ${\geq}99\%$ for all models ${\geq}160$M and is omitted.}
\label{tab:base_model_stats}
\small
\setlength{\tabcolsep}{4pt}
\begin{tabular}{l cc cc cc cc cc}
\toprule
 & \multicolumn{2}{c}{Pythia-70m} & \multicolumn{2}{c}{Pythia-160m} & \multicolumn{2}{c}{Pythia-410m} & \multicolumn{2}{c}{Pythia-1b} & \multicolumn{2}{c}{Pythia-1.4b} \\
\cmidrule(lr){2-3}\cmidrule(lr){4-5}\cmidrule(lr){6-7}\cmidrule(lr){8-9}\cmidrule(lr){10-11}
Band & Acc & $\pm$ & Acc & $\pm$ & Acc & $\pm$ & Acc & $\pm$ & Acc & $\pm$ \\
\midrule
low        & 30.1 & 5.3 & 92.9 & 0.4 & 98.2 & 0.4 & 97.6 & 0.3 & 95.9 & 1.1 \\
medium     & 40.0 & 2.5 & 95.7 & 0.3 & 98.7 & 0.8 & 99.0 & 1.0 & 97.0 & 0.3 \\
high       & 50.2 & 4.0 & 96.0 & 0.9 & 99.3 & 0.3 & 99.0 & 0.9 & 99.0 & 1.1 \\
very\_high & 67.0 & 3.2 & 96.9 & 0.4 & 99.1 & 0.8 & 99.1 & 0.4 & 99.4 & 0.7 \\
control    & 59.6 & 3.6 & 97.8 & 0.8 & 99.0 & 0.3 & 99.0 & 0.5 & 98.4 & 0.5 \\
\bottomrule
\end{tabular}
\end{table}

The frequency effect is significant only for Pythia-70m
(Kruskal-Wallis $H = 13.5$, $\eta^2 = 0.95$, $p_{\text{BH}} = 0.036$).
For the larger models, $p_{\text{BH}} \geq 0.054$.

\subsection{Circuit Size and Minimality}
\label{app:circuit_size_detail}

\begin{figure}[t]
\centering
\includegraphics[width=0.8\linewidth]{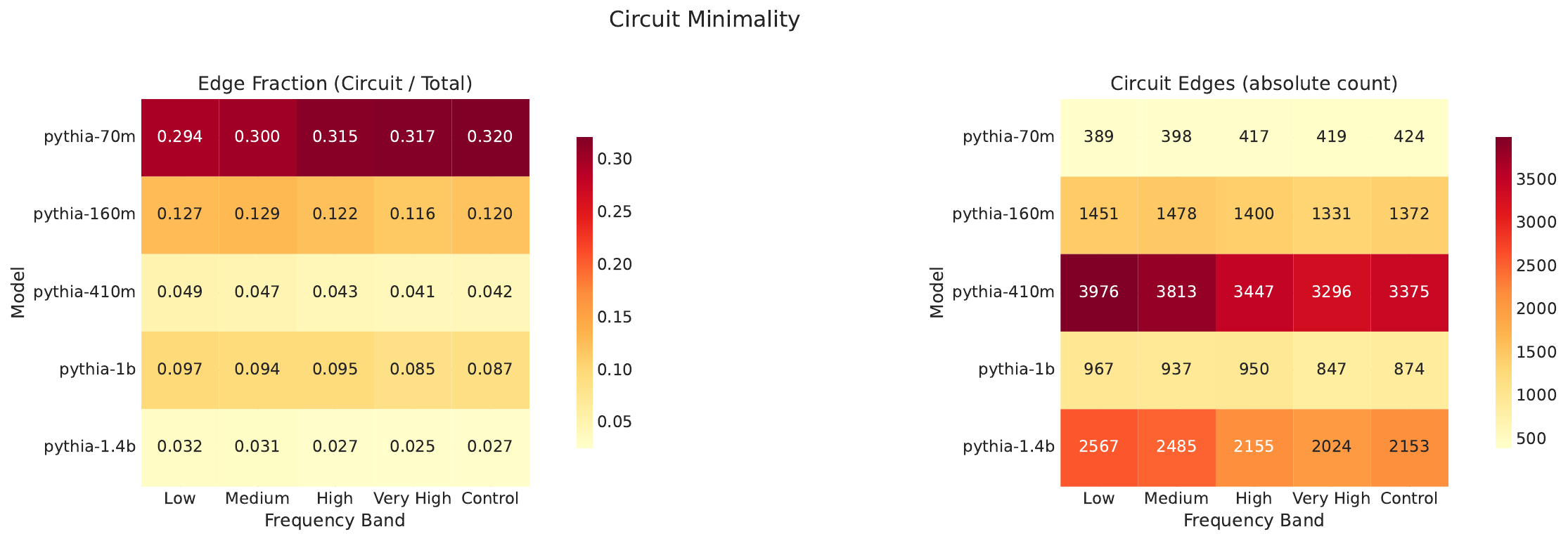}
\caption{Circuit size (edge fraction) across models and frequency bands.
Variation across bands within each model is not significant after FDR correction.}
\label{fig:circuit_size_heatmap}
\end{figure}

The fraction of retained edges decreases with model scale, from 29--32\%
for Pythia-70m to 2.5--3.2\% for Pythia-1.4b
(Figure~\ref{fig:circuit_size_heatmap}, Table~\ref{tab:circuit_size_stats}).

\begin{table}[t]
\centering
\caption{Circuit size by model and frequency band.
Values report $n_{\text{edges}}$ $\pm$ the standard deviation over three draws. The rightmost column for each model reports the edge fraction as a percentage of the full graph.}
\label{tab:circuit_size_stats}
\small
\setlength{\tabcolsep}{3.5pt}
\begin{tabular}{l rrr rrr rrr rrr rrr}
\toprule
 & \multicolumn{3}{c}{Pythia-70m} & \multicolumn{3}{c}{Pythia-160m} & \multicolumn{3}{c}{Pythia-410m} & \multicolumn{3}{c}{Pythia-1b} & \multicolumn{3}{c}{Pythia-1.4b} \\
\cmidrule(lr){2-4}\cmidrule(lr){5-7}\cmidrule(lr){8-10}\cmidrule(lr){11-13}\cmidrule(lr){14-16}
Band & $n$ & $\pm$ & \% & $n$ & $\pm$ & \% & $n$ & $\pm$ & \% & $n$ & $\pm$ & \% & $n$ & $\pm$ & \% \\
\midrule
low        & 389 &  8 & 29.4 & 1451 &  39 & 12.7 & 3976 & 138 & 4.9 & 967 & 14 & 9.7 & 2567 &  52 & 3.2 \\
medium     & 398 &  7 & 30.0 & 1478 &  34 & 12.9 & 3813 & 106 & 4.7 & 937 & 61 & 9.4 & 2485 &  57 & 3.1 \\
high       & 417 & 13 & 31.5 & 1400 &  49 & 12.2 & 3447 &  59 & 4.3 & 950 & 50 & 9.5 & 2155 &  31 & 2.7 \\
very\_high & 419 &  4 & 31.7 & 1331 &  26 & 11.6 & 3296 & 173 & 4.1 & 847 & 56 & 8.5 & 2024 &  84 & 2.5 \\
control    & 424 & 13 & 32.0 & 1372 &  24 & 12.0 & 3375 &  54 & 4.2 & 874 & 53 & 8.7 & 2153 & 128 & 2.7 \\
\bottomrule
\end{tabular}
\end{table}

Circuit size does not vary significantly across bands after FDR correction
(all $p_{\text{BH}} \geq 0.064$).

\subsection{Same-Band Faithfulness and Sufficiency}
\label{app:faithfulness_detail}

Same-band circuit accuracy ranges from 24--55\% (Pythia-70m, retention
80--89\%) to 80--97\% for larger models (retention ${\geq}$93\% for
160m--1b; Pythia-1.4b retains 83--94\%).

KL~divergence increases with model size, from 0.23--0.27 for Pythia-70m
to 0.51--0.65 for Pythia-1.4b. A band effect on circuit accuracy is
significant only for Pythia-70m ($H = 12.7$, $\eta^2 = 0.87$,
$p_{\text{BH}} = 0.047$).

\subsection{Circuit Completeness (Necessity)}
\label{app:completeness_detail}

\begin{figure}[t]
\centering
\includegraphics[width=\linewidth]{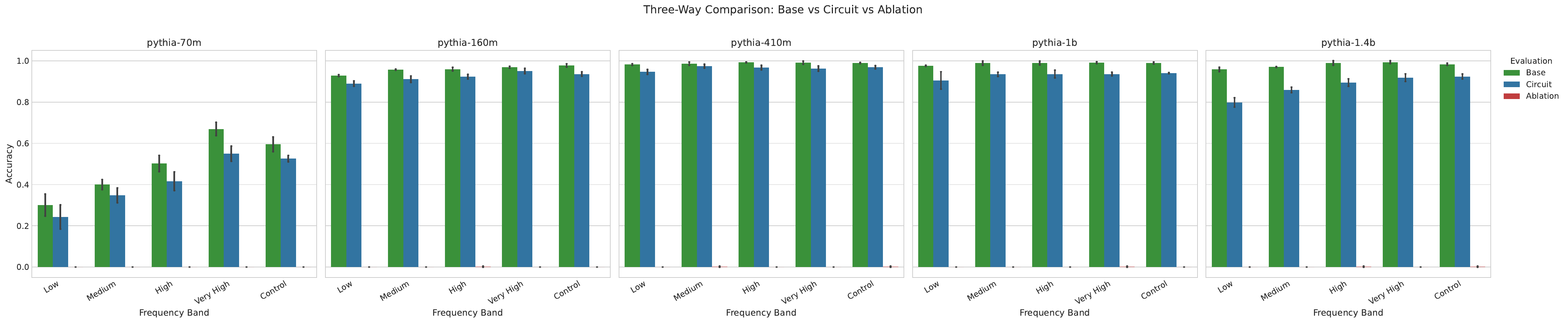}
\caption{Base model, circuit, and ablation accuracy across models and bands.
Ablation accuracy is at or near zero in all conditions.}
\label{fig:three_way_comparison}
\end{figure}

Ablation accuracy is 0.0\% in 69/75 circuits
(Figure~\ref{fig:three_way_comparison}).
The remaining six circuits have an ablation accuracy of 0.44\% and
completeness of 0.998. They are Pythia-160m high (draw~3), Pythia-410m
medium and control (draw~2), Pythia-1b \texttt{very\_high} (draw~2), and
Pythia-1.4b high (draw~2) and control (draw~1).

The Wilcoxon signed-rank results support necessity in every model, with
$p_{\text{BH}} = 0.0024$ and $d = 3.4$--$73.9$. Completeness does not
vary by band ($p > 0.40$).

\subsection{Cross-Band Generalization}
\label{app:generalization_detail}

\begin{figure}[p]
\centering
\includegraphics[width=0.8\linewidth]{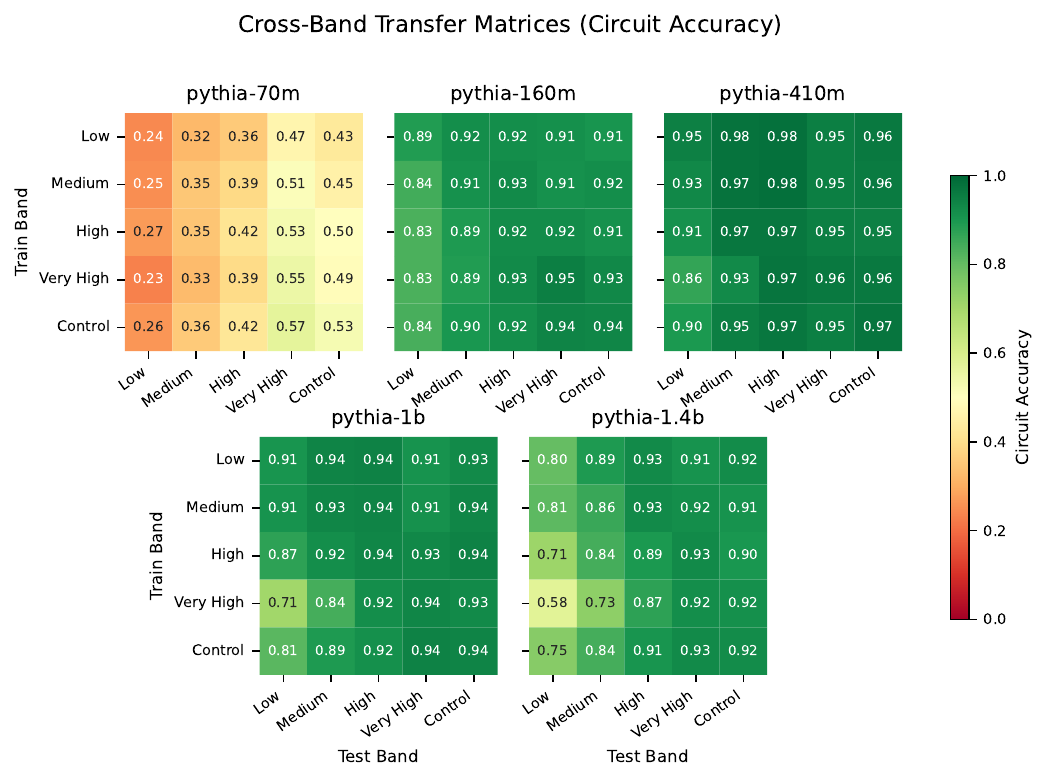}
\caption{Cross-band transfer matrices for all five models (averaged over draws).
Rows show the extraction band, and columns show the test band. The matrices are largely uniform, and low-frequency circuits transfer well to high-frequency data.}
\label{fig:cross_band_transfer}
\end{figure}

Generalization gaps are small and often negative
(Figure~\ref{fig:cross_band_transfer}, Table~\ref{tab:generalization_gap}).
Appendix~\ref{app:targeted_transfer} provides the corresponding
targeted ablation analysis.

\begin{table}[t]
\centering
\caption{Cross-band generalization gap by model and training band.
Same and Cross report accuracy in percent. Gap is Same minus Cross in percentage points, with positive values indicating higher same-band accuracy.}
\label{tab:generalization_gap}
\scriptsize
\setlength{\tabcolsep}{3pt}
\begin{tabular}{l ccc ccc ccc ccc ccc}
\toprule
 & \multicolumn{3}{c}{Pythia-70m} & \multicolumn{3}{c}{Pythia-160m} & \multicolumn{3}{c}{Pythia-410m} & \multicolumn{3}{c}{Pythia-1b} & \multicolumn{3}{c}{Pythia-1.4b} \\
\cmidrule(lr){2-4}\cmidrule(lr){5-7}\cmidrule(lr){8-10}\cmidrule(lr){11-13}\cmidrule(lr){14-16}
Band & S & C & Gap & S & C & Gap & S & C & Gap & S & C & Gap & S & C & Gap \\
\midrule
low        & 24.3 & 39.6 & $-$15.3 & 89.0 & 91.4 & $-$2.4 & 94.7 & 96.7 & $-$2.0 & 90.5 & 93.0 & $-$2.5 & 79.9 & 91.3 & $-$11.4 \\
medium     & 34.8 & 40.1 & $-$5.3 & 91.1 & 90.1 & $+$1.0 & 97.5 & 95.6 & $+$1.9 & 93.5 & 92.3 & $+$1.2 & 85.9 & 89.1 & $-$3.2 \\
high       & 41.6 & 41.1 & $+$0.5 & 92.3 & 89.0 & $+$3.3 & 96.7 & 94.4 & $+$2.3 & 93.6 & 91.7 & $+$1.9 & 89.5 & 84.6 & $+$4.9 \\
very\_high & 55.0 & 36.0 & $+$19.0 & 95.1 & 89.5 & $+$5.6 & 96.3 & 93.1 & $+$3.2 & 93.6 & 84.8 & $+$8.8 & 91.9 & 77.5 & $+$14.4 \\
control    & 52.6 & 40.2 & $+$12.4 & 93.6 & 89.9 & $+$3.7 & 96.9 & 94.2 & $+$2.7 & 94.1 & 89.2 & $+$4.9 & 92.4 & 85.6 & $+$6.8 \\
\bottomrule
\end{tabular}
\end{table}

The difference between same-band and cross-band is significant only for Pythia-1b
(Mann-Whitney $r = 0.41$, $p_{\text{BH}} = 0.035$); other models
do not survive FDR correction. The gap does not differ across models
($p_{\text{BH}} = 0.92$).

\subsection{Power Analysis for Cross-Band Transfer Test}
\label{app:transfer_power}
We estimate the sensitivity of the functional cross-band transfer test. The test uses 15 paired comparisons per model (5~conditions $\times$ 3~draws). Each comparison pairs the circuit's accuracy on its own condition with the mean accuracy of circuits from other conditions on that same test condition, so the shared baseline cancels. Significance is assessed with a one-sided paired Wilcoxon signed-rank test.
Under a calculation based on paired means and a working independence assumption, the nominal 80\%-power threshold at $\alpha{=}0.05$ (one-sided) is 0.013--0.026 accuracy points, depending on the model-specific standard deviation of the paired differences (0.019--0.039).
This is a nominal sensitivity estimate rather than a guarantee because the 15~differences are clustered within three draws and share source circuits. The working independence assumption is therefore approximate.
The pipeline detects the residual same-band advantage of 0.016--0.029 accuracy points with $p \leq 0.028$ for all five models (Table~\ref{tab:transfer_efficiency}), consistent with this sensitivity estimate.
The detected advantage amounts to 3--19\% of the same-band boost, equivalent to transfer efficiency of 0.81--0.97. Under zero ablation, the advantage vanishes entirely (Cohen's $d \leq 0.18$; Appendix~\ref{app:zero_ablation}).
Detection of \emph{genuine} specialization when present is shown separately by the two-route positive control (\autoref{sec:two_route_control}).
The structural Jaccard gap analysis rests on fewer pairwise comparisons and has weaker power (Table~\ref{tab:power_analysis}).

\subsection{Asymmetric Transfer}
\label{app:asymmetry_detail}

\begin{figure}[t]
\centering
\includegraphics[width=0.8\linewidth]{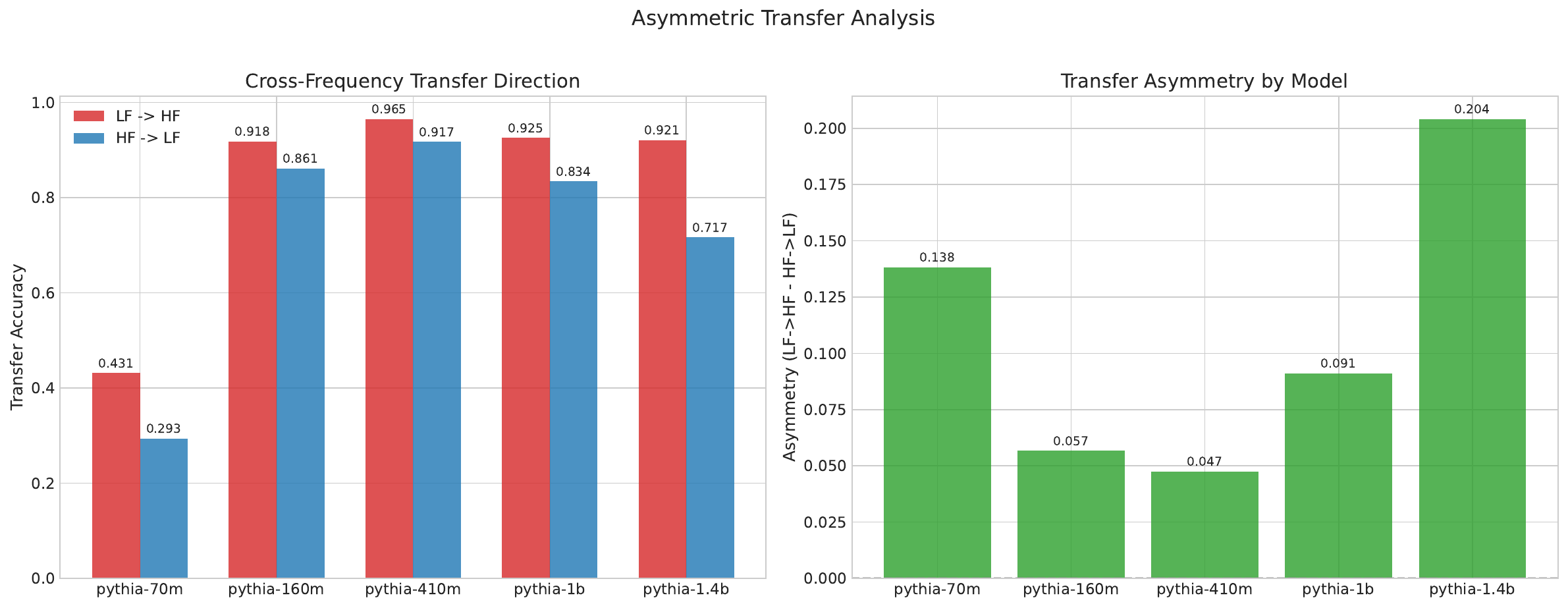}
\caption{Low-frequency (LF) circuits achieve higher accuracy on high-frequency (HF) data than high-frequency circuits achieve on low-frequency data.}
\label{fig:asymmetric_transfer}
\end{figure}

Transfer from low-frequency circuits to high-frequency data exceeds transfer
from high-frequency circuits to low-frequency data by 4.7--20.4~percentage points
(Figure~\ref{fig:asymmetric_transfer}; Table~\ref{tab:asymmetry_summary}
in the main text). Pythia-70m has the largest ratio (1.47), Pythia-1.4b has
the largest absolute asymmetry ($+$20.4~percentage points), and Pythia-410m
has the smallest asymmetry (1.05).

The asymmetry is significant for all models
(all $p_{\text{BH}} \leq 0.037$; $r = 0.65$--$0.88$) and increases with
frequency distance ($\rho = 0.93$, $p_{\text{BH}} = 0.036$).

\subsection{Model Scaling}
\label{app:functional_scaling}

Circuit accuracy peaks at Pythia-410m (96.4\%), while edge fraction
decreases from 30.9\% to 2.8\% and retention reaches 97.5\%
(Table~\ref{tab:scaling_summary}).
Pythia-1b and Pythia-1.4b do not continue the accuracy trend. Their accuracy
(93.1\% and 87.9\%) and retention (94.3\% and 89.7\%) fall below
Pythia-410m, while KL divergence continues to increase (0.52 and 0.56).

\begin{figure}[t]
\centering
\includegraphics[width=0.5\linewidth]{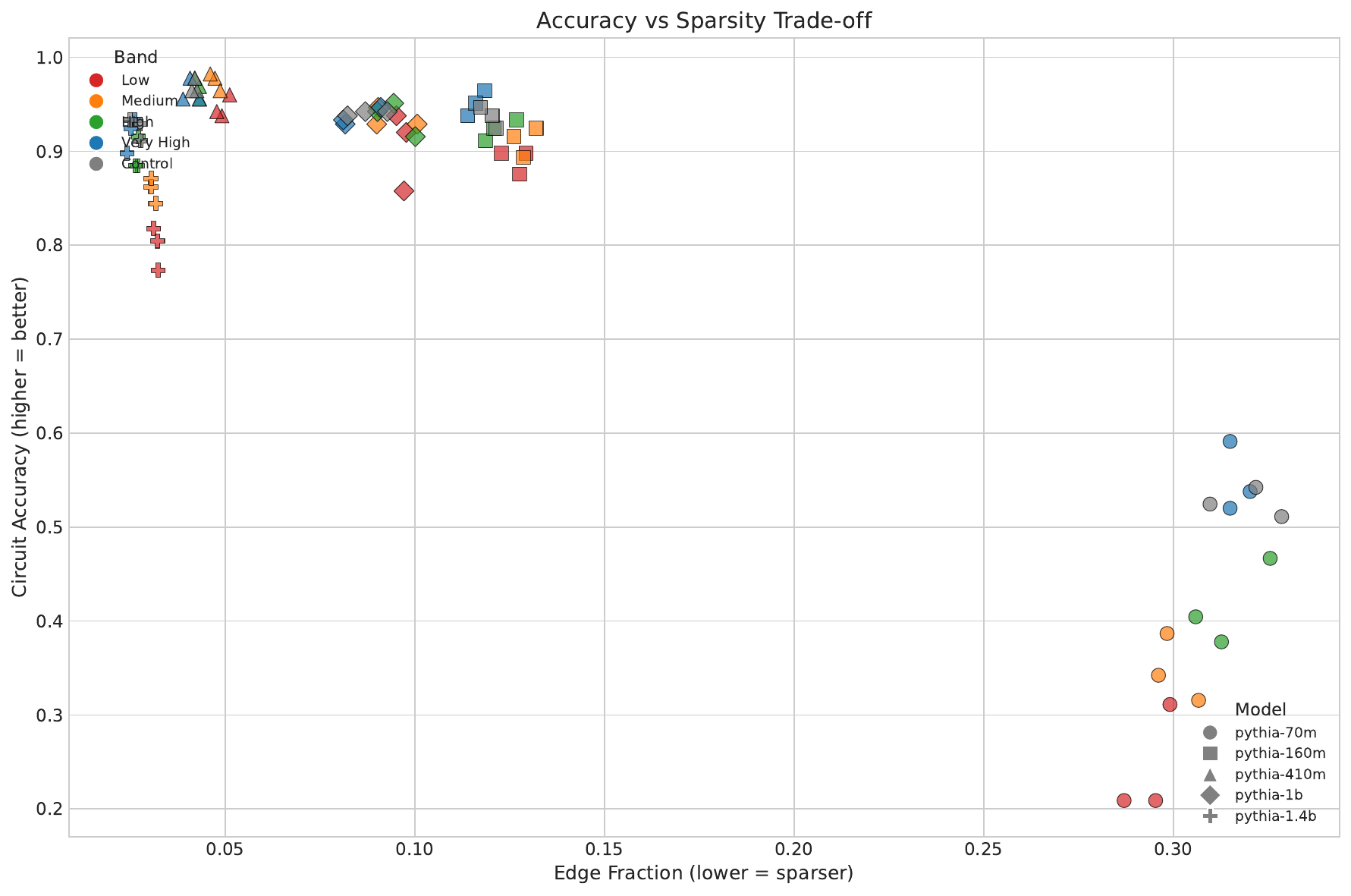}
\caption{Circuit accuracy versus circuit size (edge fraction) for all
75~circuits.
Larger models achieve higher accuracy with smaller circuits, occupying
the region of the plot with high accuracy and a low edge fraction.
Within each model, circuits from different frequency bands overlap,
reflecting the absence of a band effect on circuit size or accuracy.}
\label{fig:accuracy_vs_sparsity}
\end{figure}

Figure~\ref{fig:accuracy_vs_sparsity} shows circuit accuracy against edge
fraction for all 75 circuits. Models separate clearly, while frequency bands
overlap within each model.

\begin{table}[t]
\centering
\caption{Circuit properties by model size, averaged over all bands and
draws. The $\pm$ column is the standard deviation of circuit accuracy
across those circuits.}
\label{tab:scaling_summary}
\small
\begin{tabular}{rcccccc}
\toprule
Size (M) & Base (\%) & Circ.\ (\%) & $\pm$ & Edge~\% & Retention & KL \\
\midrule
70   & 49.4 & 41.7 & 12.3 & 30.9 & 0.841 & 0.251 \\
160  & 95.9 & 92.2 & 2.4  & 12.3 & 0.962 & 0.292 \\
410  & 98.8 & 96.4 & 1.3  & 4.4  & 0.975 & 0.320 \\
1000 & 98.7 & 93.1 & 2.3  & 9.1  & 0.943 & 0.523 \\
1400 & 97.9 & 87.9 & 5.0  & 2.8  & 0.897 & 0.559 \\
\bottomrule
\end{tabular}
\end{table}

Circuit accuracy varies strongly across models ($H = 47.0$,
$\eta^2 = 0.79$, $p_{\text{BH}} < 10^{-8}$). Edge fraction
decreases with size ($\rho = -0.77$, $p < 10^{-11}$) and retention
improves ($\rho = 0.41$, $p = 0.006$). The generalization gap
does not scale with model size ($p_{\text{BH}} = 0.92$).

\subsection{Variance Decomposition}
\label{app:variance_detail}

Table~\ref{tab:variance_decomposition} reports the proportion of variance in
circuit accuracy attributable to model, frequency band, draw, and residual
variation. The model $\times$ band term measures whether the effect of
frequency band differs across models. Appendix~\ref{app:integration_variance}
reports the corresponding analysis across structural, functional, and
representational metrics.

\begin{table}[t]
\centering
\caption{Variance decomposition of LSC circuit accuracy. The table reports $\eta^2$ from a full factorial ANOVA and the corresponding percentage of variance.}
\label{tab:variance_decomposition}
\small
\begin{tabular}{lcc}
\toprule
Factor & $\eta^2$ & Variance (\%) \\
\midrule
Model             & 0.923 & 92.3 \\
Band              & 0.034 &  3.4 \\
Model $\times$ Band & 0.035 &  3.5 \\
Draw              & 0.001 &  0.1 \\
Residual          & 0.007 &  0.7 \\
\bottomrule
\end{tabular}
\end{table}

\subsection{Control Band Analysis}
\label{app:control_detail}

\begin{figure}[p]
\centering
\includegraphics[width=0.8\linewidth]{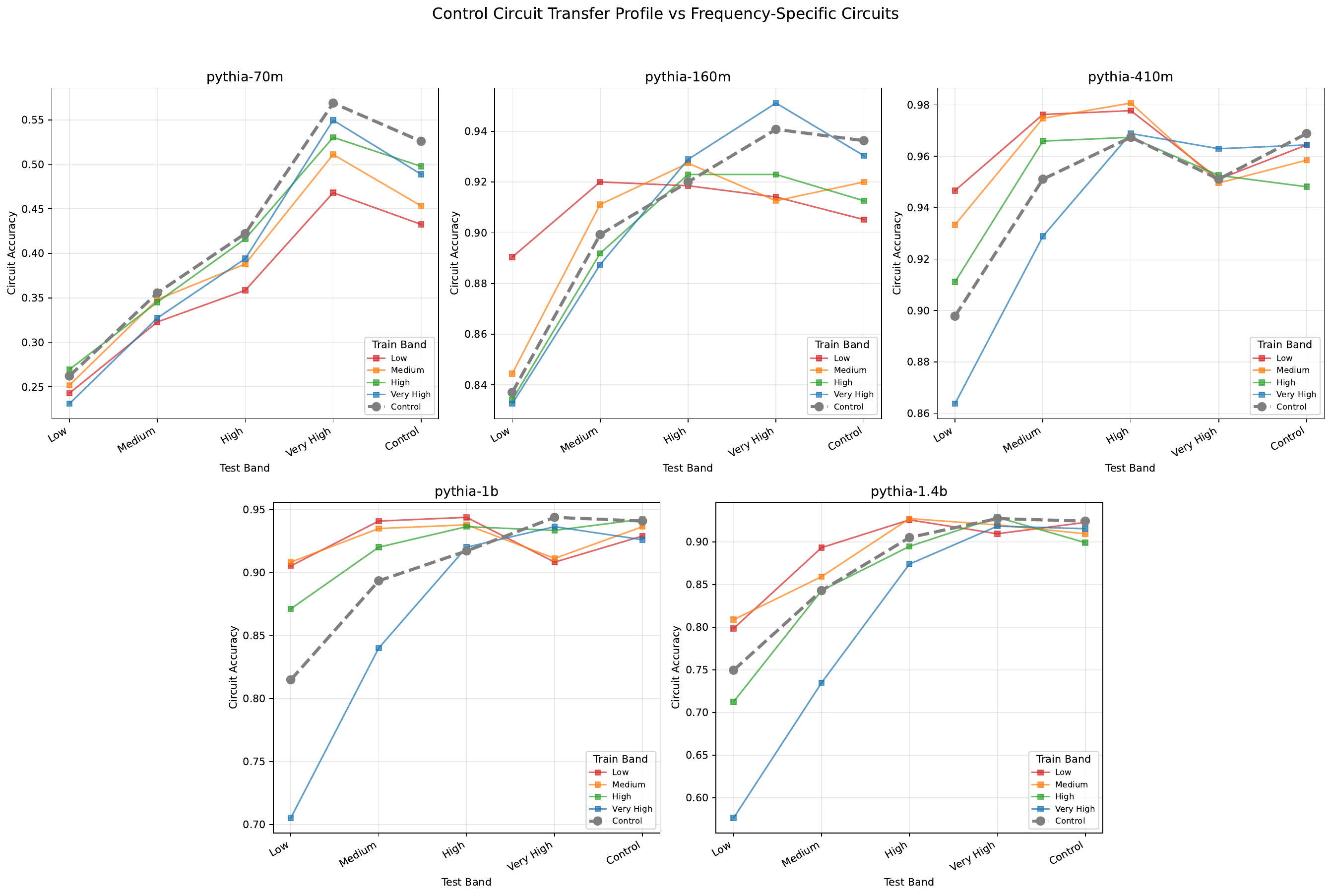}
\caption{Accuracy and transfer of the circuit extracted from the control condition. The control circuit, sampled according to token frequency, performs similarly to circuits extracted for individual frequency bands.}
\label{fig:control_average}
\end{figure}

The control circuit does not differ from frequency-specific circuits in
accuracy or size ($p_{\text{BH}} > 0.17$,
Figure~\ref{fig:control_average}). For Pythia-70m, it achieves the
best cross-band transfer on 4/5 test bands (binomial test~\mbox{\citep{arbuthnot1710ii}}, $p_{\text{BH}} = 0.033$);
for larger models it is not significantly better. Jaccard similarity
to band-specific circuits is nearly uniform (CV 1--3\%).

\subsection{Per-Example Failure Analysis}
\label{app:failure_analysis}

Bimodality is rare. Hartigan's dip test identifies 24 of 375 conditions as
bimodal, comprising 2 of 75 same-band and 22 of 300 cross-band conditions.
Most conditions instead show a general decrease across examples rather than a
split into two groups. Degradation is uniform across examples. The probability
assigned to the correct answer under cross-band transfer correlates with the
corresponding probability on the extraction band (Spearman
$\rho = 0.37$--$0.77$). Examples assigned a higher probability on the
extraction band also tend to receive a higher probability under transfer. For Pythia-70m, 168 examples are answered
incorrectly by every circuit. The larger models have no such examples, and
every circuit answers 421--810 examples correctly.

Per-example prediction agreement over all circuit pairs is 78--93\% for
models ${\geq}160$M (54--66\% at Pythia-70m), with cross-band pairs
agreeing at least as often as same-band pairs (same-band minus
cross-band, $-0.03$ to $-0.01$).
Chance-corrected agreement (Cohen's $\kappa$) is near zero for same-band
pairs and 0.15--0.27 for cross-band pairs. This indicates that prediction
errors vary across extractions rather than following a consistent pattern by
frequency band.

\FloatBarrier\section{Additional Structural Results}
\label{app:structural}
This appendix provides structural analyses of all 75~LSC circuits (5~models $\times$ 5~conditions $\times$ 3~draws). Structural results for the SVA replication are reported in \autoref{sec:sva_replication}.
Statistical tests use the nonparametric framework described in \autoref{app:stat_summary}. Model effects explain most of the variance (\autoref{app:structural_scaling_detail}).

\subsection{Edge Statistics and Circuit Size}
\label{app:edge_stats_detail}
Edge fraction generally decreases with model size, from 30.9\% in Pythia-70m to 2.8\% in Pythia-1.4b. The exception is Pythia-1b, whose edge fraction is 9.1\%, compared with 4.4\% for Pythia-410m (Table~\ref{tab:structural_edge_stats}).

\subsection{Component Composition}
\label{app:component_detail}
Attention edges dominate all circuits (53.6--62.3\%), followed by MLP (29.2--38.4\%) and residual (6.8--12.9\%; Table~\ref{tab:structural_component}).

\begin{table}[t]
\begin{minipage}[t]{0.46\linewidth}\centering
\caption{Circuit edge statistics by model, reported as the range over bands after averaging across draws.}\label{tab:structural_edge_stats}\small
\begin{tabular}{lrc}\toprule
Model & $n_{\text{edges}}$ & Edge~\% \\\midrule
Pythia-70m  & 389--424   & 30.9 \\
Pythia-160m & 1{,}331--1{,}478 & 12.3 \\
Pythia-410m & 3{,}296--3{,}976 &  4.4 \\
Pythia-1b   & 847--967   &  9.1 \\
Pythia-1.4b & 2{,}024--2{,}567 & 2.8 \\\bottomrule
\end{tabular}
\end{minipage}\hfill
\begin{minipage}[t]{0.50\linewidth}\centering
\caption{Component composition (\%) by model (ranges over bands).}\label{tab:structural_component}\small
\begin{tabular}{lccc}\toprule
Model & Attn & MLP & Resid \\\midrule
Pythia-70m  & 54.8--58.4 & 29.2--32.2 & 12.3--12.9 \\
Pythia-160m & 54.9--57.9 & 33.6--35.9 &  8.5--9.3 \\
Pythia-410m & 53.6--57.5 & 35.6--38.4 &  6.9--8.1 \\
Pythia-1b   & 59.6--61.9 & 30.9--32.7 &  6.9--8.1 \\
Pythia-1.4b & 61.0--62.3 & 30.6--31.5 &  6.8--7.6 \\\bottomrule
\end{tabular}
\end{minipage}
\end{table}

\subsection{Edge Categories}
\label{app:layer_flow_detail}
We describe edges by their source and destination, and these categories are not mutually exclusive. Input edges originate at the embedding, output edges terminate at the final residual stream, local edges remain within one layer, forward edges connect adjacent layers, and skip edges bypass at least one layer. Skip edges account for 62.2\% of edges in Pythia-70m and 88.7\% in Pythia-410m, while the fraction of forward edges generally decreases as the fraction of skip edges increases. Local edges account for less than 2\%, input edges for 0.4--4.5\%, and output edges for 6.9--12.9\%.

\subsection{Head Participation}
\label{app:head_participation_detail}
\begin{figure}[t]\centering
\includegraphics[width=\linewidth]{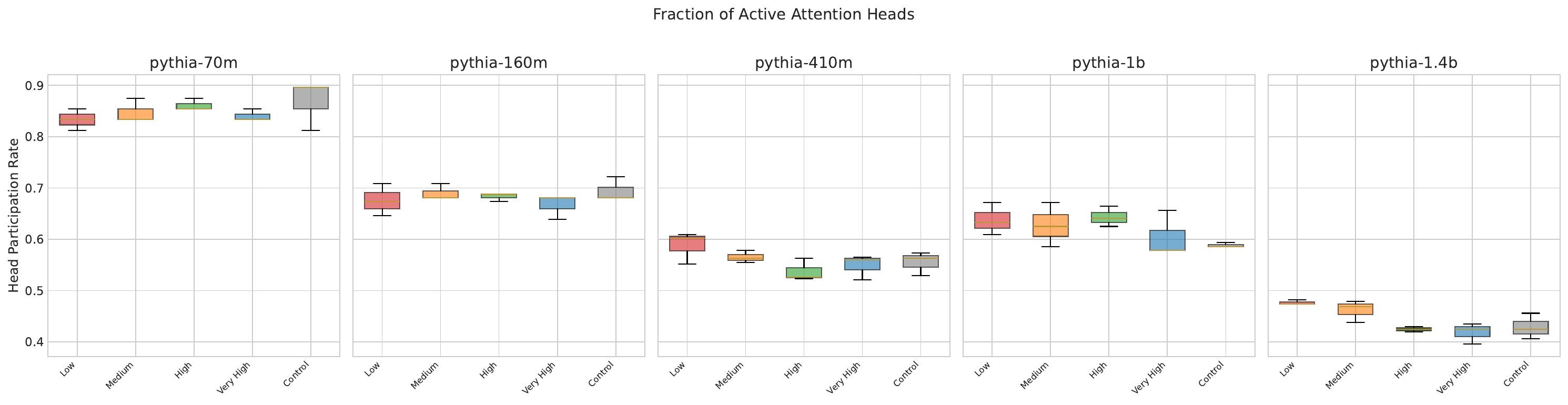}
\caption{Head participation rate by model and band.}\label{fig:structural_head_rate}
\end{figure}

Head participation is generally lower in larger models. It is 85.0\% in Pythia-70m, 68.2\% in 160m, 55.9\% in 410m, 62.0\% in 1b, and 44.2\% in 1.4b (Figure~\ref{fig:structural_head_rate}).

\subsection{Universal vs.\ Band-Specific Edges}
\label{app:universal_edges_detail}
\begin{figure}[t]\centering
\includegraphics[width=0.8\linewidth]{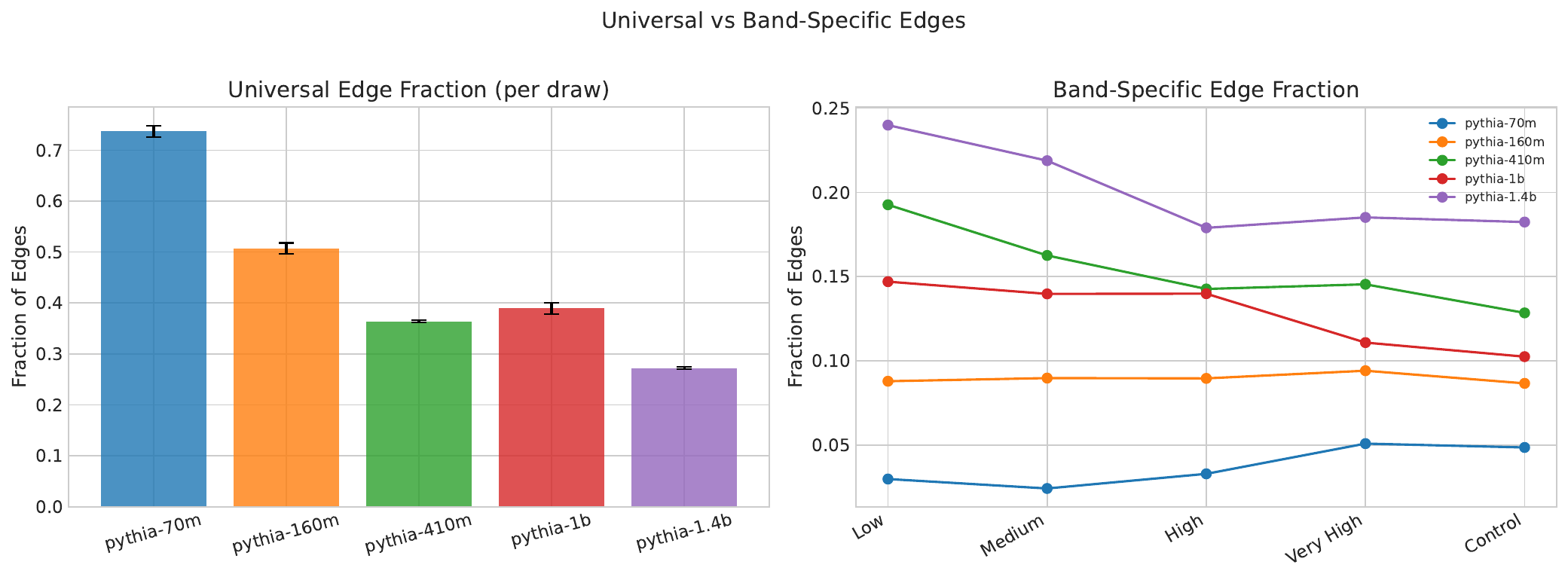}
\caption{Universal vs.\ band-specific edge counts per model.}\label{fig:structural_universal}
\end{figure}

Universal fraction decreases from 65.5\% (Pythia-70m) to 15.4\% (Pythia-1.4b; Figure~\ref{fig:structural_universal}, Table~\ref{tab:structural_universal}).
Attention edges account for 75--93\% of band-specific edges. Universal edges have larger contributions from MLPs (37--50\%) and the residual stream (14--17\%).
In every draw, the universal edges form one weakly connected component when edge direction is ignored. Across draws, the nodes incident to universal edges comprise 48--91\% of the nodes in the union of the five band circuits.

\begin{table}[t]
\centering\caption{Universal edge counts and fractions per model.}\label{tab:structural_universal}\small
\begin{tabular}{lrrc}\toprule
Model & Universal & Mean edges & Univ.\ \% \\\midrule
Pythia-70m  & 268   & 409   & 65.5 \\
Pythia-160m & 531   & 1{,}407 & 37.8 \\
Pythia-410m & 872   & 3{,}581 & 24.3 \\
Pythia-1b   & 223   & 915   & 24.4 \\
Pythia-1.4b & 350   & 2{,}277 & 15.4 \\\bottomrule
\end{tabular}\end{table}

\subsection{Edge Marginality Under an Independent Importance Measure}
\label{app:edge_marginality}

Section~\ref{sec:why_phantom} proposes that ACDC can select different,
individually dispensable edges across frequency bands. When many edges can be
removed with little effect, small input changes can alter which edges ACDC
retains.
We cannot test this directly from ACDC's own scores, because AutoCircuit
saves them only in thresholded form (retained edges as infinity, pruned
edges at a threshold sentinel). We therefore use the EAP-IG attribution scores
saved for the full computational graph in every condition as an independent
importance measure (\autoref{app:method_comparison}).
For each edge selected by ACDC we compute $\rho = r / k$, where $r$ is the
edge's rank when all graph edges are ordered by absolute EAP-IG score and
$k$ is the ACDC circuit size. A value of $\rho = 1$ is the cutoff at the same number of edges.
Values below 1 indicate that an edge ranks within the cutoff, and values above
1 indicate that it ranks outside the cutoff.

\begin{figure}[t]
\centering
\includegraphics[width=0.85\linewidth]{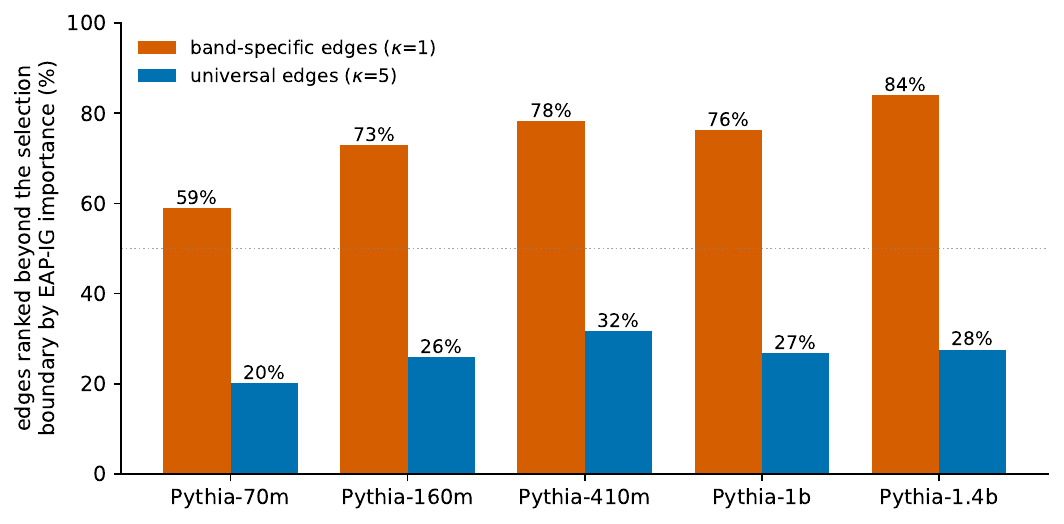}
\caption{Percentage of edges selected by ACDC that EAP-IG ranks beyond the selection boundary. An edge is beyond the boundary if its EAP-IG rank exceeds the ACDC circuit size ($\rho > 1$).
Most band-specific edges fall outside this boundary, and the fraction grows with model scale, whereas universal edges mostly fall inside it.}
\label{fig:edge_marginality}
\end{figure}

Band-specific edges ($\kappa{=}1$) tend to rank outside the cutoff at the same number of edges.
Median $\rho$ ranges from 1.11 in Pythia-70m to 3.25 in Pythia-1.4b, and
59--84\% of these edges rank beyond the cutoff.
Universal edges ($\kappa{=}5$) rank well within the cutoff at the same number of edges in
every model. Median $\rho$ is 0.39--0.50, and 20--32\% of these edges rank
beyond the cutoff (Table~\ref{tab:edge_marginality},
Figure~\ref{fig:edge_marginality}).
Because edges within a circuit are not independent, we treat the per-edge
Mann--Whitney comparison as descriptive and base inference on a conservative
test across conditions. In each of the 15 band--draw conditions per model, the
median $\rho$ of band-specific edges exceeds that of universal edges. The
resulting Wilcoxon test gives $p = 3.1 \times 10^{-5}$ for every model.
EAP-IG is independent of ACDC's selection, and the two methods overlap only
partially (Jaccard 0.28--0.60; \autoref{app:method_comparison}). This makes the
comparison strict because the marginal candidates predicted in
Section~\ref{sec:why_phantom} may be important under only one measure, whereas
both methods rank the universal core as important.

\begin{table}[t]
\centering
\caption{Edge marginality under EAP-IG importance. $\rho$ is the edge's EAP-IG rank divided by the ACDC circuit size. A value of $\rho = 1$ marks the selection boundary with the same number of edges as ACDC. Beyond~\% is the fraction of edges with $\rho > 1$.
The $p$ value comes from a Wilcoxon test of the difference in median $\rho$ between band-specific and universal edges across 15 band--draw conditions per model.}
\label{tab:edge_marginality}
\small
\begin{tabular}{lccccc}
\toprule
 & \multicolumn{2}{c}{Band-specific ($\kappa{=}1$)} & \multicolumn{2}{c}{Universal ($\kappa{=}5$)} & \\
\cmidrule(lr){2-3}\cmidrule(lr){4-5}
Model & Median $\rho$ & Beyond \% & Median $\rho$ & Beyond \% & $p$ \\
\midrule
Pythia-70m  & 1.11 & 59 & 0.44 & 20 & $3.1 \times 10^{-5}$ \\
Pythia-160m & 1.70 & 73 & 0.47 & 26 & $3.1 \times 10^{-5}$ \\
Pythia-410m & 2.29 & 78 & 0.50 & 32 & $3.1 \times 10^{-5}$ \\
Pythia-1b   & 1.95 & 76 & 0.42 & 27 & $3.1 \times 10^{-5}$ \\
Pythia-1.4b & 3.25 & 84 & 0.39 & 28 & $3.1 \times 10^{-5}$ \\
\bottomrule
\end{tabular}
\end{table}

\subsection{Circuit Overlap (Jaccard Similarity)}
\label{app:jaccard_detail}
Within-band Jaccard exceeds between-band in all models (gaps 0.013--0.032; Table~\ref{tab:structural_jaccard} in the main text).

\subsection{Power Analysis for Jaccard Gap Detection}
\label{app:power_analysis}
Effect sizes are large for models from 70m to 410m ($d \geq 1.0$, power ${\geq}0.998$). Pythia-1b has lower power ($d = 0.53$, power $= 0.69$) and requires five draws to reach 80\% power. CLES is the probability that a randomly selected within-band pair has a higher Jaccard value than a randomly selected between-band pair. It ranges from 0.65 to 0.85 (Table~\ref{tab:power_analysis}), showing why comparisons based on a single extraction per condition are unreliable.

\begin{table}[t]
\centering\caption{Power analysis for detecting the Jaccard gap. $n_{\min}$ is the number of draws required for 80\% power at $\alpha{=}0.05$, and CLES is the Common Language Effect Size.}\label{tab:power_analysis}\small
\begin{tabular}{lcccccc}\toprule
Model & Gap & Pooled SD & $d$ & CLES & $n_{\min}$ & Power (3) \\\midrule
Pythia-70m & 0.032 & 0.026 & 1.22 & 0.81 & 3 & 1.00 \\
Pythia-160m & 0.032 & 0.022 & 1.45 & 0.85 & 2 & 1.00 \\
Pythia-410m & 0.016 & 0.016 & 1.02 & 0.76 & 3 & 1.00 \\
Pythia-1b & 0.013 & 0.024 & 0.53 & 0.65 & 5 & 0.69 \\
Pythia-1.4b & 0.019 & 0.026 & 0.72 & 0.69 & 4 & 0.89 \\\bottomrule
\end{tabular}\end{table}

\subsection{Position-Resolved Attribution Check}
\label{app:positional_attribution}

Circuit discovery assigns each edge a single decision across all token
positions (Section~\ref{sec:pipeline}). It therefore cannot detect whether
different bands use the same edge at different positions. We test for such
differences by recomputing EAP-IG attribution over a position-resolved
graph (one score per edge per token position), with settings otherwise
identical to the aggregated scores of \autoref{app:edge_marginality}, for
all five models, five conditions, and three draws.
Two implementation checks support this analysis. Attribution mass at the 17
positions preceding the repeated prefix is exactly zero in all 75~runs because
the clean and corrupted prompts coincide there. The signed sum of positional
scores also reproduces the saved aggregated scores (cosine
${\geq}0.9999997$).
Signed aggregation retains 87--93\% of the absolute positional mass,
corresponding to 7--13\% sign cancellation across positions.
Attribution mass concentrates at the prediction position (64--74\% across
models), decaying toward the start of the repeated prefix.

For each pair of runs, we take the edges that both ACDC circuits selected
and compare how each edge spreads its attribution over the five positions.
We give greater weight to edges with more attribution by weighting each edge
by the smaller of its two normalized masses.
Positional use is similar but not identical across bands. For edges shared by
two circuits, attribution profiles are slightly more similar between runs from
the same band than between runs from different bands.
Weighted profile cosine is 0.981--0.995 within band and 0.978--0.990
between bands, and weighted total variation is 0.032--0.066 within band
and 0.047--0.074 between bands
(Table~\ref{tab:positional_attribution}).
These gaps are small, corresponding to 0.002--0.006 cosine points or one to
two percentage points of reallocated positional mass. An exact blocked
relabeling test permuting band labels within draws detects the difference
across 14{,}400 relabelings. The two-sided $p$-value is 0.0011 for Pythia-70m
and 0.0001 for the four larger models.
The position of the largest decrease depends on the model. For Pythia-1b and
Pythia-1.4b, it occurs at the prediction position (edge vector cosine 0.947
and 0.952, compared with 0.995 and 0.993 within bands). For Pythia-70m, it
occurs at the low-mass positions at the start of the repeated prefix.

This result indicates a small structural difference in where shared edges
place their attribution mass. It is not evidence of functional specialization
because no cross-condition transfer test was performed with position-specific
masks.
The check has two limits.
First, this check uses attribution rather than rerunning ACDC with
position-specific masks. Circuit discovery also used only the first training
batch (\autoref{app:circuit_extraction}), so the settings match ACDC but the
data differ for the three larger models.
It also cannot separate the two things that could differ between bands,
which edges a circuit uses and where along the sequence it uses them.
Summing signed scores over positions and then taking absolute values gives a
between-band cosine of 0.948--0.987, compared with 0.949--0.982 when edges and
positions are evaluated jointly. Signed scores can cancel across positions,
so these quantities cannot be used to divide the between-band difference into
separate edge and position contributions. We therefore report both.
Position-aware discovery~\mbox{\citep{haklay2025position}} remains future
work (Section~\ref{sec:limitations}).

\begin{table}[ht]
\centering
\caption{Position-resolved attribution check. For edges shared by two
ACDC circuits, we compare how their EAP-IG attribution is distributed
over the five corrupted positions, within band (across draws) versus
between bands (same draw). Profiles are very similar in both cases, but
between-band pairs are slightly less similar. The $p$ value comes from an exact blocked relabeling test with 14{,}400 two-sided relabelings on the weighted profile cosine. Applied to total variation, the same test gives $p = 0.0013$ for Pythia-70m and $p = 0.0001$ for the other models.}
\label{tab:positional_attribution}
\small
\begin{tabular}{lccccc}
\toprule
 & \multicolumn{2}{c}{Profile cosine} & \multicolumn{2}{c}{Profile TV} & \\
\cmidrule(lr){2-3} \cmidrule(lr){4-5}
Model & Within & Between & Within & Between & $p$ (cosine) \\
\midrule
Pythia-70m  & 0.981 & 0.978 & 0.066 & 0.074 & 0.0011 \\
Pythia-160m & 0.990 & 0.987 & 0.041 & 0.050 & 0.0001 \\
Pythia-410m & 0.993 & 0.988 & 0.036 & 0.051 & 0.0001 \\
Pythia-1b   & 0.995 & 0.989 & 0.034 & 0.054 & 0.0001 \\
Pythia-1.4b & 0.994 & 0.990 & 0.032 & 0.047 & 0.0001 \\
\bottomrule
\end{tabular}
\end{table}

\subsection{Component-Level Jaccard}
\label{app:component_jaccard_detail}
Within-band Jaccard exceeds between-band Jaccard for each component type, with attention showing the largest gap (0.014--0.042, significant in all models). The MLP gap is significant in three models, and the residual stream gap is significant in two (Table~\ref{tab:structural_comp_jaccard}).

\begin{table}[t]
\centering\caption{Jaccard gaps by component. $^{*}$Significant after Benjamini--Hochberg FDR correction.}\label{tab:structural_comp_jaccard}\small
\begin{tabular}{llcccc}\toprule
Model & Component & $J_{\text{within}}$ & $J_{\text{between}}$ & Gap & $d$ \\\midrule
\multirow{3}{*}{Pythia-70m}
  & Attn$^{*}$  & 0.699 & 0.657 & 0.042 & 0.99 \\
  & MLP$^{*}$   & 0.925 & 0.910 & 0.014 & 0.74 \\
  & Resid$^{*}$ & 0.985 & 0.971 & 0.013 & 0.77 \\\midrule
\multirow{3}{*}{Pythia-160m}
  & Attn$^{*}$  & 0.478 & 0.440 & 0.038 & 1.30 \\
  & MLP$^{*}$   & 0.715 & 0.695 & 0.019 & 0.96 \\
  & Resid$^{*}$ & 0.957 & 0.928 & 0.029 & 1.24 \\\midrule
\multirow{3}{*}{Pythia-410m}
  & Attn$^{*}$  & 0.344 & 0.328 & 0.017 & 0.74 \\
  & MLP$^{*}$   & 0.552 & 0.540 & 0.012 & 0.98 \\
  & Resid$^{*}$ & 0.869 & 0.844 & 0.025 & 1.03 \\\midrule
\multirow{3}{*}{Pythia-1b}
  & Attn$^{*}$  & 0.421 & 0.407 & 0.014 & 0.51 \\
  & MLP         & 0.536 & 0.526 & 0.010 & 0.38 \\
  & Resid       & 0.786 & 0.777 & 0.009 & 0.20 \\\midrule
\multirow{3}{*}{Pythia-1.4b}
  & Attn$^{*}$  & 0.321 & 0.303 & 0.018 & 0.64 \\
  & MLP$^{*}$   & 0.456 & 0.437 & 0.019 & 0.66 \\
  & Resid$^{*}$ & 0.744 & 0.714 & 0.029 & 0.76 \\\bottomrule
\end{tabular}\end{table}

\subsection{Band Affinity and Containment}
\label{app:band_affinity_detail}
\begin{figure}[t]\centering
\includegraphics[width=0.8\linewidth]{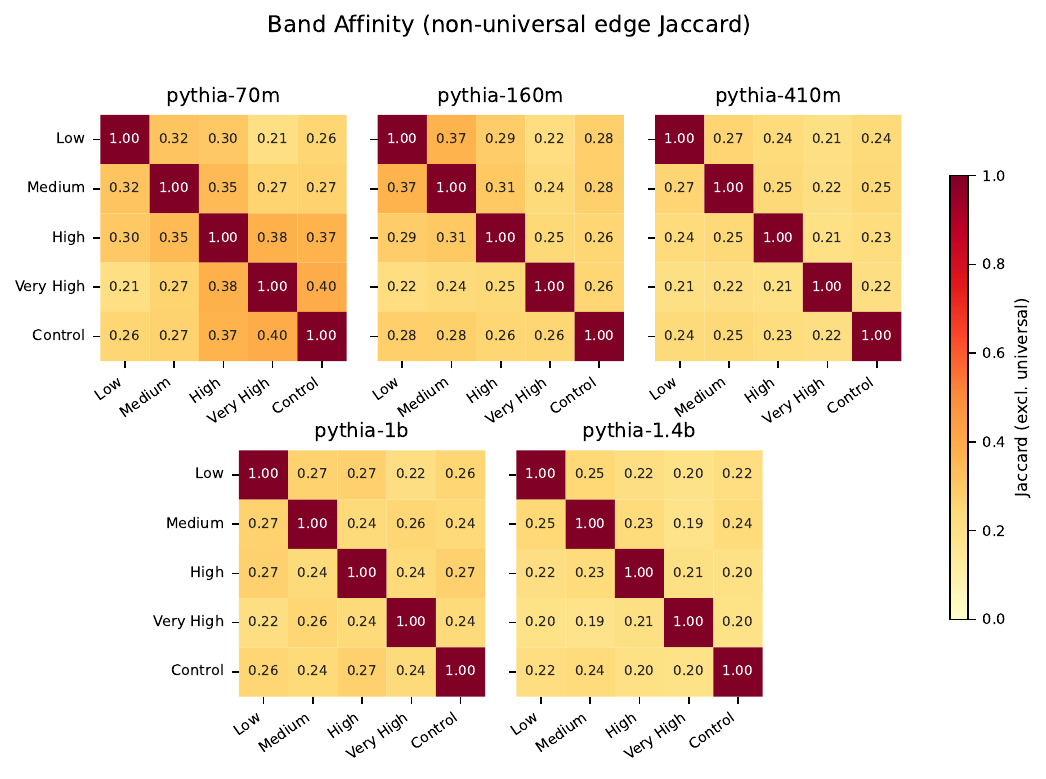}
\caption{Band affinity (non-universal edge Jaccard) per model. Diagonal entries
show self-affinity, and off-diagonal values decrease with band
distance.}\label{fig:structural_affinity}
\end{figure}

Shared non-universal structure decreases with band distance, significantly for Pythia-70m ($\rho = -0.926$, $p_{\text{BH}} = 0.033$; Figure~\ref{fig:structural_affinity}).

Low-frequency circuits contain more edges from high-frequency circuits than high-frequency circuits contain from low-frequency circuits. This difference is significant for 160m, 410m, and 1b, while Pythia-70m shows the opposite direction. The pattern parallels the functional transfer asymmetry (Appendix~\ref{app:asymmetry_detail}).

\subsection{Layer Sensitivity}
\label{app:layer_sensitivity_detail}
Neither layer sensitivity nor per-layer universal fraction correlates with depth (all $|\rho| \leq 0.38$, $p_{\text{BH}} > 0.08$).

\subsection{Head Discrimination Entropy}
\label{app:head_universality_detail}
Normalized entropy of head participation decreases with model size, from 0.907 to 0.621, indicating that head participation becomes more concentrated by frequency band in larger models.

\subsection{Graph Topology}
\label{app:graph_topology_detail}
We treat each circuit as a directed graph. Density is the fraction of possible directed edges that are present. Clustering is the average fraction of each node's neighbor pairs that are connected after ignoring edge direction. Diameter is the longest shortest path in the largest weakly connected component, and path length is the average shortest path in that component, both after ignoring edge direction. Model identity explains 93\% of the variation in density, 68\% of the variation in clustering, and 64\% of the variation in diameter ($\eta^2 = 0.93$, 0.68, and 0.64, respectively). All model effects have $p_{\text{BH}} < 10^{-7}$, while no band effects are significant (Table~\ref{tab:structural_graph_metrics}).

\begin{table}[t]
\centering\caption{Graph topology metrics per model (averaged over bands and draws).}\label{tab:structural_graph_metrics}\small
\begin{tabular}{lcccc}\toprule
Model & Density & Diameter & Clustering & Path length \\\midrule
Pythia-70m  & 0.150 & 2.6 & 0.714 & 1.71 \\
Pythia-160m & 0.079 & 3.3 & 0.703 & 1.89 \\
Pythia-410m & 0.041 & 3.9 & 0.685 & 2.02 \\
Pythia-1b   & 0.065 & 3.9 & 0.564 & 2.12 \\
Pythia-1.4b & 0.036 & 4.1 & 0.543 & 2.22 \\\bottomrule
\end{tabular}\end{table}

\subsection{Draw Stability}
\label{app:draw_stability_detail}
Universal edges are much more stable across draws than band-specific edges, which appear in only one or two of three draws. Cram\'{e}r's $V$ is 0.53--0.58, and rank-biserial $r$ is 0.91--0.93 (Table~\ref{tab:draw_stability_main} in the main text).

\subsection{Scaling and Variance Decomposition}
\label{app:structural_scaling_detail}
Table~\ref{tab:structural_scaling} summarizes model size trends.

\begin{table}[t]
\centering\caption{Structural metrics by model size (averaged over bands and draws).}\label{tab:structural_scaling}\small
\begin{tabular}{rccccc}\toprule
Size (M) & Edge~\% & Skip~\% & Head part.\ \% & Attn~\% \\\midrule
70   & 30.9 & 62.2 & 85.0 & 57.2 \\
160  & 12.3 & 80.2 & 68.2 & 56.2 \\
410  &  4.4 & 88.7 & 55.9 & 55.6 \\
1000 &  9.1 & 79.2 & 62.0 & 61.1 \\
1400 &  2.8 & 84.4 & 44.2 & 61.7 \\\bottomrule
\end{tabular}\end{table}

\paragraph{Variance decomposition.} Table~\ref{tab:structural_variance} decomposes the variance in structural metrics. Appendix~\ref{app:integration_variance} presents the combined analysis across structural, functional, and representational metrics.

\begin{table}[t]
\centering\caption{Variance decomposition ($\eta^2$) for structural metrics.}\label{tab:structural_variance}\small
\begin{tabular}{llcc}\toprule
Metric & Factor & $\eta^2$ & Var.\ (\%) \\\midrule
\multirow{4}{*}{Edge fraction}
  & Model             & 0.995 & 99.5 \\
  & Band              & 0.000 &  0.0 \\
  & Model $\times$ Band & 0.004 &  0.4 \\
  & Residual          & 0.001 &  0.1 \\\midrule
\multirow{4}{*}{Skip fraction}
  & Model             & 0.993 & 99.3 \\
  & Band              & 0.002 &  0.2 \\
  & Model $\times$ Band & 0.002 &  0.2 \\
  & Residual          & 0.003 &  0.3 \\\midrule
\multirow{4}{*}{Head part.\ rate}
  & Model             & 0.962 & 96.2 \\
  & Band              & 0.005 &  0.5 \\
  & Model $\times$ Band & 0.011 &  1.1 \\
  & Residual          & 0.022 &  2.2 \\\bottomrule
\end{tabular}\end{table}

\FloatBarrier\section{Additional Representational Results}
\label{app:representational}
This appendix details representational analyses of all 75~LSC circuits (5~models $\times$ 5~conditions $\times$ 3~draws). These analyses were not repeated on the SVA task (\autoref{sec:sva_replication}).
Statistical tests use the nonparametric framework described in \autoref{app:stat_summary}.

\subsection{Embedding Space Geometry}
\label{app:embedding_detail}
\begin{figure}[t]\centering
\includegraphics[width=\linewidth]{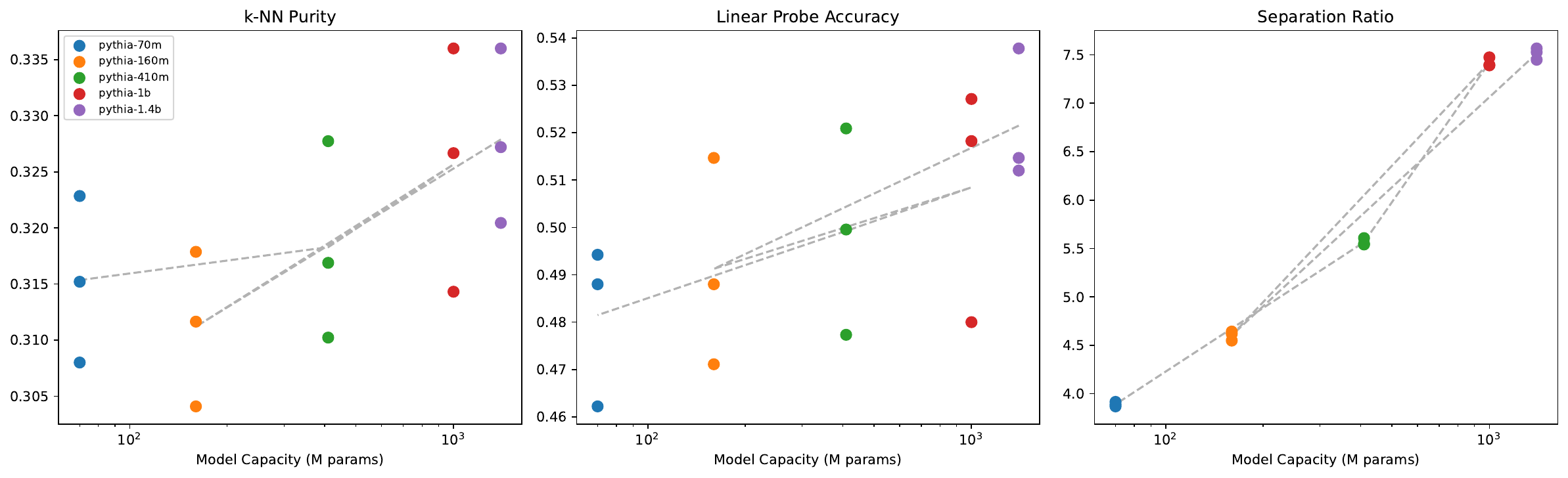}
\caption{Representational metrics at the embedding layer by model size.}\label{fig:repr_embedding_scaling}
\end{figure}

Purity based on $k$ nearest neighbors (KNN purity) is the mean proportion of a token's 10 nearest embedding neighbors that belong to the same frequency band. Probe accuracy is the accuracy under cross-validation with five folds of a linear classifier trained to predict frequency band from the embeddings. KNN purity (0.304--0.336) and probe accuracy (0.462--0.538) exceed their chance levels of 0.214 and 0.2, respectively. Both differences are significant in all five models ($p_{\text{adj}} \leq 0.0033$; Table~\ref{tab:repr_embedding}, Figure~\ref{fig:repr_embedding_scaling}).
The separation ratio is the mean distance between band centroids divided by the mean spread within bands. It increases from 3.89 in Pythia-70m to 7.51 in Pythia-1.4b.
Linear centered kernel alignment (CKA) measures the similarity between the centered embedding geometries of two bands. It ranges from 0.45 to 0.69, and adjacent frequency bands are no more similar than other pairs.
Intrinsic dimensionality estimates how many dimensions are needed to describe the embedding distribution. The estimate based on nearest neighbors, reported relative to $d_{\text{model}}$, ranges from 64--115\% in Pythia-70m to 26--47\% in Pythia-1.4b. Values above 100\% reflect estimation error rather than a dimension larger than the ambient space.

\subsection{Residual Stream Trajectories}
\label{app:residual_detail}
We report normalized layer depth as the layer number divided by the model's total number of transformer layers. Probe accuracy peaks at normalized depths ranging from 0.50 in Pythia-70m to 0.96 in Pythia-1.4b. The peak separation ratio increases from 12.1 to 19.6 (Table~\ref{tab:repr_residual}).
RSA yields $\rho = 0.77$--$0.93$.
A single linear frequency direction explains up to $R^{2} = 0.79$, with a peak from early to middle layers.
When the model is restricted to an extracted circuit, its residual stream geometry remains similar to that of the full model. Median CKA is at least 0.98 for models up to 1B and 0.93 for Pythia-1.4b, with values of approximately 0.88 for its high-frequency bands.

\begin{table}[t]
\centering
\begin{minipage}[t]{0.44\linewidth}\centering
\caption{Embedding metrics by model, averaged over draws.}\label{tab:repr_embedding}\small
\begin{tabular}{lccc}\toprule
Model & KNN Pur.\ & Probe & Sep.\ R.\ \\\midrule
Pythia-70m  & 0.315 & 0.481 & 3.89 \\
Pythia-160m & 0.311 & 0.491 & 4.60 \\
Pythia-410m & 0.318 & 0.499 & 5.57 \\
Pythia-1b   & 0.326 & 0.509 & 7.42 \\
Pythia-1.4b & 0.328 & 0.521 & 7.51 \\\bottomrule
\end{tabular}
\end{minipage}\hfill
\begin{minipage}[t]{0.54\linewidth}\centering
\caption{Peak metrics for the residual stream, averaged over draws.}\label{tab:repr_residual}\small
\begin{tabular}{lccc}\toprule
Model & Peak Probe & Norm.\ Depth & Peak Sep.\ R.\ \\\midrule
Pythia-70m  & 0.59 & 0.50 & 12.1 \\
Pythia-160m & 0.65 & 0.56 & 12.8 \\
Pythia-410m & 0.69 & 0.94 & 13.7 \\
Pythia-1b   & 0.69 & 0.94 & 16.2 \\
Pythia-1.4b & 0.74 & 0.96 & 19.6 \\\bottomrule
\end{tabular}
\end{minipage}
\end{table}

\subsection{Logit Lens Convergence}
\label{app:logit_lens_detail}
\begin{figure}[t]\centering
\includegraphics[width=0.5\linewidth]{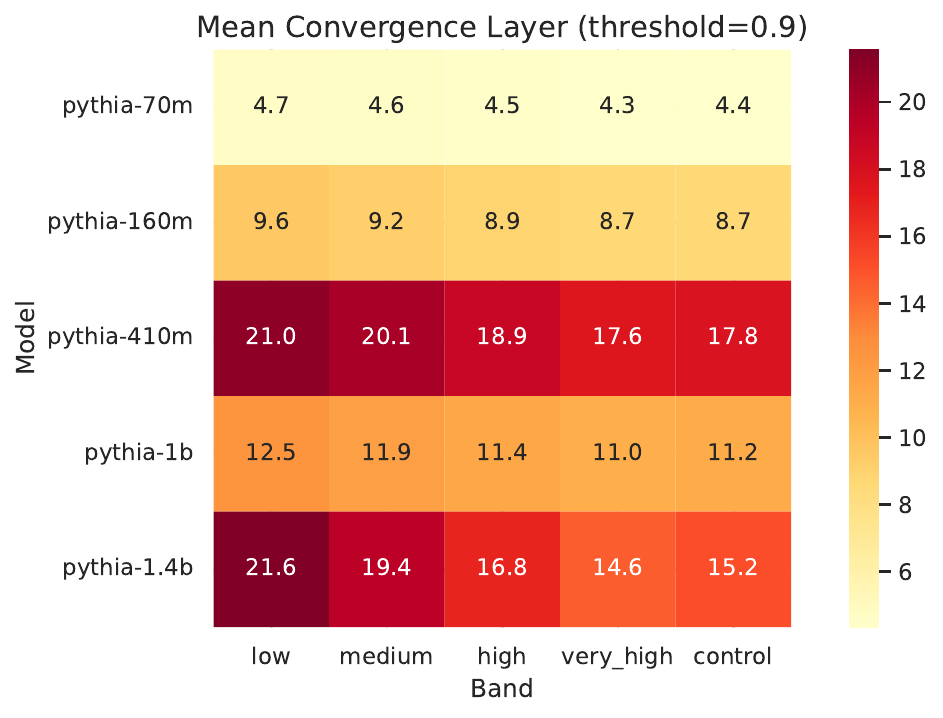}
\caption{Logit lens convergence layer by model and frequency band.}\label{fig:repr_logit_convergence}
\end{figure}

Output distributions converge at normalized depths of 0.61--0.90, with low-frequency bands converging later in all models (Table~\ref{tab:repr_logit_lens}, Figure~\ref{fig:repr_logit_convergence}).
At the final layer, the probability assigned to the correct token, $P(\text{correct})$, ranges from 0.10--0.30 in Pythia-70m to 0.58--0.71 in Pythia-1b (Figure~\ref{fig:repr_logit_prob}).

Component attribution varies by only 1.0--2.5~percentage points across bands for models ${\geq}$160M. Embedding norms do not predict convergence depth (pooled Spearman $\rho = -0.26$, $p = 0.26$; within-model mean $\rho = -0.2$).
The attention share in Table~\ref{tab:repr_logit_lens} is the absolute attention contribution to the logit for the correct token divided by the sum of the absolute attention and MLP contributions.
The convergence gap scales from $+0.4$~layers (Pythia-70m) to $+7.0$ (Pythia-1.4b).

\paragraph{Tuned Lens validation.}
\label{app:tuned_lens_detail}
To test whether the convergence delay is an artifact of logit lens misalignment, we trained Tuned Lenses~\mbox{\citep{belrose2023eliciting}} for all five Pythia models.
Each tuned lens consists of per-layer affine translators ($h \mapsto h + Wh + b$, $W \in \mathbb{R}^{d \times d}$) trained to minimize $\mathrm{KL}(p_{\text{final}} \| p_{\text{lens}})$ over 500~steps of random token sequences, then applied to the same LSC test data used in the standard logit lens analysis.
Because the translators are trained on random tokens rather than LSC sequences, a distribution mismatch may affect calibration. Nevertheless, low-frequency tokens converge later in all five models, and the ratio between the Tuned Lens and logit lens gaps ranges from 0.73 to 1.25 (Table~\ref{tab:tuned_lens_gap}).

\begin{table}[t]
\begin{minipage}[t]{0.47\linewidth}\centering
\caption{Logit lens convergence by model, reported as the range over bands after averaging across draws.}\label{tab:repr_logit_lens}\footnotesize
\begin{tabular}{lccc}\toprule
Model & Conv.\ depth & $P(\text{correct})$ & Attn share \\\midrule
Pythia-70m  & .72--.79 & .10--.30 & .44--.54 \\
Pythia-160m & .71--.80 & .41--.56 & .39--.41 \\
Pythia-410m & .73--.88 & .51--.65 & .52--.54 \\
Pythia-1b   & .68--.78 & .58--.71 & .59--.60 \\
Pythia-1.4b & .61--.90 & .55--.56 & .54--.57 \\\bottomrule
\end{tabular}
\end{minipage}\hfill
\begin{minipage}[t]{0.50\linewidth}\centering
\caption{Convergence gap between the low and very\_high bands, measured as a fraction of model depth, under the logit lens and Tuned Lens.}\label{tab:tuned_lens_gap}\footnotesize
\begin{tabular}{lccc}\toprule
Model & Logit gap & Tuned gap & Ratio \\\midrule
Pythia-70m  & $+$.088 & $+$.073 & 0.83 \\
Pythia-160m & $+$.076 & $+$.055 & 0.73 \\
Pythia-410m & $+$.151 & $+$.152 & 1.01 \\
Pythia-1b   & $+$.111 & $+$.138 & 1.25 \\
Pythia-1.4b & $+$.277 & $+$.209 & 0.76 \\\bottomrule
\end{tabular}
\end{minipage}
\end{table}

Restricting the model to an extracted circuit changes the estimated convergence point by less than 0.05 of the model's depth.

\subsection{Attention Patterns}
\label{app:attention_detail}

\paragraph{Head role classification.}
Roles are assigned from attention statistics computed on LSC inputs rather
than taken from previous work.
For each head we average the attention from the prediction position over
examples and compute the \emph{previous token attention} (the mean attention
weight on the position immediately preceding the prediction), the \emph{BOS
attention} (the mean attention weight on position~0), the \emph{induction
score} (the mean attention weight on the
previous occurrence of the current token, the prefix match position),
and the attention \emph{entropy} in bits.
The induction score measures the duplicate token pattern. The canonical
induction pattern of \mbox{\citet{olsson2022context}} attends to the
token \emph{following} the previous occurrence, which in LSC is the
answer position and is measured separately as target attention
(analyzed in Section~\ref{sec:mechanism}).
A head is labeled \texttt{previous\_token}, \texttt{bos\_sink}, or
\texttt{induction} if the corresponding statistic exceeds its threshold
(0.20, 0.30, and 0.15, respectively) and, after normalization by that
threshold, exceeds the next-highest statistic by a factor of at least 1.5.
Heads with no dominant pattern are labeled \texttt{diffuse} when their
entropy is at least 3.0~bits. If entropy is lower, the role with the highest score is
used when it passes its threshold; otherwise the head is also labeled
\texttt{diffuse}.
We separately compute an OV copy score,
$W_U[\text{target}]^{\top} W_{OV} W_E[\text{source}]$, where $W_E$ is the
embedding matrix, $W_{OV}=W_VW_O$ is the head's value--output map, and $W_U$
is the unembedding matrix. The score is averaged over 200 random token pairs
and measures whether the head maps a source token embedding toward the target
token logit direction. It is not used to assign roles.
We classify heads on LSC inputs because their functional roles depend on the
input distribution. \mbox{\citet{tigges2024llm}} document functional roles of
attention heads in Pythia models.
The tables in this appendix use the duplicate token labels. Classification by
the canonical position instead (induction statistic $:=$ target attention)
agrees for 86.9\% of heads. The 100 of 1{,}088 heads that show the canonical induction pattern
are analyzed in Section~\ref{sec:mechanism}.
The downstream enrichment result (Section~\ref{sec:mechanism}) does not depend
on the exact threshold values. Halving or doubling the three role thresholds
changes 7--35\% of labels, but \texttt{bos\_sink} depletion remains significant
in all 25 model--multiplier combinations. \texttt{previous\_token} enrichment
remains in the same direction in all 25 combinations and is significant in 20.
In the five exceptions, extreme thresholds leave too few heads with the label
for the test to retain power.
The labels are also 77--87\% stable across draws and similar across bands
(Table~\ref{tab:repr_head_roles}).

Induction scores range from 0.031 to 0.054, and copy scores are at most 0.12
(Tables~\ref{tab:repr_attention}--\ref{tab:repr_head_roles},
Figure~\ref{fig:repr_bos_layers}).
Although LSC is an induction task, these scores are low because the metric
measures attention to the previous occurrence of the current token. Heads
carrying out the copy instead attend to the token following that occurrence,
which is the answer position.
Attention to the answer position identifies canonical
heads showing the canonical induction attention pattern (100 of 1{,}088 heads across models at the
0.15 threshold), which concentrate in the universal core
(Section~\ref{sec:mechanism}).
BOS-sink heads account for 35--63\% of heads. Only 2--6\% pass the
duplicate token criterion. Role labels agree across draws for 77--87\% of heads
and are similar across bands.
Circuit extraction preserves attention entropy.

\begin{table}[t]
\begin{minipage}[t]{0.46\linewidth}\centering
\caption{Attention metrics by model (ranges over bands).}\label{tab:repr_attention}\small
\begin{tabular}{lccc}\toprule
Model & Ind.\ score & BOS attn. & Entropy \\\midrule
Pythia-70m  & .050--.055 & .285--.299 & 1.81--1.93 \\
Pythia-160m & .031--.034 & .363--.384 & 1.65--1.67 \\
Pythia-410m & .032--.034 & .456--.476 & 2.05--2.14 \\
Pythia-1b   & .036--.044 & .353--.360 & 2.53--2.61 \\
Pythia-1.4b & .027--.030 & .490--.495 & 1.94--1.96 \\\bottomrule
\end{tabular}
\end{minipage}\hfill
\begin{minipage}[t]{0.52\linewidth}\centering
\caption{Head role distribution (\%) by model. Labels follow the classification above, and \texttt{induction} uses the duplicate token criterion.}\label{tab:repr_head_roles}\small
\begin{tabular}{lcccc}\toprule
Model & BOS & Diff.\ & Prev & Ind.\ \\\midrule
Pythia-70m  & 35.4 & 35.4 & 22.9 & 6.2 \\
Pythia-160m & 56.9 & 25.7 & 13.9 & 3.5 \\
Pythia-410m & 62.8 & 28.9 &  6.2 & 2.1 \\
Pythia-1b   & 56.2 & 32.0 &  8.6 & 3.1 \\
Pythia-1.4b & 67.2 & 26.6 &  4.4 & 1.8 \\\bottomrule
\end{tabular}
\end{minipage}
\end{table}

\subsection{MLP Contributions}
\label{app:mlp_detail}
A linear probe trained on MLP outputs predicts frequency band with peak accuracy of 0.55--0.68 (Table~\ref{tab:repr_mlp}, Figure~\ref{fig:repr_mlp_sep}). Neuron selectivity is the percentage of MLP neurons whose preactivations differ across bands in a one-way ANOVA at $p < 0.01$. It averages 62--68\% across layers. The Gini coefficient measures how concentrated the absolute preactivations are across neurons, with larger values indicating that fewer neurons dominate. It ranges from 0.37 to 0.43 and peaks near the middle of the model. The MLP share in Table~\ref{tab:repr_mlp} is $\|\text{MLP output}\|/(\|\text{MLP output}\|+\|\text{attention output}\|)$, averaged across layers and examples.
Circuit extraction preserves the layerwise selectivity pattern (Spearman $\rho = 0.84$--$0.99$ for models ${\geq}160$M; $\rho = 0.60$, not significant, for the six-layer Pythia-70m).

\begin{table}[t]
\centering\caption{MLP metrics by model. Ranges of the separation ratio are calculated across draws.}\label{tab:repr_mlp}\small
\begin{tabular}{lcccc}\toprule
Model & Peak Probe & Sep.\ Ratio & MLP share & Selectivity~\% \\\midrule
Pythia-70m  & 0.55 & 12.1--18.9 & 0.57 & 66.9 \\
Pythia-160m & 0.60 & 15.5--16.4 & 0.63 & 67.6 \\
Pythia-410m & 0.66 & 25.6--28.4 & 0.62 & 63.8 \\
Pythia-1b   & 0.64 & 20.3--20.8 & 0.55 & 62.4 \\
Pythia-1.4b & 0.68 & 29.5--29.8 & 0.57 & 65.0 \\\bottomrule
\end{tabular}\end{table}

\subsection{Information-Theoretic Measures}
\label{app:info_theoretic_detail}
\begin{figure}[t]
\centering
\begin{subfigure}[t]{0.48\linewidth}
\centering
\includegraphics[width=\linewidth]{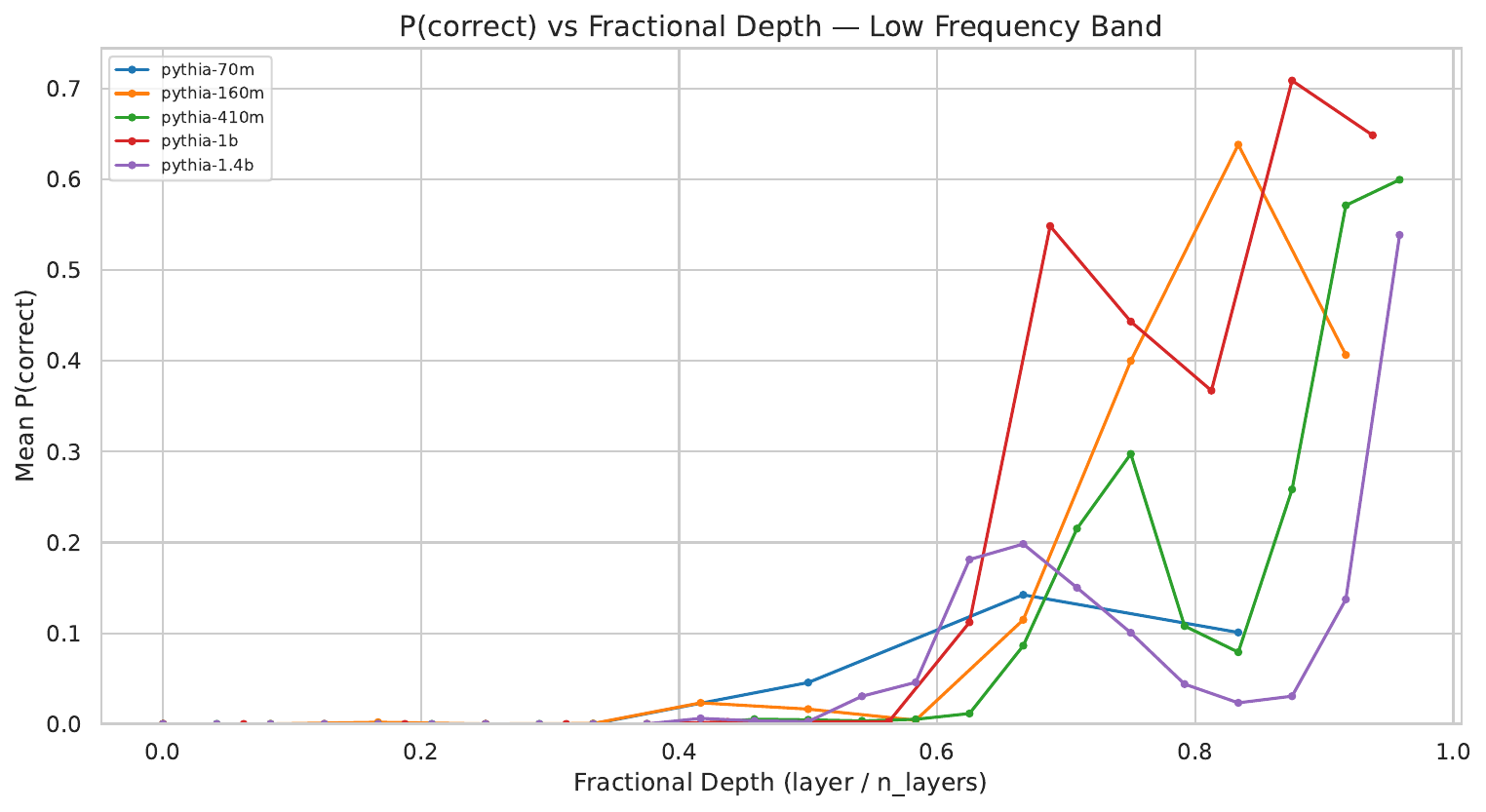}
\caption{Probability assigned to the correct token by normalized layer depth.}
\label{fig:repr_logit_prob}
\end{subfigure}\hfill
\begin{subfigure}[t]{0.48\linewidth}
\centering
\includegraphics[width=\linewidth]{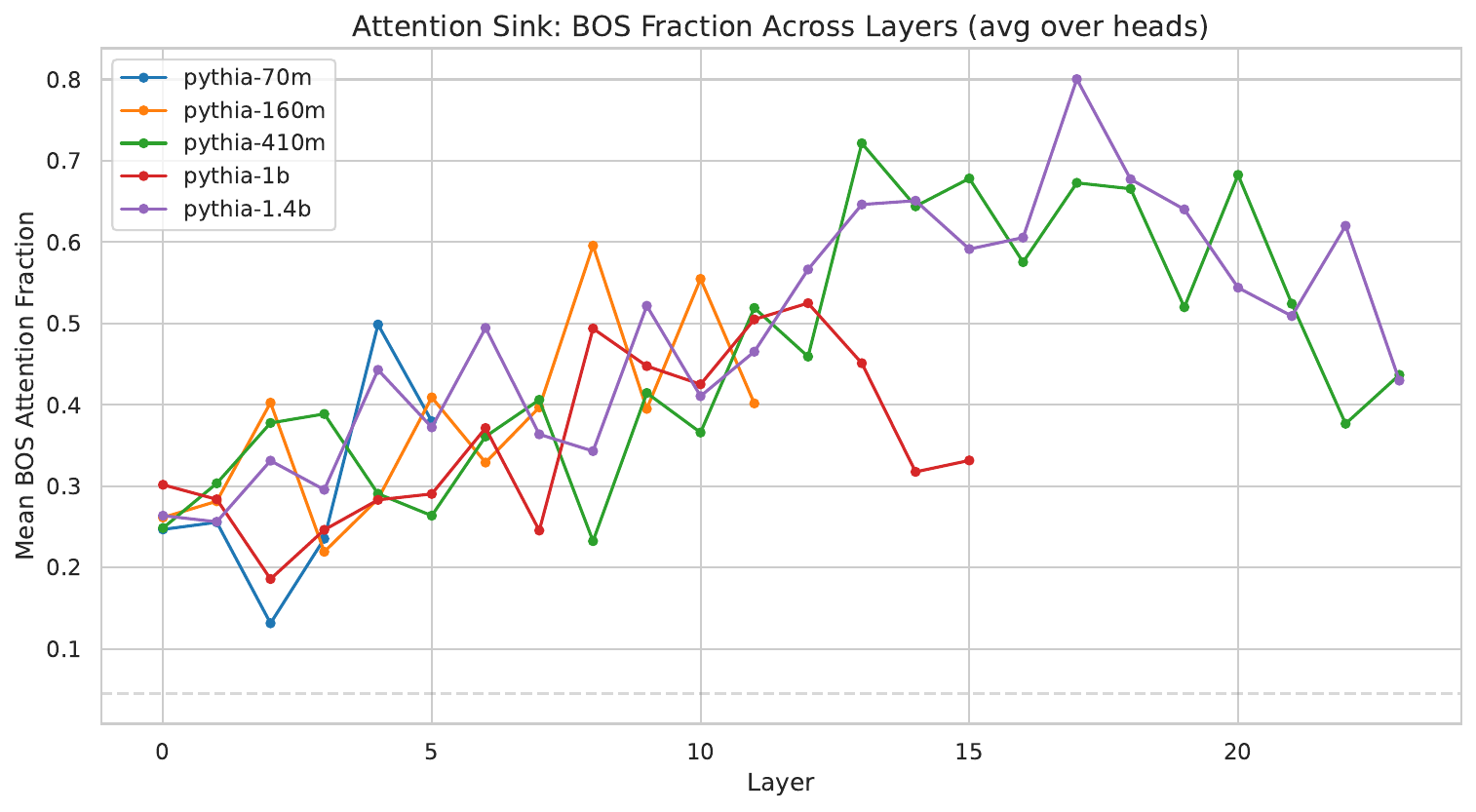}
\caption{Mean attention weight on the BOS token across layers.}
\label{fig:repr_bos_layers}
\end{subfigure}

\vspace{0.5em}

\begin{subfigure}[t]{0.48\linewidth}
\centering
\includegraphics[width=\linewidth]{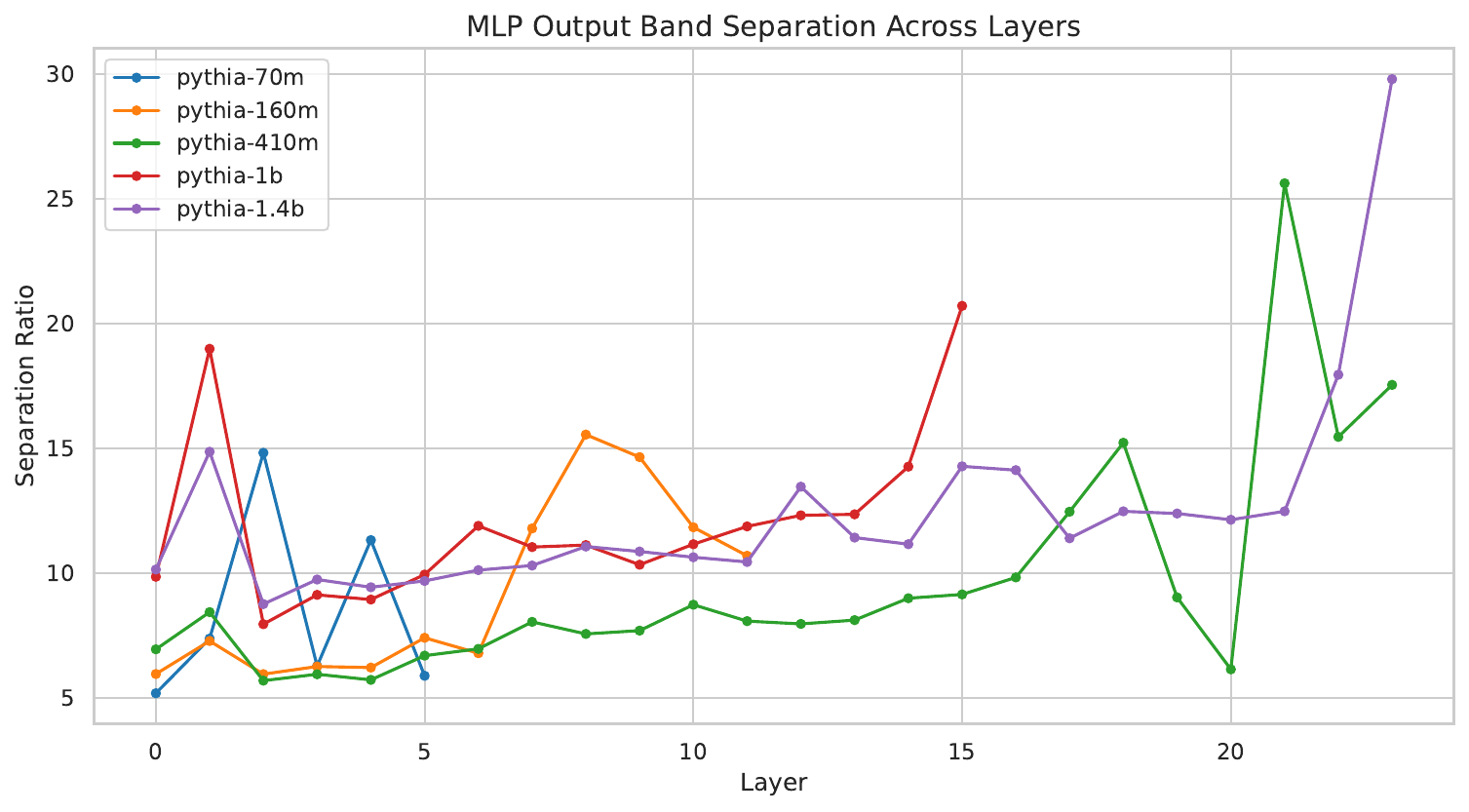}
\caption{MLP separation ratio by layer.}
\label{fig:repr_mlp_sep}
\end{subfigure}\hfill
\begin{subfigure}[t]{0.48\linewidth}
\centering
\includegraphics[width=\linewidth]{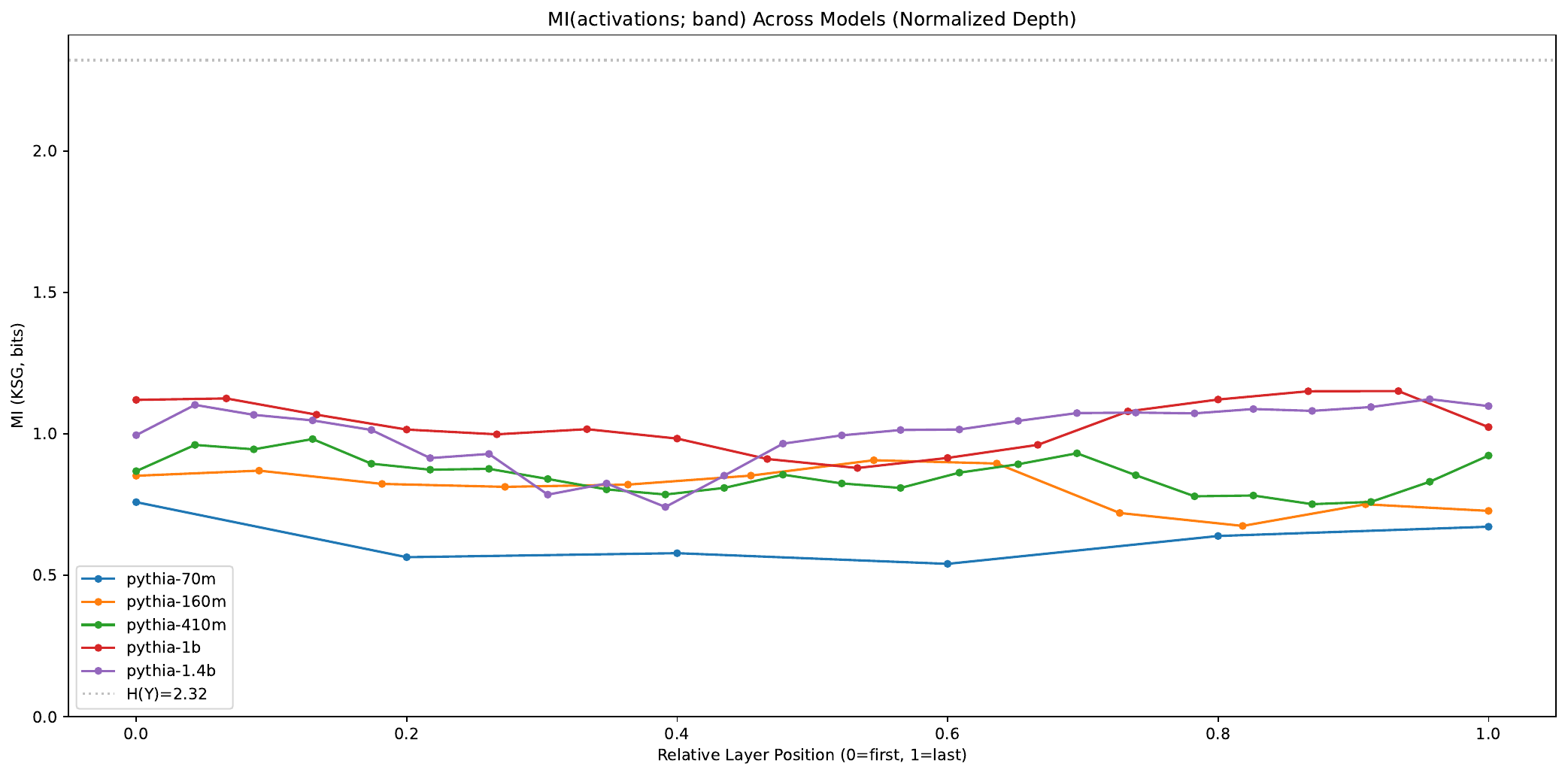}
\caption{Mutual information (KSG) by layer.}
\label{fig:repr_mi_trajectory}
\end{subfigure}
\caption{Representational metrics across layers for all five Pythia models. Panel~(a) shows logit lens convergence by normalized layer depth. Panel~(b) shows the mean attention weight on the BOS token, panel~(c) the MLP separation ratio across bands, and panel~(d) the mutual information between activations and band identity.}
\label{fig:repr_layer_grid}
\end{figure}

Mutual information (MI) measures how much the residual stream activations reveal about the frequency band label. We estimate it in bits with an estimator based on nearest neighbors for continuous activations and discrete labels after reducing the activations to 10 principal components. Peak MI increases from 0.73 in Pythia-70m to 1.16 in Pythia-1b. Coding efficiency, defined as MI divided by $d_{\text{model}}$, decreases from 0.0014 to 0.0006 bits per dimension (Table~\ref{tab:repr_info}, Figure~\ref{fig:repr_mi_trajectory}). The normalized band information in the final column of Table~\ref{tab:repr_info} is $MI/H(\text{band})$, the proportion of uncertainty in the band label removed by observing the activations.
Attention carries more MI about frequency band than MLPs (0.70--1.07 vs.\ 0.46--0.74 bits). The largest increase in MI between successive layers is at the embedding layer (0.69--1.08 bits).
Circuit extraction reduces MI by less than 10\% while preserving its trajectory across layers.

\begin{table}[t]
\centering\caption{Information-theoretic metrics by model (ranges over draws).}\label{tab:repr_info}\small
\begin{tabular}{lcccc}\toprule
Model & Peak MI & Peak Layer & Bits/dim.\ & Band info.\ \\\midrule
Pythia-70m  & 0.69--0.76 &  0   & 0.0013--0.0015 & 0.30--0.33 \\
Pythia-160m & 0.91       &  6   & 0.0012         & 0.39 \\
Pythia-410m & 0.97--1.01 & 1--3 & 0.0010         & 0.42--0.44 \\
Pythia-1b   & 1.15--1.17 & 1--14 & 0.0006        & 0.50 \\
Pythia-1.4b & 1.11--1.16 & 1--22 & 0.0005--0.0006 & 0.48--0.50 \\\bottomrule
\end{tabular}\end{table}

\subsection{Cross-Band Transfer}
\label{app:transfer_detail}
Low-frequency circuits transfer better to high-frequency bands than high-frequency circuits transfer to low-frequency bands. The asymmetry is as large as 0.237 in Pythia-70m and below 0.087 for models ${\geq}$160M.
Early-layer residual CKA is the strongest predictor ($r = 0.895$); a combined regression achieves $R^{2} = 0.990$ (CV $0.978$), with representational features adding $\Delta R^{2} = 0.130$ ($F = 35.9$, $p < 10^{-8}$; Table~\ref{tab:repr_transfer_corr}).

\begin{table}[t]
\centering\caption{Correlations between representational metrics and transfer across five models. The sample size varies from 30 to 50 by metric.}\label{tab:repr_transfer_corr}\small
\begin{tabular}{lrrr}\toprule
Metric & $r$ & $p_{\text{FDR}}$ & Sig.\ \\\midrule
Resid CKA (early)  &  0.895 & $< 10^{-16}$ & \checkmark \\
Jaccard             & $-$0.829 & $< 10^{-12}$ & \checkmark \\
Resid CKA (mid)    &  0.806 & $< 10^{-11}$ & \checkmark \\
Embedding CKA      &  0.790 & $< 10^{-8}$  & \checkmark \\
Resid CKA (final)  &  0.680 & $< 10^{-7}$  & \checkmark \\
Convergence diff   &  0.228 & 0.13 & \\
Freq distance      & $-$0.119 & 0.53 & \\\bottomrule
\end{tabular}\end{table}

\subsection{Scaling Summary}
\label{app:repr_scaling}
Table~\ref{tab:repr_scaling} summarizes the trends across model sizes.

\begin{table}[t]
\centering\caption{Representational metrics by model size, averaged over draws.}\label{tab:repr_scaling}\small
\begin{tabular}{rcccccc}\toprule
Size (M) & Emb Sep & Peak Probe & Peak Sep & Conv.\ Depth & Peak MI & Coding Eff.\ \\\midrule
70   & 3.89  & 0.59 & 12.1 & 0.75 & 0.73  & 0.0014 \\
160  & 4.60  & 0.65 & 12.8 & 0.75 & 0.91  & 0.0012 \\
410  & 5.57  & 0.69 & 13.7 & 0.80 & 0.99  & 0.0010 \\
1000 & 7.42  & 0.69 & 16.2 & 0.73 & 1.16  & 0.0006 \\
1400 & 7.51  & 0.74 & 19.6 & 0.73 & 1.13  & 0.0006 \\\bottomrule
\end{tabular}\end{table}

\subsection{Causal Interventions: Interchange Patching and Boundless DAS}
\label{app:causal_interventions}

The preceding correlational analyses show that structurally distinct circuits produce equivalent outputs (Sections~\ref{sec:generic_boost}--\ref{sec:cross_draw}) and that the universal core preserves base model representational geometry (Section~\ref{sec:representational_summary}).
This subsection provides causal evidence via interchange patching and Boundless DAS~\mbox{\citep{wu2023boundless}}, applied at the residual stream of the full (unpruned) model.

\subsubsection{Interchange Intervention (Activation Patching)}
\label{app:interchange_patching}

\paragraph{Method.}
For each ordered pair of bands, we select 100 random indices in the test set and pair the example at each index from the base band with the example at the same index from the source band. The first example supplies the base input, and the second supplies the source activation and target token.
We cache source activations at the residual stream (\texttt{hook\_resid\_post}) and run the base input with source activations patched in at a single position.
Interchange Intervention Accuracy (IIA) is the proportion of patched examples for which the model predicts the target token from the source example.
High cross-band IIA indicates that bands share a representational format. Low cross-band IIA would instead indicate representations specific to each band.

\begin{figure}[ht]
\centering
\includegraphics[width=\linewidth]{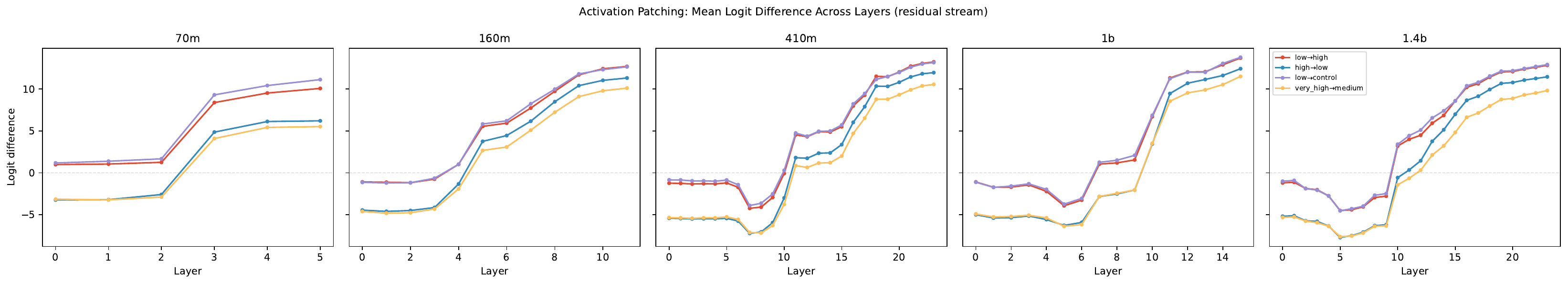}
\caption{Interchange intervention logit difference by layer (residual stream,
prediction position).
The patched model's logit for the source target rises sharply in upper
layers, with all band pairs converging to similar peak values.}
\label{fig:layer_logit_diff}
\end{figure}

\begin{figure}[ht]
\centering
\includegraphics[width=0.62\linewidth]{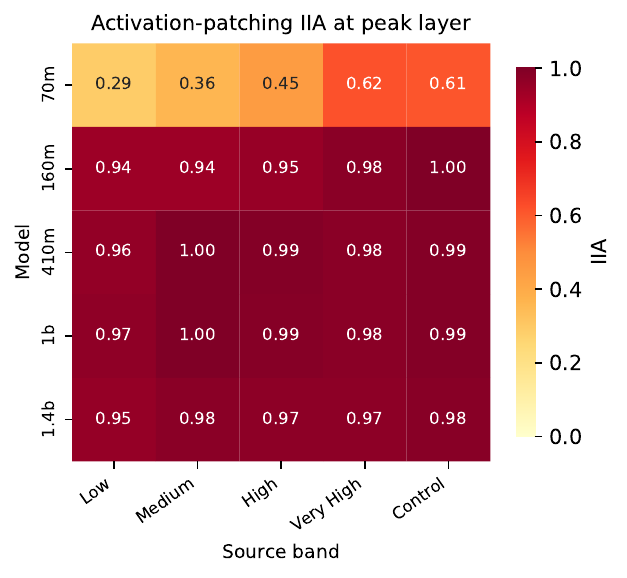}
\caption{Activation patching IIA at the peak layer, averaged over base bands. Rows show models, and columns show source bands. IIA varies by at most 0.03 across base bands for each model and source band, so the figure reports their mean. In Pythia-70m, IIA changes systematically with source band, whereas all larger models have uniformly high IIA across bands.}
\label{fig:iia_matrix}
\end{figure}

\paragraph{Layer sweep and full IIA matrix.}
We sweep all layers at the prediction position (position~21, including BOS) for
four representative band pairs. IIA rises sharply in the upper layers and
peaks at L5 for 70m, L11 for 160m, L23 for 410m, L13 for 1b, and L21 for 1.4b
(Figure~\ref{fig:layer_logit_diff}).
At each model's peak layer, we evaluate all 25~band pairs (Figure~\ref{fig:iia_matrix}).

For models ${\geq}160$M, cross-band IIA is uniformly high (Table~\ref{tab:iia_matrix_summary}): mean IIA ranges from 0.96 (Pythia-160m) to 0.99 (Pythia-1b).
The same-band vs.\ cross-band difference is at most 0.004, with no systematic advantage for same-band pairs.
Pythia-70m achieves lower IIA overall (mean 0.47), consistent with its limited task capacity.

\paragraph{Position and component analysis.}
At the peak layer, we test 14 patching positions covering the source prefix, target, distractors, and repeated prefix. Only the final prediction position (position~21) yields nonzero IIA.
Decomposing the intervention into residual stream, attention, and MLP components shows that attention heads contribute more than MLP at the peak layer, consistent with the attention-mediated copy mechanism.

\paragraph{Draw robustness.}
Repeating the evaluation at the peak layer across all three draws yields virtually identical IIA (standard deviation 0.000--0.008 for all model--band-pair combinations).

\begin{figure}[ht]
\centering
\includegraphics[width=\linewidth]{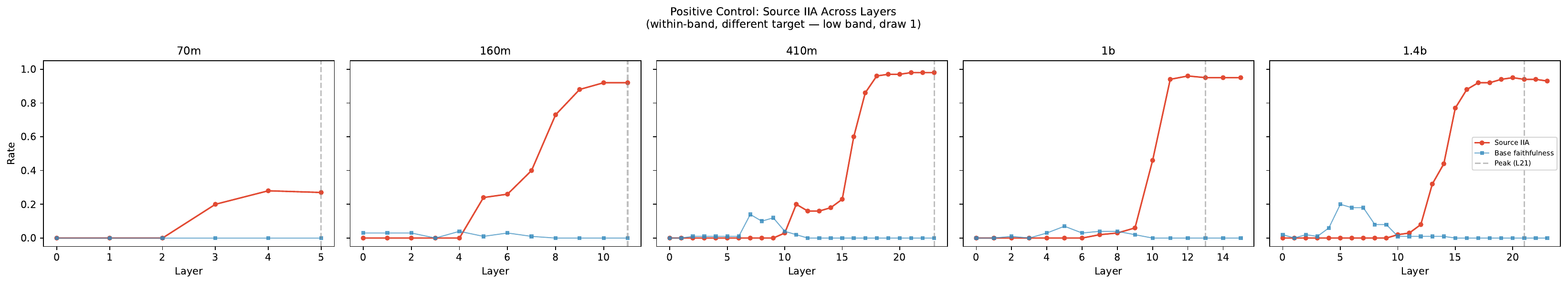}
\caption{Layer sweep of source IIA using interchange patching between examples from the same band with different target tokens. IIA is~0 at early layers, where patching should fail. At the peak layer, it rises to $0.92$--$0.98$ for models ${\geq}160$M and reaches each model's unpatched ceiling in all five models.
Dashed lines mark the peak layer for each model.}
\label{fig:layer_sweep}
\end{figure}

\paragraph{Sensitivity check across layers.}
\label{app:positive_control}
To verify that interchange patching has sufficient power to detect causal structure when it exists, we perform a layer sweep using within-band example pairs with distinct target tokens.
In all five models, source~IIA is exactly~0 at layers~0--2. At the peak layer, it reaches $0.92$--$0.98$ for models ${\geq}160$M and $0.28$ for Pythia-70m, matching that model's unpatched accuracy on the same pairs ($0.27$). At the peak layer every model recovers $96$--$104\%$ of its unpatched accuracy (Figure~\ref{fig:layer_sweep}).
The transition is sharp: for models ${\geq}160$M, IIA jumps from ${\leq}0.06$ to ${\geq}0.73$ within two to seven layers.
Comparing within-band and cross-band patching at the peak layer, the IIA gap is ${\leq}0.010$ for all models ${\geq}160$M ($\Delta = +0.006, +0.010, -0.003, +0.004$ for 160M, 410M, 1B, 1.4B respectively).
This check shows that the method detects target identity at the layers where it is present and returns zero IIA at early layers where patching should fail. The same analysis finds no evidence that band identity affects the prediction causally.
This check tests the sensitivity of interchange patching specifically. The two-route positive control evaluates the full circuit comparison pipeline separately (\autoref{sec:two_route_control}).

\subsubsection{Boundless DAS: Minimal Subspace for Band Identity}
\label{app:boundless_das}

\paragraph{Method.}
To test whether information that distinguishes frequency bands concentrates in a low-dimensional subspace, we apply Boundless DAS~\mbox{\citep{wu2023boundless}}.
Boundless DAS jointly learns a rotation matrix~$R \in \mathbb{R}^{k \times d}$ and a boundary parameter~$b \in [0, 1]$ determining the effective subspace dimension, with $L_1$~regularization encouraging the smallest sufficient subspace.
We train at each model's peak layer on the low$\to$high and high$\to$low band pairs (500~steps, $\lambda_{\text{boundary}} = 0.05$, warmup 100~steps, max candidate dimension $\min(128,\, d_{\text{model}})$).

\begin{figure}[ht]
\centering
\includegraphics[width=0.8\linewidth]{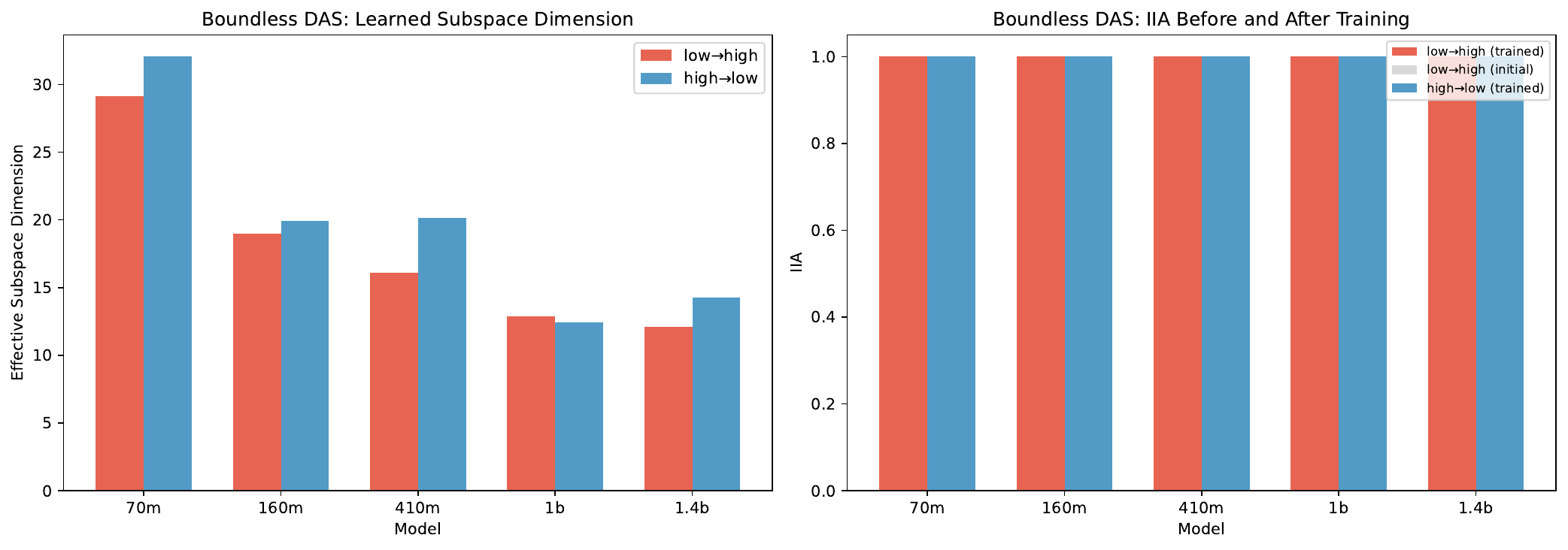}
\caption{Boundless DAS effective subspace dimension across models.
The left panel shows the absolute dimension, and the right panel shows the fraction of $d_{\text{model}}$.
Both intervention directions produce nearly identical dimensions.}
\label{fig:bdas_dimensions}
\end{figure}

\paragraph{Results.}
Boundless DAS achieves IIA~$= 1.0$ for all models and both directions.
The effective dimension decreases with model scale, from approximately 29 dimensions in Pythia-70m (5.7\% of $d_{\text{model}}$) to approximately 12 in Pythia-1.4b (0.6\%; Table~\ref{tab:bdas_dimensions}, Figure~\ref{fig:bdas_dimensions}).
The number of dimensions encoding band identity decreases from 29 to 12 ($2.4{\times}$), while the ambient dimension increases from 512 to 2048 ($4{\times}$). Together, these changes produce an approximately $10{\times}$ decrease in the fraction of the representation used to encode band identity, consistent with compression rather than proportional scaling.
Both directions (low$\to$high vs.\ high$\to$low) agree to within 1--4~dimensions for every model.

\subsubsection{Interpretation}
\label{app:causal_interpretation}

Together, interchange patching and Boundless DAS provide causal evidence that
representations are interchangeable across frequency bands. Cross-band
patching redirects predictions with at least 0.95 accuracy for models
${\geq}160$M, band identity occupies only 12--32 dimensions (0.6--6.3\% of
$d_{\text{model}}$), and neither patching direction nor the full $5{\times}5$
IIA matrix shows pair-specific asymmetry.

\FloatBarrier\section{Integration Analysis}
\label{app:integration}

Key results are summarized in Section~\ref{sec:variance_decomposition} of the main text.
Most analyses in this appendix concern all 75~LSC circuits. The unified analyses in
Tables~\ref{tab:app_integration_edge_strength} and~\ref{tab:app_integration_variance},
together with the scaling and disagreement results below, use the 60~circuits from the
four frequency bands.

\subsection{Structure--Function Correlations}
\label{app:structure_function_detail}

Edge fraction correlates negatively with accuracy ($\rho = -0.460$, $p < 10^{-4}$) and retention ($\rho = -0.241$). Skip fraction correlates positively with accuracy ($\rho = 0.608$; Table~\ref{tab:structural_function}).
Within-model correlations are generally nonsignificant except for Pythia-70m and 160m edge fraction ($p = 0.0008$ and $p = 0.0028$).

\begin{table}[t]
\centering
\caption{Pooled Spearman correlations between structure and function over all 75~LSC circuits. A $p$ value is reported for each row. Pooling across models inflates these associations, as correlations within individual models are generally not significant (see text).}
\label{tab:structural_function}
\small
\begin{tabular}{llcc}
\toprule
Structural & Functional & $\rho$ & $p$ \\
\midrule
Edge fraction & Accuracy  & $-$0.460 & $< 10^{-4}$ \\
Edge fraction & Retention & $-$0.241 & $0.037$ \\
Edge fraction & KL div.   & $-$0.708 & $< 10^{-11}$ \\
Skip fraction & Accuracy  & $+$0.608 & $< 10^{-8}$ \\
Skip fraction & Retention & $+$0.531 & $< 10^{-6}$ \\
Head part.\ rate & Accuracy  & $-$0.436 & $< 10^{-4}$ \\
Head part.\ rate & Retention & $-$0.235 & $0.042$ \\
Total edges   & Accuracy  & $+$0.594 & $< 10^{-7}$ \\
Total edges   & Retention & $+$0.541 & $< 10^{-6}$ \\
\bottomrule
\end{tabular}
\end{table}

\subsection{Structure--Representation Mantel Tests}
\label{app:mantel_detail}

Per-model Mantel tests~\mbox{\citep{mantel1967detection}} comparing structural (Jaccard) and representational (CKA) distance matrices yield nonsignificant correlations ($r = -0.41$ to $+0.39$, all $p > 0.18$).

\subsection{Similarity Triangle}
\label{app:integration_similarity_triangle}

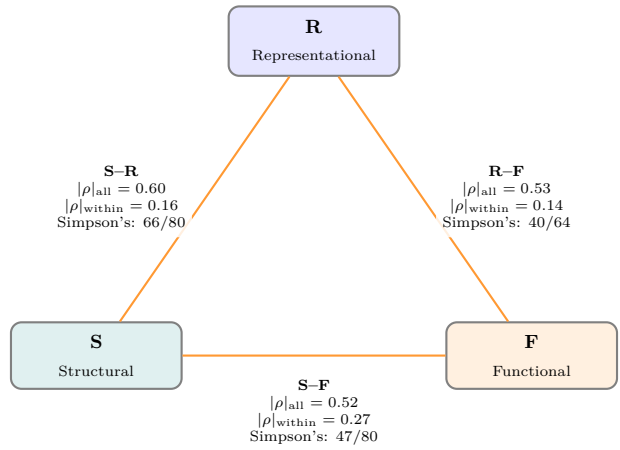
\begin{figure}[t]
\centering
\begin{tikzpicture}[
  scale=0.80, transform shape,
  vertex/.style={
    rounded corners=4pt, draw=gray, thick,
    minimum width=2.8cm,
    text width=2.4cm, align=center, font=\small,
    inner sep=6pt
  },
  elabel/.style={
    font=\scriptsize, align=center, text=black,
    fill=white, fill opacity=0.85, text opacity=1,
    rounded corners=2pt, inner sep=3pt
  },
]
  \node[vertex, fill=teal!12] (struct) at (-3.6, 0) {
    \textbf{S}\\[2pt]
    {\scriptsize Structural}
  };
  \node[vertex, fill=orange!12] (func) at (3.6, 0) {
    \textbf{F}\\[2pt]
    {\scriptsize Functional}
  };
  \node[vertex, fill=blue!10] (repr) at (0, 5.2) {
    \textbf{R}\\[2pt]
    {\scriptsize Representational}
  };

  \draw[thick, orange!80] (struct) -- (func);
  \node[elabel, below=8pt] at ($(struct)!0.5!(func)$) {
    \textbf{S--F}\\
    $|\rho|_{\text{all}} = 0.52$\\
    $|\rho|_{\text{within}} = 0.27$\\
    Simpson's: 47/80
  };

  \draw[thick, orange!80] (struct) -- (repr);
  \node[elabel, left=6pt] at ($(struct)!0.5!(repr)$) {
    \textbf{S--R}\\
    $|\rho|_{\text{all}} = 0.60$\\
    $|\rho|_{\text{within}} = 0.16$\\
    Simpson's: 66/80
  };

  \draw[thick, orange!80] (repr) -- (func);
  \node[elabel, right=6pt] at ($(repr)!0.5!(func)$) {
    \textbf{R--F}\\
    $|\rho|_{\text{all}} = 0.53$\\
    $|\rho|_{\text{within}} = 0.14$\\
    Simpson's: 40/64
  };
\end{tikzpicture}
\caption{Quantitative similarity triangle.  Mean absolute Spearman correlations computed over all metric pairs for each edge. ``Within'' is the average of the correlations computed separately for each model.  All three edges are confounded by model scale
(orange), with high Simpson's paradox rates indicating sign reversals
within individual models.  Compare with the conceptual triangle in
Figure~\ref{fig:similarity_triangle}.}
\label{fig:app_integration_triangle}
\end{figure}

Figure~\ref{fig:app_integration_triangle} and
Table~\ref{tab:app_integration_edge_strength} summarize pairwise Spearman correlations across the three triangle edges.
The structure--function edge yields the strongest individual pairwise correlations, though 47/80 pairs exhibit Simpson's paradox.
Structure and representation show the most frequent disagreement between pooled and within-model correlations, occurring in 66 of 80 metric pairs. For example, universal fraction and peak probe layer correlate strongly when models are pooled ($\rho = -0.961$) but only weakly within models (mean $\rho = -0.12$).
The representation--function edge follows a similar pattern (40/64 reversals; Figure~\ref{fig:simpsons_paradox} in main text).

\begin{table}[t]
\centering
\caption{Strongest pairwise correlations per triangle edge. Within-model $\rho$ is the mean across models in which both metrics vary. ``Simp.''\ indicates Simpson's paradox, defined here as a sign reversal within models.}
\label{tab:app_integration_edge_strength}
\small
\begin{tabular}{llccc}
\toprule
Edge & Metric pair & Overall $\rho$ & Within $\rho$ & Simp. \\
\midrule
S--F & Edge frac.\ $\leftrightarrow$ Size frac.       & $+$1.000 & $+$1.000 & No  \\
S--F & Active heads $\leftrightarrow$ $n_{\text{edges}}$ & $+$0.984 & $+$0.578 & No  \\
S--F & Head part.\ rate $\leftrightarrow$ Size frac.\ & $+$0.979 & $+$0.578 & No  \\
\midrule
S--R & Univ.\ frac.\ $\leftrightarrow$ Peak probe layer & $-$0.961 & $-$0.122 & Yes \\
S--R & Edge frac.\ $\leftrightarrow$ Peak probe layer    & $-$0.953 & $-$0.050 & Yes \\
S--R & Head part.\ rate $\leftrightarrow$ Peak probe layer & $-$0.937 & $+$0.184 & Yes \\
\midrule
R--F & Peak probe layer $\leftrightarrow$ Size frac.       & $-$0.953 & $-$0.050 & Yes \\
R--F & Peak probe acc.\ $\leftrightarrow$ Size frac.       & $-$0.906 & $+$0.003 & Yes \\
R--F & Peak efficiency $\leftrightarrow$ Circuit KL         & $-$0.887 & $-$0.036 & Yes \\
\bottomrule
\end{tabular}
\end{table}

\subsection{Unified Variance Decomposition}
\label{app:integration_variance}

Extending the structural variance decomposition (Appendix~\ref{app:structural_scaling_detail}) to all 26 unified metrics, model identity remains dominant ($\eta^2_{\text{model}} \geq 0.75$ for all metrics except completeness; Table~\ref{tab:app_integration_variance}).
Frequency band accounts for less than 5\% of variance in every metric.
Completeness is the sole exception. Replication accounts for $\eta^2 = 0.063$, while the residual accounts for 0.884.

\begin{table}[t]
\centering
\caption{Variance decomposition ($\eta^2$) for representative structural (S),
functional (F), and representational (R) metrics. Model identity explains the largest share of variance for every metric except completeness.}
\label{tab:app_integration_variance}
\small
\begin{tabular}{llcccc}
\toprule
Metric & Type & $\eta^2_{\text{model}}$ & $\eta^2_{\text{cond.}}$ & $\eta^2_{\text{repl.}}$ & Resid. \\
\midrule
Edge fraction      & S & 0.996 & ${<}0.001$ & ${<}0.001$ & 0.004 \\
Skip fraction      & S & 0.992 & 0.002 & ${<}0.001$ & 0.005 \\
Univ.\ fraction    & S & 0.997 & ${<}0.001$ & 0.002 & 0.001 \\
Base accuracy      & F & 0.911 & 0.026 & ${<}0.001$ & 0.063 \\
Circuit accuracy   & F & 0.930 & 0.030 & ${<}0.001$ & 0.039 \\
Retention ratio    & F & 0.751 & 0.043 & 0.004 & 0.202 \\
Completeness       & F & 0.018 & 0.036 & 0.063 & 0.884 \\
Convergence layer  & R & 0.944 & 0.030 & ${<}0.001$ & 0.026 \\
Peak MI probe      & R & 0.966 & ${<}0.001$ & 0.014 & 0.020 \\
Peak efficiency    & R & 0.994 & ${<}0.001$ & 0.002 & 0.005 \\
\bottomrule
\end{tabular}
\end{table}

\subsection{Agreement and Scaling Across Analysis Types}
\label{app:integration_concordance}

\begin{figure}[t]
\centering
\includegraphics[width=0.8\linewidth]{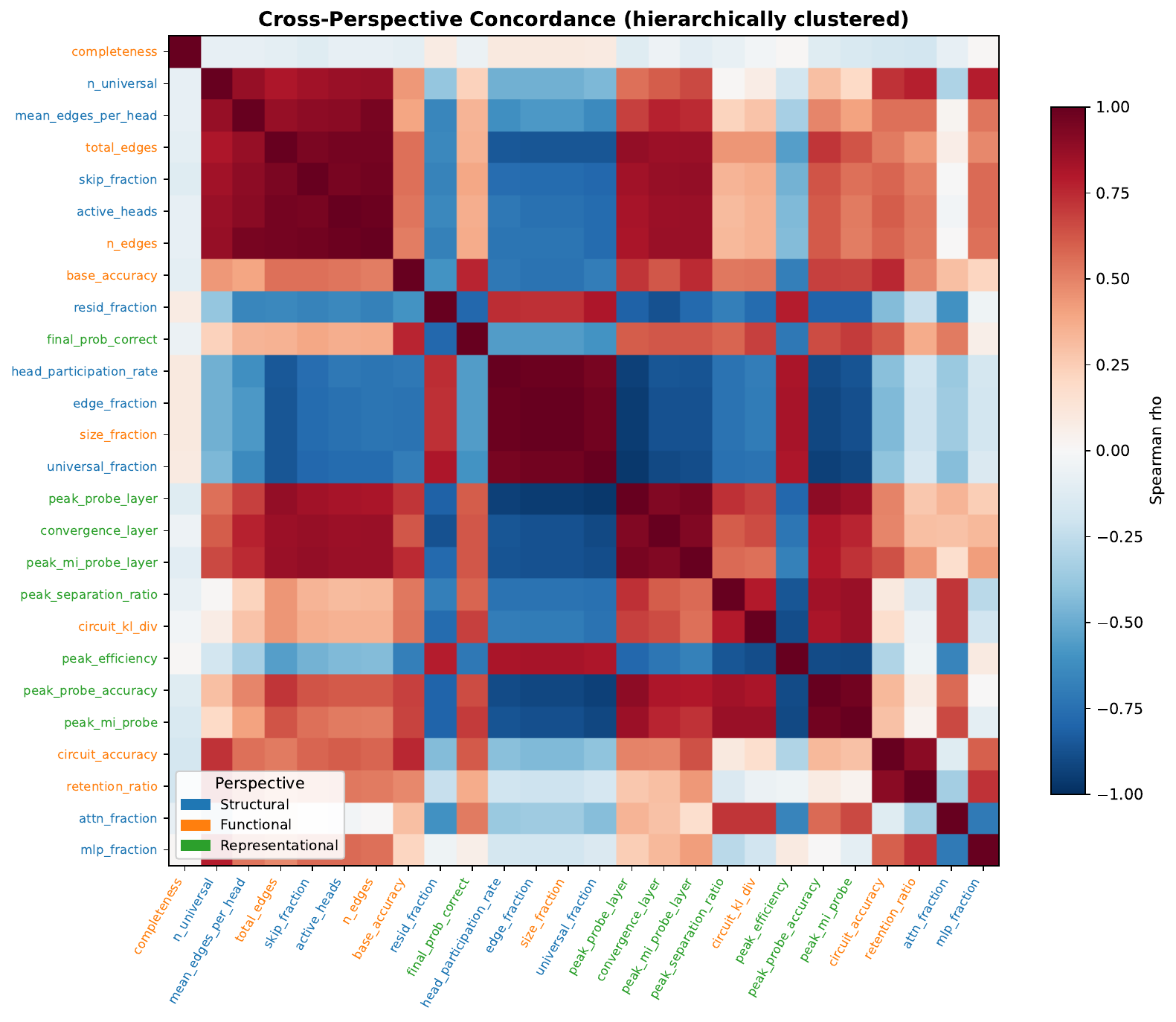}
\caption{Hierarchically clustered correlation matrix of 26 unified
metrics. The block structure aligns with the three analysis types. Blue denotes structural metrics, orange functional metrics, and green representational metrics.}
\label{fig:app_integration_concordance}
\end{figure}

Hierarchical clustering of the $26 \times 26$ Spearman correlation matrix produces three blocks corresponding to the structural, functional, and representational metrics (Figure~\ref{fig:app_integration_concordance}).

For each model, we rank the five conditions using two metrics from each analysis type. Kendall's $W$ measures agreement among those rankings, from 0 for no agreement to 1 for complete agreement. The mean is 0.202, and no frequency band consistently ranks highest or lowest across all three analysis types.
For each circuit, we also average the percentile ranks within each analysis type and sum the three pairwise absolute differences. This disagreement score averages 0.431 (range 0.136--0.730), with Pythia-1.4b showing the highest mean disagreement (0.589) and Pythia-410m the lowest (0.223).

\begin{figure}[t]
\centering
\includegraphics[width=\linewidth]{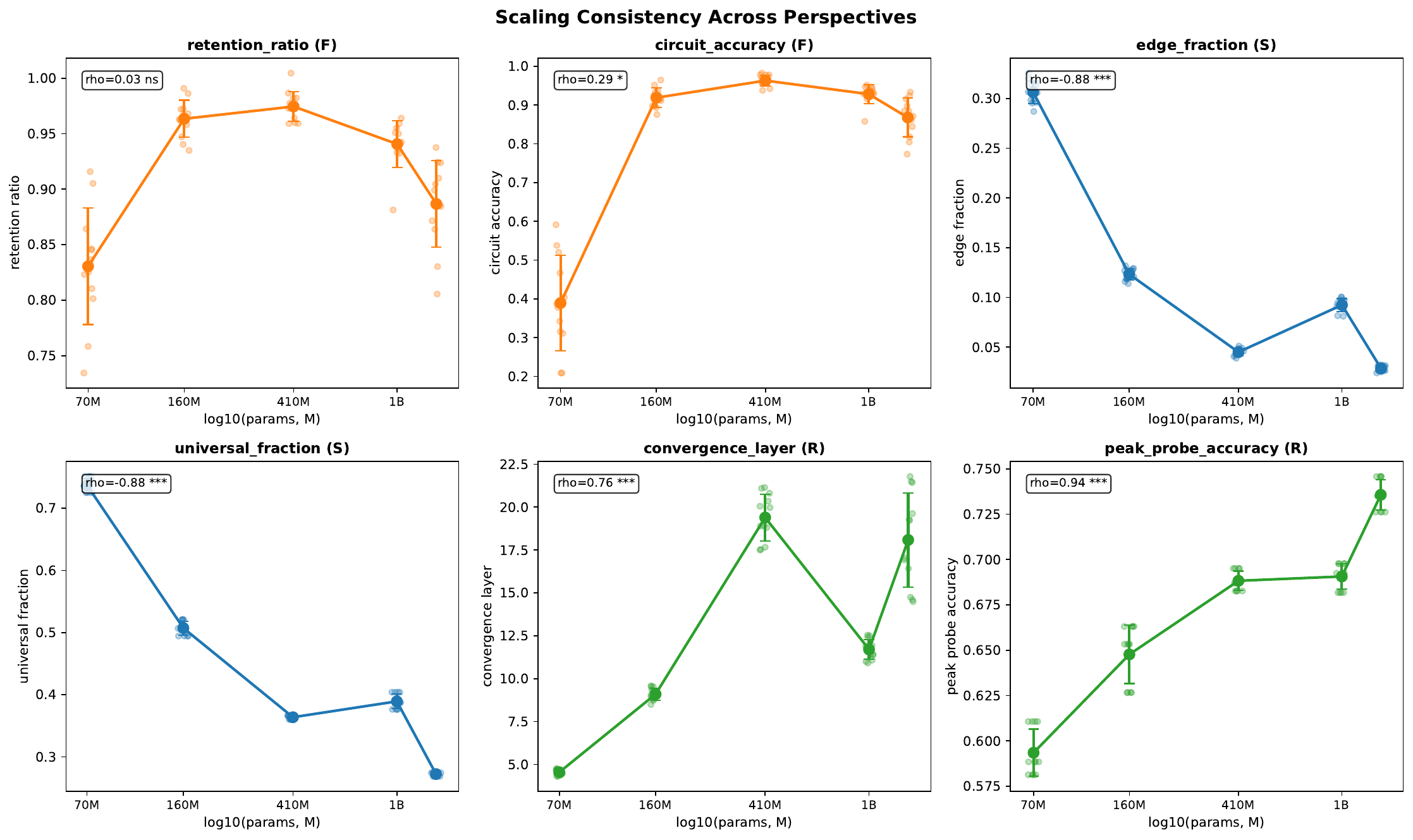}
\caption{Scaling trends across structural, functional, and representational metrics. Six representative
metrics (two per analysis type) are plotted against $\log_{10}$
model capacity. Points represent individual circuits. Lines show the mean for each model $\pm$ one standard deviation.}
\label{fig:app_integration_scaling}
\end{figure}

All three analysis types exhibit consistent scaling trends with model capacity
(Figure~\ref{fig:app_integration_scaling}).
As model size increases, functional performance improves (base accuracy $\rho = 0.689$, circuit accuracy $\rho = 0.289$), representational effects peak at later layers or become stronger (peak MI probe $\rho = 0.971$, final correct token probability $\rho = 0.715$), and circuits retain smaller edge fractions and universal cores (edge fraction $\rho = -0.882$, universal fraction $\rho = -0.884$, peak efficiency $\rho = -0.960$).
Four metrics fail to reach significance: MLP share ($\rho = -0.122$, $p = 0.354$), $n_{\text{universal}}$ ($\rho = 0.197$, $p = 0.132$), retention ratio ($\rho = 0.033$, $p = 0.804$), and completeness ($\rho = -0.094$, $p = 0.473$).

\subsection{Incremental Prediction}
\label{app:integration_prediction}

In pooled leave-one-out cross-validated regressions ($N = 60$), skip fraction is the best single structural predictor of retention ratio ($R^2 = 0.412$), while final correct token probability is the best representational predictor ($R^2 = 0.364$; Table~\ref{tab:app_integration_prediction}). The full structural and representational models reach $R^2 = 0.706$ and $0.535$, respectively.
Adding representational metrics to the full structural model yields negative incremental $R^2$ ($\Delta R^2 = -0.043$).
The per-model regressions use only 12 observations each and overfit. Their full model $R^2$ values are strongly negative, although the best single predictor retains a positive $R^2$ for Pythia-1.4b and Pythia-160m.

\begin{table}[t]
\centering
\caption{Incremental prediction of retention ratio (LOO-CV $R^2$). Per-model
regressions overfit due to small $N$.}
\label{tab:app_integration_prediction}
\small
\begin{tabular}{lcllcc}
\toprule
Scope & $N$ & Best structural & $R^2_S$ & Best repr.\ & $R^2_R$ \\
\midrule
Overall     & 60 & Skip fraction & $+$0.412 & Final prob.\ correct & $+$0.364 \\
Pythia-1.4b & 12 & Edge fraction & $+$0.452 & Convergence layer    & $+$0.650 \\
Pythia-160m & 12 & Skip fraction & $+$0.283 & Convergence layer    & $+$0.070 \\
Pythia-1b   & 12 & MLP share  & $-$0.042 & Peak efficiency      & $-$0.122 \\
Pythia-410m & 12 & MLP share  & $-$0.200 & Peak probe layer     & $-$0.288 \\
Pythia-70m  & 12 & Resid.\ share & $-$0.093 & Peak sep.\ ratio  & $-$0.258 \\
\bottomrule
\end{tabular}
\end{table}

\subsection{Statistical Testing Summary Across LSC Analyses}
\label{app:stat_summary}

All hypothesis tests across the three analysis types use BH-FDR
correction at $\alpha = 0.05$. The summaries below cover the LSC analyses.

\paragraph{Functional (Phase~1).}
Across 160 tests in 11 domains (Kruskal-Wallis, Mann-Whitney~U,
Wilcoxon signed-rank, Jonckheere-Terpstra, Spearman),
38 are significant (24\%).
The strongest effects are in random baselines (D11: 6/6) and
variance decomposition (D9: 3/3), followed by scaling (D6: 3/5),
band distance (D5: 3/6) and completeness (D8: 5/10).
The weakest are circuit quality (D2: 0/20), control vs.\ frequency (D7: 0/10),
control averaging (D10: 1/18) and generalization gap (D3: 1/12).

\paragraph{Structural (Phase~2).}
Across 371~structural hypothesis tests (229~basic, 142~deep),
the strongest effects correspond to containment asymmetry, draw stability,
and model-level variance decomposition.
Band effects are significant only for Jaccard gap and component-level
Jaccard (Appendices~\ref{app:jaccard_detail}--\ref{app:component_jaccard_detail}),
with small absolute magnitudes.

\paragraph{Representational (Phase~3).}
All 28~representational hypotheses (88 individual tests) were tested.
The results are reported in the relevant subsections of
Appendix~\ref{app:embedding_detail}--\ref{app:causal_interventions}.

\FloatBarrier\section{Targeted Ablation Studies}
\label{app:targeted}

This appendix reports ten targeted ablation analyses testing the
universal core hypothesis across all 75~LSC circuits
(5~models $\times$ 5~conditions $\times$ 3~draws). The corresponding
SVA analyses under zero ablation and using the core shared by most circuits are reported in
\autoref{sec:sva_replication}.

\subsection{Universal Core Sufficiency}
\label{app:targeted_sufficiency}

\begin{figure}[t]
\centering
\includegraphics[width=0.8\linewidth]{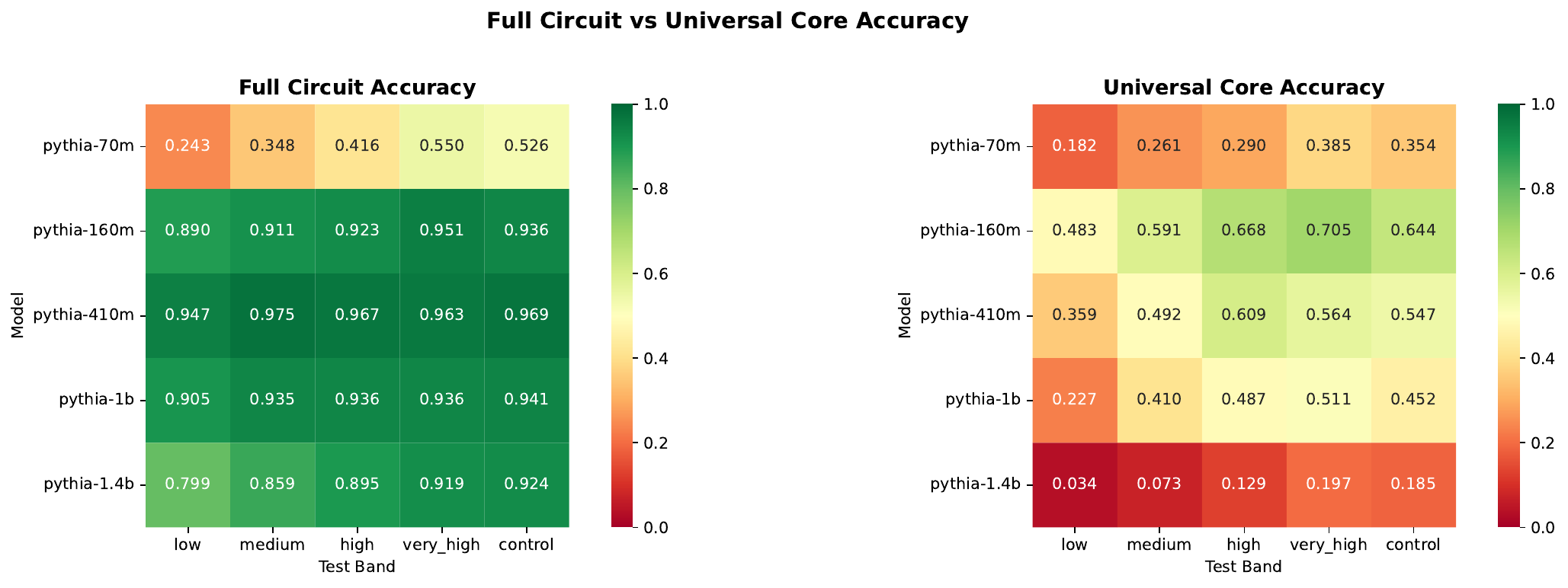}
\caption{Universal core accuracy across models and frequency bands.
Retention decreases with model scale.}
\label{fig:targeted_sufficiency_heatmap}
\end{figure}

The universal core recovers 71.4\% (Pythia-70m) to 13.7\% (Pythia-1.4b)
of full circuit accuracy
(Figure~\ref{fig:targeted_sufficiency_heatmap}, Table~\ref{tab:targeted_sufficiency}).
Band-specific edges in isolation achieve 0\% accuracy across all models.

The universal core outperforms random edge sets with the same number of edges by
209--9{,}940$\times$ (Wilcoxon $p \leq 3.3 \times 10^{-4}$ in every
model).

\begin{table}[t]
\centering
\caption{Universal core sufficiency and necessity by model (means over
bands and draws). Retention is universal core accuracy divided by full circuit accuracy. Complement accuracy is measured after retaining only the circuit complement, defined as the full circuit minus the universal core.}
\label{tab:targeted_sufficiency}
\label{tab:targeted_necessity}
\small
\begin{tabular}{lcccccc}
\toprule
Model & Full Acc & Univ.\ Acc & Retention & Band-Spec.\ Acc & Comp.\ Acc & Random Acc \\
\midrule
Pythia-70m  & 0.417 & 0.295 & 0.714 & 0.000 & 0.000 & ${<}0.001$ \\
Pythia-160m & 0.922 & 0.618 & 0.669 & 0.000 & 0.000 & ${<}0.001$ \\
Pythia-410m & 0.964 & 0.514 & 0.533 & 0.001 & 0.001 & ${<}0.001$ \\
Pythia-1b   & 0.931 & 0.417 & 0.447 & 0.000 & 0.000 & ${<}0.001$ \\
Pythia-1.4b & 0.879 & 0.124 & 0.137 & 0.001 & 0.001 & ${<}0.001$ \\
\bottomrule
\end{tabular}
\end{table}

\FloatBarrier

\subsection{Cross-Band Transfer}
\label{app:targeted_transfer}

\begin{figure}[t]
\centering
\includegraphics[width=0.8\linewidth]{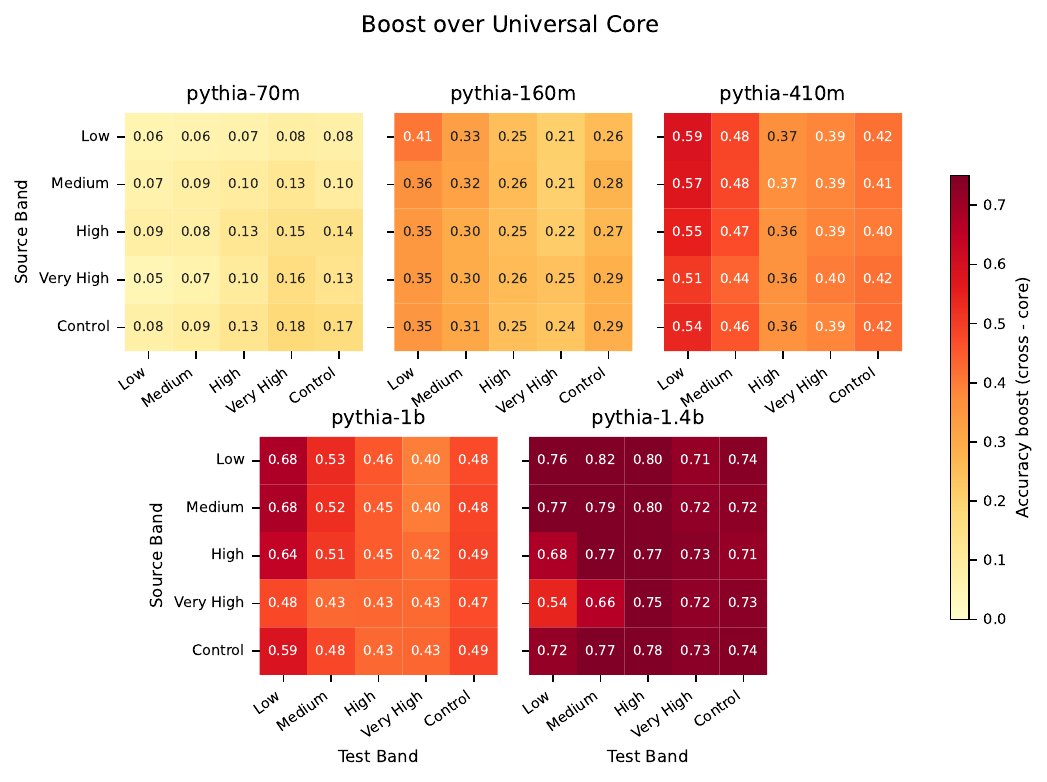}
\caption{Cross-band accuracy boost from band-specific edges, on a shared
color scale. Each panel shows one model. Rows indicate the source band whose
specific edges are added to the universal core, and columns indicate the test
band.
Similar values across each row indicate that band-specific edges help all test bands, not only the band from which the edges were extracted.}
\label{fig:boost_heatmaps}
\end{figure}

Band-specific edges transfer across bands at 81--97\% efficiency
(Table~\ref{tab:transfer_efficiency} in the main text,
Figure~\ref{fig:boost_heatmaps}).
The same-band boost significantly exceeds the cross-band boost
(all $p \leq 0.028$, $d = 0.16$--$0.51$), but the absolute advantage is only
0.016--0.029 accuracy points. This indicates a small same-band advantage
rather than strong specialization. Random edges contribute only 1--3\% boost.

\subsection{Sharing Threshold Sweep}
\label{app:targeted_threshold}

The critical threshold where accuracy reaches 95\% of the full circuit
is predominantly $k=3$ (Pythia-410m requires $k=3$--$4$;
Table~\ref{tab:targeted_threshold}).

\begin{table}[t]
\centering
\caption{Critical sharing threshold $k$ per model and band.
Target accuracy is the value at the critical $k$. Full circuit accuracy is shown in parentheses.}
\label{tab:targeted_threshold}
\small
\setlength{\tabcolsep}{3.5pt}
\begin{tabular}{l cc cc cc cc cc}
\toprule
 & \multicolumn{2}{c}{Pythia-70m} & \multicolumn{2}{c}{Pythia-160m} & \multicolumn{2}{c}{Pythia-410m} & \multicolumn{2}{c}{Pythia-1b} & \multicolumn{2}{c}{Pythia-1.4b} \\
\cmidrule(lr){2-3}\cmidrule(lr){4-5}\cmidrule(lr){6-7}\cmidrule(lr){8-9}\cmidrule(lr){10-11}
Band & $k$ & Acc (Full) & $k$ & Acc (Full) & $k$ & Acc (Full) & $k$ & Acc (Full) & $k$ & Acc (Full) \\
\midrule
low        & 3 & .231\,(.243) & 3 & .846\,(.890) & 3 & .899\,(.947) & 3 & .860\,(.905) & 3 & .759\,(.799) \\
medium     & 3 & .331\,(.348) & 3 & .866\,(.911) & 3 & .926\,(.975) & 3 & .888\,(.935) & 3 & .816\,(.859) \\
high       & 3 & .395\,(.416) & 3 & .877\,(.923) & 4 & .919\,(.967) & 3 & .889\,(.936) & 3 & .850\,(.895) \\
very\_high & 3 & .522\,(.550) & 3 & .904\,(.951) & 4 & .915\,(.963) & 3 & .889\,(.936) & 3 & .873\,(.919) \\
control    & 2 & .500\,(.526) & 3 & .889\,(.936) & 4 & .920\,(.969) & 3 & .894\,(.941) & 3 & .878\,(.924) \\
\bottomrule
\end{tabular}
\end{table}

\subsection{Core Shared by Most Bands: Per-Band Edge Composition}
\label{app:k3_composition}

The $k{\geq}3$ edge sets are similar across frequency bands
(Table~\ref{tab:k3_jaccard}). The $k{\geq}3$ majority core also recovers
similar performance on the SVA task (\autoref{sec:sva_replication}). Pairwise
Jaccard similarity exceeds the full circuit value for every model. The largest
increase occurs in Pythia-1.4b ($+$79\%), which also has the lowest
universal core retention (13.7\%). Per-band participation ratios range from
0.84 to 1.18, indicating similar band membership across conditions.

\begin{table}[ht]
\centering
\caption{Jaccard similarity between the $k{\geq}3$ edge sets for individual bands, compared with Jaccard similarity between full circuits from different bands. Values are means over three draws.}
\label{tab:k3_jaccard}
\small
\begin{tabular}{lccc}
\toprule
Model & Full circuit between-band & $k{\geq}3$ per-band & Relative increase \\
\midrule
Pythia-70m  & 0.763 & 0.876 & $+$14.8\% \\
Pythia-160m & 0.557 & 0.770 & $+$38.2\% \\
Pythia-410m & 0.430 & 0.700 & $+$62.8\% \\
Pythia-1b   & 0.465 & 0.704 & $+$51.4\% \\
Pythia-1.4b & 0.366 & 0.656 & $+$79.2\% \\
\bottomrule
\end{tabular}
\end{table}

\FloatBarrier

\subsection{Draw Variability}
\label{app:targeted_draw}

Cross-draw transfer ratios range from 0.990 to 1.017, with all 95\% CIs
spanning 1.0 (Table~\ref{tab:targeted_draw}).

Structurally, 68--73\% of Pythia-70m edges appear in all three draws, compared
with 21--27\% for Pythia-1.4b. Despite this structural variation, functional
transfer remains nearly perfect.

\FloatBarrier

\subsection{Edge and Mechanism Profiling}
\label{app:targeted_profiling}

\begin{figure}[t]
\centering
\includegraphics[width=\linewidth]{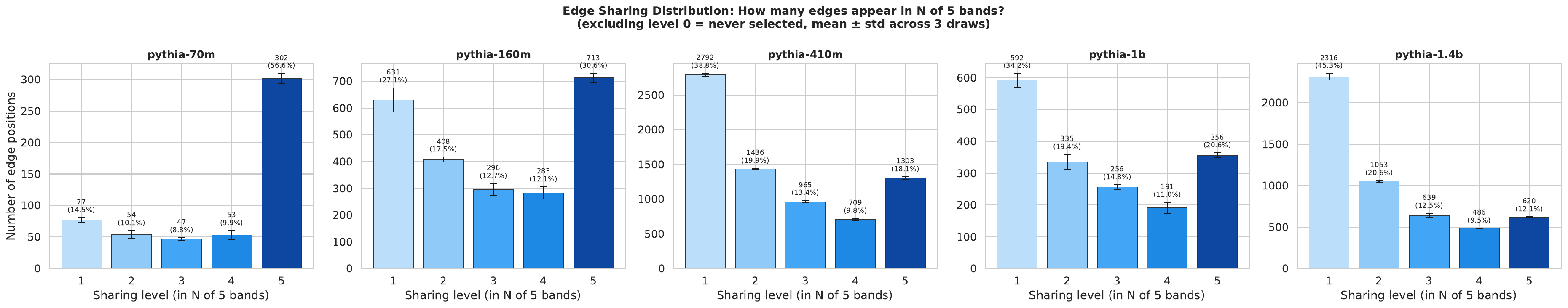}
\caption{Edge sharing distribution by model.
Smaller models have a higher fraction of edges shared by all five bands.}
\label{fig:targeted_sharing_dist}
\end{figure}

Within each band, universal edges recur in all three draws 93--98\% of the time, compared with only 1--2\% for band-specific edges (Figure~\ref{fig:targeted_sharing_dist}). When computed as a fraction over the full set of universal edges, the proportion present in all three draws is 57--82\% (Table~\ref{tab:draw_stability_main} in the main text).

\begin{figure}[t]
\centering
\includegraphics[width=\linewidth]{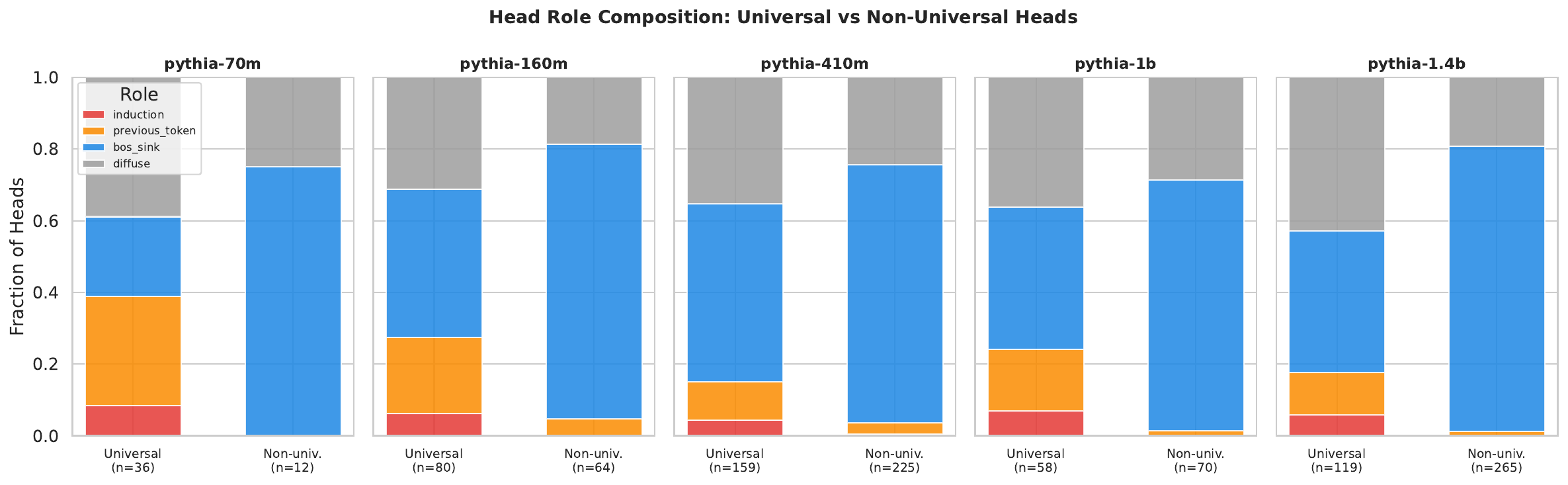}
\caption{Attention head role distribution by universality class.}
\label{fig:targeted_role_universality}
\end{figure}

The \texttt{previous\_token} role is enriched in universal heads
(OR~$= 3.7$--$14.4$, all $p \leq 0.044$; Table~\ref{tab:targeted_role_enrichment}),
while \texttt{bos\_sink} is depleted (OR~$= 0.10$--$0.38$, all $p \leq 0.0017$).
Universal head fraction decreases from 75\% in Pythia-70m to 31\% in
Pythia-1.4b, and the dominant role shifts from \texttt{diffuse} to
\texttt{bos\_sink} at larger scales (Table~\ref{tab:targeted_mechanism},
Figure~\ref{fig:targeted_role_universality}).
Table~\ref{tab:targeted_layer_mechanism} reports the same profiles for early,
middle, and late layer groups.

\begin{table}[t]
\centering
\caption{Role enrichment in universal heads (Fisher's exact test~\mbox{\citep{fisher1922interpretation}}).
Roles follow the classification in \autoref{app:attention_detail}. The \texttt{induction} rows use the duplicate token criterion, and \autoref{sec:mechanism} reports the analysis at the canonical position.
$^{*}$ denotes significance at $\alpha = 0.05$.}
\label{tab:targeted_role_enrichment}
\small
\begin{tabular}{llccl}
\toprule
Model & Role & OR & $p$ & Sig \\
\midrule
Pythia-70m  & previous\_token$^{*}$ & $\infty$ & 0.044 & Yes \\
            & bos\_sink$^{*}$       & 0.10 & 0.002 & Yes \\
            & induction             & $\infty$ & 0.563 & No \\
\midrule
Pythia-160m & previous\_token$^{*}$ & 5.49 & 0.006 & Yes \\
            & bos\_sink$^{*}$       & 0.21 & ${<}0.001$ & Yes \\
            & induction             & $\infty$ & 0.066 & No \\
\midrule
Pythia-410m & previous\_token$^{*}$ & 3.73 & 0.004 & Yes \\
            & bos\_sink$^{*}$       & 0.38 & ${<}0.001$ & Yes \\
            & induction$^{*}$       & 10.32 & 0.010 & Yes \\
\midrule
Pythia-1b   & previous\_token$^{*}$ & 14.38 & 0.002 & Yes \\
            & bos\_sink$^{*}$       & 0.28 & ${<}0.001$ & Yes \\
            & induction$^{*}$       & $\infty$ & 0.040 & Yes \\
\midrule
Pythia-1.4b & induction$^{*}$       & $\infty$ & ${<}0.001$ & Yes \\
            & previous\_token$^{*}$ & 11.64 & ${<}0.001$ & Yes \\
            & bos\_sink$^{*}$       & 0.17 & ${<}0.001$ & Yes \\
\bottomrule
\end{tabular}
\end{table}

\begin{table}[t]
\centering
\caption{Mechanism profiling summary by model. Roles follow the classification in \autoref{app:attention_detail}.}
\label{tab:targeted_mechanism}
\small
\begin{tabular}{lcllcc}
\toprule
Model & Univ.\ Heads~\% & Dom.\ Role & Dom.\ Edge & Univ.\ Stable & Band-Spec.\ Stable \\
\midrule
Pythia-70m  & 75.0 & diffuse   & attn$\to$mlp  & 97.9\% & 1.2\% \\
Pythia-160m & 55.6 & bos\_sink & attn$\to$mlp  & 95.9\% & 1.7\% \\
Pythia-410m & 41.4 & bos\_sink & attn$\to$mlp  & 94.8\% & 1.6\% \\
Pythia-1b   & 45.3 & bos\_sink & attn$\to$attn & 93.9\% & 1.1\% \\
Pythia-1.4b & 31.0 & diffuse   & attn$\to$attn & 93.2\% & 0.8\% \\
\bottomrule
\end{tabular}
\end{table}

\begin{table}[t]
\centering
\caption{Mechanism profiles grouped into early, middle, and late layers. Universal edge and head fractions are means within each group. The dominant role is the mode.}
\label{tab:targeted_layer_mechanism}
\small
\begin{tabular}{llcccc}
\toprule
Model & Position & Layers & Univ.\ Edge~\% & Univ.\ Head~\% & Dom.\ Role \\
\midrule
Pythia-70m  & early  & 0--1   & 0.79 & 0.75 & bos\_sink \\
            & middle & 2--4   & 0.48 & 0.79 & diffuse \\
            & late   & 5      & 0.68 & 0.62 & bos\_sink \\
\midrule
Pythia-160m & early  & 0--3   & 0.30 & 0.46 & bos\_sink \\
            & middle & 4--8   & 0.26 & 0.68 & bos\_sink \\
            & late   & 9--11  & 0.40 & 0.47 & bos\_sink \\
\midrule
Pythia-410m & early  & 0--7   & 0.13 & 0.21 & bos\_sink \\
            & middle & 8--16  & 0.15 & 0.65 & bos\_sink \\
            & late   & 17--23 & 0.22 & 0.35 & bos\_sink \\
\midrule
Pythia-1b   & early  & 0--5   & 0.24 & 0.40 & bos\_sink \\
            & middle & 6--10  & 0.22 & 0.53 & bos\_sink \\
            & late   & 11--15 & 0.20 & 0.45 & bos\_sink \\
\midrule
Pythia-1.4b & early  & 0--7   & 0.15 & 0.30 & bos\_sink \\
            & middle & 8--16  & 0.11 & 0.43 & bos\_sink \\
            & late   & 17--23 & 0.09 & 0.16 & bos\_sink \\
\bottomrule
\end{tabular}
\end{table}

\subsection{BOS-Sink Ablation}
\label{app:bos_sink_ablation}

We compare removal of all BOS-sink edges with removal of the same number of
randomly selected edges (Table~\ref{tab:bos_ablation}). Accuracy decreases by
67 versus 89 percentage points in Pythia-160m and by 19 versus 93 percentage
points in Pythia-410m.

\begin{table}[ht]
\centering
\caption{Accuracy after removing all BOS-sink edges compared with removing the same number of random edges. Values are means across five conditions.}
\label{tab:bos_ablation}
\small
\begin{tabular}{lccccc}
\toprule
Model & Full circuit & $-$BOS-sink & $\Delta_{\text{BOS}}$ & $-$Random & $\Delta_{\text{rand}}$ \\
\midrule
Pythia-160m & 89.9\% & 22.4\% & $-$67.5 & 0.9\% & $-$89.0 \\
Pythia-410m & 96.0\% & 76.7\% & $-$19.3 & 3.4\% & $-$92.6 \\
\bottomrule
\end{tabular}
\end{table}

BOS-sink edges contribute substantially but not preferentially to any
frequency band. Accuracy decreases by 59--77 percentage points across the five
conditions for 160m and by 13--25 points for 410m. The model size dependence
is consistent with smaller models having fewer redundant pathways.
Memory constraints prevented evaluation on Pythia-1b and Pythia-1.4b, so we
do not test whether the effect continues to decrease at those scales.
BOS-sink edges likely suppress competing attention
signals~\mbox{\citep{xiao2023efficient}} rather than propagating target
information. Their contribution is similar across frequency bands.

\FloatBarrier

\subsection{Scaling Synthesis}
\label{app:targeted_scaling}
\label{app:scaling_detail}

Table~\ref{tab:scaling_synthesis} in the main text summarizes the preceding
metrics across model sizes.

\FloatBarrier

\subsection{Complement Ablation}
\label{app:targeted_necessity}

The complement circuit (all edges not in the universal core) achieves
0\% accuracy across all models and bands, showing necessity
(Table~\ref{tab:targeted_necessity}).
Random edge sets of equal size also yield 0\%.

\FloatBarrier

\subsection{Algorithm Preservation}
\label{app:targeted_algorithm}

\begin{figure}[t]
\centering
\includegraphics[width=\linewidth]{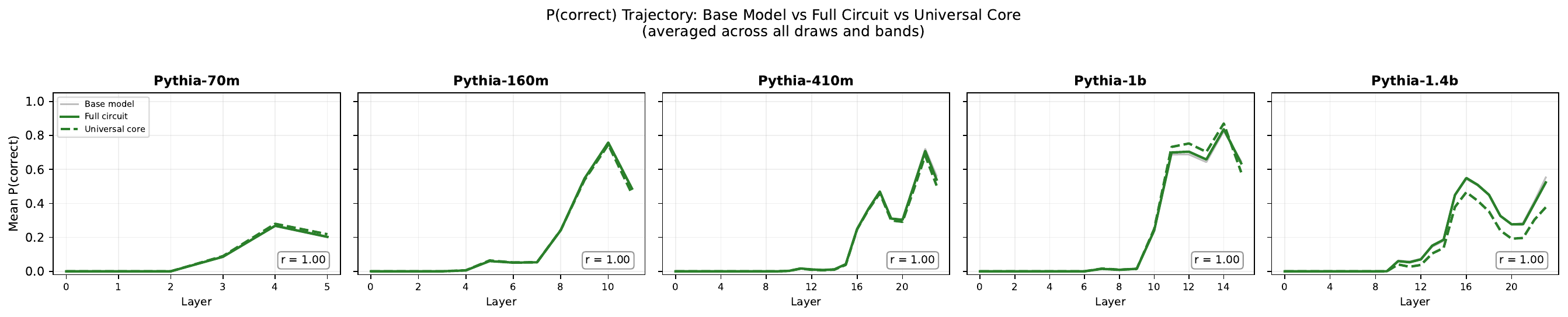}
\caption{Logit lens trajectories for the universal core and full circuit. The
universal core closely follows the full circuit, with the largest amplitude
reduction in Pythia-1.4b.}
\label{fig:targeted_trajectory}
\end{figure}

Logit lens trajectories of the universal core closely match those of the full
circuit, with $r = 0.99$--$1.00$
(Figure~\ref{fig:targeted_trajectory}, Table~\ref{tab:targeted_algorithm}).

The universal core shows no convergence delay ($-0.4$ to $0.0$~layers), and
its final probability of the correct token is 0.72--1.07 times that of the
full circuit. This is consistent with the core implementing the same
computation rather than a weaker version.

\begin{table}[t]
\begin{minipage}[t]{0.48\linewidth}\centering
\caption{Accuracy on data from the same draw compared with data from a different draw. Values are band means with 95\% confidence intervals.}
\label{tab:targeted_draw}\footnotesize
\begin{tabular}{lccc}
\toprule
Model & Same & Cross & Ratio [CI] \\
\midrule
Pythia-70m  & .417 & .413 & .990\;[.94,\,1.05] \\
Pythia-160m & .922 & .921 & .998\;[.99,\,1.01] \\
Pythia-410m & .964 & .963 & .999\;[.99,\,1.01] \\
Pythia-1b   & .931 & .934 & 1.00\;[.99,\,1.01] \\
Pythia-1.4b & .879 & .894 & 1.02\;[1.0,\,1.03] \\
\bottomrule
\end{tabular}
\end{minipage}\hfill
\begin{minipage}[t]{0.48\linewidth}\centering
\caption{Algorithm preservation (band and draw means).
Convergence shift is the difference in layers between the universal core and the full circuit, with negative values indicating earlier convergence. Amplitude ratio is the universal core amplitude divided by the full circuit amplitude.}
\label{tab:targeted_algorithm}\footnotesize
\begin{tabular}{lccc}
\toprule
Model & Pearson $r$ & Conv.\ Shift & Amp.\ Ratio \\
\midrule
Pythia-70m  & 1.000 & $0.00$ & 1.07 \\
Pythia-160m & 0.999 & $0.00$ & 0.93 \\
Pythia-410m & 1.000 & $0.00$ & 0.94 \\
Pythia-1b   & 0.997 & $-0.40$ & 0.92 \\
Pythia-1.4b & 0.993 & $-0.40$ & 0.72 \\
\bottomrule
\end{tabular}
\end{minipage}
\end{table}

\subsection{Performance Gap Decomposition}
\label{app:targeted_gap}

The gap between universal core and full circuit grows with scale
(0.12--0.76 accuracy points; Table~\ref{tab:targeted_gap}).
Including $k{\geq}3$ edges closes 99--105\% of this gap;
band-specific edges contribute negligibly.
Source-level accuracy exceeds edge-level accuracy by 0.22--0.85 points because
source-level evaluation preserves unselected outgoing edges.

\begin{table}[t]
\centering
\caption{Performance gap decomposition by model (means over bands).
Edges shared by at least three bands close more than 99\% of the performance gap for every model.}
\label{tab:targeted_gap}
\small
\begin{tabular}{lcccccc}
\toprule
Model & Full Acc & Univ.\ Acc & Gap & Closed ($k{\geq}3$) & Source Acc & Edge Acc \\
\midrule
Pythia-70m  & 0.417 & 0.295 & 0.122 & 99.3\% & 0.513 & 0.295 \\
Pythia-160m & 0.922 & 0.618 & 0.304 & 100.9\% & 0.954 & 0.618 \\
Pythia-410m & 0.964 & 0.514 & 0.450 & 100.9\% & 0.989 & 0.514 \\
Pythia-1b   & 0.931 & 0.417 & 0.513 & 105.3\% & 0.991 & 0.417 \\
Pythia-1.4b & 0.879 & 0.124 & 0.756 & 105.5\% & 0.974 & 0.124 \\
\bottomrule
\end{tabular}
\end{table}

\subsection{Robustness Under Zero and Mean Ablation}
\label{app:zero_ablation}

Under zero ablation~\mbox{\citep{olsson2022context}}, the same-band advantage
vanishes. The absolute effect size is at most 0.18 in every model, and the
direction reverses in Pythia-410m and Pythia-1b
(Table~\ref{tab:zero_ablation} in the main text). The corresponding SVA
analysis, including a direct comparison across ablation protocols, is
reported in \autoref{sec:sva_replication}. Absolute accuracy drops
substantially under zero ablation. Models ${\geq}$410M fall below 0.4\%
accuracy because zeroing 71--95\% of edges moves activations off the data
manifold~\mbox{\citep{NEURIPS2023_34e1dbe9, yu2024functional}}. Pythia-70m
retains cross-band transfer accuracies of 0.13--0.61, and Pythia-160m retains
accuracies of 0.02--0.25. All transfer entries for models ${\geq}410$M are at
most 0.02. Despite this collapse, the difference between same-band and
cross-band boosts is at most 0.012 accuracy points and at most 0.001 for three
models. This result indicates that the small same-band advantage under
resample ablation depends on the resampled activations rather than on genuine
specialization. For models ${\geq}$410M, source-level zero ablation retains
35--90\% accuracy, whereas edge-level accuracy falls to at most 0.5\%. This gap
shows that preserving unselected outgoing edges makes source-level evaluation
more permissive.

Mean ablation replaces each ablated activation with its mean at that token
position over the corrupt input distribution. The models retain circuit
accuracy of $0.14$--$0.68$ on the extraction band, whereas zero ablation
causes near-total collapse in the larger models. Under mean ablation, the same-band advantage is not
statistically significant in any model (Wilcoxon $p = 0.0938$--$0.211$),
compared with $p \leq 0.028$ in every model under resample ablation. Cross-band
transfer efficiency remains 66--96\% (Table~\ref{tab:zero_ablation} in the main
text). One caveat is that the universal core collapses under mean ablation for
models ${\geq}$410M, with accuracy at most 0.005 compared with 0.28--0.32 for
70m and 160m. The boosts for models ${\geq}$410M are therefore measured
relative to a core baseline close to zero.

\FloatBarrier\section{The Excluded Very-Low Band: Extension Without Length Matching}
\label{app:very_low}

Section~\ref{sec:data} explains why very\_low cannot enter the design in which bands are matched by length.
Here we extract and evaluate circuits using the unmatched pool, which
applies only the lowercase filter. The results below therefore do not control
for character length. All
other bands in the analyses of this section use their canonical matched
datasets, and no number reported elsewhere in the paper changes.

\paragraph{Extraction.}
We run the unmodified discovery pipeline (Section~\ref{sec:pipeline}) on the
unmatched very\_low pool. We use the same fixed per-model thresholds~$\tau^*$
and extraction settings as for all other bands.
This yields 15 circuits (5~models $\times$ 3~draws), all of which pass the
pipeline's necessity gate. Circuit sizes fall within the range of the other
frequency band circuits for each model, including 378--385 edges for
Pythia-70m and 1{,}278--1{,}334 for Pythia-160m. Base model accuracy on
very\_low is $0.871$--$0.982$ for models from 160m to 1.4b. Pythia-70m is the
exception, with accuracy of $0.258$--$0.280$, consistent with the low-accuracy
regime documented in Section~\ref{sec:results}.

\paragraph{Cross-band transfer (resample ablation).}
Extending the transfer analysis to include very\_low gives the same qualitative
result. In the $6\times6$ grid, the very\_low circuits also transfer broadly
across bands (Table~\ref{tab:very_low_transfer}). Transfer efficiency relative
to the six-band universal core is 92--96\% for models from 160m to 1.4b and
81\% for 70m. On tests from the other four frequency bands, very\_low circuits
retain 93--98\% of the corresponding same-band circuit accuracy for models
from 160m to 1.4b and 85\% for 70m. In the other direction, circuits from the
four bands retain 85--100\% accuracy on the very\_low test. The same-band
advantage is positive in all three draws for every model. The Wilcoxon values
follow the main protocol but are exploratory because the 18 observations
across bands and draws are clustered within only three draws. A sign test at
the draw level cannot produce $p<0.125$ with three draws.

\begin{table}[t]
\centering
\caption{Cross-band transfer under resample ablation with the unmatched very\_low band included ($6\times6$ grid). TE is transfer efficiency relative to the universal core shared by all six bands. The retention columns compare very\_low circuits evaluated on the core bands with core-band circuits evaluated on very\_low, reported as mean [min--max]. The Wilcoxon $p$ value is exploratory because the 18 band--draw observations are clustered within three draws. The very\_low pool contains 97 tokens and is not matched by length.}
\label{tab:very_low_transfer}
\small
\begin{tabular}{lccccc}
\toprule
Model & TE & Cohen's $d$ & $p$ & vl$\to$core ret. & core$\to$vl ret. \\
\midrule
Pythia-70m  & 81.3\% & 0.56 & .0019 & 0.85 [0.69--1.11] & 1.00 [0.86--1.10] \\
Pythia-160m & 91.9\% & 0.36 & .0001 & 0.93 [0.87--1.00] & 0.95 [0.84--1.04] \\
Pythia-410m & 96.2\% & 0.27 & .0003 & 0.98 [0.94--1.01] & 0.93 [0.83--0.99] \\
Pythia-1b   & 93.0\% & 0.42 & $<$.0001 & 0.95 [0.81--1.02] & 0.87 [0.55--1.04] \\
Pythia-1.4b & 94.2\% & 0.84 & .0003 & 0.93 [0.87--1.02] & 0.85 [0.61--0.96] \\
\bottomrule
\end{tabular}
\end{table}

\paragraph{Intervention robustness.}
Mean ablation gives a similar transfer pattern
(Table~\ref{tab:very_low_mean}). Mean same-band accuracy is 0.15--0.60, the
same-band advantage is positive in all three draws for every model, and
transfer efficiency is 64--87\%. These values are close to the five-band range
of 66--96\% under the same intervention (\autoref{app:zero_ablation}), so most
of the reduction relative to resample ablation also occurs without very\_low.
One exception occurs for the Pythia-160m very\_low circuit, whose accuracy
decreases more under mean ablation. Its retention is therefore low on tests
from the other frequency bands, although circuits from those bands still
perform well on the very\_low test.
Under zero ablation, the $6\times6$ grid reproduces the collapse documented in
\autoref{app:zero_ablation}. For models ${\geq}$410M, edge-level accuracy falls
to 0.001--0.004 for every band, including very\_low. Zero ablation therefore
cannot distinguish transfer within and across bands at those scales. Where
accuracy is measurable (70m, 160m), the six-band same-band minus cross-band
boost difference stays at most $0.017$ accuracy points (Cohen's
$d \leq 0.23$, both bounds attained by 70m), consistent with the five-band
result.

\paragraph{Cross-method check.}
Repeating the EAP-IG comparison with very\_low included does not change the
cross-method result (\autoref{app:method_comparison}). Across the six-band grid
and all ten size multipliers, the maximum same-band advantage per model is
$+1.3$ to $+4.6$ percentage points. The maximum therefore remains at most 4.6
percentage points after adding very\_low. At circuit sizes matched to ACDC,
EAP-IG circuits for very\_low have Jaccard overlap of 0.29--0.62 with the
corresponding ACDC circuits. This is comparable to the range of 0.26--0.62 for
the other frequency bands.

\begin{table}[t]
\centering
\caption{Cross-band transfer with very\_low included under mean ablation. Same-band accuracy is the mean circuit accuracy on data from the same band under this intervention. The Wilcoxon $p$ value is exploratory for the reason given in Table~\ref{tab:very_low_transfer}.}
\label{tab:very_low_mean}
\small
\begin{tabular}{lccccc}
\toprule
Model & TE & Cohen's $d$ & $p$ & Draws pos. & Same-band acc. \\
\midrule
Pythia-70m  & 84.7\% & 0.40 & .041 & 3/3 & 0.406 \\
Pythia-160m & 87.3\% & 0.32 & .123 & 3/3 & 0.603 \\
Pythia-410m & 76.0\% & 0.57 & .106 & 3/3 & 0.384 \\
Pythia-1b   & 81.3\% & 0.69 & .065 & 3/3 & 0.428 \\
Pythia-1.4b & 64.5\% & 0.63 & .041 & 3/3 & 0.149 \\
\bottomrule
\end{tabular}
\end{table}

\paragraph{Scope.}
The representational analyses and the interchange
interventions remain five-band. The SVA replication has no very-low condition
because none of the 97 available tokens form an eligible singular--plural pair
(\autoref{sec:sva_replication}).

\FloatBarrier\section{Second Task Family: Subject-Verb Agreement Replication}
\label{app:sva_replication}

\autoref{sec:sva_replication} reports the SVA replication, and
\autoref{app:sva_generation} gives the task and data construction. Here we
report sensitivity checks following the LSC protocol, along with the controls
and data integrity checks.

\paragraph{Statistical testing.}
The primary SVA analyses use directional permutation tests that block by draw,
operate on integer success counts, and apply Benjamini--Hochberg correction
across the five models (\autoref{sec:sva_replication}). We use this procedure
for the comparison across bands matched by length, the zero and mean ablation
checks, the two comparisons of the advantage on the extraction band across ablation
protocols, and the sensitivity analysis that excludes mislabeled noun pairs.
The last analysis uses exact rational accuracies. For comparison with the LSC
protocol, we also report secondary Wilcoxon signed-rank tests across conditions
and draws, using exact sign-flip enumeration with midranks.

\paragraph{Sensitivity checks.}
The test applied to each condition and draw following the LSC protocol uses
exact sign-flip enumeration of the Wilcoxon signed-rank statistic with
midranks. It is significant for Pythia-1b and Pythia-1.4b ($p = 0.025$ and
$p = 0.020$) when all five SVA conditions are included. However, that analysis
includes the unmatched very-high and control conditions, so frequency remains
confounded with token length. Restricting the same test to
observations from bands matched by length gives no significant result in any model ($p = 0.21$--$0.42$).
Both tests across conditions and draws treat dependent observations as
independent. We therefore report them only for comparison with the LSC
protocol and use the blocked permutation test in
\autoref{sec:sva_replication} as the primary analysis.

\paragraph{Controls and data integrity.}
Random circuits with the same number of edges score 0--0.004 across all five models. A
cross-task check shows that the circuits are specific to their respective
tasks. LSC circuits score at most 0.022 on SVA for 160m, 1b, and 1.4b, at most
0.071 for 410m, and at most 0.173 for 70m. The SVA circuit scores 0.0 on the
LSC test, where the LSC reference circuit scores 0.911--0.964 for 160m--1.4b
and 0.511 for 70m. This result shows that the evaluation distinguishes
circuits used for different tasks. It is not a positive control for
specialization within a task, which is addressed by the two-route control
(\autoref{sec:two_route_control}).
Repeating ACDC produces non-identical masks in all five models, with edge
counts changing from 31 to 27, 88 to 99, 140 to 144, 153 to 153, and 405 to
429 from 70m to 1.4b. This variation is already represented by the use of
multiple draws in the main design. Finally, we identified a small labeling
error in the frozen token pools. Two noun pairs (\textit{brow}, \textit{disc})
were included even though the canonical classifier labels their singular
forms as stems rather than common nouns. The error affects 195 of 22{,}500
examples (0.9\%), 34 of 3{,}375 test examples, and 21 of the 1{,}920 examples
in ACDC's effective first batches. The affected prompts are still well-formed
agreement items. A sensitivity analysis excludes the affected test prompts
while leaving the circuits unchanged. The Pythia-1.4b transfer efficiency
changes only from 0.9756 to 0.9752. Repeating the primary blocked permutation
test on the filtered test sets for bands matched by length again identifies Pythia-410m as
the only significant model (gap $+0.020$, $q = 0.023$), with all other models
at $q \geq 0.081$. Exact rational accuracies are used because the filtered
test sets contain 219--225 examples.
Because discovery used the frozen datasets, the training set
contamination is retained in the discovered circuits.

\FloatBarrier\section{Pipeline Positive Control: Within-Task Mechanistic Contrast}
\label{app:pipeline_positive_control}

The main text reports evidence that LSC uses a single mechanism and presents
the pipeline results on the learned two-route task
(\autoref{sec:two_route_control}).
This appendix records the two controlled LSC variants in detail
(\autoref{app:reverse_copy}, \autoref{app:zero_distractor}) and the
full task, training, route separation, and pipeline specification for the
two-route control (\autoref{app:two_route_control}).

\subsection{LSC Is Single-Mechanism: Reverse-Copy and Zero-Distractor Variants}
\label{app:lsc_single_mechanism}

To determine whether LSC provides a second mechanism for comparison within the task, we tested two variants on Pythia-160m that preserve the task family while changing its computational requirements.

\subsubsection{Reverse-Copy LSC}
\label{app:reverse_copy}

We placed the target token~$T$ \emph{before} the source prefix:
\begin{align*}
\text{Standard:}\quad & S_1\;S_2\;S_3\;S_4\;S_5\;\mathbf{T}\;R_1{\ldots}R_{10}\;S_1\;S_2\;S_3\;S_4\;S_5 \;\to\; \text{predict } T \\
\text{Reverse:}\quad  & \mathbf{T}\;S_1\;S_2\;S_3\;S_4\;S_5\;R_1{\ldots}R_{10}\;S_1\;S_2\;S_3\;S_4\;S_5 \;\to\; \text{predict } T
\end{align*}
Standard induction copies from offset~$+1$, whereas the reverse-copy variant requires offset~$-5$.
Pythia-160m achieves \textbf{0.0\% top-1 accuracy}, compared with 98.7\% on standard LSC, and predicts~$R_1$ (the offset~$+1$ token) in 94.2\% of cases. This result indicates that the model relies on forward-copy induction heads for this task.
Since the model cannot solve this variant, no meaningful circuit can be extracted and it cannot serve as a positive control.

\subsubsection{Zero-Distractor LSC}
\label{app:zero_distractor}

We removed all distractor tokens, reducing the sequence from 21 to 11 tokens:
\begin{align*}
\text{Standard (21 tokens):}\quad      & S_1\;S_2\;S_3\;S_4\;S_5\;\mathbf{T}\;R_1{\ldots}R_{10}\;S_1\;S_2\;S_3\;S_4\;S_5 \\
\text{Zero-distractor (11 tokens):}\quad & S_1\;S_2\;S_3\;S_4\;S_5\;\mathbf{T}\;S_1\;S_2\;S_3\;S_4\;S_5
\end{align*}
The motivation is that standard LSC requires compositional induction to bridge the 15-position gap between~$T$ and the prediction point, whereas zero-distractor places~$T$ only 5~positions away, potentially enabling a single head that copies directly without previous token composition.
Pythia-160m solves zero-distractor LSC at 88.4\% accuracy (vs.\ 98.7\% standard).

We ran ACDC at the same threshold ($\tau^* = 6.31 \times 10^{-4}$), yielding circuits of 1{,}498 (zero-distractor) and 1{,}392 (standard) edges, and computed a $2 \times 2$ cross-condition transfer matrix (Table~\ref{tab:positive_control_main} in the main text).
The transfer matrix is uniform rather than block diagonal. Cross-condition transfer efficiency is 96.7\%, comparable to the cross-band accuracy ratio of 97.8\% (\autoref{app:method_comparison}).
Jaccard similarity between the two circuits is 0.539, below the between-band Jaccard of 0.557 for Pythia-160m (Table~\ref{tab:structural_jaccard} in the main text). The circuits are therefore more structurally different but more functionally interchangeable, extending the phantom specialization pattern to sequence length.
The standard circuit outperforms on both test conditions, consistent with the asymmetric transfer pattern in Section~\ref{sec:asymmetric_transfer}. The circuit extracted from the condition with higher base accuracy (98.7\%) transfers better to the zero-distractor condition than the zero-distractor circuit transfers to the standard condition.

Both variants provide evidence that Pythia uses a single LSC mechanism. The reverse-copy variant fails completely, while the zero-distractor variant transfers well to the standard condition, indicating that both successful conditions rely on forward copying through induction heads.
A positive control for the pipeline therefore requires a task with two mechanisms, which we construct next.

\subsection{Learned Two-Route Positive Control}
\label{app:two_route_control}

\paragraph{Task.}
We construct a task in which one cue token chooses between two routes.
Each sequence has the fixed form $[\mathrm{BOS}]\;[\mathrm{CUE}]\;x_1\,{\ldots}\,x_6\;[q]$, where the~$x_i$ are distinct random content tokens (512-token content vocabulary) and the query~$q$ is one of them; the model predicts at the final position.
Under $\mathrm{CUE_A}$ the model must output the token following the earlier occurrence of~$q$, that is~$x_{i+1}$, the same computation LSC isolates.
Under $\mathrm{CUE_B}$, the model must output~$\mathrm{PERM}[q]$, a fixed vocabulary permutation held constant across training. This condition requires a memorized mapping that ignores the prompt.
The two conditions provide a simplified version of the competition between copying and recall proposed for a natural positive control. Retrieval relies on attention, whereas the memorized mapping relies on MLP lookup.
Paired prompts for the two conditions are realized from one shared base pool per draw and differ \emph{only} in the cue token, so ``condition'' plays exactly the role ``frequency band'' plays in the main experiments.
The two-way evaluation compares each rule's answer with the answer produced by the other rule, so the incorrect option identifies the competing condition.
One asymmetry is inherent to the contrast and disclosed: $\mathrm{CUE_A}$ answers appear in the prompt while $\mathrm{CUE_B}$ answers do not.

\paragraph{Models and route separation check.}
We train ten small transformers from scratch that differ only in the training
seed. Each model has 4~layers, $d_{\text{model}} = 128$, 4~heads,
$d_{\text{head}} = 32$, $d_{\text{mlp}} = 512$, GELU MLPs, layer normalization,
learned positional embeddings, context length~9, and a 515-token vocabulary
comprising BOS, two cue tokens, and 512~content tokens.
Training minimizes cross-entropy at the final position on batches of
256~examples generated with an even cue mixture. We use AdamW with learning
rate $10^{-3}$ and weight decay~0.01 for at most 8{,}000~steps. After a
200-step linear warmup, the learning rate decays linearly to one tenth of its
initial value. Training stops after step~2{,}000 once both conditions exceed
$0.995$ top-1 accuracy.
Seven of the ten reach the prespecified accuracy gate (${\geq}0.99$ top-1 on both conditions, on 4{,}000 held-out examples per condition).
Before circuit discovery, each eligible model must pass a route separation check independent of the pipeline. On held-out data, we separately apply mean ablation to all attention heads and all MLPs in layers~1--3 using component groups specified in advance. Ablating one group must reduce one condition by at least 0.5 while leaving the other within 0.1 of baseline, with the opposite pattern for the other group.
Six of the seven gated models pass; for example, in one qualified model, ablating attention drops $\mathrm{CUE_A}$ accuracy from 1.00 to 0.00 while $\mathrm{CUE_B}$ stays at 1.00, and ablating MLPs drops $\mathrm{CUE_B}$ from 1.00 to 0.20 while $\mathrm{CUE_A}$ stays at 0.99.
The pattern is confirmed on a separate validation split for all six.
The seventh eligible model fails on both splits because its memorized route also uses attention, so we exclude it. We apply the pipeline only to models for which the independent check verifies route separation.
In this task and training setup, six of the seven models that learned both rules showed the independently tested route separation; learned routing does not always separate at the component level, consistent with the degeneracy perspective in the main text.

\paragraph{Pipeline.}
We then apply the same pipeline to the six models that passed the check. It uses the same ACDC implementation, threshold grid, and \texttt{minimal\_acceptable} selection rule as the main experiments. Corruption resamples the content tokens and query under the same cue. We evaluate three conditions ($\mathrm{CUE_A}$, $\mathrm{CUE_B}$, and a 50/50 mixed control) with three draws each using the same structural, transfer, and universal core analyses.
Detection criteria were specified in advance: cross-route transfer efficiency below~0.5, within-condition cross-draw transfer at least~0.8, and a gap of at least~0.15 between within-condition and cross-condition Jaccard.
All six models met these criteria. The aggregate results and comparison with
the frequency band transfer efficiencies are reported in
\autoref{sec:two_route_control} (Table~\ref{tab:two_route_control}).

\paragraph{Scope.}
This is a learned positive control, not a planted edge benchmark. The two routes emerge under ordinary training and are identified afterwards by an independently specified intervention, so the control does not test recovery of ground-truth edges.
The check shows route separation at the level of attention and MLP groups, not at the level of individual edges.
The remaining scope limitations are stated in Section~\ref{sec:limitations}.

\FloatBarrier\section{Few-Shot Demonstrations: A Study of Accuracy}
\label{app:fewshot}

This appendix reports an evaluation of accuracy for few-shot variants of the
copying task.
This exploratory study tests behavior only. Because we do not extract
circuits, it does not test whether phantom specialization persists under
few-shot prompting.

\paragraph{Strict variant using demonstrations only.}
In a first variant, two demonstrations of the form
$a_1{..}a_4\,a_1{..}a_4$ precede a query $c_1{..}c_4$ whose copy has not
started, so the answer ($c_1$) is determined only by the few-shot
pattern.
No model solves this task reliably across bands (top-1 accuracy 0--16\%
through Pythia-1b, with two-way accuracy 0.44--0.66; Pythia-1.4b is
partially competent), and the disadvantage for rare tokens remains.
Because this fails the minimum base accuracy requirement for circuit extraction
in this paper (Section~\ref{sec:base_performance}), no circuit
variant was run.

\paragraph{Paired demonstration experiment.}
The main experiment uses the shortened copy query
$c_1\,c_2\,c_3\,c_4\,c_1\,c_2$, in which the model predicts~$c_3$ after
seven tokens including BOS. We compare three conditions with the same query,
target, and two-way distractor ($c_4$). These conditions use no prefix
($k{=}0$), two structured copying demonstrations ($k{=}2$), or a
scrambled prefix matched by length ($k{=}2s$). The scrambled prefix permutes the
16 demonstration tokens while fixing the separator positions, thereby removing
the repetition structure while preserving length, token multiset, and query
position.
Each entry pools three independent prompt draws of 225~examples ($n{=}675$).
The 12 token identities sampled for each prompt are distinct and come from the
specified band pools, although their occurrences are deliberately repeated.
Table~\ref{tab:fewshot_pilot} reports top-1 accuracy for every model, band, and condition.
Because the bare $k{=}0$ query contains six non-BOS tokens rather than the 21
used in the main experiment, its absolute accuracies are lower than those in
Section~\ref{sec:base_performance}. The ordering across bands is nevertheless
similar, with \texttt{very\_low} and low both below control in all five models
on the pooled draws.

\begin{table}[ht]
\centering
\caption{Top-1 accuracy in the few-shot experiment by model, band, and condition, pooled over three independent prompt draws. Each combination of model, band, and condition contains $n{=}675$ examples. $k{=}0$ is the bare query, $k{=}2$ includes two structured demonstrations, and $k{=}2s$ uses a scrambled prefix matched by length. $\dagger$The very\_low band uses the unmatched pool of 97 tokens because it could not be reliably matched by length under the study's pool construction (Section~\ref{sec:data}).
At Pythia-1.4b, $k{=}2$ top-1 is 0.960--0.988 across the four matched
bands, so gap narrowing may partly reflect ceiling compression.}
\label{tab:fewshot_pilot}
\small
\begin{tabular}{llcccccc}
\toprule
Model & Cond. & very\_low$^\dagger$ & low & medium & high & very\_high & control \\
\midrule
Pythia-70m  & $k{=}0$  & 0.129 & 0.126 & 0.167 & 0.172 & 0.157 & 0.199 \\
            & $k{=}2$  & 0.270 & 0.216 & 0.290 & 0.302 & 0.280 & 0.287 \\
            & $k{=}2s$ & 0.268 & 0.209 & 0.296 & 0.301 & 0.277 & 0.283 \\
\midrule
Pythia-160m & $k{=}0$  & 0.413 & 0.434 & 0.526 & 0.575 & 0.557 & 0.570 \\
            & $k{=}2$  & 0.600 & 0.630 & 0.724 & 0.766 & 0.757 & 0.763 \\
            & $k{=}2s$ & 0.410 & 0.440 & 0.545 & 0.643 & 0.599 & 0.618 \\
\midrule
Pythia-410m & $k{=}0$  & 0.683 & 0.680 & 0.744 & 0.812 & 0.781 & 0.788 \\
            & $k{=}2$  & 0.877 & 0.907 & 0.930 & 0.938 & 0.929 & 0.953 \\
            & $k{=}2s$ & 0.508 & 0.462 & 0.465 & 0.499 & 0.508 & 0.511 \\
\midrule
Pythia-1b   & $k{=}0$  & 0.775 & 0.790 & 0.846 & 0.859 & 0.855 & 0.859 \\
            & $k{=}2$  & 0.847 & 0.881 & 0.923 & 0.895 & 0.890 & 0.905 \\
            & $k{=}2s$ & 0.342 & 0.428 & 0.434 & 0.481 & 0.502 & 0.484 \\
\midrule
Pythia-1.4b & $k{=}0$  & 0.680 & 0.689 & 0.773 & 0.856 & 0.868 & 0.887 \\
            & $k{=}2$  & 0.942 & 0.960 & 0.988 & 0.982 & 0.982 & 0.984 \\
            & $k{=}2s$ & 0.453 & 0.458 & 0.584 & 0.687 & 0.665 & 0.673 \\
\bottomrule
\end{tabular}
\end{table}

\paragraph{Demonstration effects.}
The structured prefix condition improves top-1 accuracy over $k{=}0$ in
all 30~model--band tests (exact McNemar tests, all significant after
Benjamini--Hochberg correction over the 30-test family).
Demonstration structure provides a significant benefit over the scrambled
prefix in all 24 model--band tests from Pythia-160m upward. The six exceptions
are the Pythia-70m tests, where structured and scrambled prefixes do not differ
detectably (band means 0.274 and 0.272). At Pythia-160m, the scrambled prefix
alone also improves three of six bands, so demonstration structure is not the
only source of the gain in that model.
We also conduct a post-hoc comparison between the scrambled prefix and
conditions without a prefix, corrected as a separate family of 30 McNemar tests. The
scrambled prefix changes accuracy significantly in 27 of 30~tests. It reduces
performance in all 18~tests at Pythia-410m and above, including a decrease in
the band mean from 0.748 to 0.492 at 410m, while improving all six bands at
Pythia-70m and three of six at Pythia-160m.

\paragraph{Frequency gap contrasts.}
The primary contrast specified in advance uses matched bands only:
$\mathrm{gap} = \mathrm{acc(very\_high)} - \mathrm{acc(low)}$, testing
whether the gap changes from $k{=}0$ to $k{=}2$
(Table~\ref{tab:fewshot_gaps}).
The gap narrows significantly at Pythia-410m, Pythia-1b, and Pythia-1.4b,
with no detectable change at Pythia-70m or Pythia-160m. The Pythia-1b result
is close to the significance threshold ($q = 0.048$), with per-draw changes
of $+0.000$, $+0.067$, and $+0.102$.
The secondary contrast, which includes the unmatched \texttt{very\_low} band,
narrows significantly only at Pythia-1.4b ($+0.207 \to +0.041$,
$q = 5{\times}10^{-5}$), with no detectable change elsewhere after correction.
Because the demonstrations change both prompt length and the query's absolute
position relative to $k{=}0$, these results measure the combined effect of
adding two copying examples. The scrambled control tests sensitivity to prefix
organization while matching length and token multiset. However, it cannot
separate the benefit of coherent demonstrations from the cost of scrambling,
and this experiment does not identify a causal mechanism specific to the
demonstrations.

\begin{table}[ht]
\centering
\caption{Primary contrast specified before analysis: the accuracy gap between the matched very\_high and low bands at $k{=}0$ compared with $k{=}2$. $\Delta$ is the gap at $k{=}0$ minus the gap at $k{=}2$, so positive values indicate narrowing. The table reports bootstrap 95\% confidence intervals. $q$ is the Benjamini--Hochberg-adjusted $p$ value across five models from a two-sided permutation test blocked by draw with 100{,}000 permutations.}
\label{tab:fewshot_gaps}
\small
\begin{tabular}{lccccc}
\toprule
Model & Gap $k{=}0$ & Gap $k{=}2$ & $\Delta$ & 95\% CI & $q$ \\
\midrule
Pythia-70m  & $+0.031$ & $+0.064$ & $-0.033$ & $[-0.077, +0.013]$ & 0.217 \\
Pythia-160m & $+0.123$ & $+0.127$ & $-0.004$ & $[-0.059, +0.049]$ & 0.915 \\
Pythia-410m & $+0.101$ & $+0.022$ & $+0.079$ & $[+0.030, +0.127]$ & 0.006 \\
Pythia-1b   & $+0.065$ & $+0.009$ & $+0.056$ & $[+0.009, +0.104]$ & 0.048 \\
Pythia-1.4b & $+0.179$ & $+0.022$ & $+0.157$ & $[+0.113, +0.201]$ & ${<}0.001$ \\
\bottomrule
\end{tabular}
\end{table}

\paragraph{Circuit-level cost.}
A single timed ACDC run at the strict ICL prompt length took 72.6~minutes for
Pythia-160m at the selected threshold, compared with 33--43~minutes for the
standard prompt at the same model size. Standard Pythia-1.4b extraction takes
13.9~hours per circuit.
This single timing is not a scaling estimate, but it indicates that extracting
the full grid of few-shot circuits would be computationally expensive. We
therefore leave that analysis for future work (Section~\ref{sec:limitations}).

\end{document}